\documentclass{article} %
\usepackage{iclr2027_conference_preprint,times} %

\usepackage{amsmath,amsfonts,bm}

\def\eqref#1{equation~\ref{#1}}

\def\1{\bm{1}}

\DeclareMathAlphabet{\mathsfit}{\encodingdefault}{\sfdefault}{m}{sl}
\SetMathAlphabet{\mathsfit}{bold}{\encodingdefault}{\sfdefault}{bx}{n}

\usepackage{url}
\usepackage{graphicx}
\usepackage{amsmath}
\usepackage{amssymb}
\usepackage{booktabs}
\usepackage{multirow}
\usepackage{caption}
\usepackage{subcaption}
\usepackage{array}
\usepackage{makecell}
\usepackage{colortbl}
\usepackage{titletoc}
\usepackage[pagebackref=true,breaklinks=true,colorlinks,bookmarks=false,citecolor=blue,linkcolor=blue]{hyperref}      %
\usepackage{cleveref} %
\usepackage{enumitem} %
\setlist[itemize]{leftmargin=1em,topsep=0pt,itemsep=0pt}
\setlist[enumerate]{leftmargin=2em,topsep=0pt,itemsep=3pt}
\usepackage{tikz}
\usetikzlibrary{spy}

\newif\ifshowzoomgrid
\showzoomgridfalse %

\newcommand{\zWidth}{0.19\textwidth}
\newcommand{\zSize}{1.05cm}

\newcommand{\setzoom}[4]{%
  \gdef\zTargetX{#1}%
  \gdef\zTargetY{#2}%
  \gdef\zInsetPos{#3}%
  \gdef\zColor{#4}%
}

\newcommand{\zoomimg}[2][]{%
  \begin{tikzpicture}[spy using outlines={rectangle,draw=\zColor,thick,magnification=3,size=\zSize,connect spies}]%
    \node[anchor=south west,inner sep=0pt] (img) at (0,0) {%
      \if\relax\detokenize{#1}\relax
        \includegraphics[width=\zWidth]{#2}%
      \else
        \includegraphics[#1,width=\zWidth]{#2}%
      \fi
    };%
    \useasboundingbox (img.south west) rectangle (img.north east);%
    \begin{scope}[x={(img.south east)},y={(img.north west)}]%
      \coordinate (spy_target) at (\zTargetX,\zTargetY);%
      \coordinate (spy_inset) at (\zInsetPos);%
      \ifshowzoomgrid
        \foreach \i in {0.2,0.4,0.6,0.8} {%
          \draw[white,thin,opacity=0.7] (\i,0) -- (\i,1) (0,\i) -- (1,\i);%
          \draw[black,dotted,thin] (\i,0) -- (\i,1) (0,\i) -- (1,\i);%
          \node[font=\tiny\sffamily,fill=white,inner sep=0.5pt,opacity=0.85,text opacity=1] at (\i,0.06) {\i};%
          \node[font=\tiny\sffamily,fill=white,inner sep=0.5pt,opacity=0.85,text opacity=1] at (0.08,\i) {\i};%
        }%
      \fi
    \end{scope}%
    \spy on (spy_target) in node[fill=white,draw=\zColor,thick] at (spy_inset);%
  \end{tikzpicture}%
}

\newcommand{\zHeadCell}[1]{\makebox[\zWidth]{#1}}
\newcommand{\zHeadInput}{\zHeadCell{\footnotesize{Input}}}
\newcommand{\zHeadTarget}{\zHeadCell{\footnotesize{Target}}}
\newcommand{\zHeadExpert}{\zHeadCell{\makecell{\footnotesize{Single Task} \\ \footnotesize{Expert}}}}
\newcommand{\zHeadNaive}{\zHeadCell{\makecell{\footnotesize{Naive Soup} \\ \footnotesize{($\alpha=0.5$)}}}}

\newcommand{\zHeadOurs}{\zHeadCell{\makecell{\fontsize{8pt}{10pt}\selectfont{\textbf{$\text{OptIMerge}$}} \\ \fontsize{8pt}{10pt}\selectfont{\textbf{(Ours)}}}}}
\newcommand{\zHeaderRow}{\zHeadInput{} \zHeadTarget{} \zHeadExpert{} \zHeadNaive{} \zHeadOurs}

\title{OptIMerge: Optimization-Based Merging of Diffusion Models} %

\author{%
  \textbf{Viraj Shah}\thanks{Equal contribution.} \qquad
  \textbf{Ashwini Pokle}\footnotemark[1] \qquad
  \textbf{Peyman Milanfar} \qquad
  \textbf{Mauricio Delbracio} \qquad
  \textbf{Hossein Talebi} \\ [0.5ex]
  \centerline{\textbf{Google}} \\ [0.5ex]
  \texttt{\{virajshah, apokle, milanfar, mdelbra, htalebi\}@google.com}
}

\iclrfinalcopy %
\begin{document}
\maketitle

\begin{abstract}
The arithmetic merging of weights from multiple task-specific models fine-tuned from a shared base model, commonly known as model souping, has emerged as a popular strategy for multi-task learning without the computational overhead of joint retraining. 
However, naive scalar weight averaging frequently leads to severe performance degradation when the constituent models learn structurally divergent feature representations, leading to destructive parameter interference.
We study this problem in a largely unexplored setting: merging image-to-image (I2I) diffusion models fine-tuned for dense, pixel-accurate tasks such as image restoration. Unlike the large language models (LLMs) and text-to-image (T2I) diffusion models for which most merging methods were developed, I2I experts encode conflicting spatial priors (\textit{e.g.}, anisotropic motion-blur vs. isotropic defocus-blur) in their convolutional filters, where both naive soup and LLM-style element-wise merging introduce visible artifacts.
We introduce \textbf{OptIMerge}, a gradient-based optimization framework: instead of relying on heuristic scalars, OptIMerge learns highly expressive, yet parameter-efficient merger coefficients by minimizing a multi-task diffusion loss over a small set of $100\text{--}500$ images.
OptIMerge parameterizes the merge with architecture-aware transformations: for attention layers, a structured low-rank feature alignment with explicit channel biases; for convolutional layers, per-kernel merger coefficients, and a low-rank Cayley transform that aligns channels by orthogonal rotation without distorting the filters. 
By keeping the experts frozen and optimizing only these lightweight merger parameters (equivalent to $<0.4\%$ of the network footprint), OptIMerge achieves simultaneous multi-task capabilities that rival computationally expensive joint training and decisively outperform zero-shot model soups.
\end{abstract}

\section{Introduction}

Large-scale image diffusion models have become widely adopted for high-quality visual synthesis such as generating photorealistic images from open-ended text prompts~\citep{dhariwal2021diffusion,ho2020denoising,song2020denoising,rombach2022high}. 
Beyond standalone generation, the rich structural and semantic representations embedded within these networks function as powerful generative priors. 
Consequently, to harness these priors for specific downstream applications, a single pre-trained base model is frequently fine-tuned to master specific downstream domains,yielding specialized expert models tailored to distinct tasks. 
While these individual experts achieve excellent single-task performance, deploying a separate multi-gigabyte backbone, such as a U-Net, for every distinct capability incurs a prohibitive memory and computational footprint. 

Model merging~\citep{wortsman2022model, ilharco2022task} presents a practical solution to this inefficiency, allowing the unification of weights from multiple fine-tuned experts into a single, multi-task architecture without the extensive overhead of joint retraining. 
While model merging has seen significant success in standard text-to-image (T2I) generation~\citep{biggs2024diffusion}, its application to dense image-to-image (I2I) translation tasks, such as image restoration, remains relatively under-explored. 
To adapt a base T2I model for I2I applications, 
the input image is incorporated as an additional conditioning signal to the base model
~\citep{rombach2022high,brooks2023instructpix2pix,lin2024diffbir} which is then independently fine-tuned to resolve distinct physical degradations. This process yields specialized restoration experts---such as one model tailored for anisotropic motion deblurring, and another for isotropic defocus deblurring.

\begin{figure}
    \centering
    \includegraphics[width=\linewidth]{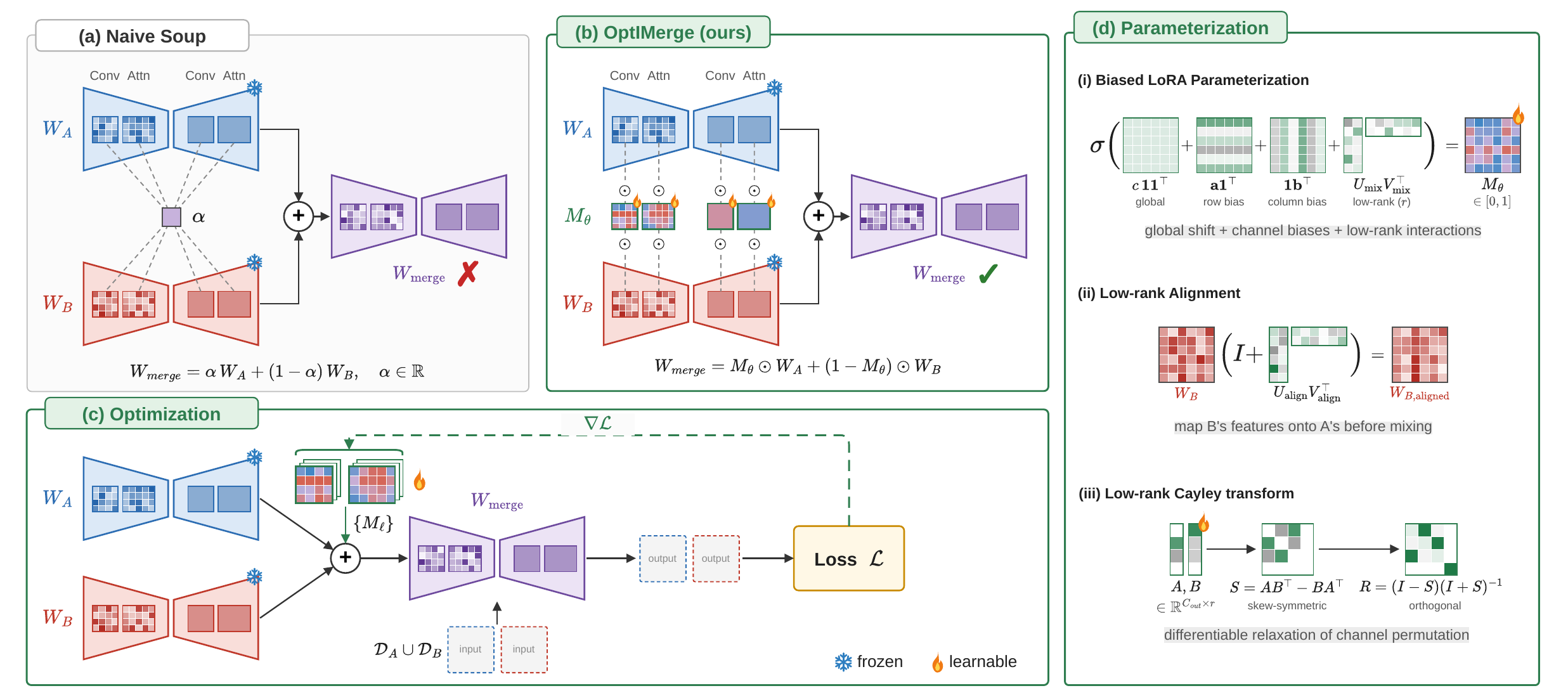}
    \caption{\small\textbf{Overview of OptIMerge.} \textbf{(a)}~Naive souping merges experts $W_A$ and $W_B$ with one scalar $\alpha$ shared by all layers. \textbf{(b)}~OptIMerge learns a merger coefficient matrix $M_\ell$ per layer, as large as that layer's weights. \textbf{(c)}~The coefficients are optimized with the diffusion loss on a small calibration set while the experts stay frozen. \textbf{(d)}~Architecture-aware parameterizations keep this under $0.4\%$ of the U-Net parameters: a biased LoRA mask with low-rank alignment for attention layers, and a low-rank Cayley rotation for convolutional layers.}
    \label{fig:overview}
    \vspace{-1.2em}
\end{figure}

One of the prevailing methods for unifying such experts is zero-shot weight averaging~\citep{wortsman2022model, biggs2024diffusion}, also known as souping (Fig.~\ref{fig:overview}(a)), which interpolates the constituent network weights using a single, heuristically chosen global scalar (e.g., $\alpha=0.5$). Such uniform scalar approach rests on two assumptions: all layers contribute equally to the merged multi-task objective, and the experts' features are aligned. However, diffusion U-Net is a deeply heterogeneous architecture: attention modules govern global semantics and cross-channel interactions, while convolutional filters dictate continuous spatial geometry and high-frequency edge detection~\citep{tumanyan2023plug,zhang2023tale,frenkel2024implicit}. 
Secondly, I2I experts are obtained by independent, full-parameter fine-tuning of the entire U-Net, which can misalign their representations throughout the network and store physically conflicting spatial priors (\textit{e.g.}, anisotropic motion-blur versus isotropic defocus kernels) in the same convolutional filters. Uniformly averaging such experts therefore causes severe destructive interference. Merging methods from other domains do not resolve this: LLM merging algorithms~\citep{yadav2023ties,yu2024language} rely on element-wise pruning, or random dropout that fracture the spatial filters, while T2I merging methods~\citep{shah2024ziplora,shenaj2025lora,wu2024mixture} compose concept-specific LoRAs that typically modify only the attention layers. Consequently, there is critical need to transition from manually guessing a single heuristic scalar to systematically learning 
expressive, architecture-aware merging parameters.

Motivated by this need for structural granularity, we introduce \textbf{OptIMerge}, an optimization-based model merging framework tailored specifically for I2I diffusion models. 
By utilizing a small calibration dataset ($100\text{--}500$ images) and keeping the constituent expert networks strictly frozen, OptIMerge optimizes highly expressive multi-task merger parameters via standard gradient descent (\Cref{fig:overview}(c)).

To ensure that this merging process remains lightweight, stable, and expressive, we propose architecture-aware parameterizations that update less than $0.4\%$ of the network footprint. 
As attention and convolutional layers have distinct functional roles, we parameterize their merging distinctly (Fig.~\ref{fig:overview}(b,d)). For attention layers, we propose a modified low-rank parameterization, which we refer to as \textit{biased LoRA}, to smoothly blend semantic features. For convolutional layers, preserving spatial integrity requires a kernel-centric approach that merges filters as indivisible units. However, because this dense filter-level fusion incurs a prohibitive parameter overhead, we introduce a novel \textit{low-rank Cayley transform} to efficiently align divergent feature spaces prior to interpolation.

We comprehensively evaluate OptIMerge with a primary focus on complex, structurally conflicting I2I restoration tasks (e.g., reconciling Motion and Defocus deblurring, and Super-Resolution). 
Our empirical evaluations demonstrate that learning structured, geometry-aware coefficients successfully resolves clashing spatial priors to achieve Pareto-optimal multi-task retention, vastly outperforming heuristic souping. Finally, as a proof of concept beyond our primary I2I setting, we show that OptIMerge can also be applied to merge subject- and style-personalized T2I models such that both concepts are retained in resulting merged model. %

\section{Related Work}

\noindent\textbf{LLM Task Arithmetic and Model Merging.} \citet{wortsman2022model} introduced ``model souping'' via uniform weight averaging ($\alpha=0.5$), which \citet{ilharco2022task} and \citet{ortiz2023task} extended to task arithmetic, and \textit{Diffusion Soup}~\citep{biggs2024diffusion} adapted for zero-shot style mixing in text-to-image generation. While model merging has matured rapidly for Large Language Models (LLMs)---producing numerous zero-shot routing, weighting, and pruning algorithms~\citep{matena2022merging, yadav2023ties, yu2024language}---merging strategies for dense Image-to-Image (I2I) diffusion models remain significantly under-explored. Unlike text generation, I2I tasks ingest concatenated high-dimensional image latents and rely heavily on the continuous 2D spatial geometry of convolutional filters to reconstruct high-fidelity pixels. Consequently, adapting state-of-the-art LLM merging algorithms presents fundamental architectural roadblocks. Leading LLM methods such as \textit{TIES-Merging}~\citep{yadav2023ties} and \textit{DARE}~\citep{yu2024language} rely on aggressive element-wise operations: magnitude thresholding, sign-consensus voting, and unconstrained random parameter dropout (e.g., $p=0.9$). While effective for massively redundant 1D linear layers in Transformers, these operations are fundamentally incompatible with physical $K \times K$ spatial filters. Applying unconstrained element-wise pruning or random dropout to a 4D vision U-Net tensor treats a $3 \times 3$ convolutional kernel as an unorganized bag of independent scalars, destroying the continuous 2D geometry required to represent visual features like edge detectors or directional blurs and inducing severe high-frequency artifacts. Recent methods for concept composition, such as MoLE~\citep{wu2024mixture}, ZipLoRA~\citep{shah2024ziplora}, and LoRA.rar~\citep{shenaj2025lora} explore the T2I tasks focusing specifically on merging Low-Rank Adaptations (LoRAs). Because LoRA fine-tuning in diffusion models is conventionally applied only to attention layers, these merging techniques naturally do not account for spatial convolutions, and cannot address the clashing 2D priors in I2I mergers. %

\noindent\textbf{Feature Alignment and Optimization-Based Merging.} Computer vision alignment algorithms such as \textit{Git Re-Basin}~\citep{ainsworth2022git} and \textit{ZipIt!}~\citep{stoica2024zipit} attempt to resolve interference by matching unaligned channels, but introduce severe architectural and computational bottlenecks for diffusion models. Designed primarily for discriminative classification networks preserving lower-dimensional decision boundaries, they fail to preserve the sensitive, continuous spatial image manifold of generative models. Specifically, \textit{ZipIt!} merges unaligned channels by altering intermediate layer dimensionalities, natively breaking the strict symmetrical channel-matching required by the dense skip-connections of a U-Net, whereas \textit{Git Re-Basin} relies on discrete bipartite matching and optimal transport that are prohibitively expensive to scale to an $890\text{M}$-parameter generative U-Net. Meanwhile, existing optimization-based merging frameworks such as \textit{AdaMerging}~\citep{yang2024adamerging} and evolutionary recipe search~\citep{akiba2025evolutionary} optimize coarse layer- or block-wise scalars, struggling to scale to high-dimensional matrices. In contrast, OptIMerge directly parameterizes structured low-rank channel routing and exact orthogonal rotations (via the Cayley transform) within a fast, differentiable gradient-descent loop to dynamically align and merge diverging feature spaces.
\vspace{-0.5em}
\section{Methodology: The OptIMerge Framework}
\vspace{-0.5em}
Let $W_A$ and $W_B$ represent the frozen parameter tensors of two expert diffusion models, independently fine-tuned for Task A and Task B. 
Our objective is to construct a unified weight tensor, $W_{\text{merge}}$, that achieves high performance on both tasks without the overhead of joint retraining. 
To achieve this, we introduce a set of learnable parameters $\theta$ that formulate a dense merger coefficient matrix $M_\theta \in [0, 1]$. The merged weights are computed via an element-wise convex combination:
\begin{equation}
    W_{\text{merge}}(\theta) = M_\theta \odot W_A + (\mathbf{1} - M_\theta) \odot W_B
\end{equation}
where $\odot$ denotes the Hadamard product. 
We optimize the parameters $\theta$ by minimizing the standard conditional diffusion denoising loss over a combined calibration
dataset $\mathcal{D} = \mathcal{D}_A \cup \mathcal{D}_B$. 
Keeping the base expert weights strictly frozen, the optimization objective is defined as:
\begin{equation}
    \min_{\theta} \mathbb{E}_{z, c \sim D,\, \epsilon \sim \mathcal{N}(\mathbf{0}, \mathbf{I}),\, t} \left[ \left\| \epsilon - \epsilon_\phi\left(z_t, t, c; W_{merge}(\theta)\right) \right\|_2^2 \right]
\end{equation}
where $z_t$ is the noisy latent at timestep $t$, $c$ represents the task-specific conditioning (e.g., text prompts or conditioning images), and $\epsilon_\phi$ is the denoising network parameterized by the dynamically merged weights $W_{merge}(\theta)$.
Here, $\mathcal{D}_A$ and $\mathcal{D}_B$ represent small, held-out 
in-domain datasets (e.g., $\sim$100-500 samples)
for Task A and Task B, respectively. %

To provide a stable optimization landscape and ground our method against standard task arithmetic baselines, we carefully initialize $\theta$ such that $M_\theta = 0.5 \cdot \mathbf{1}$. Consequently, the optimization trajectory begins exactly at the naive arithmetic model soup, allowing the network to learn continuous, structure-aware deviations from the uniform average.

\subsection{The Inadequacy of Scalar Interpolation}\label{sec:scalar_inadequacy}
The most constrained formulation of the merging objective limits $M_\theta$ to a single global scalar, functionally equivalent to optimizing the interpolation coefficient in standard task arithmetic. 
Relaxing this constraint to intermediate dimensionalities, such as learning a distinct scalar per architectural block (e.g., U-Net down-blocks, mid-block, and up-blocks) or a distinct scalar per layer, expands the hypothesis space but fundamentally restricts the routing mechanism to uniform interpolation across all feature channels simultaneously. 

Empirically, we observe that optimizing such low-dimensional scalar coefficients yields a highly constrained optimization landscape. Gradient descent frequently drives these scalar parameters to binary extremes ($0$ or $1$), effectively inducing a hard routing between experts rather than discovering a complementary blend. 
More critically, scalar interpolation implicitly assumes that the learned feature manifolds of both models are perfectly aligned and independently composable. 
It inherently lacks the representational capacity to resolve permutation drift or facilitate complex cross-channel feature compositions. 
These limitations necessitated a shift from uniform scalar routing to structured, high-dimensional matrix parameterizations.

To effectively route information across a latent diffusion U-Net, we must recognize that the architecture relies on two fundamentally different mathematical topologies: dense attention projections and localized spatial convolutions. 
Applying a uniform matrix parameterization across both layer types is suboptimal and prone to failure. 

\begin{itemize}
    \item \textbf{Attention modules} (e.g. $W_Q, W_K, W_V$ projections) govern global context and cross-channel feature interactions. Because they lack a physical spatial dimension, they are highly sensitive to representation misalignment and permutation symmetries.
    Therefore, they demand parameterizations that can explicitly shift global channel distributions and apply unconstrained low-rank projections to reconcile misaligned feature spaces.
    \item \textbf{Convolutional modules} (e.g. $K \times K$ spatial filters) govern the physical geometry, edges, and localized textures of the generated image. Applying unconstrained alignments here risks compounding magnitude scaling and fracturing the physical shape of the kernels. Therefore, they strictly require parameterizations that treat spatial filters as atomic units and leverage (orthogonal) transformations to align feature spaces without distorting weight magnitudes.
\end{itemize}

Motivated by these two distinct requirements, we introduce tailored matrix parameterizations for attention and convolutional modules in the next section.

\subsection{Diffusion Model Merging with OptIMerge}
\subsubsection{Parameterizing Attention Modules}
Attention modules ($W_Q, W_K, W_V, W_O \in \mathbb{R}^{d_{out} \times d_{in}}$) represent dense, high-dimensional parameter spaces that require fine-grained routing. 
While a standard Low-Rank Adaptation (LoRA) formulation, $M_\theta = \sigma(U_{mix} V_{mix}^T)$, offers parameter efficiency, it is topologically restrictive for model merging. 
Optimal merger matrices are rarely strictly low-rank; they typically require a global baseline shift (e.g., anchoring near $0.5$) coupled with high-frequency, localized cross-channel perturbations. Forcing a strictly low-rank factorization to simultaneously encapsulate both the global shift and the local interactions is mathematically inefficient and empirically smears the routing coefficients.

To decouple these objectives, we introduce a \textbf{biased LoRA} strategy for the mixing mask:
\begin{equation}\label{eqn:biased-lora-mask}
    M_\theta = \sigma(c + \mathbf{a}\mathbf{1}^T + \mathbf{1}\mathbf{b}^T + U_{mix} V_{mix}^T)
\end{equation}
where $c \in \mathbb{R}$ is a global scalar, $\mathbf{a} \in \mathbb{R}^{d_{out}}$ is a row bias vector, $\mathbf{b} \in \mathbb{R}^{d_{in}}$ is a column bias vector, and $U_{mix} \in \mathbb{R}^{d_{out} \times r}, V_{mix} \in \mathbb{R}^{d_{in} \times r}$ are trainable low-rank matrices. The sigmoid function $\sigma$ bounds the elements to $[0, 1]$. 
The explicit bias vectors $\mathbf{a}$ and $\mathbf{b}$ act as a powerful architectural inductive bias, enabling the optimizer to independently shift or scale entire feature channels, thereby reserving the capacity of $U_{mix}V_{mix}^T$ exclusively for modeling complex cross-channel interactions.

Furthermore, independent fine-tuning often induces representation misalignment, rendering element-wise blending of unaligned attention spaces fundamentally flawed. 
To correct this prior to interpolation, we introduce an unconstrained low-rank projection adapter to dynamically align Model B's feature space to Model A:
\begin{equation}\label{eqn:feat-align}
    W_{B, aligned} = W_B (I + U_{align} V_{align}^T)
\end{equation}
where $U_{align}$ and $V_{align}$ are trainable low-rank matrices. 
By learning this adapter simultaneously, the network can smoothly reconcile misaligned representations before mixing. The final merged attention weights are thus computed as:
\begin{equation}\label{eqn:biased-lora-merge}
    W_{merge} = M_\theta \odot W_A + (\mathbf{1} - M_\theta) \odot W_{B, aligned}
\end{equation}

\subsubsection{Parameterizing Convolutional Modules}
The core representational capacity of an Unet in LDM lies in its 4D convolutional tensors $W \in \mathbb{R}^{K \times K \times C_{in} \times C_{out}}$. 
Applying unconstrained element-wise mixing destroys the structural integrity of the $K \times K$ spatial filters and fractures the useful anisotropic and isotropic priors. 
We propose two geometry-aware strategies to resolve this:

\noindent\textbf{Kernel-Centric Merging (Spatial Broadcasting).} 
To construct a spatial routing mechanism that preserves the integrity of the filters, we enforce a single learnable coefficient per input-output channel path. 
Treating the $K \times K$ kernel as an atomic unit, we learn a matrix $P \in \mathbb{R}^{C_{in} \times C_{out}}$, apply the sigmoid to obtain $M = \sigma(P)$, and reshape it to $(1, 1, C_{in}, C_{out})$. We then broadcast it across the spatial dimensions:
\begin{equation}\label{eqn:kc-merge}
    W_{merge} = M \odot W_A + (\mathbf{1} - M) \odot W_B
\end{equation}
By treating the kernel as an indivisible structural unit, the optimizer can seamlessly scale entire directional or radial filters without fracturing them. 
However, while empirically effective, this dense parameterization incurs a massive memory overhead of $\mathcal{O}(C_{in} C_{out})$ %
for the convolutional layers.

\noindent\textbf{Low-Rank Cayley Transform (Orthogonal Alignment).} 
To achieve feature alignment with high parameter efficiency, we transition from dense scalar routing to structural basis alignment.
Independent fine-tuning of models often induces permutation drift, where models learn to store similar visual features (e.g., edge detectors) in unaligned channels. 
Standard weight matching algorithms treat this as a discrete permutation problem, requiring non-differentiable and computationally expensive bipartite matching heuristics. 
Because permutation matrices are a discrete subgroup of orthogonal matrices, we propose a continuous, differentiable relaxation: dynamically rotating Model B's feature space into alignment with Model A using an orthogonal rotation matrix $R \in \mathbb{R}^{C_{out} \times C_{out}}$.

To learn $R$ without parameter explosion ($O(C_{out}^2)$), we parameterize the Cayley Transform via a low-rank bottleneck. 
We learn two matrices $A, B \in \mathbb{R}^{C_{out} \times r}$ to construct a skew-symmetric matrix $S = A B^T - B A^T$.
The exact orthogonal rotation matrix is computed as:
\begin{equation}
    R = (I - S)(I + S)^{-1}
\end{equation}
Model B's filters are orthogonally rotated along the output channel dimension. %
For a filter tensor $W_B \in \mathbb{R}^{K \times K \times C_{in} \times C_{out}}$, the rotated weights are computed as:
\begin{equation}
    (W_{B, rotated})_{x, y, c, i} = \sum_{j=1}^{C_{out}} (W_B)_{x, y, c, j} R_{j, i}
\end{equation}
where $x, y$ are spatial indices, $c$ is the input channel, and $i, j$ index the output channels.
Finally, we blend the aligned weights using a layer-wise interpolation coefficient $\alpha$. To ensure $\alpha$ remains strictly bounded within $(0, 1)$ during unconstrained gradient descent, we parameterize it as $\alpha = \sigma(p)$, where $p \in \mathbb{R}$ is a learnable scalar parameter for that specific layer (initialized to $0$ such that $\alpha = 0.5$), and $\sigma$ is the sigmoid function:
\begin{equation}\label{eqn:cayley-merge}
    W_{merge} = \alpha W_A + (1 - \alpha) W_{B, rotated}
\end{equation}
This mathematically guarantees an isometric transformation, perfectly preserving the $L_2$ norm and spatial geometry of the filters while smoothly swapping misaligned channels. 
By aligning the feature channels directly rather than relying on dense matrix parameterization, this approach reduces the convolutional parameter footprint from $\sim109$M down to roughly $\sim2.4$M.%

We initialize the low-rank matrices $A$ and $B$ near zero. Consequently, the skew-symmetric matrix $S$ initializes at the zero matrix, yielding $R = I$. 
The optimization trajectory begins without any feature rotation, defaulting exactly to the naive arithmetic blend. 
The optimizer only deviates from the identity matrix if an orthogonal rotation minimizes the multi-task objective, ensuring that the transformation  resolves representation misalignment without scrambling already aligned features.

\noindent\textbf{Auxiliary Layers.} To maximize parameter efficiency, OptIMerge parameterizes only the high-impact components of the U-Net: the attention projections ($W_Q, W_K, W_V, W_O$) and the 4D spatial convolutions. Auxiliary parameters (e.g. normalization layers, time embeddings) are interpolated using a uniform scalar average ($\alpha = 0.5$).
\vspace{-1em}
\section{Experiments}
\vspace{-1em}
\subsection{Experimental Setup}
\noindent\textbf{Task \& Datasets.} We evaluate OptIMerge on pairs of image restoration tasks whose experts learn conflicting spatial priors: (1)~super-resolution (DF2K dataset~\citep{lim2017enhanced,agustsson2017ntire,zhou2025unires}) paired with motion deblurring (Open Images Dataset (OID) with motion deblurring, \citep{kuznetsova2020open}); (2)~super-resolution (DF2K) paired with defocus deblurring (DPDD dataset~\citep{abuolaim2020defocus}); and (3)~motion deblurring (OID) paired with defocus deblurring (DPDD). The first two pairs test merging a shared anchor task against two different degradation physics, while the third is the most challenging setting for our spatial parameterizations: the merged network must reconcile strictly anisotropic (directional) motion-blur filters with strictly isotropic (radial) defocus filters. We also report a proof-of-concept experiment on T2I personalization in Section~\ref{sec:t2i}.

\noindent\textbf{Optimization.} For the restoration tasks, we construct unseen held-out validation sets of 100-500 images per task. To mitigate gradient and loss imbalance (e.g., heavy motion blur loss vastly eclipsing defocus blur loss), we dynamically compute loss normalizing factors calculated over the first 30 steps (e.g., OID: 0.70, DPDD: 0.55). We optimize for $\sim$10k steps. %

\noindent\textbf{Architecture.} For the image restoration case, we build our framework upon a large-scale Image-Conditioned Latent Diffusion Model (LDM)~\citep{rombach2022high} parameterized for $v$-prediction~\citep{salimans2022progressive}. 
The base Text-to-Image (T2I) diffusion model is adapted into Image-to-Image (I2I) expert models by modifying the first convolutional layer to accept concatenated conditioning latents (expanding input channels to $C_{in} = 16$ and output channels to $C_{out} = 8$). 
All expert models operate at a high-fidelity $1024 \times 1024$ spatial resolution, which is compressed into $128 \times 128$ latent representations by the variational autoencoder.%

\noindent\textbf{Inference and Evaluation.} 
During inference on the held-out validation sets, we utilize a DDIM sampler \citep{song2020denoising} executed over 32 inference steps. For checkpoint selection, we evaluate models periodically during their respective training or optimization phases. 
For the Single-Task Experts, we select the checkpoint that achieves peak performance on its corresponding task-specific validation set. 
For all OptIMerge configurations, we select the checkpoint that achieves the best simultaneous multi-task performance on a validation sets of each of the tasks. 
We report standard pixel-wise and perceptual metrics: Peak Signal-to-Noise Ratio (PSNR), Structural Similarity Index Measure (SSIM), and Learned Perceptual Image Patch Similarity (LPIPS) on 300 test images for image restoration tasks.

\subsection{Results: Multi-Task Image Restoration}
\subsubsection{Baselines and OptIMerge Variants}

To isolate the efficacy of our proposed parameterizations, we benchmark against three standard training paradigms:
\begin{itemize}
    \item \textbf{Single-Task Experts:} The original independently fine-tuned models. This serves as the reference for single-task performance.
    \item \textbf{Joint Multi-Task Training:} A model fine-tuned on a unified dataset ($\mathcal{D}_A \cup \mathcal{D}_B$), updating all 890M parameters. This is the traditional, computationally expensive target bound for multi-task learning.
    \item \textbf{Naive Arithmetic Soup:} The baseline $W_{merge} = 0.5 W_A + 0.5 W_B$ configuration, requiring zero trainable parameters.
\end{itemize}

\begin{table*}[t]
\centering
\caption{\small Quantitative comparison of model merging strategies on pairwise merger across three restoration tasks: Motion Deblur (OID), Defocus Deblur (DPDD), and Super-Resolution (DF2K). [\textbf{best}, \underline{second-best}].}
\vspace{-0.8em}
\resizebox{\textwidth}{!}{
\begin{tabular}{lccccccc}
\toprule
 \multicolumn{2}{c}{\cellcolor{blue!15}(a) Motion Deblur (OID) + Defocus Deblur (DPDD)} & \multicolumn{3}{c}{\textbf{OID (300 samples)}} & \multicolumn{3}{c}{\textbf{DPDD (300 samples)}} \\
\cmidrule(lr){3-5} \cmidrule(lr){6-8}
\textbf{Method / Strategy} & \textbf{Trainable Params} & \textbf{LPIPS $\downarrow$} & \textbf{PSNR $\uparrow$} & \textbf{SSIM $\uparrow$} & \textbf{LPIPS $\downarrow$} & \textbf{PSNR $\uparrow$} & \textbf{SSIM $\uparrow$} \\
\midrule
\rowcolor{gray!12} \textit{Single-Task Experts} & --- & 0.2113 & 22.92 & 0.6266 & 0.2023 & 23.39 & 0.6392 \\
\rowcolor{gray!12} \textit{Joint Multi-Task Training} & 890M (100\%) & 0.2154 & 22.59 & 0.6173 & 0.2091 & 23.37 & 0.6404 \\
\midrule
\textit{Naive Arithmetic Soup ($\alpha=0.5$)} & 0 (0\%) & 0.2899 & 20.88 & 0.5601 & 0.2641 & 23.15 & 0.6398 \\ %
\textit{TIES-Merging}~\citep{yadav2023ties} & 0 (0\%) & 0.3252 & 19.86 & 0.5367 & 0.3898 & 21.59 & 0.6062 \\ %
\textit{DARE}~\citep{yu2024language} & 0 (0\%) & 0.3042 & 20.25 & 0.5364 & 0.3032 & 22.30 & 0.6156 \\ %
\midrule
$\text{OptIMerge}_{\text{Attn}}$ & $\sim$1.9M & 0.2265 & 22.43 & 0.6049 & 0.2213 & 23.59 & 0.6423 \\ %
$\text{OptIMerge}_{\text{Conv-KC}}$ & $\sim$109.2M& 0.2297 & 22.41 & 0.6051 & 0.2334 & 23.34 & 0.6368 \\ %
$\text{OptIMerge}_{\text{Conv-Cayley}}$ & $\sim$2.4M& \underline{0.2171} & \underline{22.66} & \underline{0.6165} & \underline{0.2186} & \underline{23.86} & \underline{0.6540} \\ %
$\text{OptIMerge}$ (\textbf{Ours}) & $\sim$3.2M ($<0.4\%$) & \textbf{0.2168} & \textbf{22.72} & \textbf{0.6190} & \textbf{0.2191} & \textbf{23.91} & \textbf{0.6561} \\ %
\bottomrule
\end{tabular}
}
\resizebox{\textwidth}{!}{
\begin{tabular}{lccccccc}
\toprule
\multicolumn{2}{c}{\cellcolor{orange!14} (b) Defocus Deblur (DPDD) + Super-Resolution (DF2K)}  & \multicolumn{3}{c}{\textbf{DPDD}} & \multicolumn{3}{c}{\textbf{DF2K SR}} \\
\cmidrule(lr){3-5} \cmidrule(lr){6-8}
\textbf{Method / Strategy} & \textbf{Trainable Params} & \textbf{LPIPS $\downarrow$} & \textbf{PSNR $\uparrow$} & \textbf{SSIM $\uparrow$} & \textbf{LPIPS $\downarrow$} & \textbf{PSNR $\uparrow$} & \textbf{SSIM $\uparrow$} \\
\midrule
\rowcolor{gray!12} \textit{Single-Task Experts} & --- & 0.2023 & 23.39 & 0.6392 & 0.3131 & 21.26 & 0.5323 \\
\rowcolor{gray!12} \textit{Joint Multi-Task Training} & 890M (100\%) & 0.2185 & 23.11 & 0.6293 & 0.3160 & 21.24 & 0.5255 \\
\midrule
\textit{Naive Arithmetic Soup ($\alpha=0.5$)} & 0 (0\%) & 0.2774 & 22.57 & 0.6121 & 0.3447 & 21.10 & 0.5259 \\
\textit{TIES-Merging}~\citep{yadav2023ties} & 0 (0\%) & 0.3744 & 21.55 & 0.5986 & 0.3761 & 19.76 & 0.5032 \\ %
\textit{DARE}~\citep{yu2024language} & 0 (0\%) & 0.3064 & 21.92 & 0.5978 & 0.3777 & 20.35 & 0.4860 \\ %
\midrule
$\text{OptIMerge}_{\text{Attn}}$ & $\sim$1.9M & \underline{0.2506} & \underline{22.66} & 0.6113 & \underline{0.3208} & 20.88 & 0.5193 \\
$\text{OptIMerge}_{\text{Conv-KC}}$ & $\sim$109.2M & 0.2814
 & 22.37 & 0.6038 & 0.3501 & 20.50 & 0.4949 \\
$\text{OptIMerge}_{\text{Conv-Cayley}}$ & $\sim$2.4M & 0.2606 & 22.55 & \underline{0.6121} & \textbf{0.3143} & \textbf{21.38} & \textbf{0.5333} \\
$\text{OptIMerge}$ (\textbf{Ours}) & $\sim$3.2M ($<0.4\%$) & \textbf{0.2365} & \textbf{23.13} & \textbf{0.6258} & 0.3210 & \underline{21.12} & \underline{0.5231} \\
\bottomrule
\end{tabular}
}
\resizebox{\textwidth}{!}{
\begin{tabular}{lccccccc}
\toprule
\multicolumn{2}{c}{\cellcolor{teal!15}(c) Motion Deblur (OID) + Super-Resolution (DF2K)} & \multicolumn{3}{c}{\textbf{OID (300 samples)}} & \multicolumn{3}{c}{\textbf{DF2K SR (300 samples)}} \\
\cmidrule(lr){3-5} \cmidrule(lr){6-8}
\textbf{Method / Strategy} & \textbf{Trainable Params} & \textbf{LPIPS $\downarrow$} & \textbf{PSNR $\uparrow$} & \textbf{SSIM $\uparrow$} & \textbf{LPIPS $\downarrow$} & \textbf{PSNR $\uparrow$} & \textbf{SSIM $\uparrow$} \\
\midrule
\rowcolor{gray!12} \textit{Single-Task Experts} & --- & 0.2113 & 22.92 & 0.6266 & 0.3131 & 21.26 & 0.5323 \\
\rowcolor{gray!12} \textit{Joint Multi-Task Training} & 890M (100\%) & 0.2386 & 22.03 & 0.5867 & 0.2993 & 21.49 & 0.5407 \\
\midrule
\textit{Naive Arithmetic Soup ($\alpha=0.5$)} & 0 (0\%) & 0.2938 & 20.31 & 0.5292 & 0.3501 & 20.99 & 0.4970 \\ %
\textit{TIES-Merging}~\citep{yadav2023ties} & 0 (0\%) & 0.3480 & 19.34 & 0.5025 & 0.3805 & 19.44 & 0.4797 \\ %
\textit{DARE}~\citep{yu2024language} & 0 (0\%) & 0.3023 & 20.02 & 0.5175 & 0.3949 & 20.14 & 0.4548 \\ %
\midrule
$\text{OptIMerge}_{\text{Attn}}$ & $\sim$1.9M & 0.2454 & 21.93 & 0.5821 & 0.3191 & 21.42 & 0.5271  \\ %
$\text{OptIMerge}_{\text{Conv-KC}}$ & $\sim$109.2M & 0.2428 & \textbf{22.16} & \underline{0.5907} & \textbf{0.3087} & \underline{21.43} & \underline{0.5327} \\ %
$\text{OptIMerge}_{\text{Conv-Cayley}}$ & $\sim$2.4M & 0.2428 & 21.63 & 0.5742 & \underline{0.3175} & 21.27 & 0.5283 \\ %
$\text{OptIMerge}$ (\textbf{Ours}) & $\sim$3.2M ($<0.4\%$) & \textbf{0.2298} & \textbf{22.16} & \textbf{0.5980} & 0.3207 & \textbf{21.45} & \textbf{0.5338} \\ %
\bottomrule
\end{tabular}
}
\label{tab:main_results}
\end{table*}

To ablate the specific contributions of our geometric and semantic alignment strategies, we evaluate three distinct OptIMerge configurations. 
\begin{itemize}
    \item $\text{\textbf{OptIMerge}}_{\textbf{Attn}}$: Applies biased LoRA strategy to the attention modules, utilizing both the unconstrained feature alignment projection (\Cref{eqn:feat-align}) and the biased LoRA routing mask (\Cref{eqn:biased-lora-mask}). 
    All convolutional modules are left at a uniform $\alpha=0.5$ average.
    \item $\text{\textbf{OptIMerge}}_{\textbf{Conv-KC}}$: Applies \textit{Kernel-Centric} spatial broadcasting to the convolutional modules. 
    For the attention modules, it applies the base biased LoRA routing mask (\Cref{eqn:biased-lora-mask}) but omits the unconstrained feature alignment projection (\Cref{eqn:feat-align}).
    \item $\text{\textbf{OptIMerge}}_{\textbf{Conv-Cayley}}$: Applies the \textit{Low-Rank Cayley Transform} for orthogonal alignment to the convolutional modules. 
    For the attention modules, it applies the base biased LoRA routing mask but strictly omits the unconstrained feature alignment projection.
    \item $\text{\textbf{OptIMerge}}$ \textbf{(Ours)}: The complete framework. 
    Simultaneously applies the Cayley Transform to orthogonally align convolutional modules, and the full biased LoRA (\Cref{eqn:biased-lora-merge}) to attention modules.
\end{itemize}

\begin{figure*}[t]
    \centering
    \setlength{\tabcolsep}{1pt}
    
    \renewcommand{\zWidth}{0.126\linewidth}
    \renewcommand{\zSize}{0.72cm}
    
    \begin{tabular}{cc ccccccc}
        & & \footnotesize{Input} & \footnotesize{Target} & \makecell{\footnotesize{Single Task} \\ \footnotesize{Expert}}
        & \makecell{\footnotesize{Naive Soup} \\ \footnotesize{($\alpha=0.5$)}}& \footnotesize{TIES} & \footnotesize{DARE} & \makecell{\fontsize{8pt}{10pt}\selectfont{\textbf{$\text{OptIMerge}$}} \\ \fontsize{8pt}{10pt}\selectfont{\textbf{(Ours)}}} \\
        \cmidrule(lr){3-9}
        
        \noalign{\setzoom{0.38}{0.58}{0.79, 0.21}{yellow}}
         \cellcolor{blue!12} &
        \rotatebox{90}{\makecell{\small ~~~~~~DPDD}} &
        \zoomimg{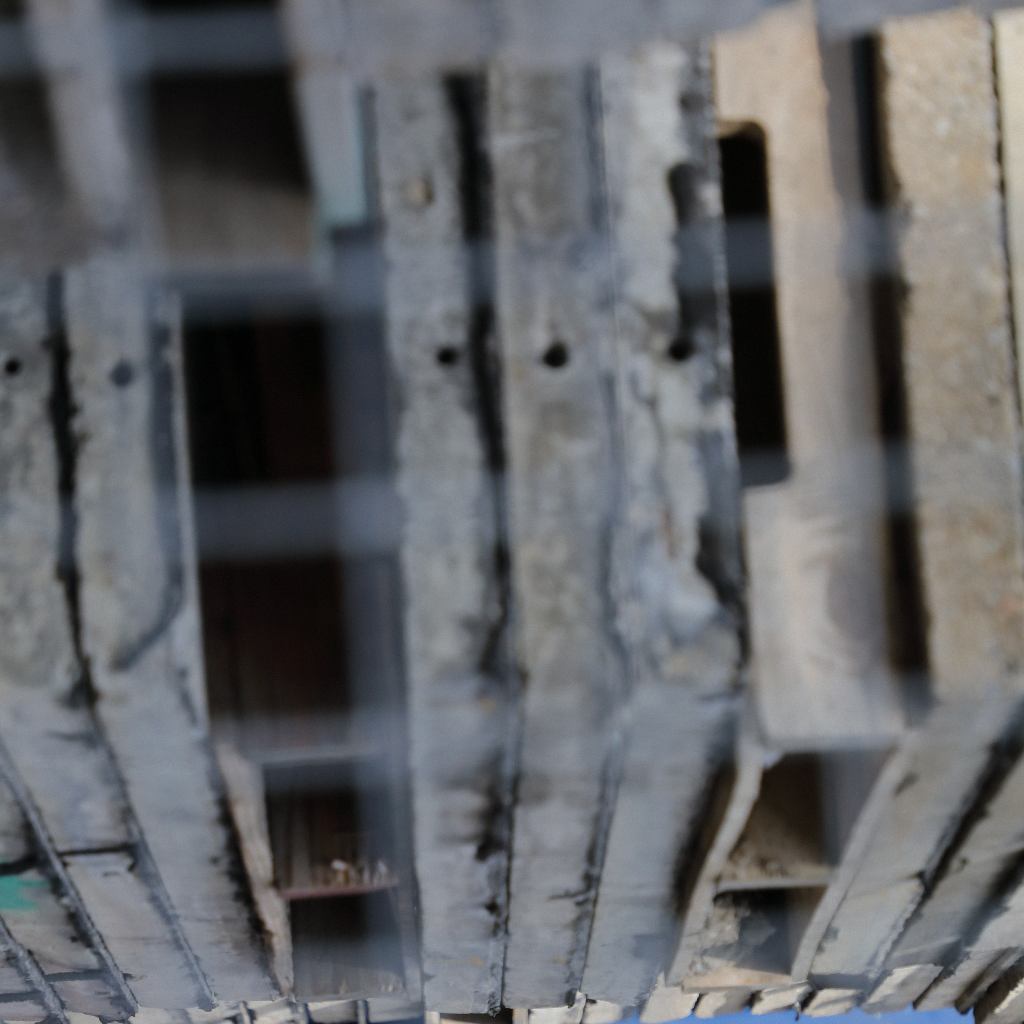} &
        \zoomimg{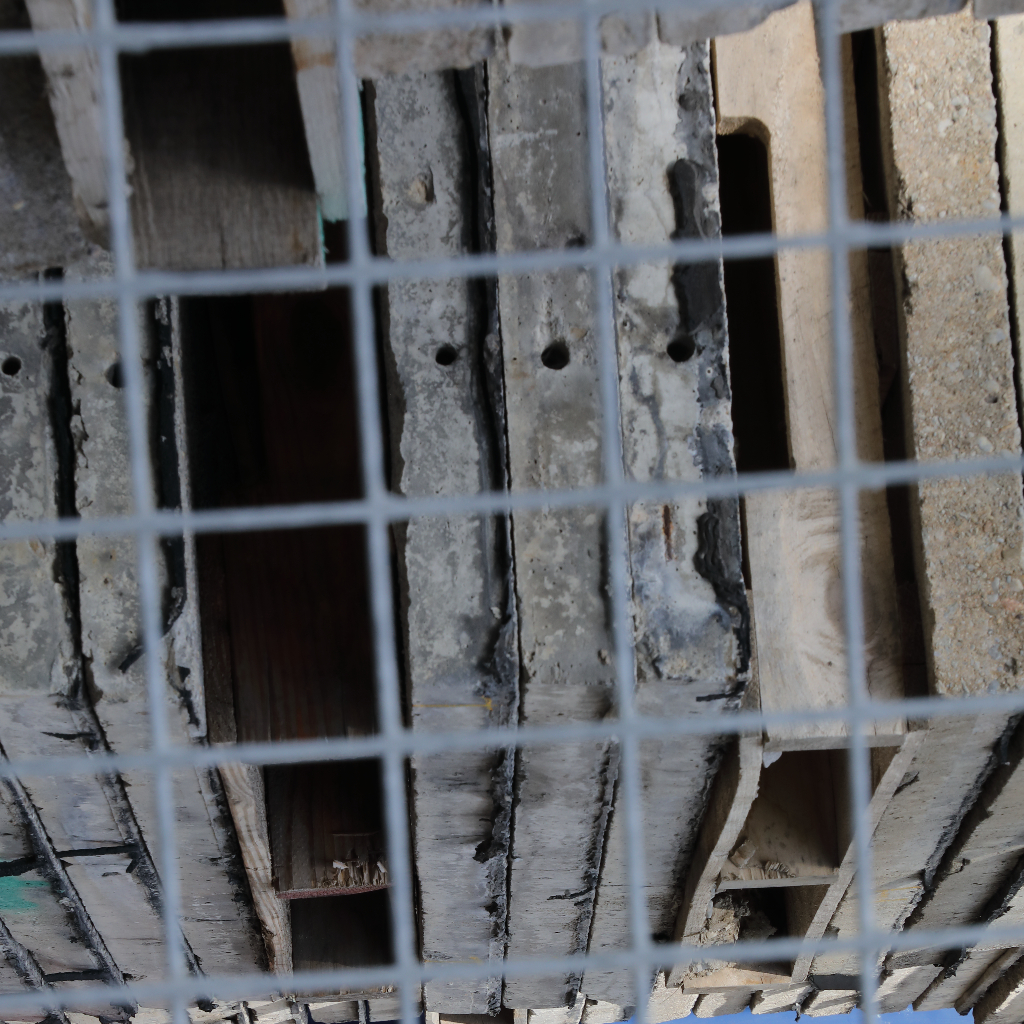} &
        \zoomimg{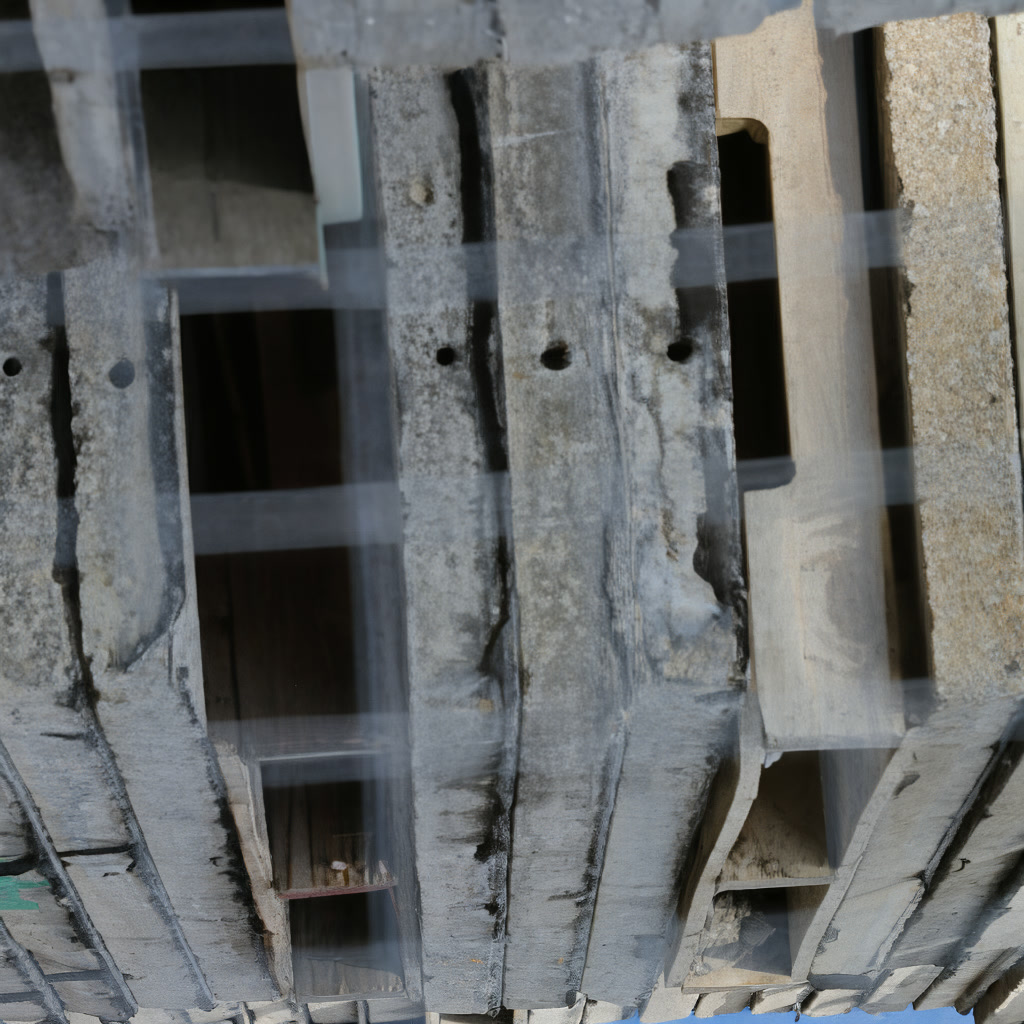} &
        \zoomimg{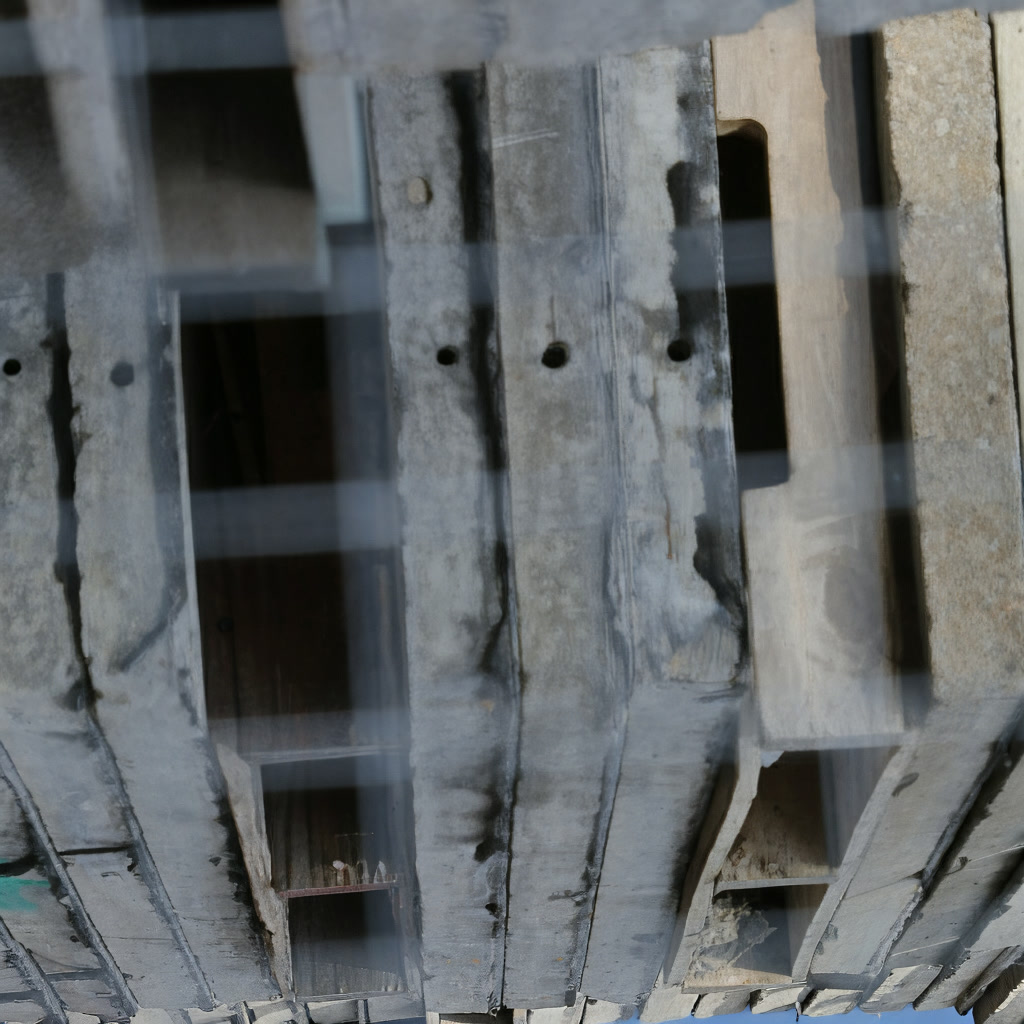} &
        \zoomimg{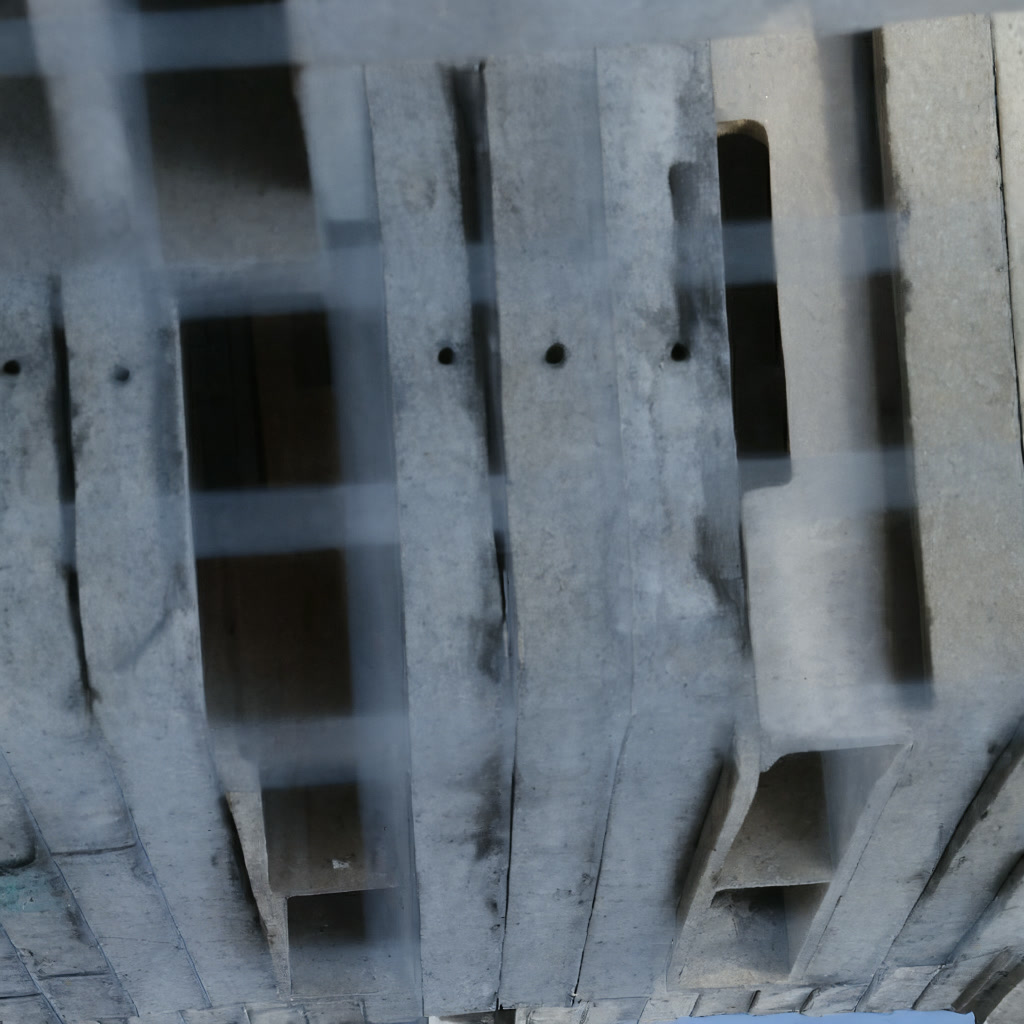} &
        \zoomimg{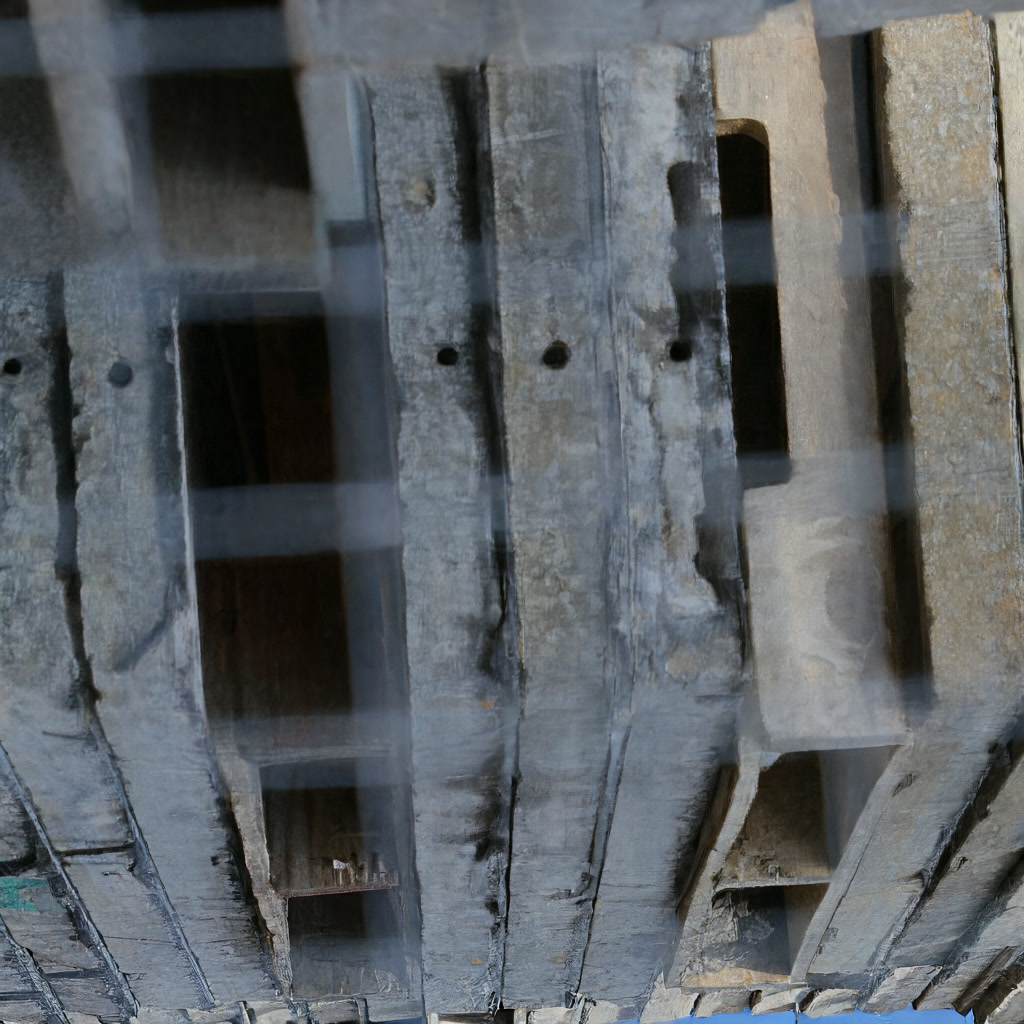} &
        \zoomimg{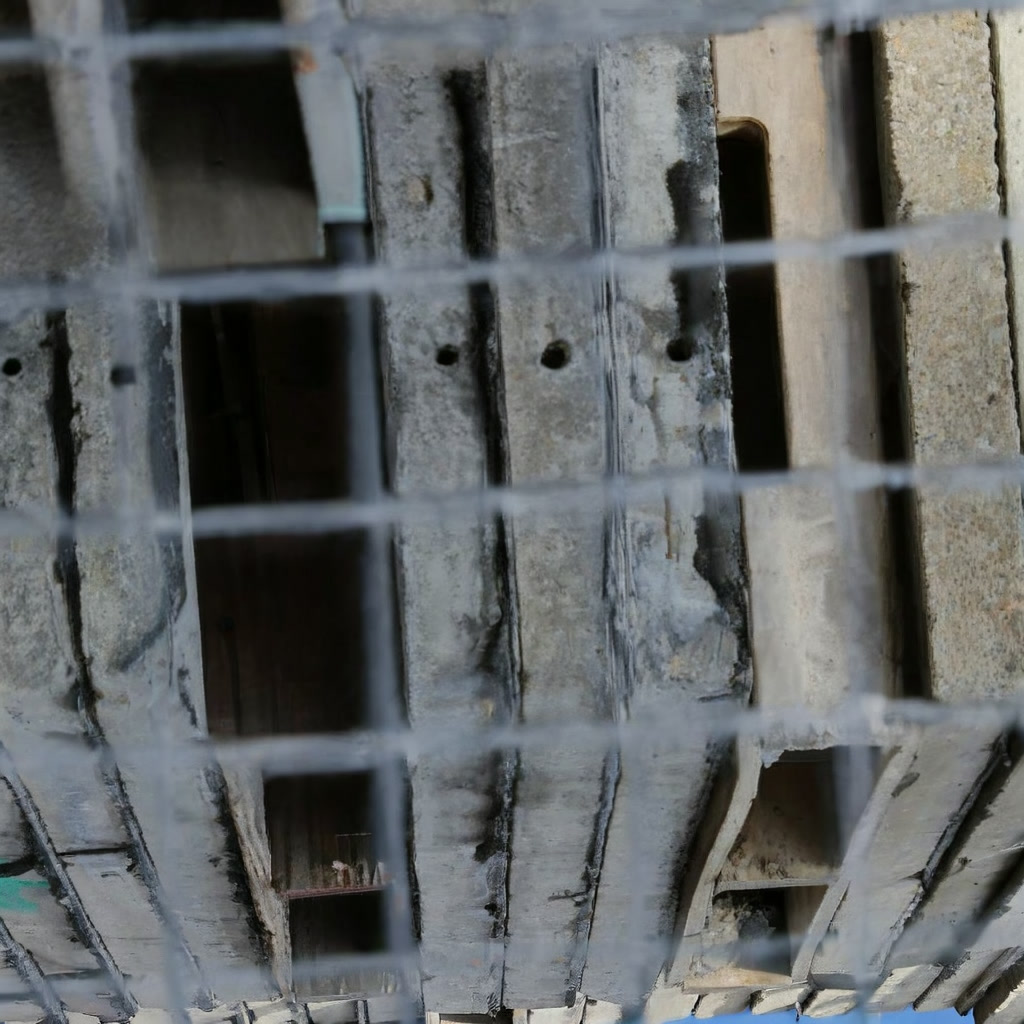} \\

        \noalign{\setzoom{0.30}{0.50}{0.79, 0.21}{yellow}}
        \multirow[c]{-10}{1.2em}{~\rotatebox{90}{\cellcolor{blue!12}\footnotesize \textbf{(a) OID + DPDD}}} &
        \rotatebox{90}{\makecell{\small ~~~~~~~~OID}} &
        \zoomimg{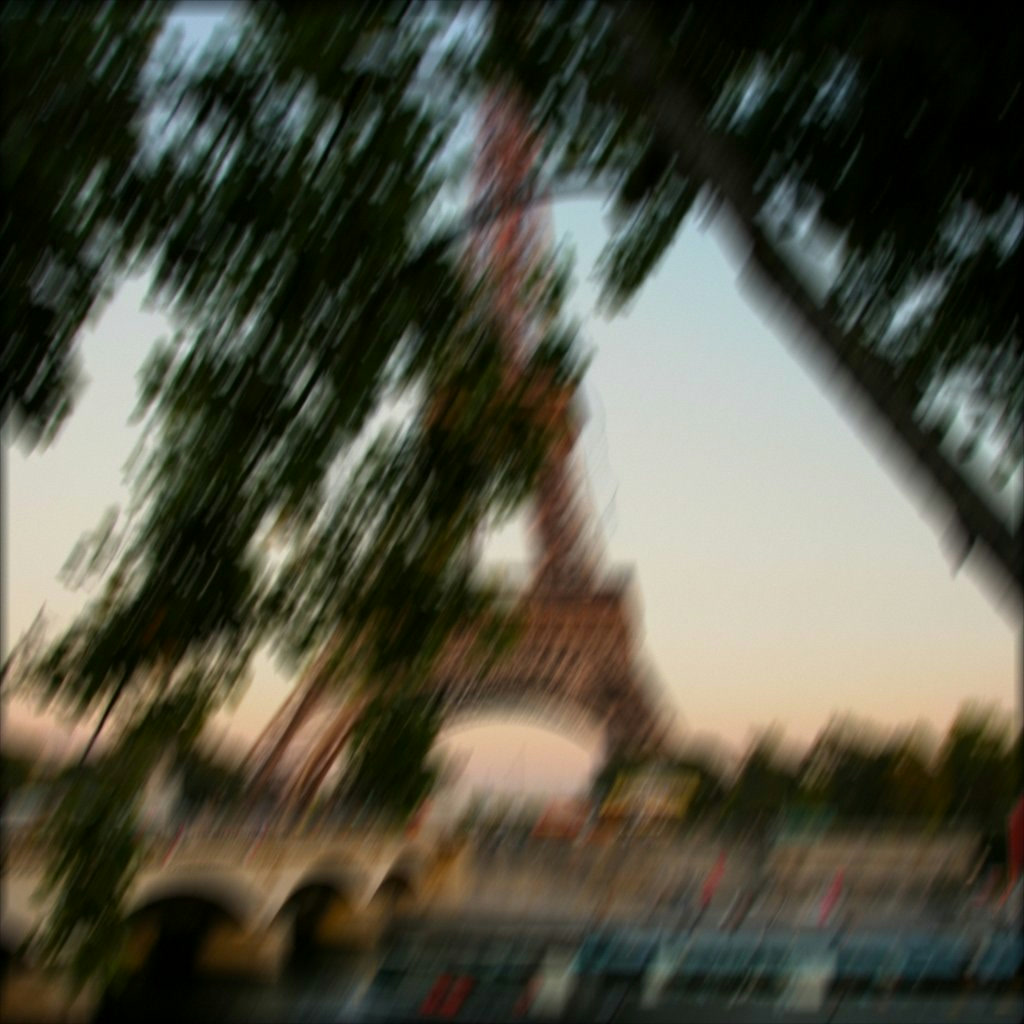} &
        \zoomimg{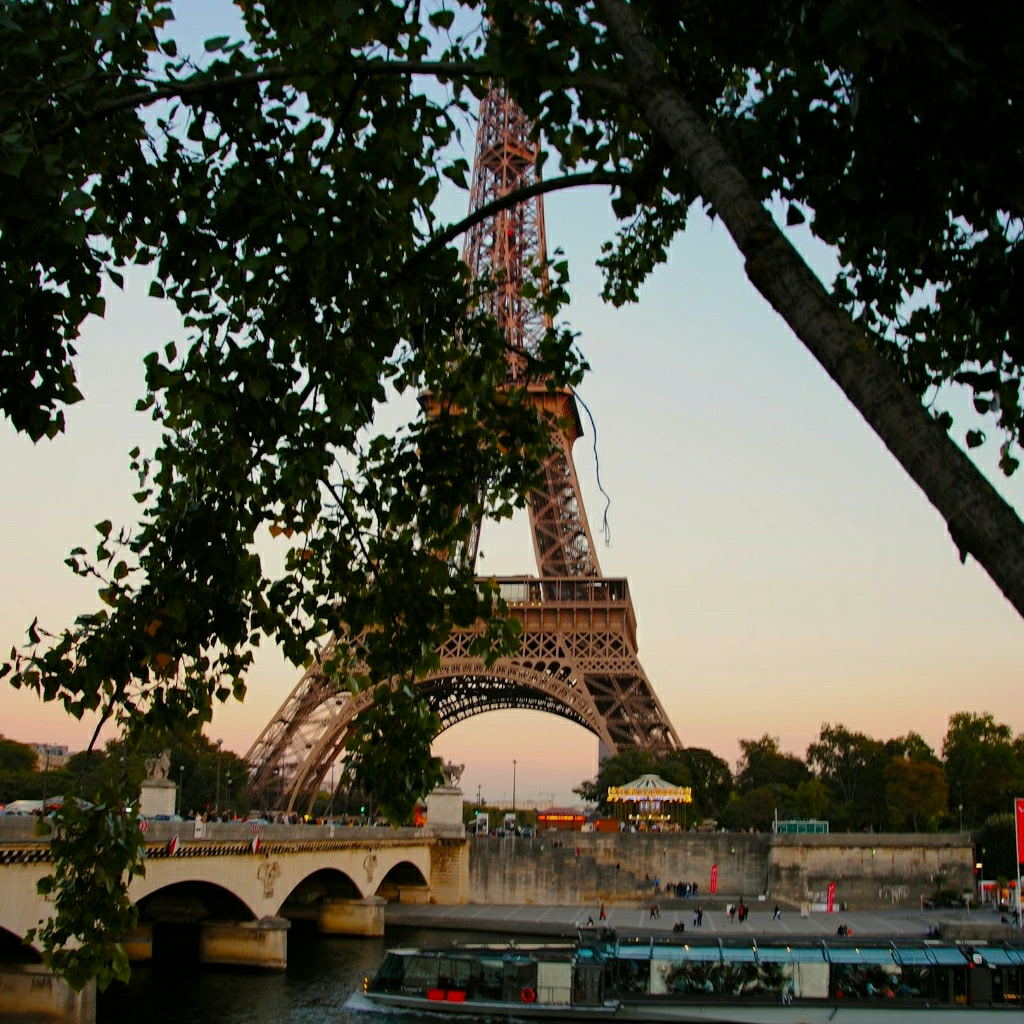} &
        \zoomimg{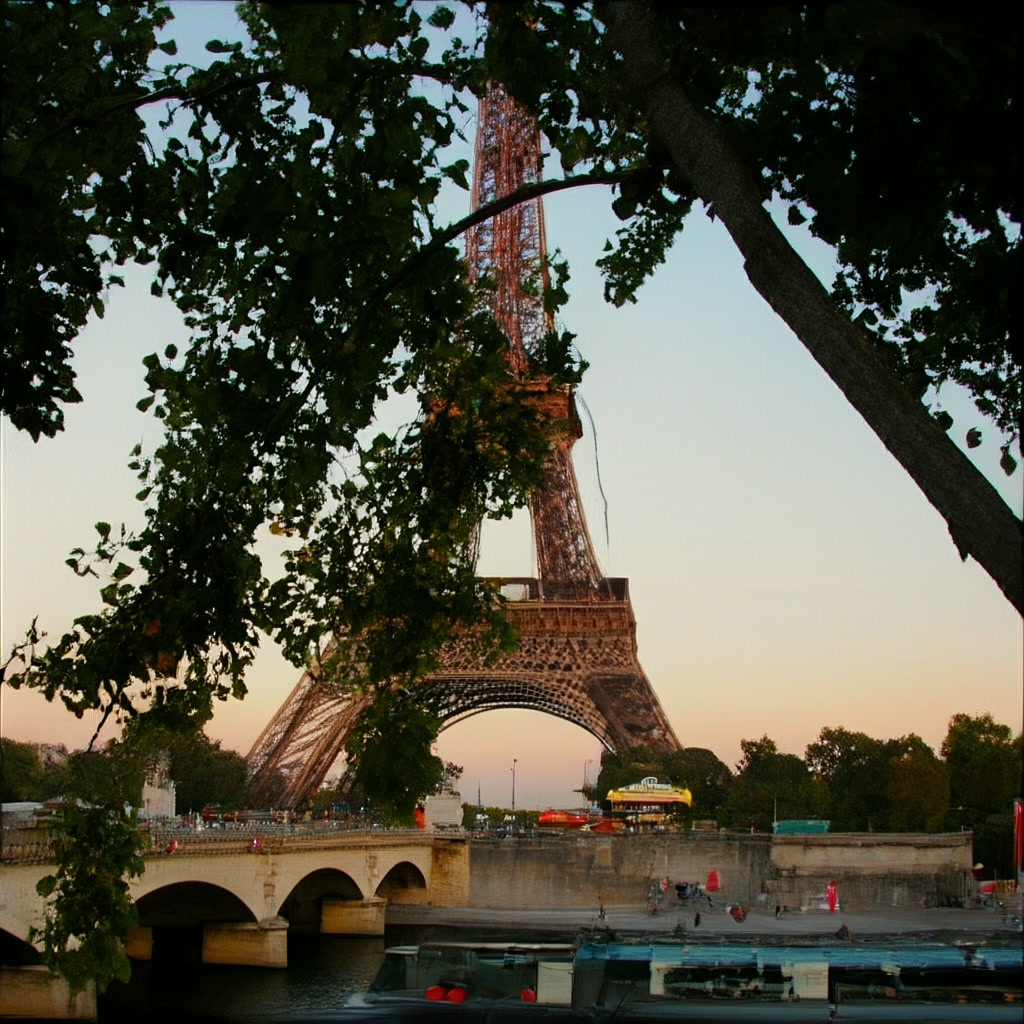} &
        \zoomimg{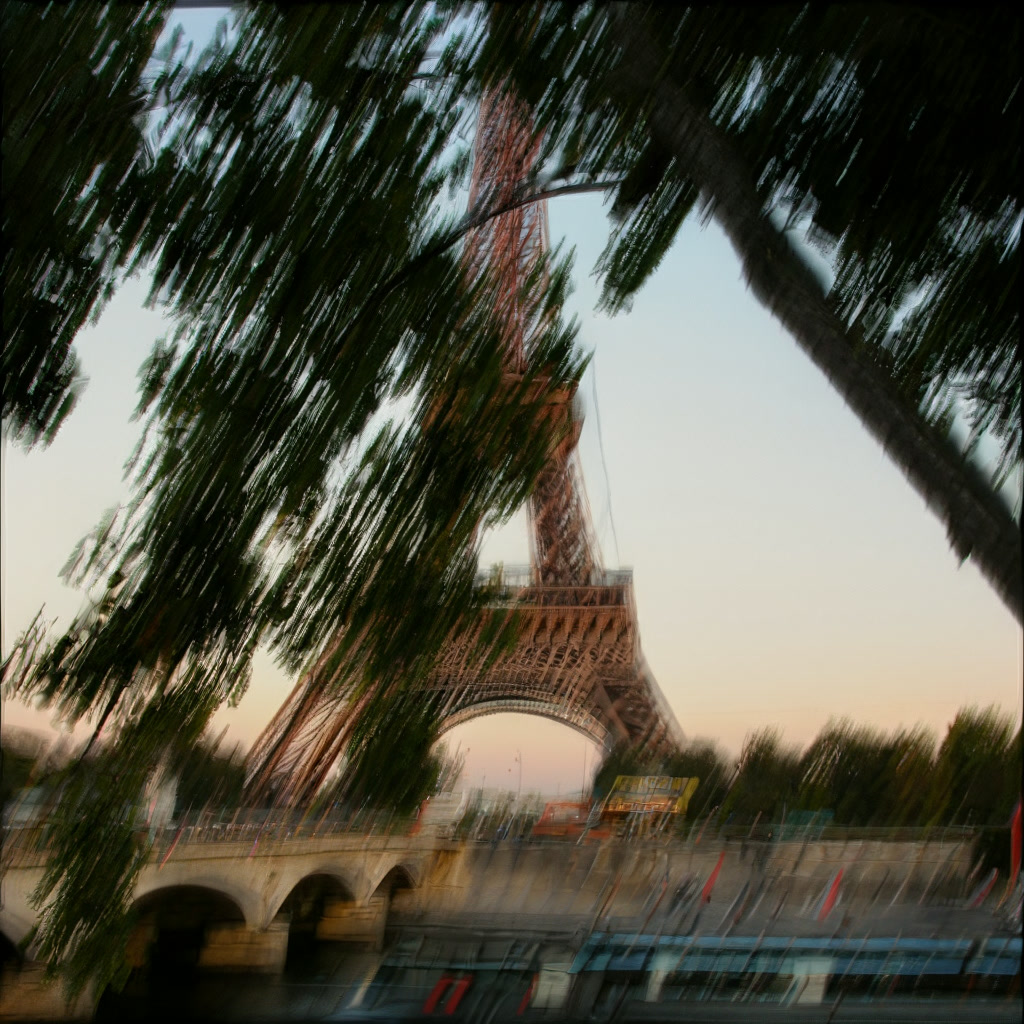} &
        \zoomimg{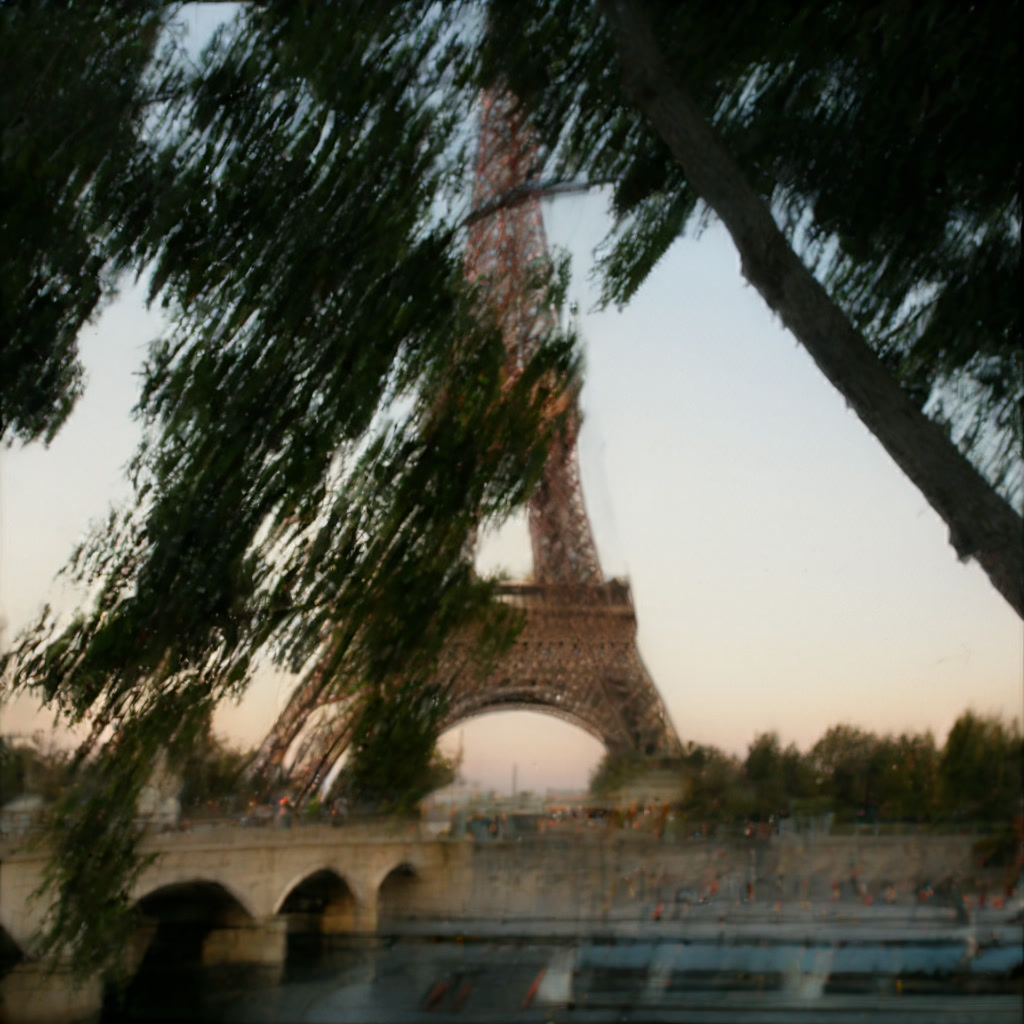} &
        \zoomimg{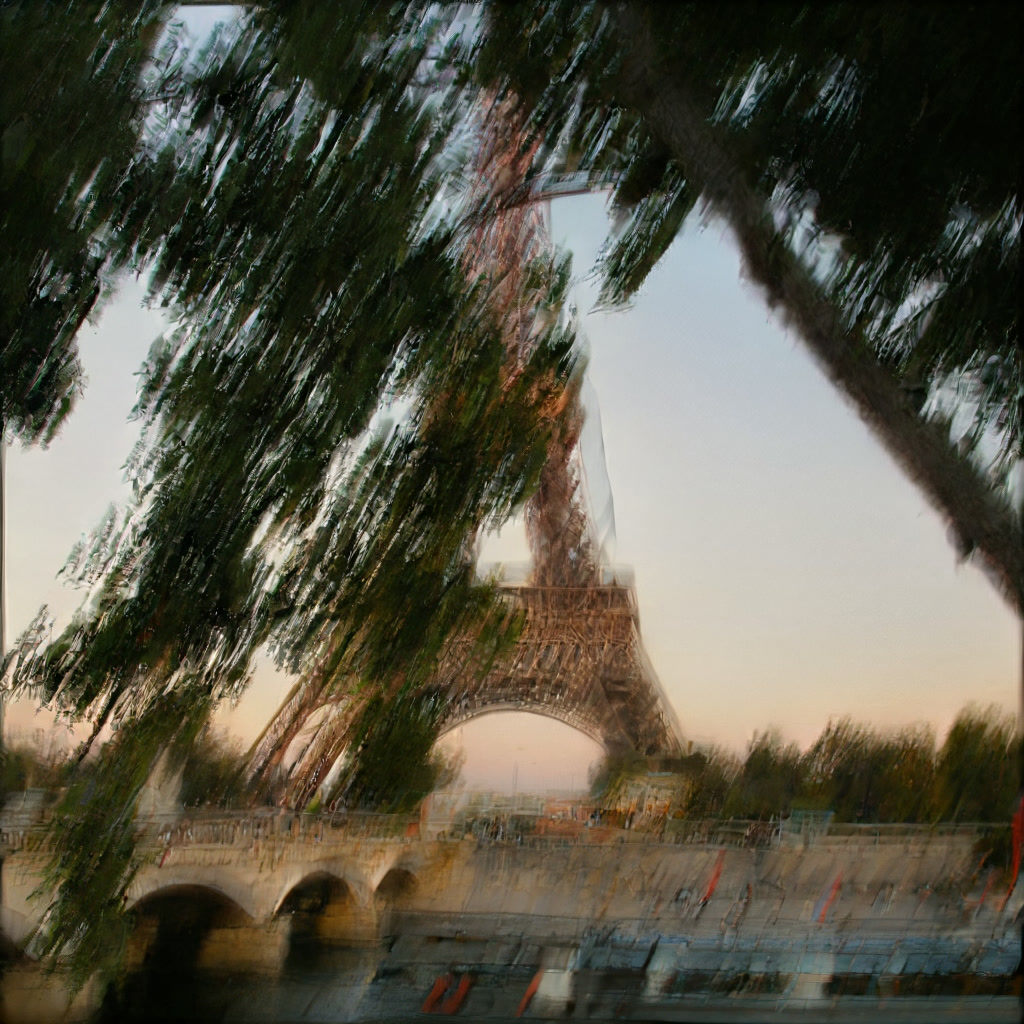} &
        \zoomimg{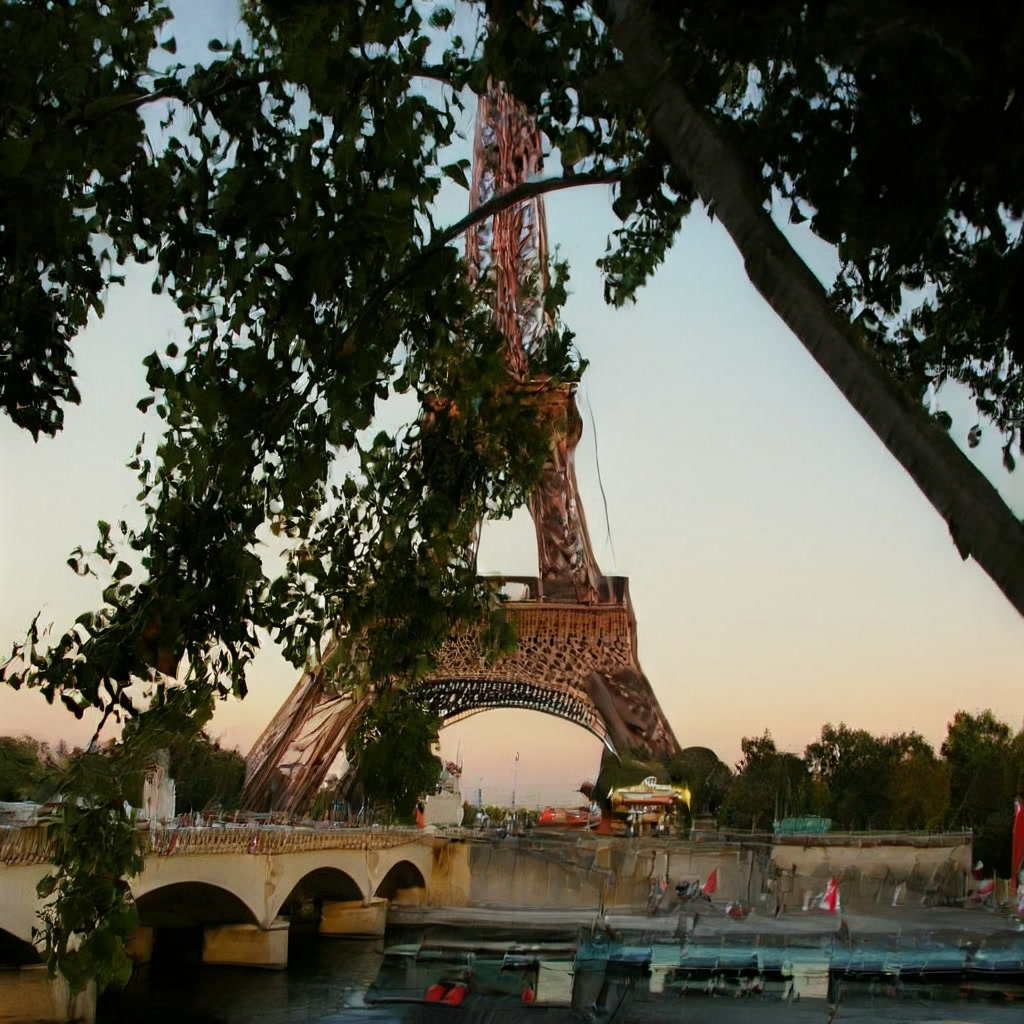} \\
        
        \noalign{\vspace{3pt}}
        \midrule
        \noalign{\vspace{3pt}}
        
        \noalign{\setzoom{0.38}{0.58}{0.79, 0.21}{yellow}}
        \cellcolor{orange!14} &
        \rotatebox{90}{\makecell{\small ~~~~~~DPDD}} &
        \zoomimg{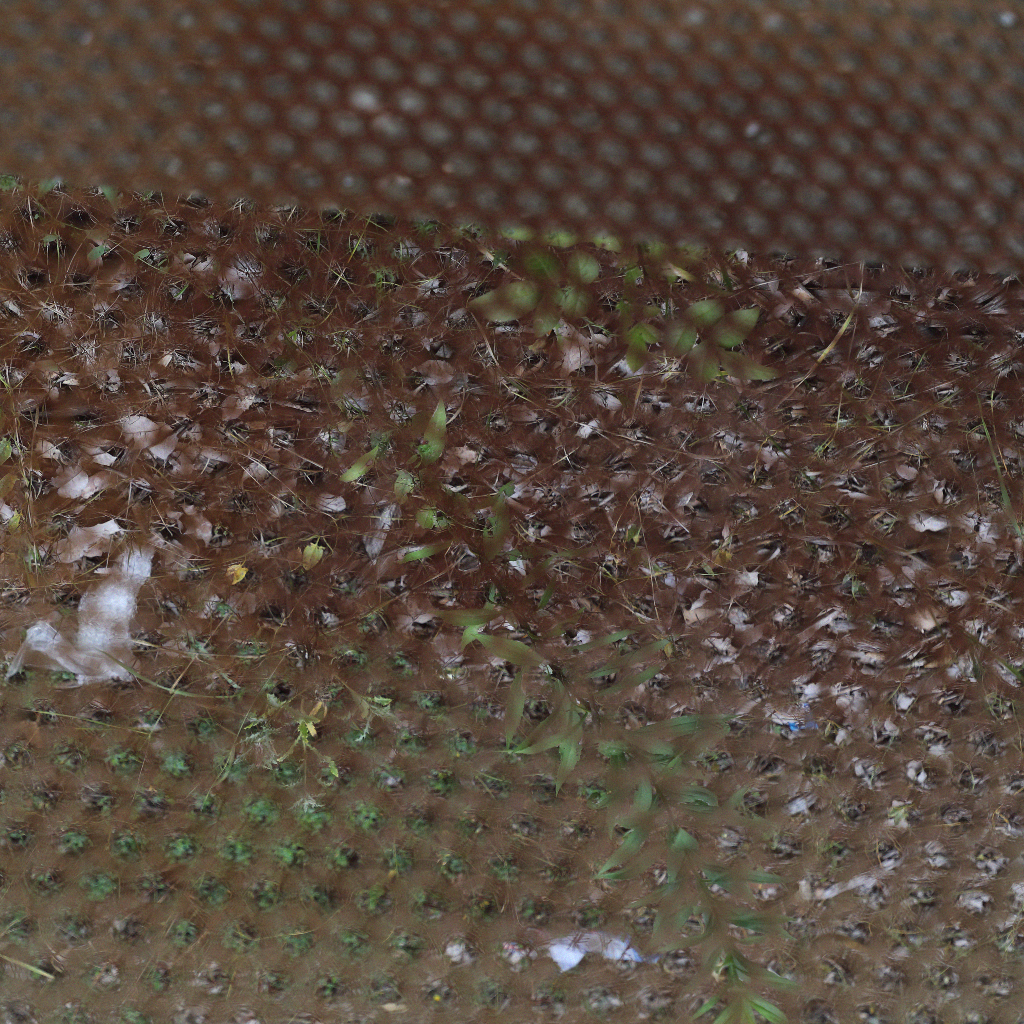} &
        \zoomimg{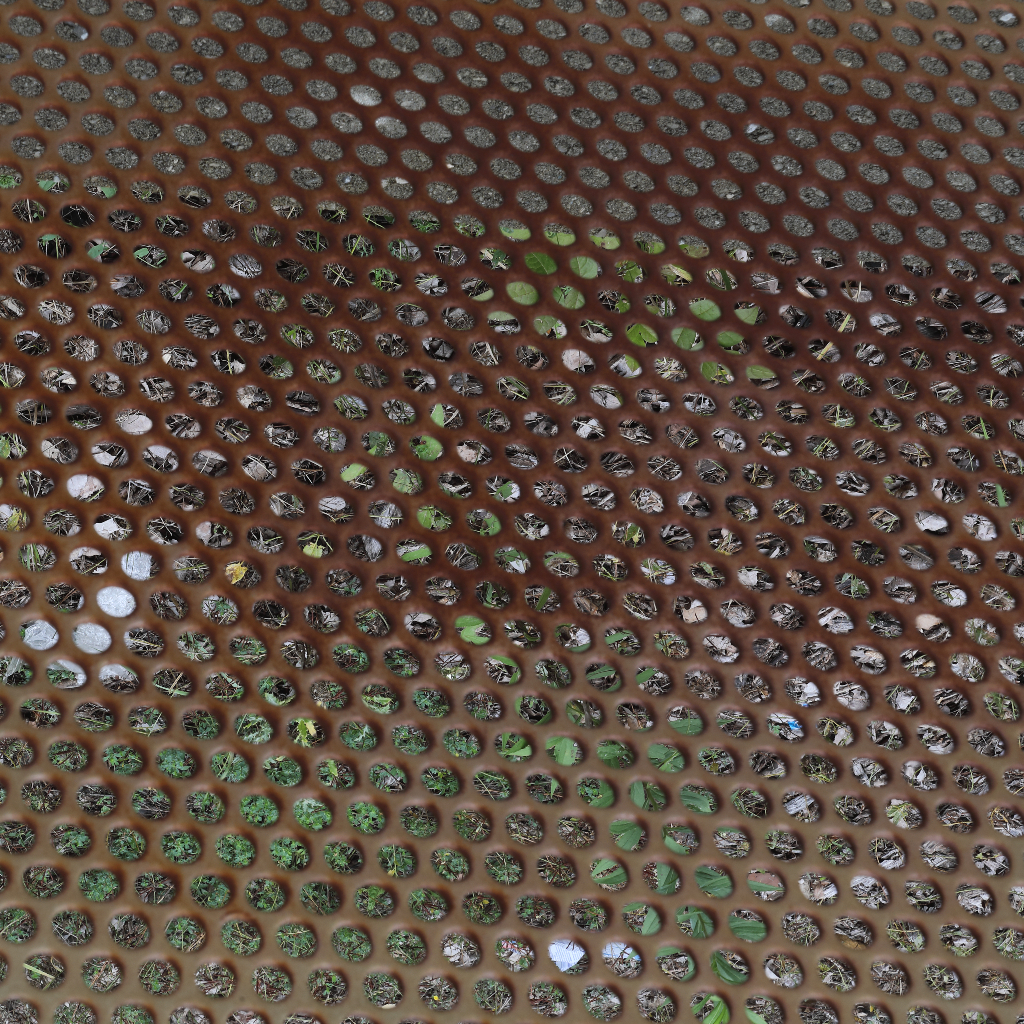} &
        \zoomimg{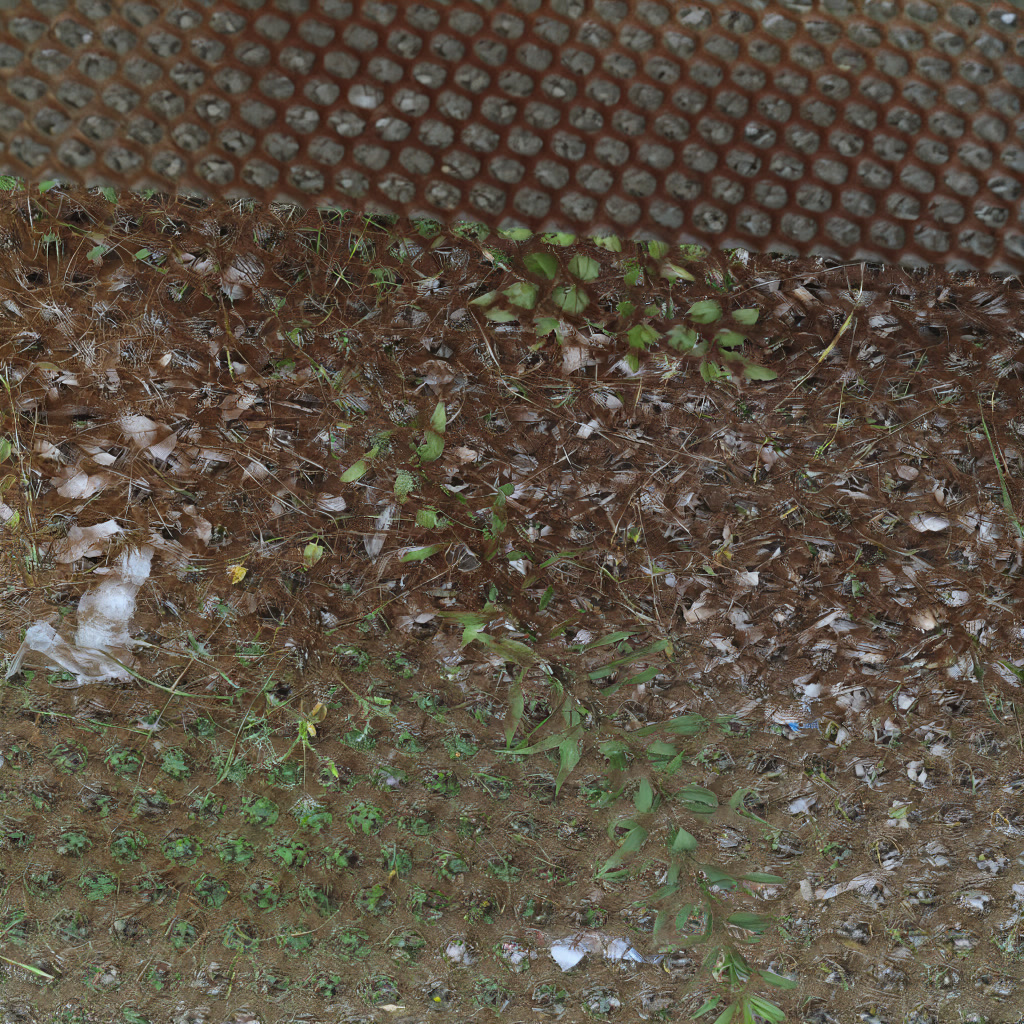} &
        \zoomimg{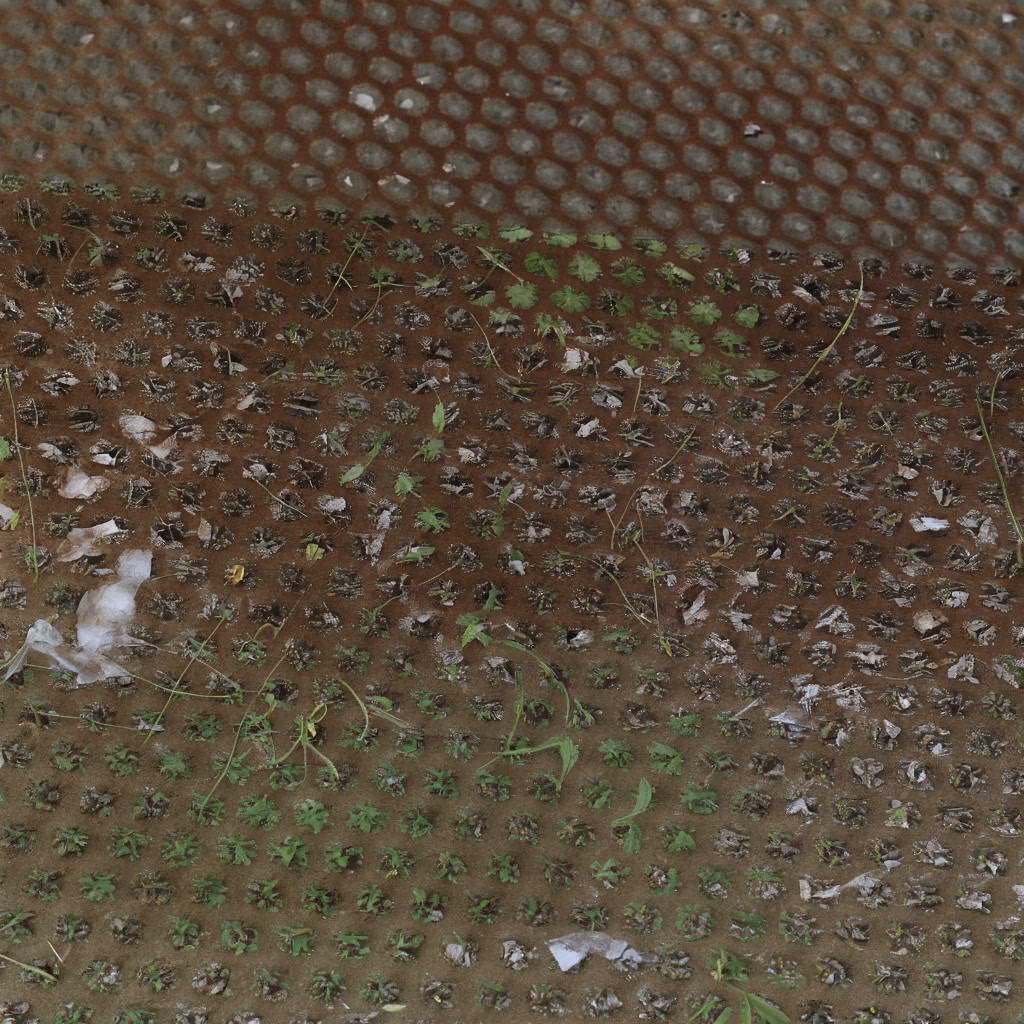} &
        \zoomimg{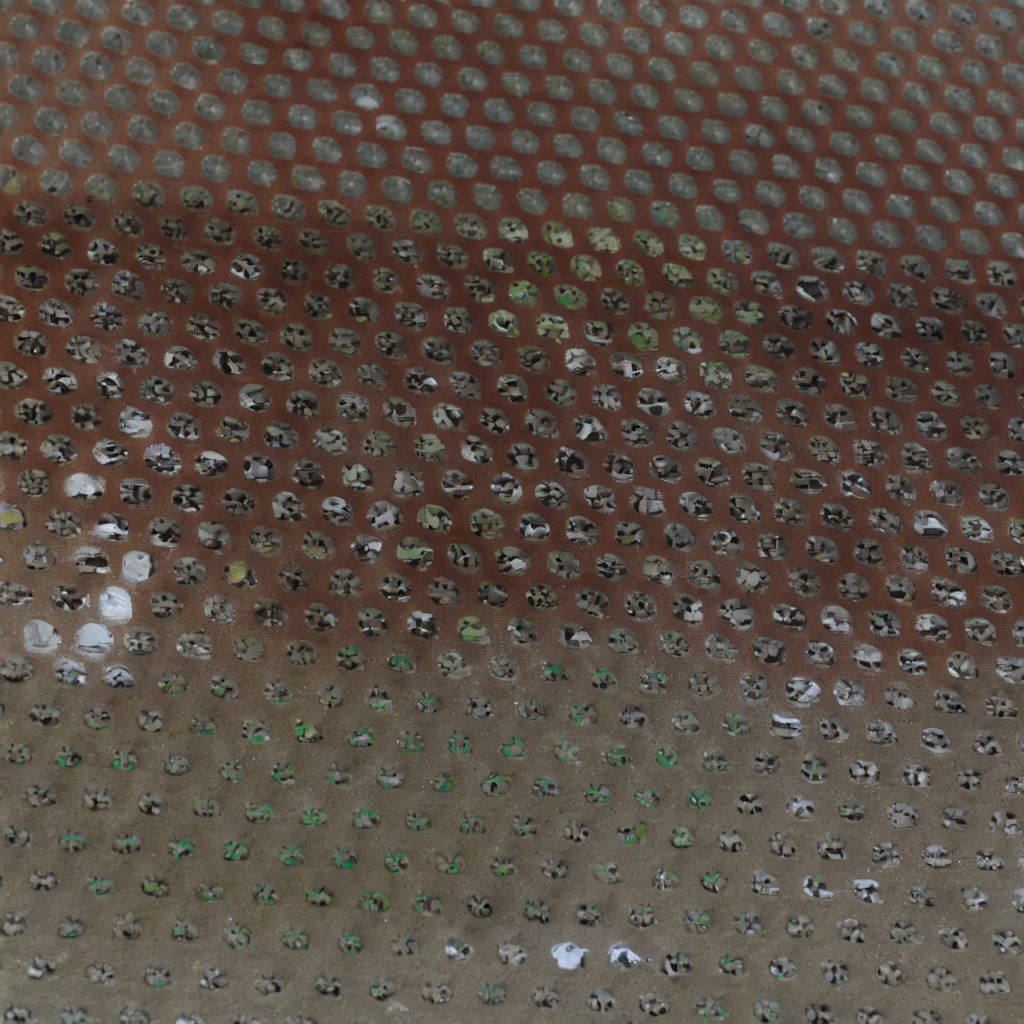} &
        \zoomimg{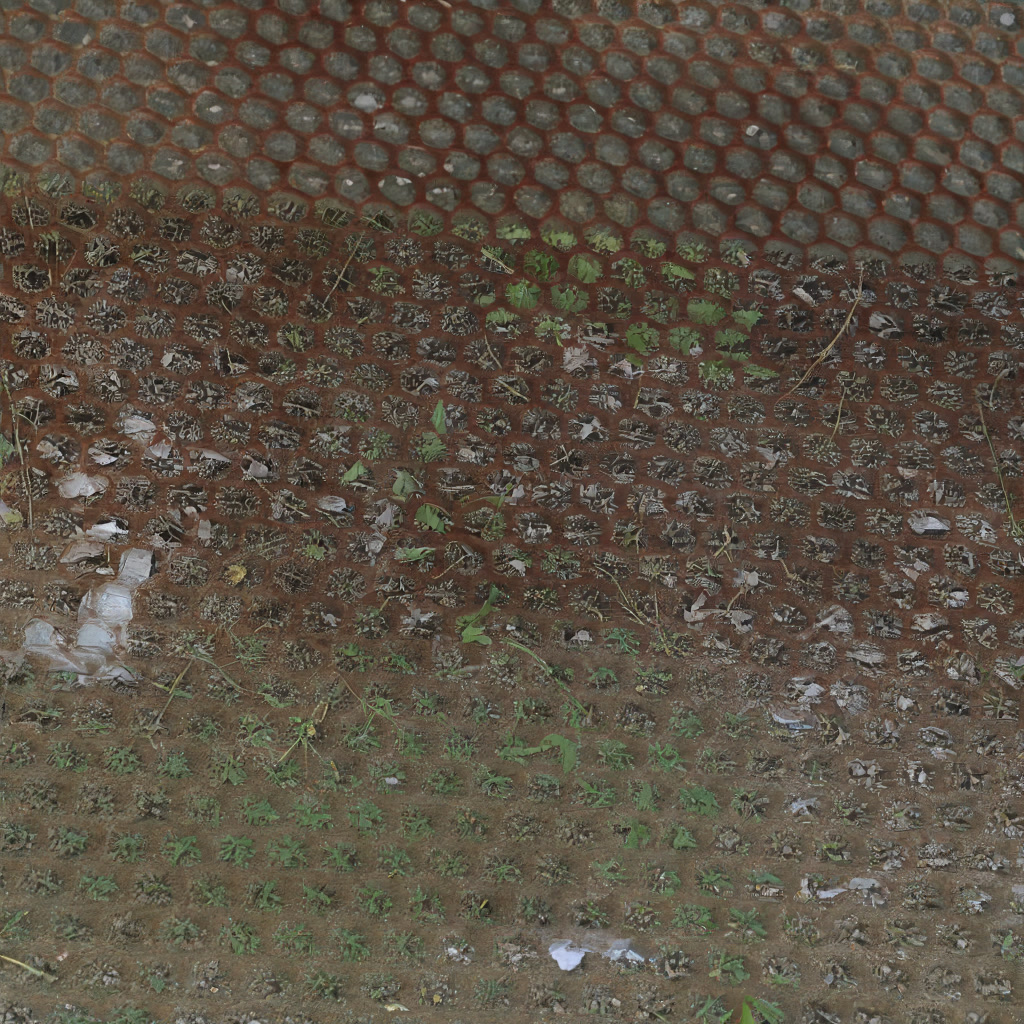} &
        \zoomimg{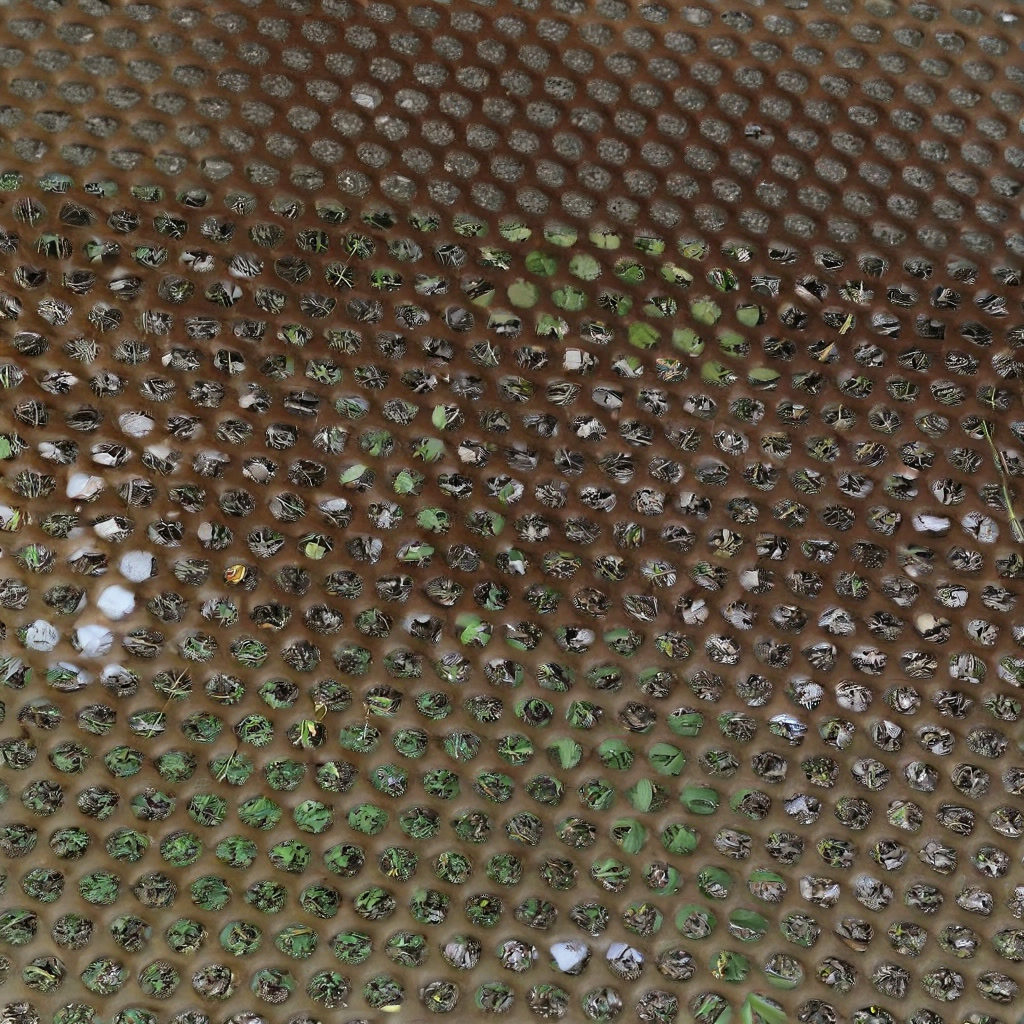} \\

        \noalign{\setzoom{0.40}{0.40}{0.79, 0.21}{yellow}}
        \multirow[c]{-10}{1.2em}{~\rotatebox{90}{\cellcolor{orange!14}\footnotesize \textbf{(b) DPDD + DF2K}}} &
        \rotatebox{90}{\makecell{\small ~~~~~~~~DF2K}} &
        \zoomimg{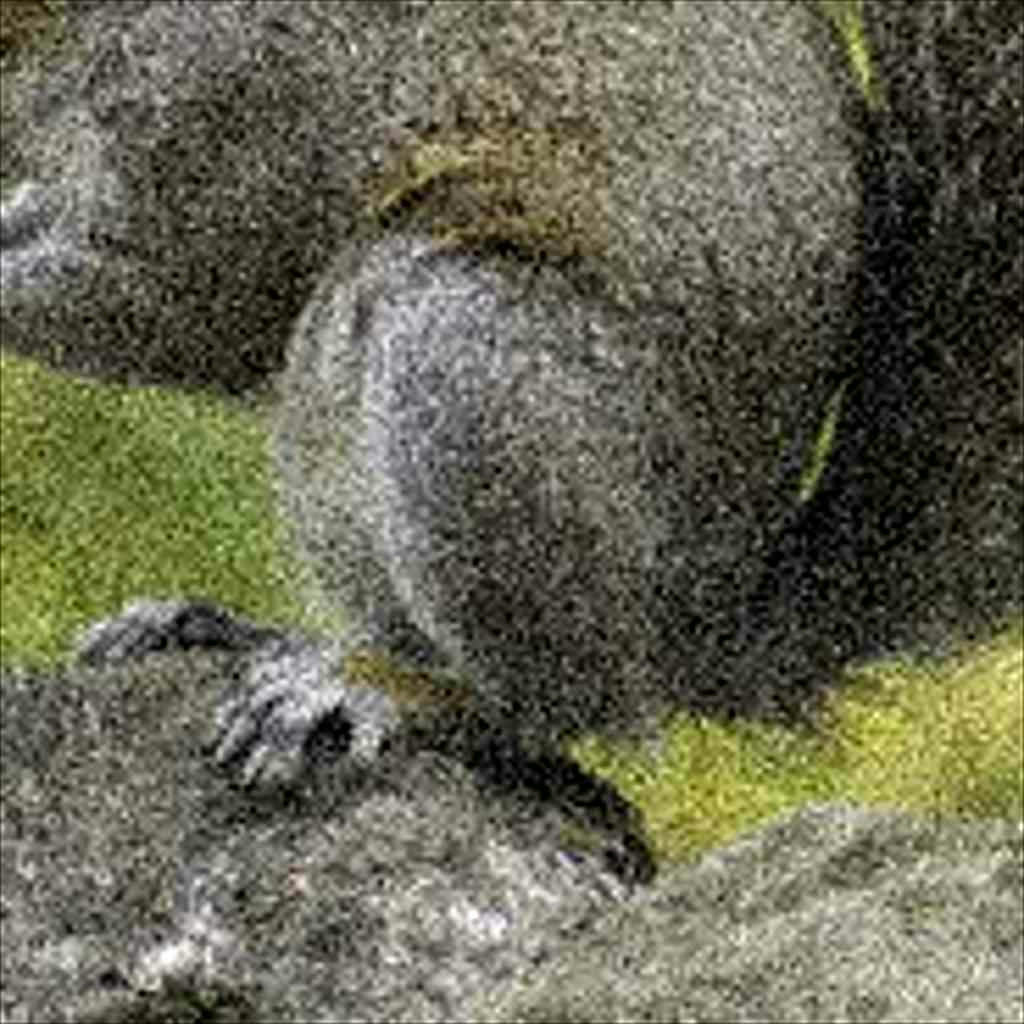} &
        \zoomimg{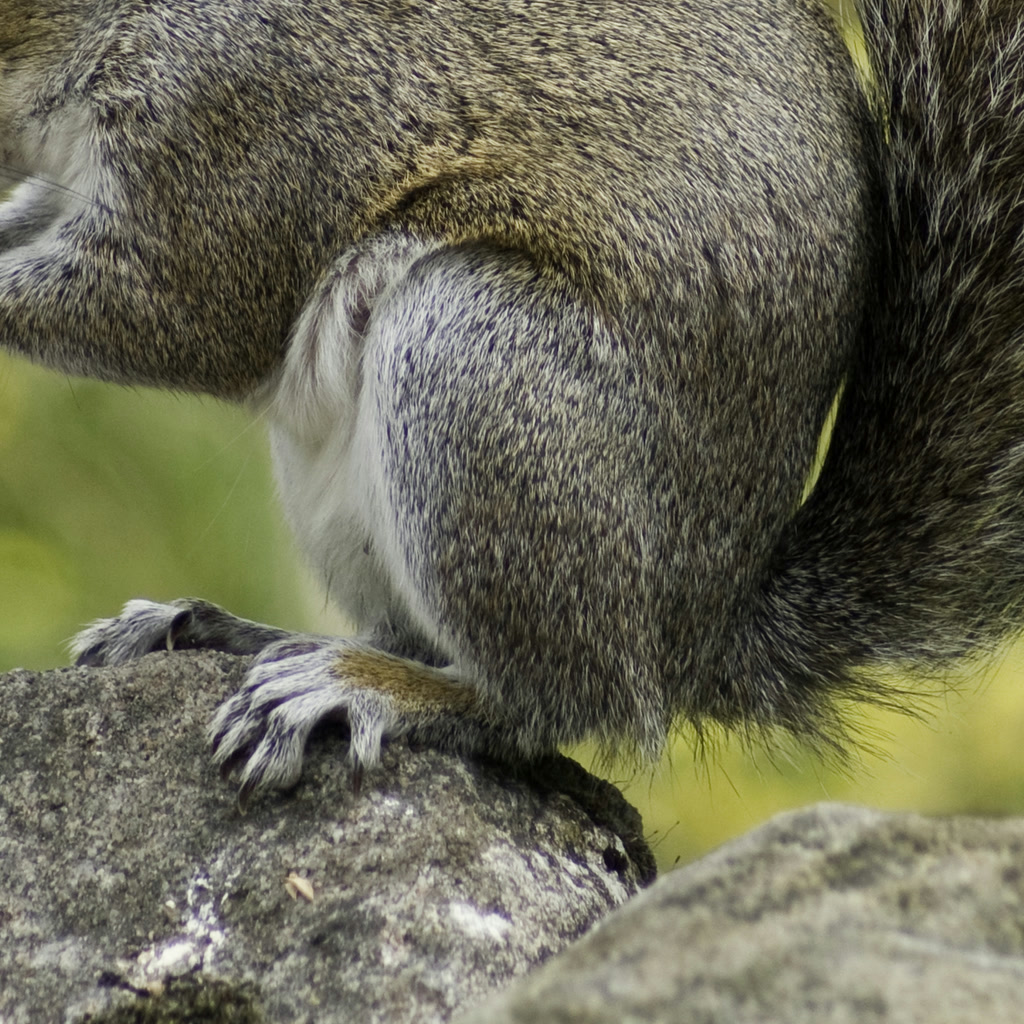} &
        \zoomimg{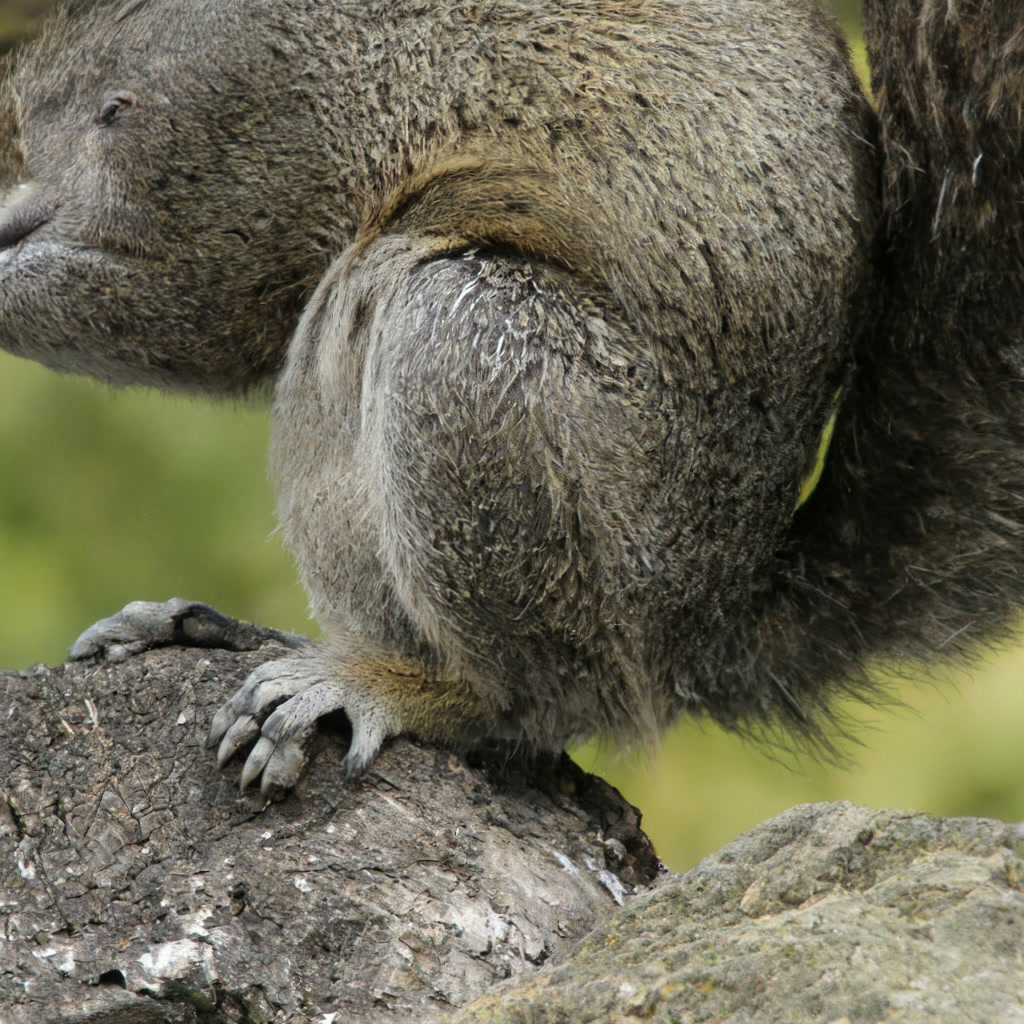} &
        \zoomimg{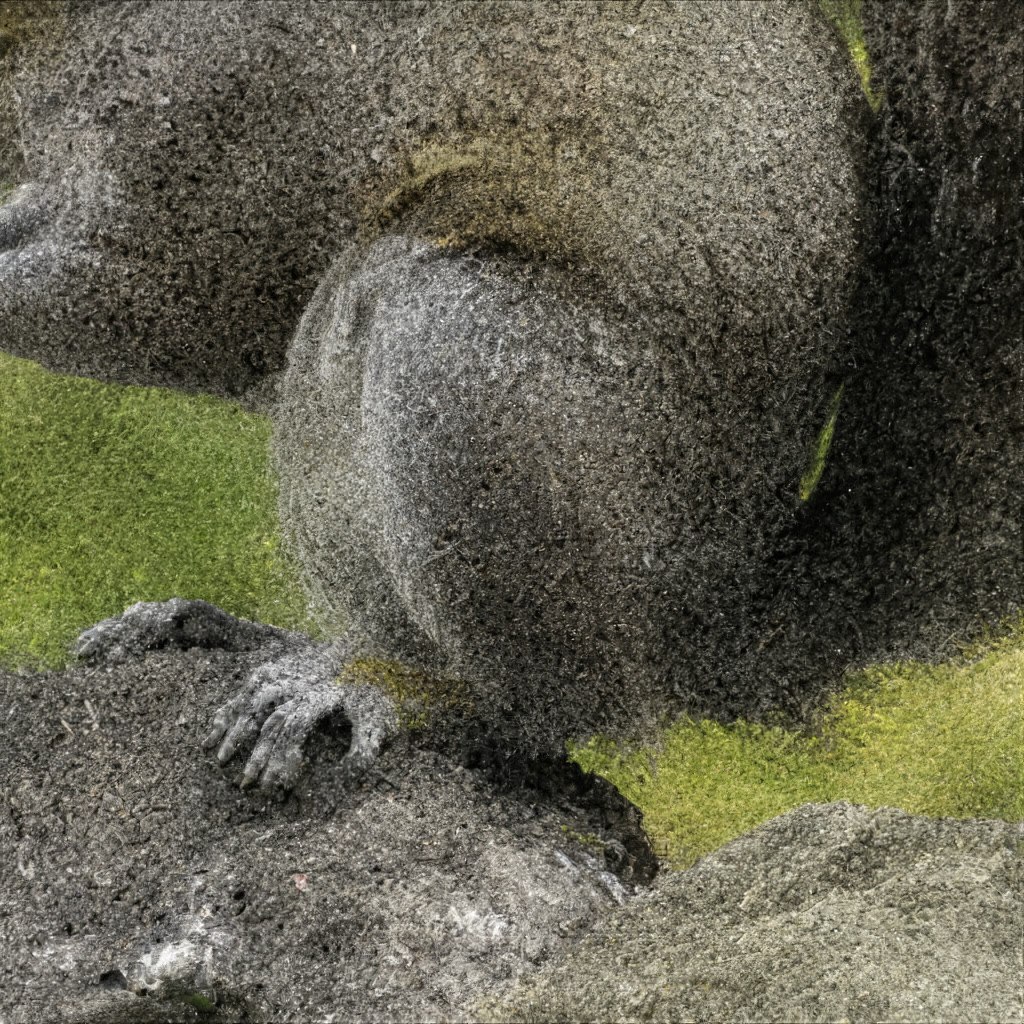} &
        \zoomimg{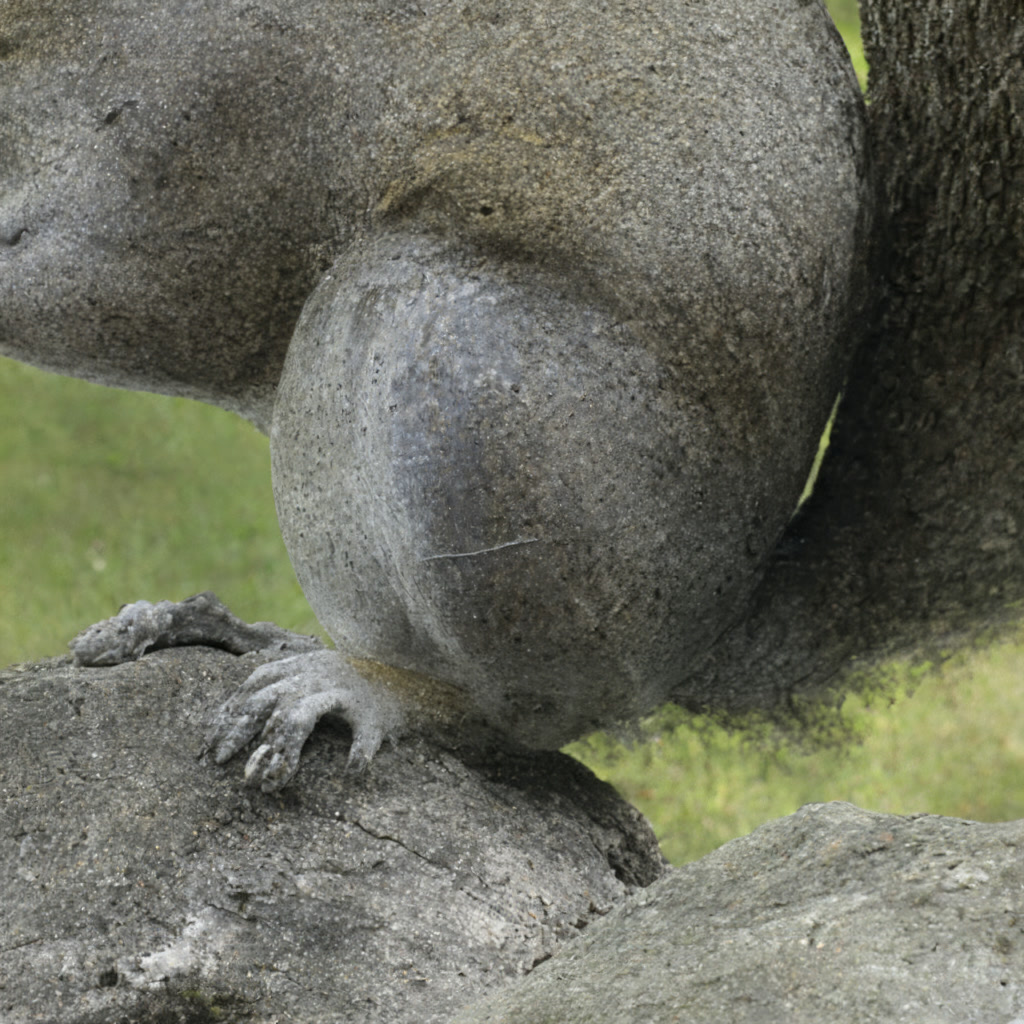} &
        \zoomimg{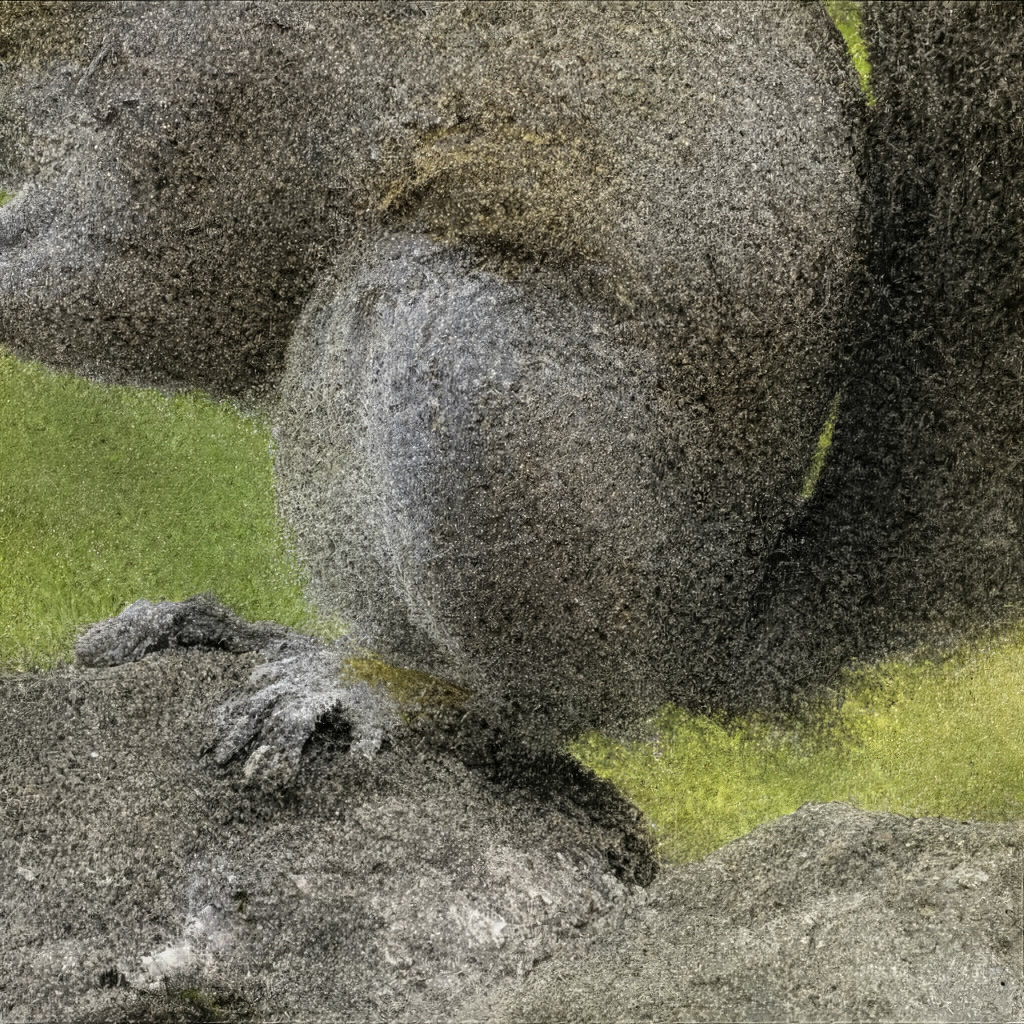} &
        \zoomimg{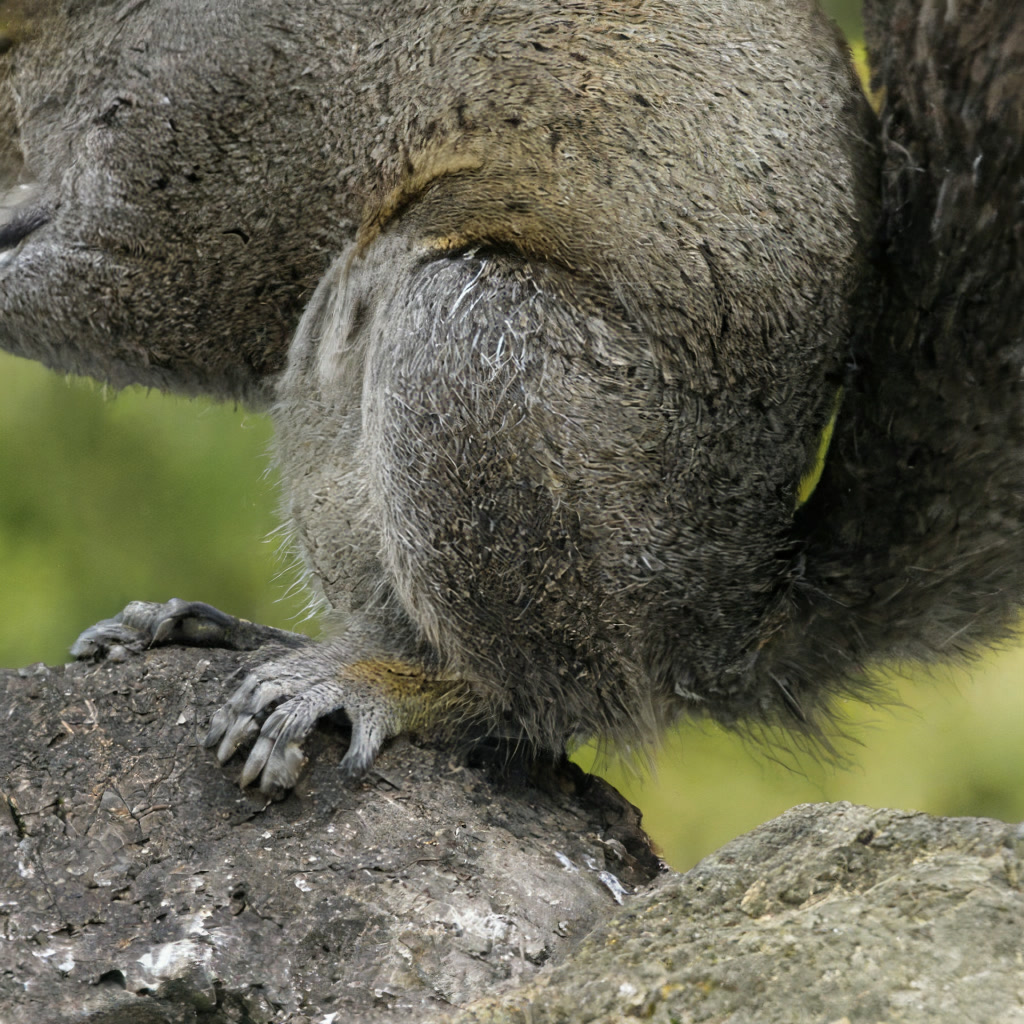} \\
        
        \noalign{\vspace{3pt}}
        \midrule
        \noalign{\vspace{3pt}}
        
        \noalign{\setzoom{0.50}{0.90}{0.79, 0.21}{yellow}}
         \cellcolor{teal!15} &
        \rotatebox{90}{\makecell{\small ~~~~~~~~OID}} &
        \zoomimg{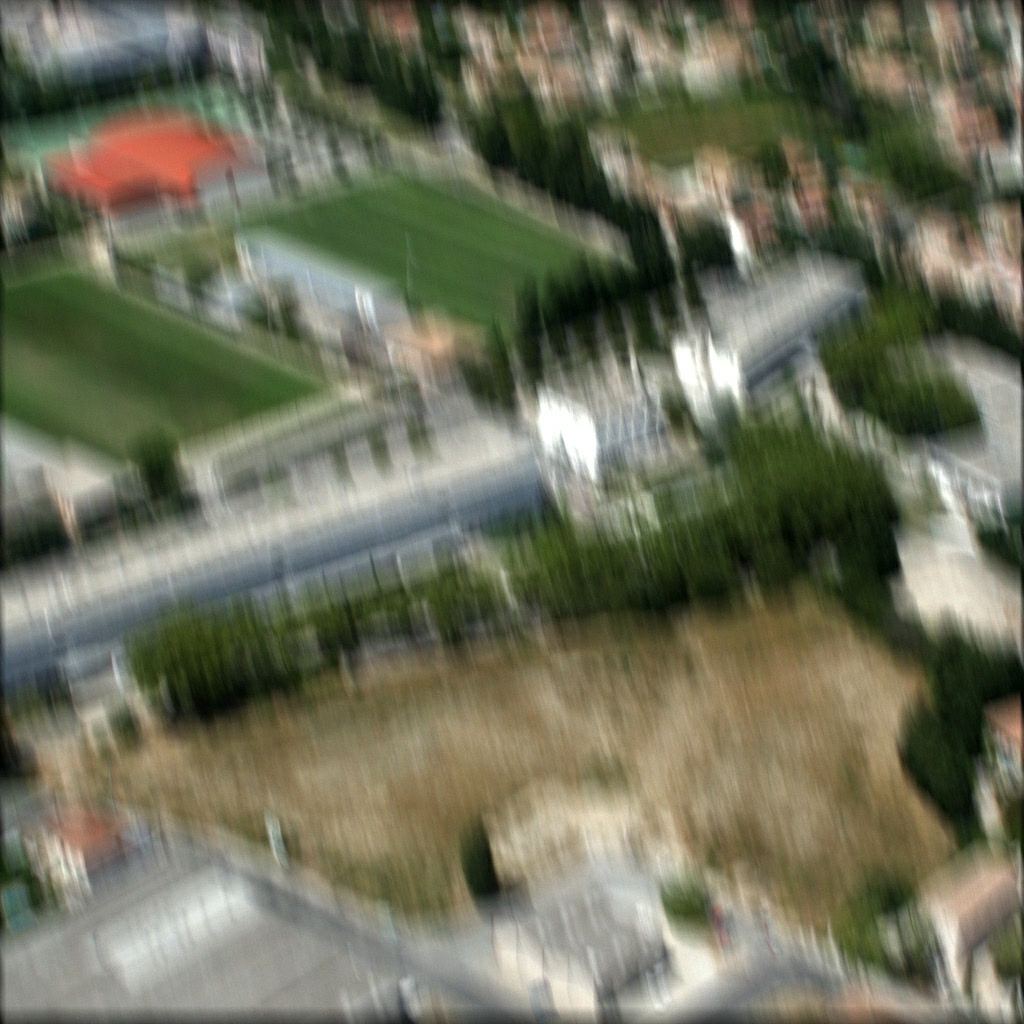} &
        \zoomimg{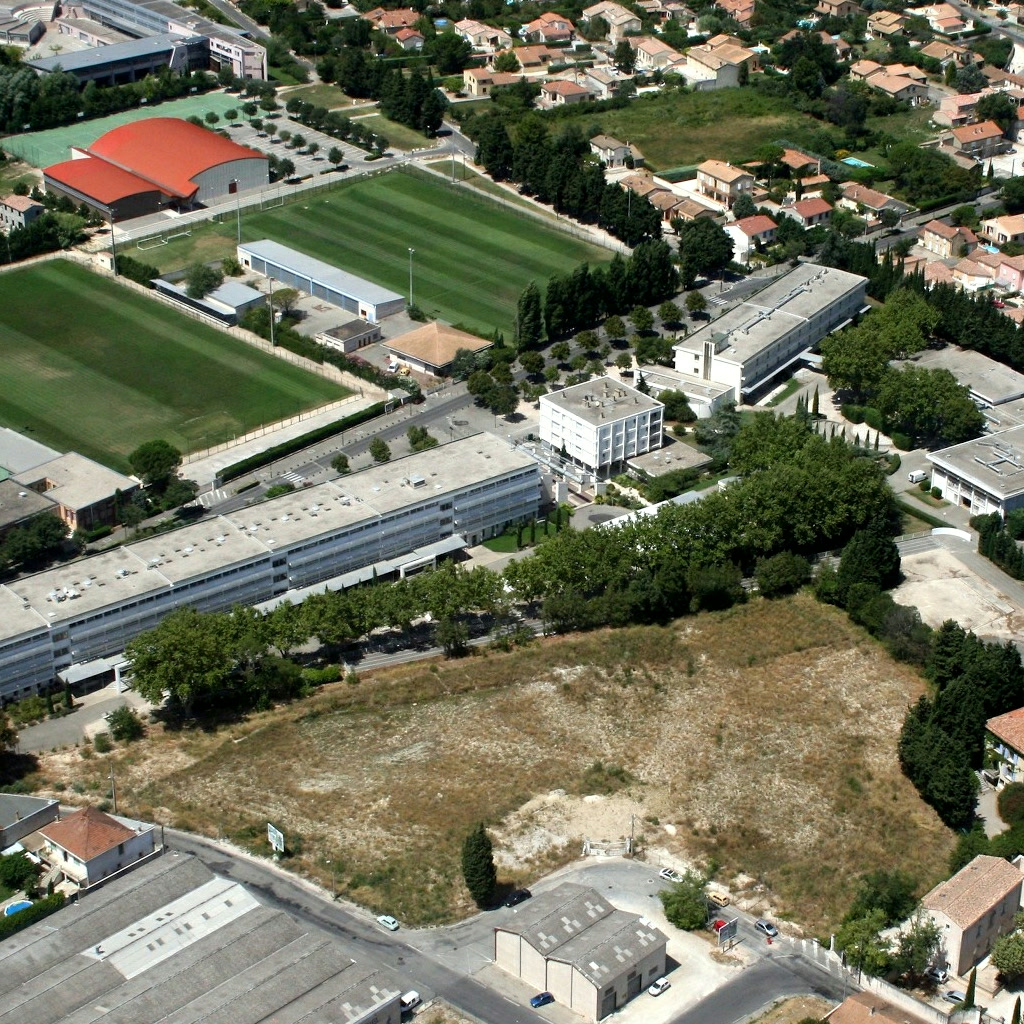} &
        \zoomimg{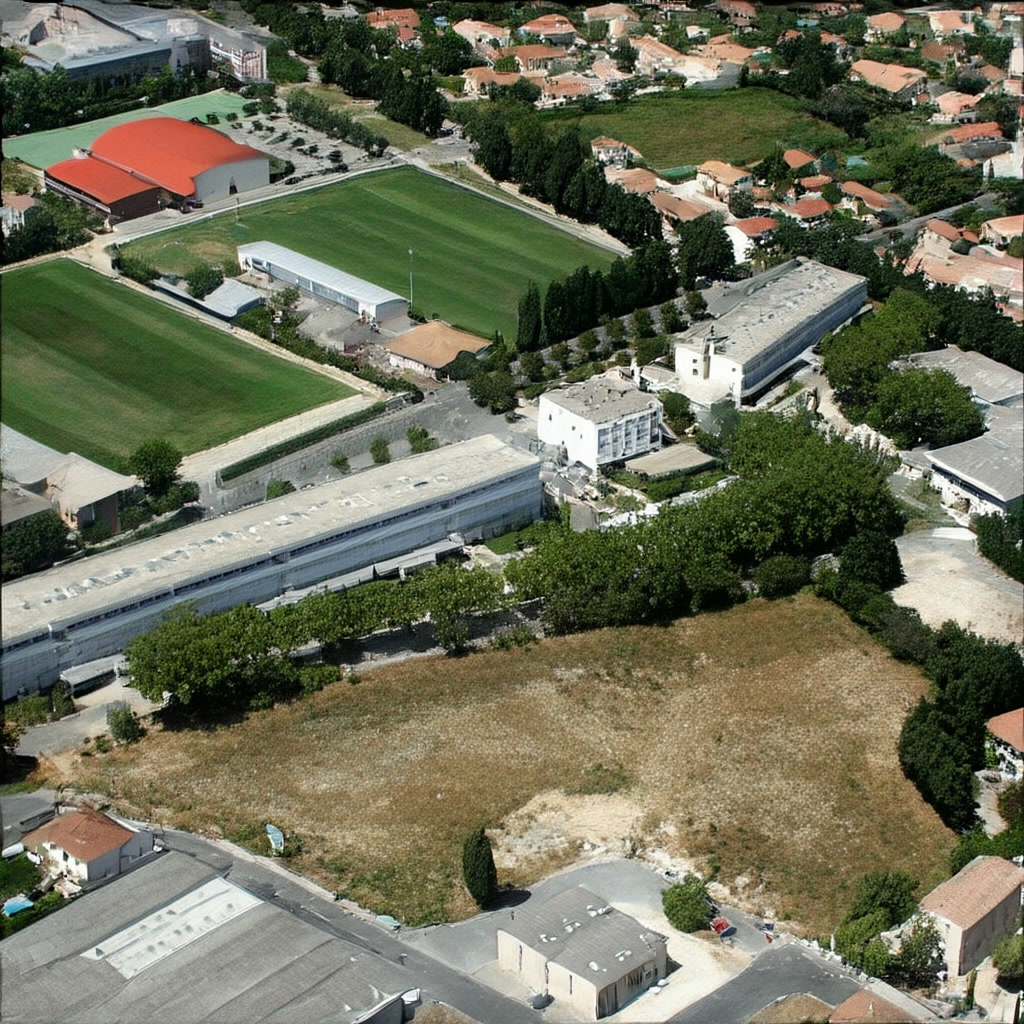} &
        \zoomimg{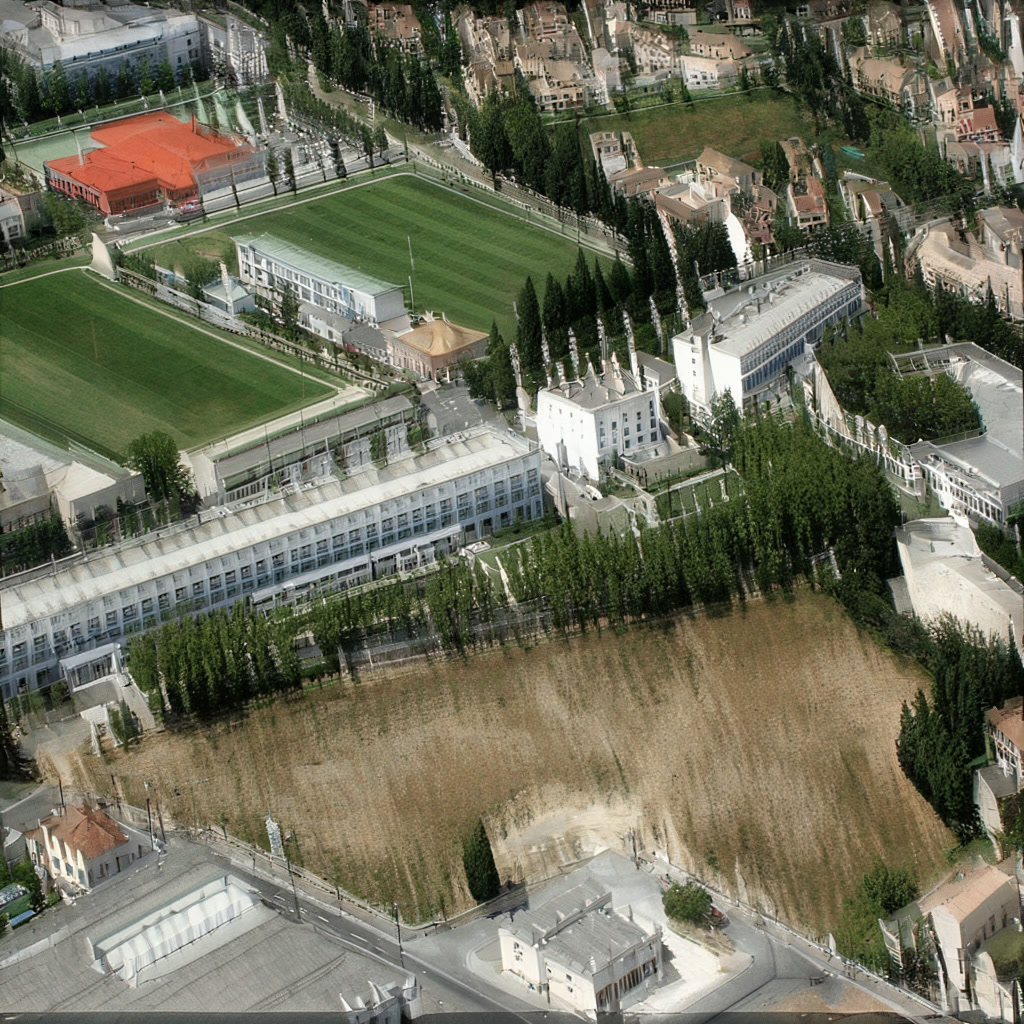} &
        \zoomimg{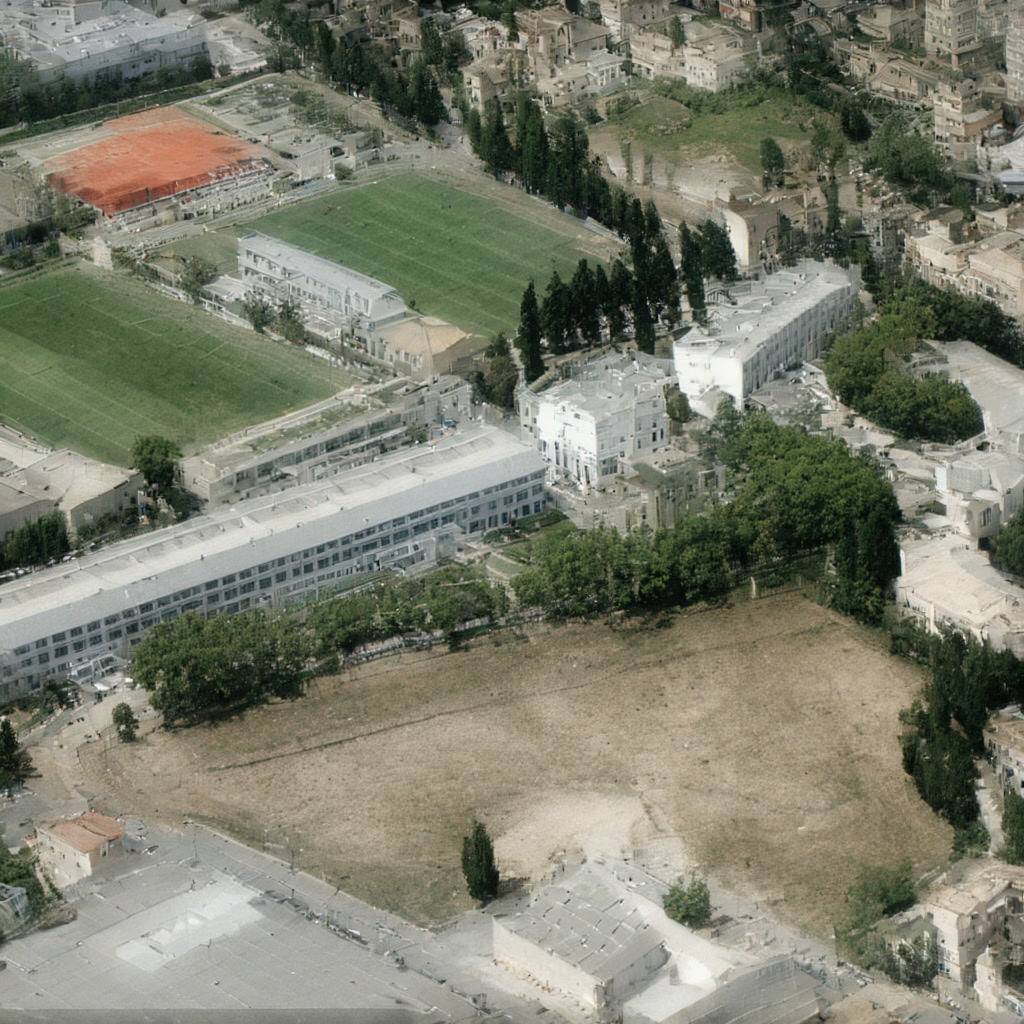} &
        \zoomimg{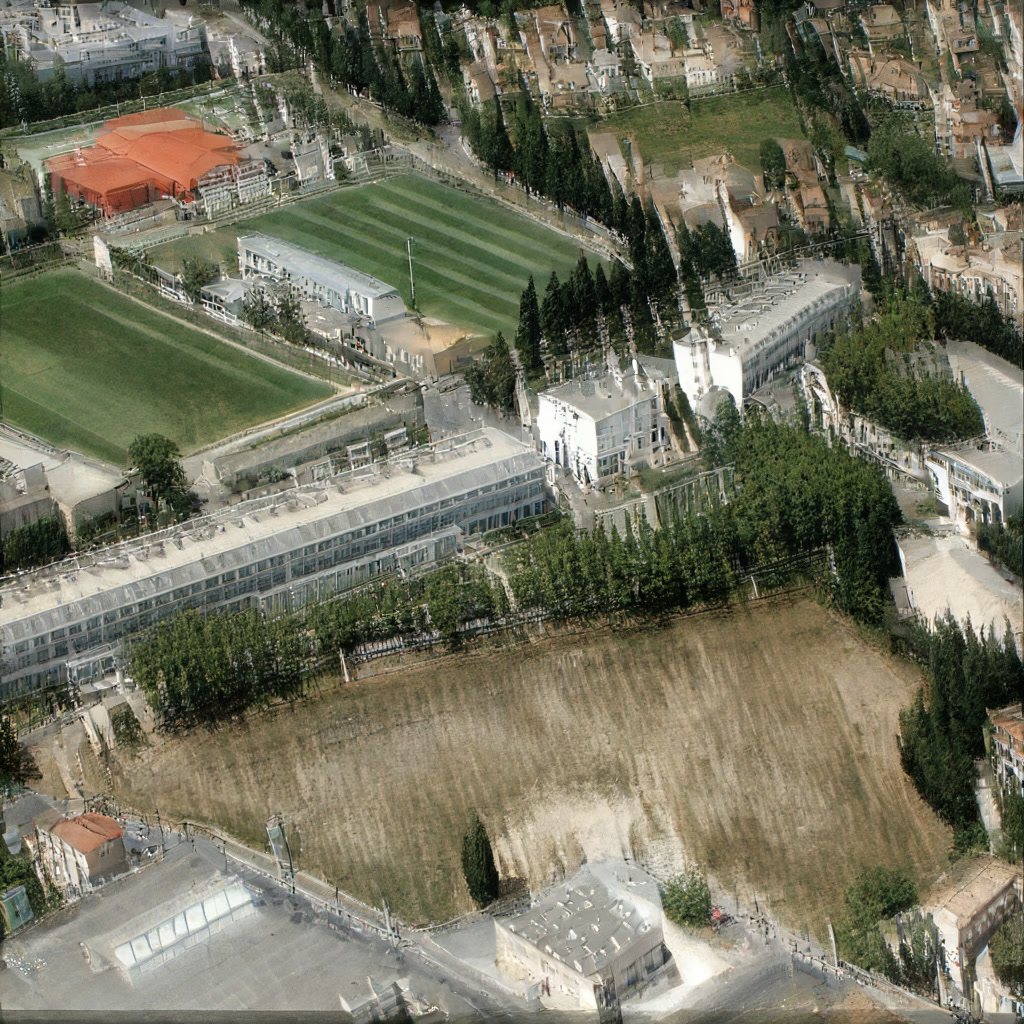} &
        \zoomimg{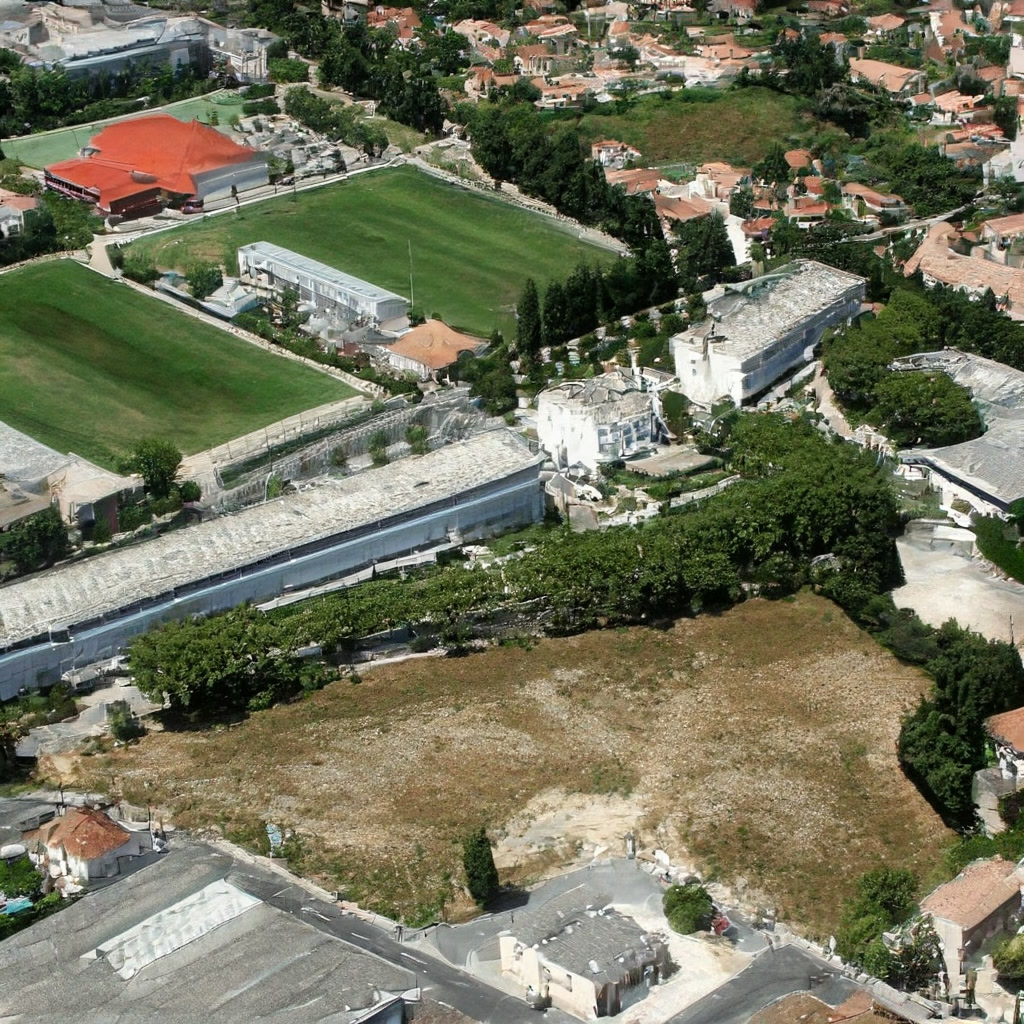} \\

        \noalign{\setzoom{0.90}{0.60}{0.79, 0.21}{yellow}}
        \multirow[c]{-10}{1.2em}{~\rotatebox{90}{\cellcolor{teal!15}\footnotesize \textbf{(c) OID + DF2K}}} &
        \rotatebox{90}{\makecell{\small ~~~~~~~~DF2K}} &
        \zoomimg{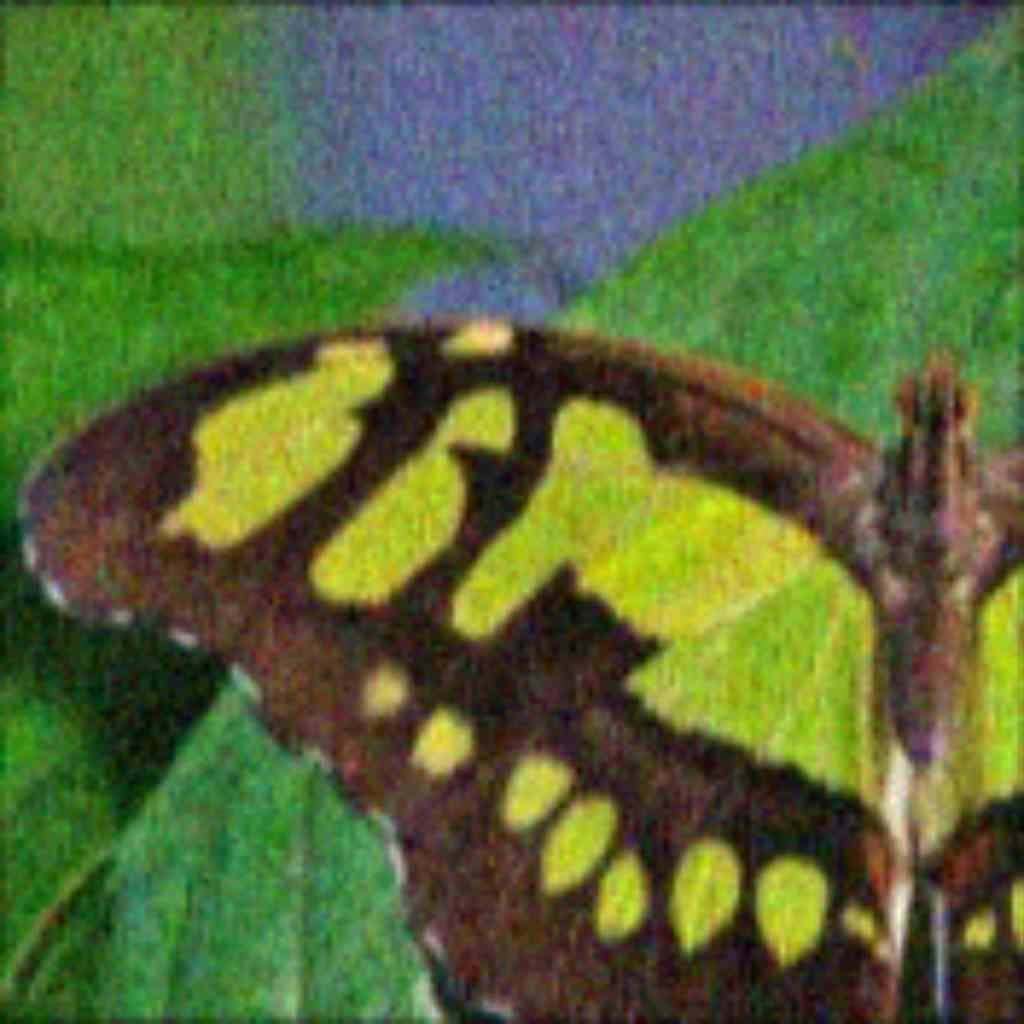} &
        \zoomimg{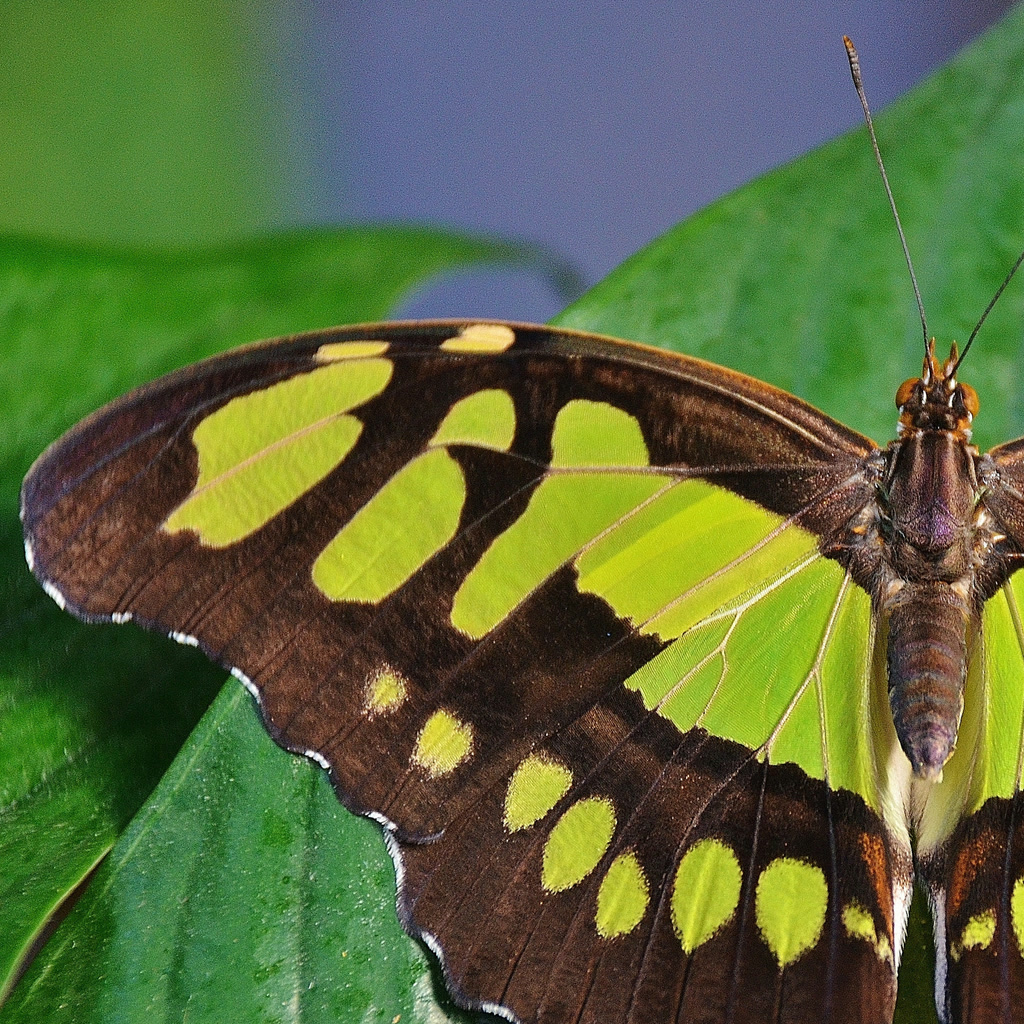} &
        \zoomimg{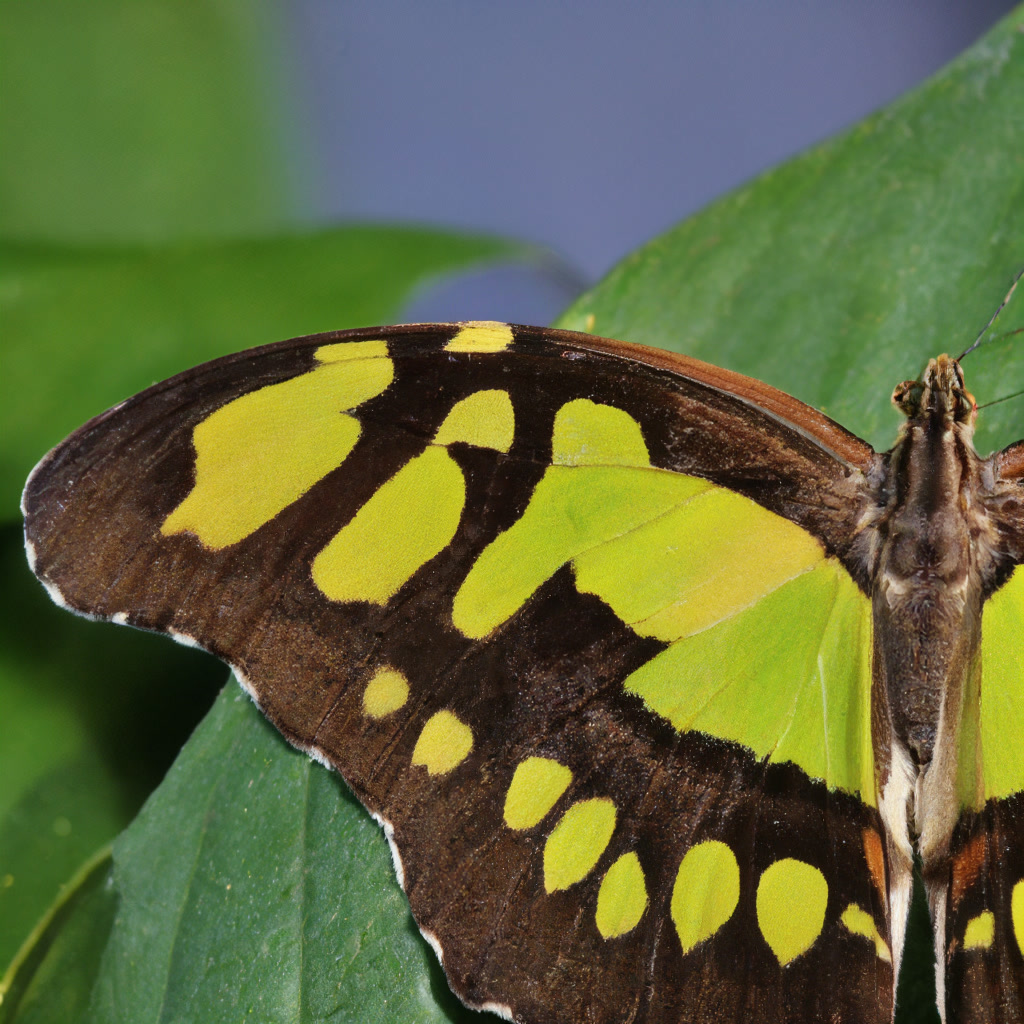} &
        \zoomimg{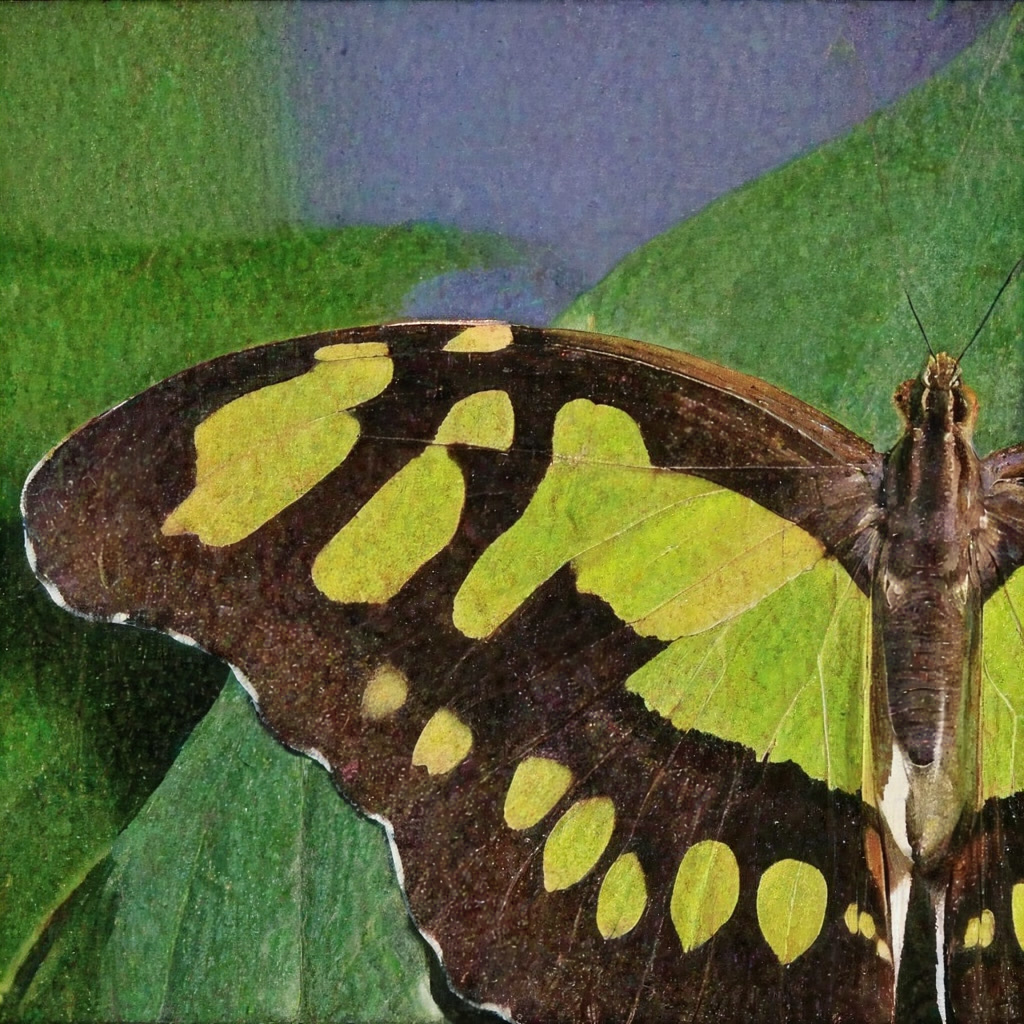} &
        \zoomimg{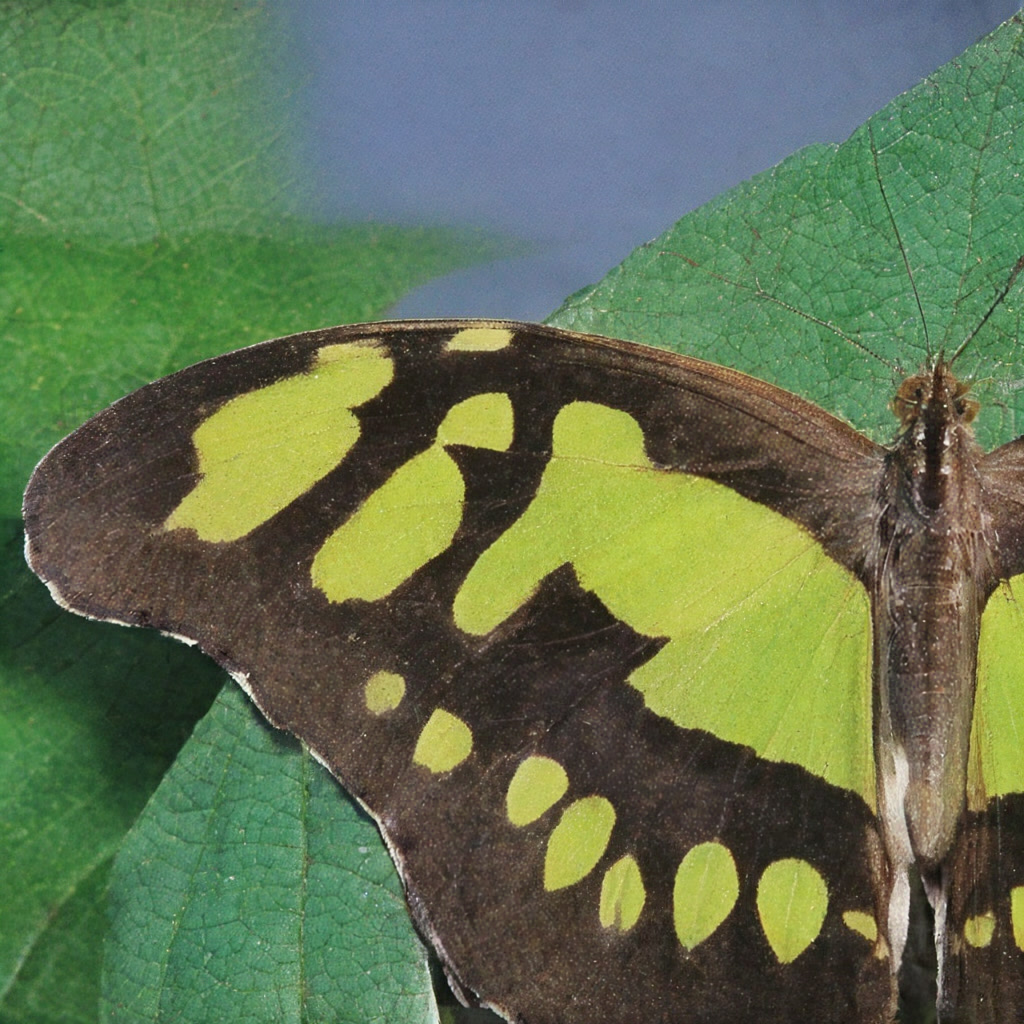} &
        \zoomimg{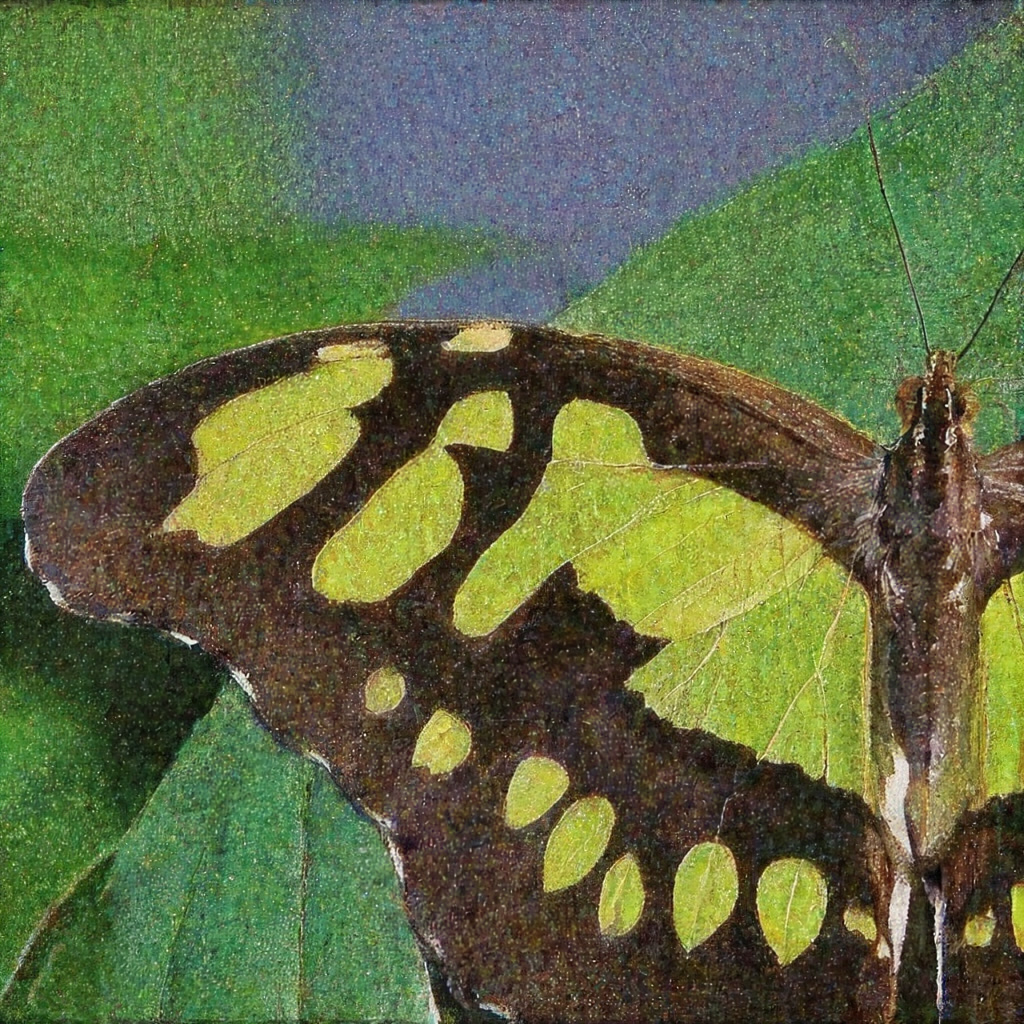} &
        \zoomimg{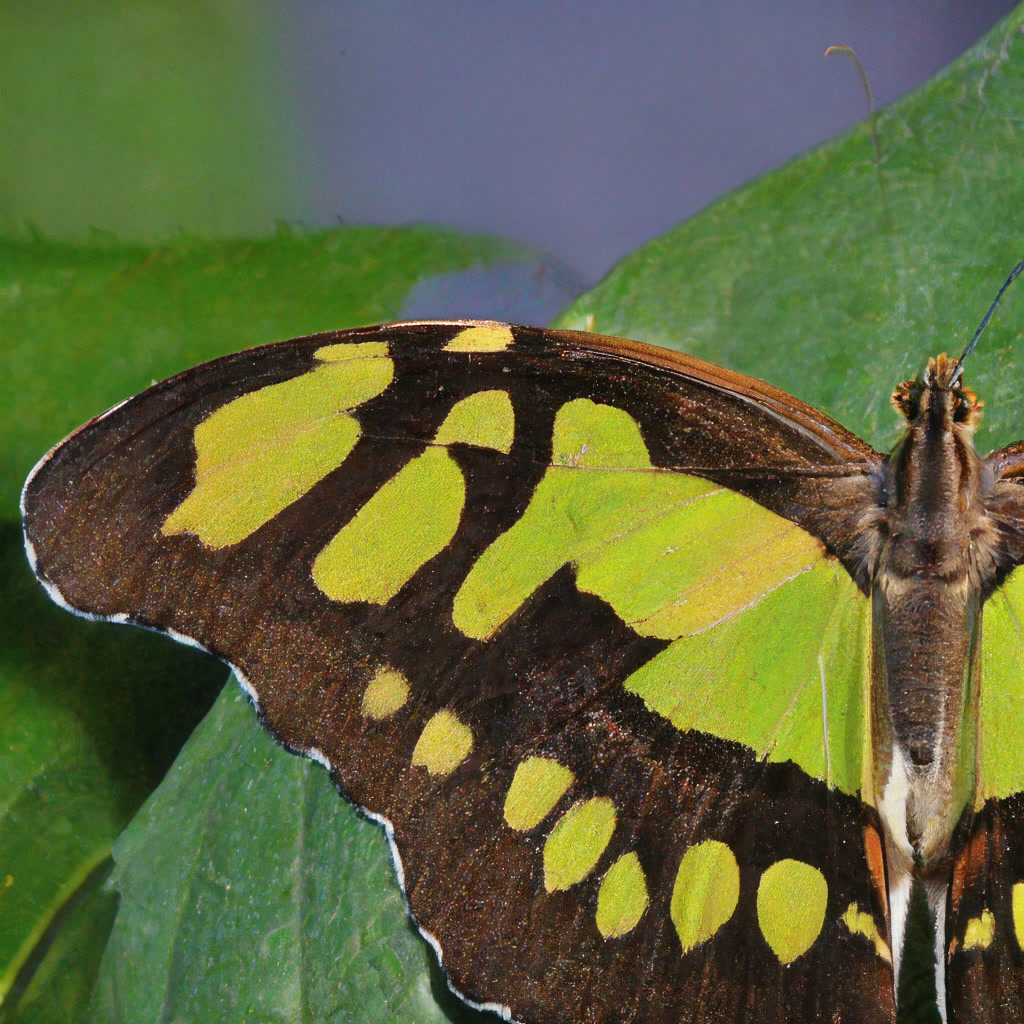} \\
        
    \end{tabular}
    
    \caption{\small \textbf{Qualitative comparison of pairwise model merging across restoration tasks:} Each two-row block evaluates a single merged model on both of its constituent tasks: \textbf{(a)} Motion Deblur (OID) + Defocus Deblur (DPDD), \textbf{(b)} Defocus Deblur (DPDD) + Super-Resolution (DF2K), and \textbf{(c)} Motion Deblur (OID) + Super-Resolution (DF2K). While heuristic scalar averaging and element-wise LLM pruning methods introduce ringing and spatial artifacts, \textbf{OptIMerge} preserves structural fidelity and high-frequency details across all pairwise combinations.}
    \label{fig:pairwise_soup_results}
    \label{fig:oid_dpdd_soup_results}
    \label{fig:dpdd_sr_soup_results}
    \label{fig:oid_sr_soup_results}
    \vspace{-2em}
\end{figure*}

Our quantitative findings, assessed via the standard metrics such as LPIPS, PSNR, and SSIM (Table \ref{tab:main_results}), indicate that the naive arithmetic soup strategy generally has subpar performance. 
As hypothesized, a rigid global scalar significantly degrades the performance of both restoration tasks due to severe interference in the convolutional filters and unaligned attention bases. We provide additional ablations and experiments in \Cref{appendix:additional_ablations}.

By applying OptIMerge, we observe improvements in representation fidelity. 
The explicit preservation of spatial geometries via rotational alignment (Cayley Transforms) combined with cross-channel semantic routing (Biased LoRA) effectively mitigates the destructive interference inherent to naive weight averaging. 
Notably, in the highly constrained OID + DPDD stress test, $\text{OptIMerge}$ successfully reconciles the conflicting isotropic and anisotropic spatial priors without fracturing the underlying  features. 
Across all evaluated pairs, our merged model consistently recovers most of the gap between the naive soup and the
single-task experts, and matches or exceeds joint multi-task training on the majority of metrics, while
optimizing $<0.4\%$ of the parameters. %

\subsection{Merging Personalized T2I Models}
\label{sec:t2i}
While OptIMerge is designed for I2I experts, as a proof of concept, we also extend it to T2I subject/style personalization models fine-tuned from Stable Diffusion~v1.5. Specifically, a \emph{subject} expert is fine-tuned on 4--5 images of a specific subject (e.g. coorgi dog) using a trigger prompt ``A [V$_1$] dog" (where [V$_1$] is unique token)~\citep{ruiz2023dreambooth}. The \emph{style} expert is fine-tuned on a single reference image of a style (e.g. watercolor painting) using the trigger prompt ``Image in $<$style description$>$ style"~\citep{sohn2023styledrop}. Because only a handful of images is available, we run OptIMerge on the experts' own training images for 100 steps, using  with rank $r=4$. For more details, see~\Cref{sec:t2i_tasks}.

\begin{figure*}[t]
    \centering
    \begin{minipage}[c]{0.50\textwidth}
        \centering
        \setlength{\tabcolsep}{1.5pt}
        \begin{tabular}{c cc | cc}
            & \scriptsize{Subject} & \scriptsize{Style} 
            & \scriptsize{Subject} & \scriptsize{Style} \\

            \rotatebox{90}{\makecell{\scriptsize ~~~Reference}} &
            \includegraphics[width=0.215\linewidth]{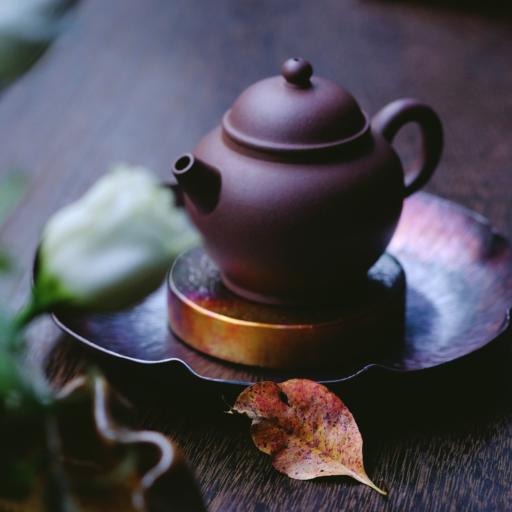} &
            \includegraphics[width=0.215\linewidth]{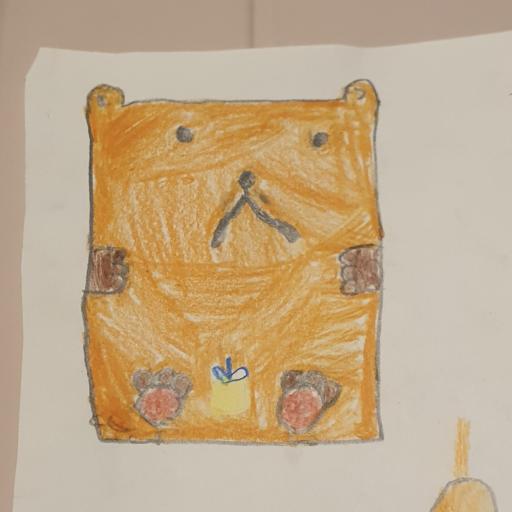} &
            \includegraphics[width=0.215\linewidth]{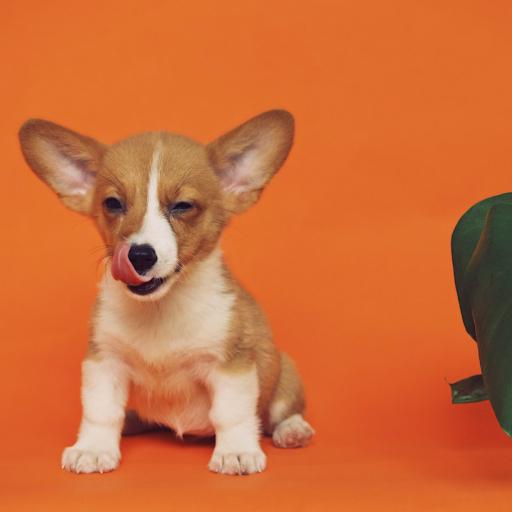} &
            \includegraphics[width=0.215\linewidth]{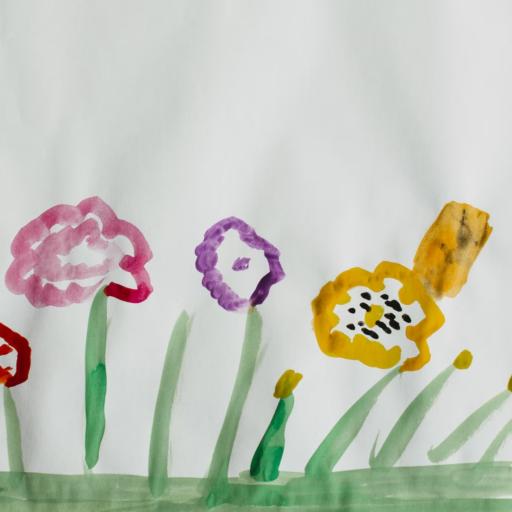} \\

            \rotatebox{90}{\makecell{\scriptsize Single Task \\ \scriptsize ~~~Expert}} &
            \includegraphics[width=0.215\linewidth]{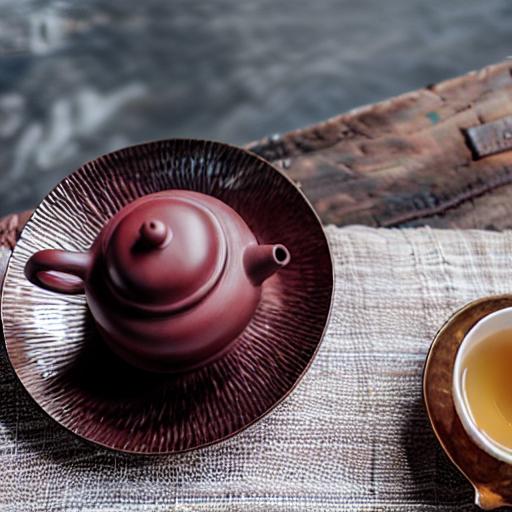} &
            \includegraphics[width=0.215\linewidth]{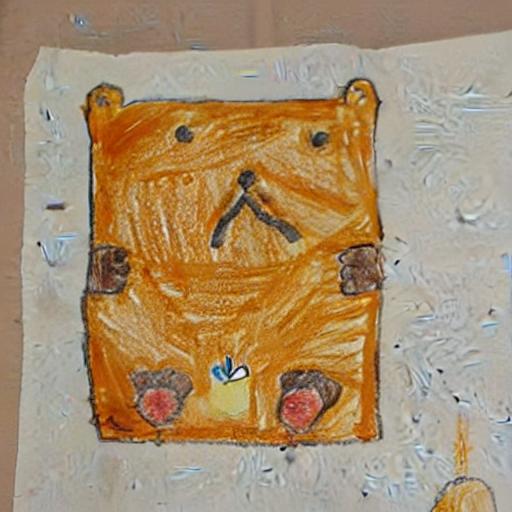} &
            \includegraphics[width=0.215\linewidth]{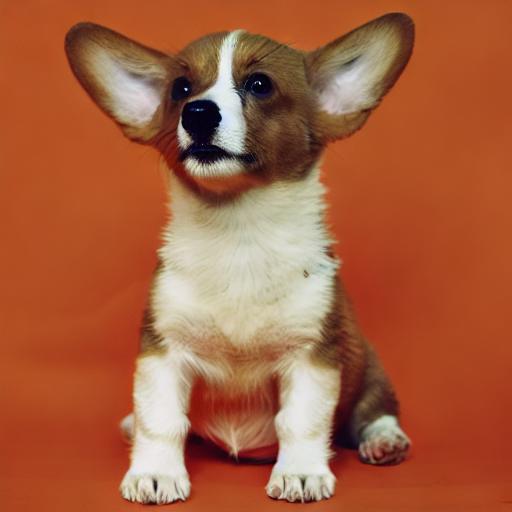} &
            \includegraphics[width=0.215\linewidth]{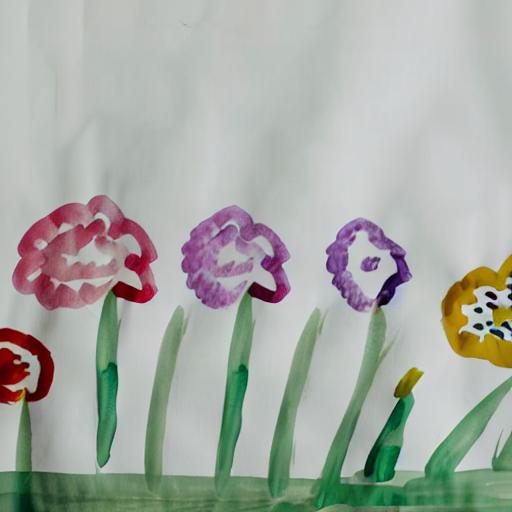} \\

            \rotatebox{90}{\makecell{\scriptsize ~Naive Soup}} &
            \includegraphics[width=0.215\linewidth]{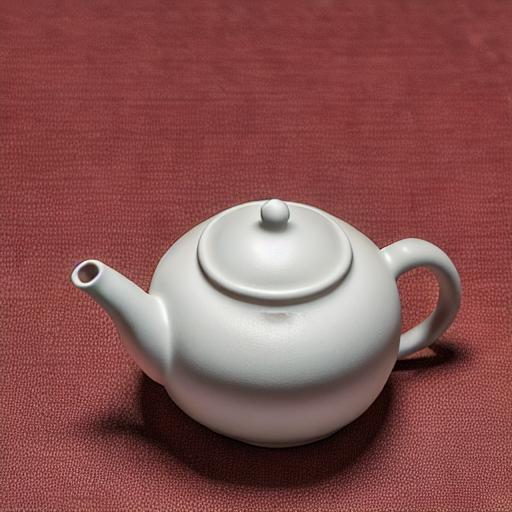} &
            \includegraphics[width=0.215\linewidth]{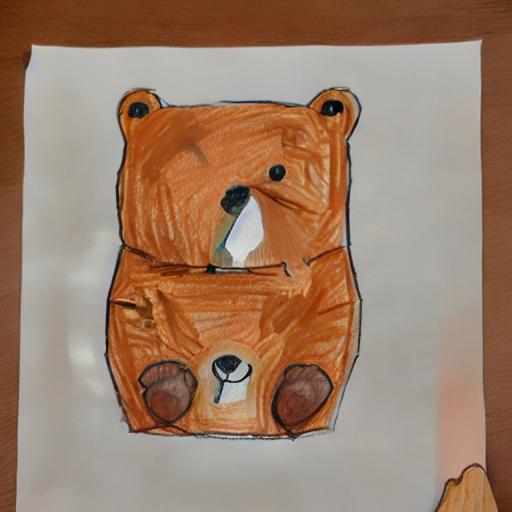} &
            \includegraphics[width=0.215\linewidth]{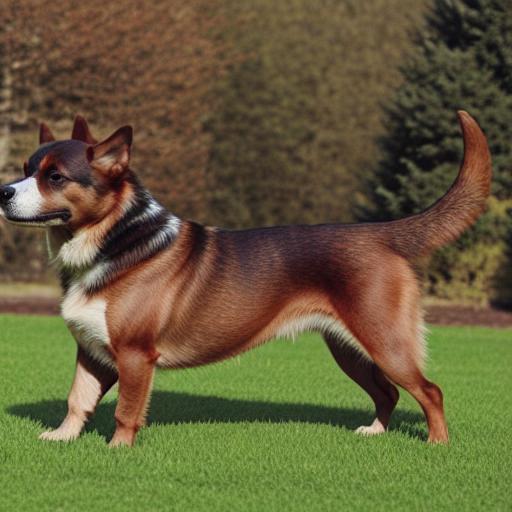} &
            \includegraphics[width=0.215\linewidth]{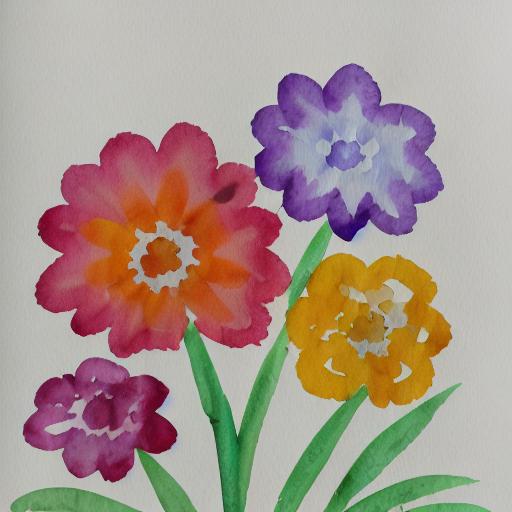} \\

            \rotatebox{90}{\makecell{\scriptsize \textbf{OptIMerge} \\ \scriptsize \textbf{~(Ours)}}} &
            \includegraphics[width=0.215\linewidth]{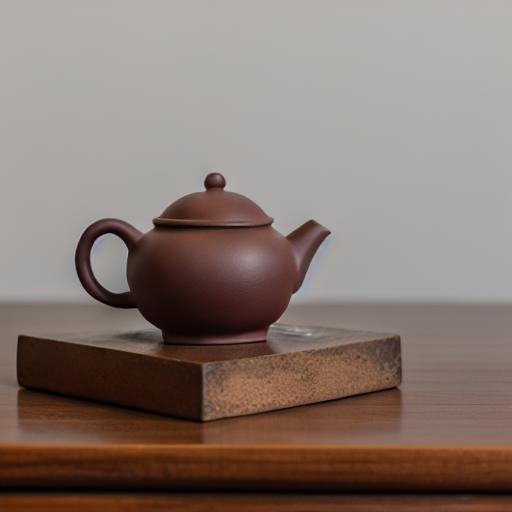} &
            \includegraphics[width=0.215\linewidth]{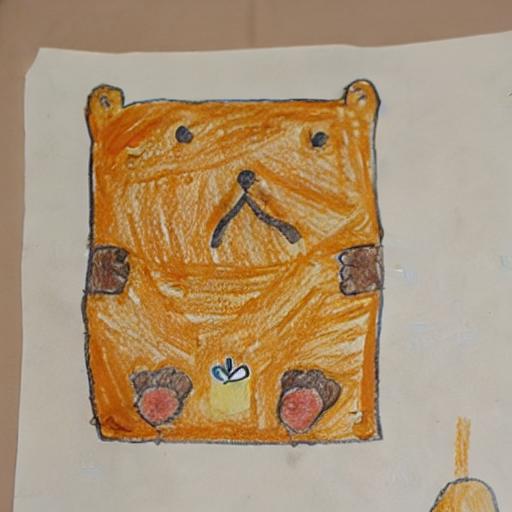} &
            \includegraphics[width=0.215\linewidth]{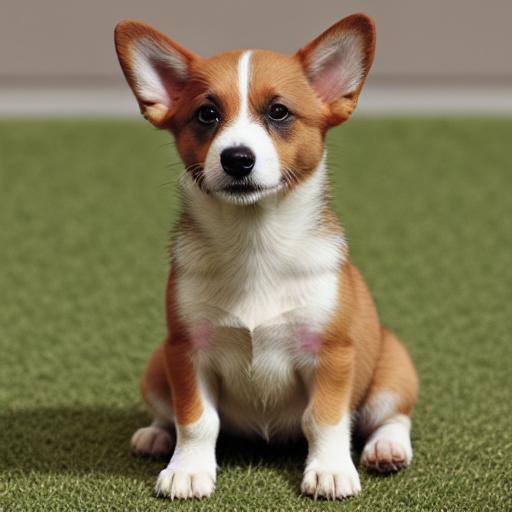} &
            \includegraphics[width=0.215\linewidth]{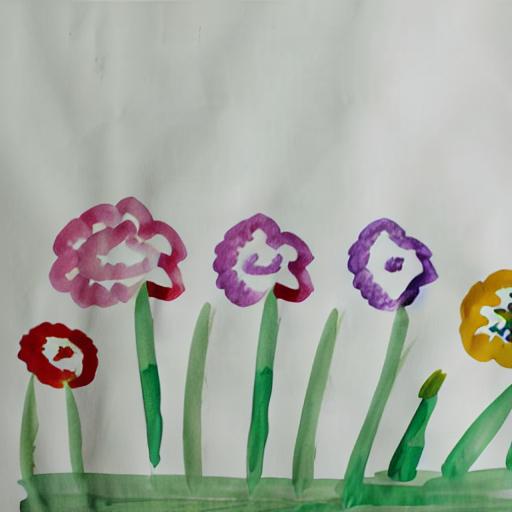} \\

            &
            \parbox[t]{0.215\linewidth}{\centering \scriptsize [V$_1$] Teapot} &
            \parbox[t]{0.215\linewidth}{\centering \scriptsize Bear in crayon drawing style} &
            \parbox[t]{0.215\linewidth}{\centering \scriptsize [V$_1$] dog} &
            \parbox[t]{0.215\linewidth}{\centering \scriptsize Flowers in watercolor painting style} \\
        \end{tabular}
    \end{minipage}%
    \hfill
    \begin{minipage}[c]{0.45\textwidth}
        \centering
        \begingroup
            \captionsetup{name={$\Leftarrow$ Figure}}
            \noindent\captionof{figure}{\footnotesize \textbf{Qualitative results for merging subject- and style-personalization models.} Naive arithmetic merging dilutes subject identity and style fidelity, whereas \textbf{OptIMerge} preserves the ability to produce both the concepts individually.}
            \label{fig:subject_style}
        \endgroup
        \vspace{4pt}
        \captionof{table}{\footnotesize \textbf{Quantitative results for merging subject- and style-personalization models} Subject/Style alignment is the average CLIP-I image similarity to the subject/style references. Scores are averaged across 16 merger combinations (4 subjects $\times$ 4 styles). [\textbf{best}, \underline{second-best}]}
        \label{tab:subject_style_merge}
        \vspace{-4pt}
        \resizebox{\linewidth}{!}{
        \begin{tabular}{lccc}
        \toprule
        \multirow{2}{*}{\textbf{Method / Strategy}} & \multirow{2}{*}{\makecell{\textbf{Trainable} \\ \textbf{Params}}} & \multicolumn{1}{c}{\textbf{Subject}} & \multicolumn{1}{c}{\textbf{Style}} \\
        & & \textbf{alignment $\uparrow$} & \textbf{alignment $\uparrow$} \\
        \midrule
        \rowcolor{gray!12} \textit{Single-Task Experts} & --- & 0.8245 & 0.7983 \\
        \midrule
        \textit{Naive Soup ($\alpha=0.5$)} & 0 (0\%) & 0.5781 & 0.5374 \\
        \midrule
        $\text{OptIMerge}_{\text{Attn}}$ & $\sim$1.9M & 0.6721 & 0.6434 \\
        $\text{OptIMerge}_{\text{Conv-KC}}$ & $\sim$109.2M & 0.7384 & \underline{0.7256} \\
        $\text{OptIMerge}_{\text{Conv-Cayley}}$ & $\sim$2.4M & \underline{0.7520} & 0.7186 \\
        $\text{OptIMerge}$ (\textbf{Ours}) & $\sim$3.2M & \textbf{0.8174} & \textbf{0.7801} \\
        \bottomrule
        \end{tabular}
        }
    \end{minipage}
    \vspace{-0.8em}
\end{figure*}

Mirroring the multi-task restoration setting, the goal of this experiment is \emph{concept retention}: after merging, a single model should reproduce \emph{each} concept as faithfully as its dedicated expert when prompted with that concept's trigger prompt, i.e., ``A [V$_1$] dog'' or a style prompt such as ``A bear in $<$style description$>$ style''. %
 As shown in Figure~\ref{fig:subject_style}, the naive soup dilutes both concepts: the subject loses its identity and the style is washed out. OptIMerge retains both. In Table~\ref{tab:subject_style_merge}, it raises subject alignment from 0.578 to 0.817 and style alignment from 0.537 to 0.780, close to the single-task experts (0.825 and 0.798). %

\vspace{-0.35em}
\section{Conclusion}
\vspace{-0.35em}
In this work, we introduced \textit{OptIMerge}, a parameter-efficient framework that formalizes diffusion model merging as a optimization-based alignment problem rather than a static, heuristic averaging process.
We demonstrated that naive arithmetic model soups systematically fail on dense pixel-prediction and synthesis tasks due to severe representation misalignment and the fracturing of clashing spatial geometries (e.g., isotropic versus anisotropic physical filters). 
To overcome these limitations without incurring the massive computational cost of joint multi-task training, OptIMerge introduces parameterizations tailored to the distinct roles of the U-Net architecture. For attention modules, we proposed \textit{Biased LoRA} parameterization, which effectively decouples feature space alignment from highly expressive, factored merging masks. 
For spatial convolutions, we introduced \textit{Kernel-Centric Merging} and the \textit{Low-Rank Cayley Transform}. 
By learning an orthogonal rotation matrix, the Cayley formulation allows an isometric transformation that resolves representation clashing while preserving weight norms. Empirically, our evaluations on pairwise mergers of restoration experts confirm that OptIMerge mitigates destructive interference, and a proof-of-concept experiment shows that the same framework also retains both concepts when merging personalized T2I models. 

\noindent\textbf{Limitations and Future Work.} While our method extends to three experts (Appendix~\ref{appendix:additional_ablations}), the residual gap to the experts grows with the number of tasks, and scaling to large ensembles introduces combinatorial complexity in the alignment pathways. Future work will explore adapting the Cayley parameterization to the emerging class of Diffusion Transformers (DiTs). 
Ultimately, OptIMerge provides a highly scalable and mathematically grounded foundation for consolidating specialized generative experts into unified, versatile foundation models.

\subsection*{AI use statement}

In this work, we used generative AI tools to design and provide feedback on research methodology or experiments, and we used coding agents to implement and verify methods. 
We have not used generative AI tools to support qualitative and thematic data analysis, interpret results. 
Additional tasks that require disclosure such as generation of synthetic data sets, cleaning and reformatting the datasets, formulation and proving of mathematical claims, and assistance in writing proofs are not applicable to this work as this is empirical work.
Additionally, we used generative AI tools to draft parts of this research paper. We have reviewed all AI-assisted work. We have carefully verified LLM-generated code for correctness. We have also proofread this work to ensure its correctness. We take responsibility for the final content of this work, including text, claims or artifacts produced with the aid of generative AI.

\subsubsection*{Acknowledgments}
We thank Jack Paparian and Mojtaba Ardakani, our colleagues at Google, for insightful discussions, valuable feedback, and manuscript review.

\bibliography{references}

@inproceedings{wortsman2022model,
  title={Model soups: averaging weights of multiple fine-tuned models improves accuracy without increasing inference time},
  author={Wortsman, Mitchell and Ilharco, Gabriel and Gadre, Samir Ya and Roelofs, Rebecca and Gontijo-Lopes, Raphael and Morcos, Ari S and Namkoong, Hongseok and Farhadi, Ali and Carmon, Yair and Kornblith, Simon and others},
  booktitle={International conference on machine learning},
  pages={23965--23998},
  year={2022},
  organization={Pmlr}
}

@article{ilharco2022task,
  title={Editing models with task arithmetic},
  author={Ilharco, Gabriel and Ribeiro, Marco Tulio and Wortsman, Mitchell and Gururangan, Suchin and Schmidt, Ludwig and Hajishirzi, Hannaneh and Farhadi, Ali},
  journal={International Conference on Learning Representations (ICLR)},
  year={2023}
}

@article{akiba2025evolutionary,
  title={Evolutionary optimization of model merging recipes},
  author={Akiba, Takuya and Shing, Makoto and Tang, Yujin and Sun, Qi and Ha, David},
  journal={Nature Machine Intelligence},
  volume={7},
  number={2},
  pages={195--204},
  year={2025},
  publisher={Nature Publishing Group UK London}
}

@inproceedings{rombach2022high,
  title={High-resolution image synthesis with latent diffusion models},
  author={Rombach, Robin and Blattmann, Andreas and Lorenz, Dominik and Esser, Patrick and Ommer, Bj{\"o}rn},
  booktitle={2022 IEEE/CVF conference on computer vision and pattern recognition (CVPR)},
  pages={10674--10685},
  year={2022},
  organization={ieee}
}

@inproceedings{ruiz2023dreambooth,
  title={{DreamBooth}: Fine tuning text-to-image diffusion models for subject-driven generation},
  author={Ruiz, Nataniel and Li, Yuanzhen and Jampani, Varun and Pritch, Yael and Rubinstein, Michael and Aberman, Kfir},
  booktitle={2023 IEEE/CVF Conference on Computer Vision and Pattern Recognition (CVPR)},
  pages={22500--22510},
  year={2023},
  organization={IEEE}
}

@article{ainsworth2022git,
  title={Git re-basin: Merging models modulo permutation symmetries},
  author={Ainsworth, Samuel K and Hayase, Jonathan and Srinivasa, Siddhartha},
  journal={International Conference on Learning Representations (ICLR)},
  year={2023}
}

@article{sohn2023styledrop,
  title={Styledrop: Text-to-image synthesis of any style},
  author={Sohn, Kihyuk and Jiang, Lu and Barber, Jarred and Lee, Kimin and Ruiz, Nataniel and Krishnan, Dilip and Chang, Huiwen and Li, Yuanzhen and Essa, Irfan and Rubinstein, Michael and others},
  journal={Advances in Neural Information Processing Systems},
  volume={36},
  pages={66860--66889},
  year={2023}
}

@inproceedings{biggs2024diffusion,
  title={Diffusion soup: Model merging for text-to-image diffusion models},
  author={Biggs, Benjamin and Seshadri, Arjun and Zou, Yang and Jain, Achin and Golatkar, Aditya and Xie, Yusheng and Achille, Alessandro and Swaminathan, Ashwin and Soatto, Stefano},
  booktitle={European Conference on Computer Vision},
  pages={257--274},
  year={2024},
  organization={Springer}
}

@article{salimans2022progressive,
  title={Progressive distillation for fast sampling of diffusion models},
  author={Salimans, Tim and Ho, Jonathan},
  journal={arXiv preprint arXiv:2202.00512},
  year={2022}
}

@article{song2020denoising,
  title={Denoising diffusion implicit models},
  author={Song, Jiaming and Meng, Chenlin and Ermon, Stefano},
  journal={International Conference on Learning Representations (ICLR)},
  year={2021}
}

@article{yadav2023ties,
  title={Ties-merging: Resolving interference when merging models},
  author={Yadav, Prateek and Tam, Derek and Choshen, Leshem and Raffel, Colin A and Bansal, Mohit},
  journal={Advances in neural information processing systems},
  volume={36},
  pages={7093--7115},
  year={2023}
}

@inproceedings{yu2024language,
  title={Language Models are Super Mario: Absorbing Abilities from Homologous Models as a Free Lunch.},
  author={Yu, Le and Yu, Bowen and Yu, Haiyang and Huang, Fei and Li, Yongbin},
  booktitle={ICML},
  year={2024}
}

@inproceedings{stoica2024zipit,
  title={{ZipIt!} merging models from different tasks without training},
  author={Stoica, George and Bolya, Daniel and Bjorner, Jakob and Ramesh, Pratik and Hearn, Taylor and Hoffman, Judy},
  booktitle={International Conference on Learning Representations},
  volume={2024},
  pages={29215--29237},
  year={2024}
}

@inproceedings{yang2024adamerging,
  title={Adamerging: Adaptive model merging for multi-task learning},
  author={Yang, Enneng and Wang, Zhenyi and Shen, Li and Liu, Shiwei and Guo, Guibing and Wang, Xingwei and Tao, Dacheng},
  booktitle={International Conference on Learning Representations},
  volume={2024},
  pages={22743--22763},
  year={2024}
}

@article{matena2022merging,
  title={Merging models with fisher-weighted averaging},
  author={Matena, Michael S and Raffel, Colin},
  journal={Advances in Neural Information Processing Systems},
  volume={35},
  pages={17703--17716},
  year={2022}
}

@article{ortiz2023task,
  title={Task arithmetic in the tangent space: Improved editing of pre-trained models},
  author={Ortiz-Jimenez, Guillermo and Favero, Alessandro and Frossard, Pascal},
  journal={Advances in Neural Information Processing Systems},
  volume={36},
  pages={66727--66754},
  year={2023}
}

@article{ho2020denoising,
  title={Denoising diffusion probabilistic models},
  author={Ho, Jonathan and Jain, Ajay and Abbeel, Pieter},
  journal={Advances in neural information processing systems},
  volume={33},
  pages={6840--6851},
  year={2020}
}

@article{dhariwal2021diffusion,
  title={Diffusion models beat gans on image synthesis},
  author={Dhariwal, Prafulla and Nichol, Alexander},
  journal={Advances in neural information processing systems},
  volume={34},
  pages={8780--8794},
  year={2021}
}

@inproceedings{brooks2023instructpix2pix,
  title={{InstructPix2Pix}: Learning to follow image editing instructions},
  author={Brooks, Tim and Holynski, Aleksander and Efros, Alexei A},
  booktitle={2023 IEEE/CVF Conference on Computer Vision and Pattern Recognition (CVPR)},
  pages={18392--18402},
  year={2023},
  organization={IEEE}
}

@inproceedings{lin2024diffbir,
  title={{DiffBIR}: Toward blind image restoration with generative diffusion prior},
  author={Lin, Xinqi and He, Jingwen and Chen, Ziyan and Lyu, Zhaoyang and Dai, Bo and Yu, Fanghua and Qiao, Yu and Ouyang, Wanli and Dong, Chao},
  booktitle={European conference on computer vision},
  pages={430--448},
  year={2024},
  organization={Springer}
}

@article{zhang2023tale,
  title={A tale of two features: Stable diffusion complements {DINO} for zero-shot semantic correspondence},
  author={Zhang, Junyi and Herrmann, Charles and Hur, Junhwa and Polania Cabrera, Luisa and Jampani, Varun and Sun, Deqing and Yang, Ming-Hsuan},
  journal={Advances in Neural Information Processing Systems},
  volume={36},
  pages={45533--45547},
  year={2023}
}

@inproceedings{tumanyan2023plug,
  title={Plug-and-play diffusion features for text-driven image-to-image translation},
  author={Tumanyan, Narek and Geyer, Michal and Bagon, Shai and Dekel, Tali},
  booktitle={2023 IEEE/CVF Conference on Computer Vision and Pattern Recognition (CVPR)},
  pages={1921--1930},
  year={2023},
  organization={IEEE}
}

@inproceedings{frenkel2024implicit,
  title={Implicit style-content separation using b-lora},
  author={Frenkel, Yarden and Vinker, Yael and Shamir, Ariel and Cohen-Or, Daniel},
  booktitle={European Conference on Computer Vision},
  pages={181--198},
  year={2024},
  organization={Springer}
}

@inproceedings{shah2024ziplora,
  title={{ZipLoRA}: Any subject in any style by effectively merging {LoRAs}},
  author={Shah, Viraj and Ruiz, Nataniel and Cole, Forrester and Lu, Erika and Lazebnik, Svetlana and Li, Yuanzhen and Jampani, Varun},
  booktitle={European Conference on Computer Vision},
  pages={422--438},
  year={2024},
  organization={Springer}
}

@article{wu2024mixture,
  title={Mixture of lora experts},
  author={Wu, Xun and Huang, Shaohan and Wei, Furu},
  journal={International Conference on Learning Representations (ICLR)},
  year={2024}
}

@inproceedings{shenaj2025lora,
  title={{LoRA.rar}: Learning to merge LoRAs via hypernetworks for subject-style conditioned image generation},
  author={Shenaj, Donald and Bohdal, Ondrej and Ozay, Mete and Zanuttigh, Pietro and Michieli, Umberto},
  booktitle={2025 IEEE/CVF International Conference on Computer Vision (ICCV)},
  pages={16132--16142},
  year={2025},
  organization={IEEE}
}

@inproceedings{abuolaim2020defocus,
  title={Defocus deblurring using dual-pixel data},
  author={Abuolaim, Abdullah and Brown, Michael S},
  booktitle={European conference on computer vision},
  pages={111--126},
  year={2020},
  organization={Springer}
}

@article{kuznetsova2020open,
  title={The open images dataset v4: Unified image classification, object detection, and visual relationship detection at scale},
  author={Kuznetsova, Alina and Rom, Hassan and Alldrin, Neil and Uijlings, Jasper and Krasin, Ivan and Pont-Tuset, Jordi and Kamali, Shahab and Popov, Stefan and Malloci, Matteo and Kolesnikov, Alexander and others},
  journal={International journal of computer vision},
  volume={128},
  number={7},
  pages={1956--1981},
  year={2020},
  publisher={Springer}
}

@inproceedings{agustsson2017ntire,
  title={Ntire 2017 challenge on single image super-resolution: Dataset and study},
  author={Agustsson, Eirikur and Timofte, Radu},
  booktitle={2017 IEEE conference on computer vision and pattern recognition workshops (CVPRW)},
  pages={1122--1131},
  year={2017},
  organization={IEEE}
}

@inproceedings{lim2017enhanced,
  title={Enhanced deep residual networks for single image super-resolution},
  author={Lim, Bee and Son, Sanghyun and Kim, Heewon and Nah, Seungjun and Lee, Kyoung Mu},
  booktitle={2017 IEEE conference on computer vision and pattern recognition workshops (CVPRW)},
  pages={1132--1140},
  year={2017},
  organization={IEEE}
}

@inproceedings{zhou2025unires,
  title={{UniRes}: Universal image restoration for complex degradations},
  author={Zhou, Mo and Ye, Keren and Delbracio, Mauricio and Milanfar, Peyman and Patel, Vishal M and Talebi, Hossein},
  booktitle={2025 IEEE/CVF International Conference on Computer Vision (ICCV)},
  pages={13237--13247},
  year={2025},
  organization={IEEE}
}

@inproceedings{wang2021real,
  title={Real-esrgan: Training real-world blind super-resolution with pure synthetic data},
  author={Wang, Xintao and Xie, Liangbin and Dong, Chao and Shan, Ying},
  booktitle={2021 IEEE/CVF International Conference on Computer Vision Workshops (ICCVW)},
  pages={1905--1914},
  year={2021},
  organization={IEEE}
}
\bibliographystyle{iclr2027_conference}

\newpage
\appendix
\section*{Appendix}
\subsection*{Table of Contents}
\startcontents
\printcontents{}{1}{} %
\section{Implementation Details} \label{appendix:additional_impl_details}
\subsection{Image Restoration tasks}
\textbf{Expert Fine-Tuning and Joint Training Baselines.} 
Both our Single-Task Experts and the Joint Multi-Task baseline are fine-tuned from the same adapted base text-to-image checkpoint for the specific image restoration task. 
We utilize the AdamW optimizer ($\beta_1 = 0.9, \beta_2 = 0.999$, weight decay = $0.01$) with a constant learning rate of $8 \times 10^{-5}$, batch size of $128$, and maintain an Exponential Moving Average (EMA) of the weights with a decay rate of $0.999$ for stability. 

\begin{itemize}
    \item \textit{Single-Task Experts:} Fine-tuned exclusively on their respective singular datasets (e.g., DPDD for defocus or OID for motion blur).
    \item \textit{Joint Multi-Task Training:} Fine-tuned on a unified dataset where batches are constructed by sampling equally ($0.5 : 0.5$ ratio) from both constituent task datasets. This baseline updates all 890M U-Net parameters simultaneously.
\end{itemize}

\textbf{OptIMerge Optimization Setup.} 
In stark contrast to full fine-tuning, during the OptIMerge phase, the weights of the constituent expert models are kept strictly frozen, and gradient descent is applied exclusively to the structural alignment and routing parameterizations. 
For all low-rank formulations (both biased LoRA in the attention layers and the Cayley Transforms in the convolutional layers), we set the rank bottleneck to $r = 4$. The merging coefficients are optimized for $15,000$ steps using AdamW using a peak learning rate of $0.1$, paired with a 500-step linear warmup and subsequent linear decay schedule. To ensure stable and diverse gradient updates when sampling from the constituent multi-task calibration datasets (mixed at a $0.5 : 0.5$ sampling ratio), we utilize a per-device batch size of $4$ combined with $64$ gradient accumulation steps. 
For experiments requiring explicit task balancing, loss normalization factors are dynamically calibrated over the initial 30 optimization steps to prevent gradient domination by high-magnitude tasks.
\noindent\textbf{Parameter Efficiency and Auxiliary Layers.} To maximize parameter efficiency, OptIMerge parameterizes only the high-impact components of the U-Net: the attention projections ($W_Q, W_K, W_V, W_O$) and the 4D spatial convolutions. Auxiliary parameters (e.g. Fully connected layers (MLPs), normalization layers, GroupNorm, time embeddings etc.) are interpolated using a uniform scalar average ($\alpha = 0.5$). This selective parameterization is what restricts the total optimization footprint to less than $0.4\%$ of the network size while still resolving geometric and cross-channel interference.

\textbf{Details of Image-restoration Datasets.}
For super-resolution task, we use the degradation pipeline from RealESRGAN~\citep{wang2021real} on a combination of DIV2K~\citep{agustsson2017ntire} and Flickr2K~\citep{lim2017enhanced} datasets which we refer to as DF2K following~\citet{zhou2025unires}. For motion-deblur task, we simulate camera shake blur by applying random blur kernels of different intensities and sizes to individual object within an image, mimicking the blur that occurs due to motion movement using images from Open Image Dataset~\citep{kuznetsova2020open}. For defocus deblur task,  we use images from DPDD dataset~\citep{abuolaim2020defocus}. The expert models are trained on entire training splits. For calibration dataset that are used to train merger parameters used in OptIMerge, we use a small subset of 100-500 indistribution images from these datasets. 

\subsection{Text-to-Image Subject-Style Personalization Tasks}
\label{sec:t2i_tasks}
\noindent \textbf{Obtaining Personalized Subject/Style Models.}We independently fine-tune a Stable Diffusion v1.5 model on subject and style reference images using the DreamBooth method to obtain subject/style personalized expert models. During the fine-tuning, we update the U-net weights using Adam optimizer for $600$ steps with batch size of $1$ and learning rate of $0.00005$ while keeping the text encoder frozen. We choose the subject images from DreamBooth dataset, where each dataset contains $4-5$ images along with its unique token ([V], e.g. ``monadikos") and ``subject category" (e.g. dog, stuffed animal etc.). For example, for the coorgi dog in Fig.~\ref{fig:subject_style}, the subject prompt becomes ``A photo of [V] dog". Similarly, for the style-learning, we use StyleDrop dataset which offers one reference image of each style along with the corresponding style descriptor phrase ([S], e.g. ``watercolor painting"). ``An image in $<$style description$>$ style" prompt that uses only a style descriptor phrase but does not use unique token as advocated in StyleDrop paper. 

\noindent \textbf{OptIMerge Optimization Setup. }
Since the personalization is data-scarce setup where only a handful subject/style reference images are available, we use the images from the original subject/style training set for performing the OptIMerge. We set the rank bottleneck for all low-rank parameterizations to $r=4$, and optimize the merger coefficients only for $100$ steps with the batch size of $2$ and learning rate of $0.001$. The rest of the setup remains similar to image restoration tasks.

\section{Additional Ablations and Experiments for OptIMerge}
\label{appendix:additional_ablations}
\subsection{Ablation on parametrization of Attention Module}

We evaluate the impact of our attention module parameterizations in Table \ref{tab:ablation_attention_module}, keeping the convolutional layers frozen at a naive arithmetic average ($\alpha=0.5$) to isolate the role of semantic alignment. 
While standard low-rank routing ($\text{OptIMerge}_{\text{LoRA}}$) yields limited improvements over the zero-shot baseline, explicitly factoring global blending shifts into dedicated biases ($\text{OptIMerge}_{\text{Biased\_LoRA}}$, \cref{eqn:biased-lora-mask}) provides consistent, measurable gains across all task pairs and metrics, justifying the adoption of this factored architectural variant. 
However, optimizing interpolation weights alone remains insufficient when constituent expert models have drifted into misaligned permutation spaces. Introducing the unconstrained projection adapter ($\text{OptIMerge}_{\text{Attn}}$) to dynamically align these feature bases prior to routing triggers substantial performance leaps. 
For instance, in the heavily misaligned OID + DF2K configuration, $\text{OptIMerge}_{\text{Attn}}$ improves OID PSNR by 1.47 dB and reduces LPIPS by 0.044 compared to unaligned biased routing. These results empirically validate that dynamic feature alignment is a strict prerequisite for resolving destructive interference and bridging the performance gap toward the joint-training upper bound.
\begin{table*}[h]
\centering
\resizebox{\textwidth}{!}{
\begin{tabular}{lccccccc}
\toprule
\multirow{2}{*}{\textbf{Method / Strategy}} & \multirow{2}{*}{\textbf{Trainable Params}} & \multicolumn{3}{c}{\textbf{OID}} & \multicolumn{3}{c}{\textbf{DPDD}} \\
\cmidrule(lr){3-5} \cmidrule(lr){6-8}
& & \textbf{LPIPS $\downarrow$} & \textbf{PSNR $\uparrow$} & \textbf{SSIM $\uparrow$} & \textbf{LPIPS $\downarrow$} & \textbf{PSNR $\uparrow$} & \textbf{SSIM $\uparrow$} \\
\midrule
\textit{Single-Task Experts} & --- & 0.2113 & 22.92 & 0.6266 & 0.2023 & 23.39 & 0.6392 \\
\textit{Joint Multi-Task Training} & 890M (100\%) & 0.2154 & 22.59 & 0.6173 & 0.2091 & 23.37 & 0.6404 \\
\textit{Naive Arithmetic Soup ($\alpha=0.5$)} & 0 (0\%) & 0.2899 & 20.88 & 0.5601 & 0.2641 & 23.15 & 0.6398 \\ %
\midrule
$\text{OptIMerge}_{\text{LoRA}}$ & 887.7K & 0.2647 & 21.68 & 0.5774 & 0.2562 &  23.13 & 0.6289 \\
$\text{OptIMerge}_{\text{Biased\_LoRA}}$ & 1.1M & \underline{0.2609} & 
2\underline{1.78} & \underline{0.5808} & \underline{0.2551} & \underline{23.16} & \underline{0.6301} \\
$\text{OptIMerge}_{\text{Attn}}$ & $\sim$1.9M & \textbf{0.2265} & \textbf{22.43} & \textbf{0.6049} & \textbf{0.2213} & \textbf{23.59} & \textbf{0.6423} \\ %
\bottomrule
\end{tabular}
}
\resizebox{\textwidth}{!}{
\begin{tabular}{lccccccc}
\toprule
\multirow{2}{*}{\textbf{Method / Strategy}} & \multirow{2}{*}{\textbf{Trainable Params}} & \multicolumn{3}{c}{\textbf{DPDD}} & \multicolumn{3}{c}{\textbf{DF2K SR}} \\
\cmidrule(lr){3-5} \cmidrule(lr){6-8}
& & \textbf{LPIPS $\downarrow$} & \textbf{PSNR $\uparrow$} & \textbf{SSIM $\uparrow$} & \textbf{LPIPS $\downarrow$} & \textbf{PSNR $\uparrow$} & \textbf{SSIM $\uparrow$} \\
\midrule
\textit{Single-Task Experts} & --- & 0.2023 & 23.39 & 0.6392 & 0.3131 & 21.26 & 0.5323 \\
\textit{Joint Multi-Task Training} & 890M (100\%) & 0.2185 & 23.11 & 0.6293 & 0.3160 & 21.24 & 0.5255 \\
\textit{Naive Arithmetic Soup ($\alpha=0.5$)} & 0 (0\%) & 0.2774 & 22.57 & 0.6121 & 0.3447 & 21.10 & 0.5259 \\
\midrule
$\text{OptIMerge}_{\text{LoRA}}$ & 887.7K & 0.2806 & \textbf{22.72} & 0.6028 & 0.3608 & 20.43 & 0.4872 \\
$\text{OptIMerge}_{\text{Biased\_LoRA}}$ & 1.1M & \underline{0.2791} & 22.34 & \underline{0.6041} & \underline{0.3529} & \underline{20.57} & \underline{0.4964} \\
$\text{OptIMerge}_{\text{Attn}}$ & $\sim$1.9M & \textbf{0.2506} & \underline{22.66} &  \textbf{0.6113} &  \textbf{0.3208} &  \textbf{20.88} &  \textbf{0.5193} \\
\bottomrule
\end{tabular}
}
\resizebox{\textwidth}{!}{
\begin{tabular}{lccccccc}
\toprule
\multirow{2}{*}{\textbf{Method / Strategy}} & \multirow{2}{*}{\textbf{Trainable Params}} & \multicolumn{3}{c}{\textbf{OID}} & \multicolumn{3}{c}{\textbf{DF2K SR}} \\
\cmidrule(lr){3-5} \cmidrule(lr){6-8}
& & \textbf{LPIPS $\downarrow$} & \textbf{PSNR $\uparrow$} & \textbf{SSIM $\uparrow$} & \textbf{LPIPS $\downarrow$} & \textbf{PSNR $\uparrow$} & \textbf{SSIM $\uparrow$} \\
\midrule
\textit{Single-Task Experts} & --- & 0.2113 & 22.92 & 0.6266 & 0.3131 & 21.26 & 0.5323 \\
\textit{Joint Multi-Task Training} & 890M (100\%) & 0.2386 & 22.03 & 0.5867 & 0.2993 & 21.49 & 0.5407 \\
\textit{Naive Arithmetic Soup ($\alpha=0.5$)} & 0 (0\%) & 0.2938 & 20.31 & 0.5292 & 0.3501 & 20.99 & 0.4970 \\ %
\midrule
$\text{OptIMerge}_{\text{LoRA}}$ & 887.7K & 0.293 & 20.37 & 0.5286 & \underline{0.3468} & 21.01 & 0.4986 \\
$\text{OptIMerge}_{\text{Biased\_LoRA}}$ & 1.1M & \underline{0.2897} & \underline{20.46} & \underline{0.531} & 0.3476
& \underline{21.06} & \underline{0.5} \\
$\text{OptIMerge}_{\text{Attn}}$ & $\sim$1.9M & \textbf{0.2454} & \textbf{21.93} & \textbf{0.5821} & \textbf{0.3191} & \textbf{21.42} & \textbf{0.5271}  \\ %

\bottomrule
\end{tabular}
}
\caption{\textbf{Ablation over model merge parametrization of Attention module} on (a) Motion Deblurring (OID) + Defocus Deblurring (DPDD), (b) Defocus Deblurring (DPDD) + Super-Resolution (DF2K), and (c) Motion Deblurring (OID) + Super-Resolution (DF2K). Best merging performance is in \textbf{bold}, second best is \underline{underlined}.}
\label{tab:ablation_attention_module}
\end{table*}

\subsection{Ablation over low rank $r$ in Attention Modules and Conv Modules}
We investigate the sensitivity of our framework to the low-rank bottleneck capacity by evaluating $\text{OptIMerge}_{\text{Attn}}$ and $\text{OptIMerge}$ across ranks $r \in \{4, 16, 32\}$ on the combined OID and DPDD restoration tasks. We summarize our results in \Cref{tab:low-rank-capacity-ablation}.
For the attention-only formulation, expanding the rank capacity from $r=4$ to $r=16$ yields  performance gains, improving OID PSNR from 22.43 to 22.63 and DPDD PSNR from 23.59 to 24.00. 
Conversely, the complete framework incorporating convolutional spatial alignment ($\text{OptIMerge}$) exhibits remarkable parameter efficiency. 
The highly constrained $r=4$ configuration, utilizing only 3.2M trainable parameters, already achieves peak performance on the OID dataset with a PSNR of 22.72 and maintains strong fidelity on DPDD, rendering higher ranks largely redundant. 

\begin{table*}[h]
\centering
\resizebox{\textwidth}{!}{
\begin{tabular}{lccccccc}
\toprule
\multirow{2}{*}{\textbf{Method / Strategy}} & \multirow{2}{*}{\textbf{Trainable Params}} & \multicolumn{3}{c}{\textbf{OID}} & \multicolumn{3}{c}{\textbf{DPDD}} \\
\cmidrule(lr){3-5} \cmidrule(lr){6-8}
& & \textbf{LPIPS $\downarrow$} & \textbf{PSNR $\uparrow$} & \textbf{SSIM $\uparrow$} & \textbf{LPIPS $\downarrow$} & \textbf{PSNR $\uparrow$} & \textbf{SSIM $\uparrow$} \\
\midrule
\textit{Single-Task Experts} & --- & 0.2113 & 22.92 & 0.6266 & 0.2023 & 23.39 & 0.6392 \\
\textit{Joint Multi-Task Training} & 890M (100\%) & 0.2154 & 22.59 & 0.6173 & 0.2091 & 23.37 & 0.6404 \\
\textit{Naive Arithmetic Soup ($\alpha=0.5$)} & 0 (0\%) & 0.2899 & 20.88 & 0.5601 & 0.2641 & 23.15 & 0.6398 \\ %
\midrule
$\text{OptIMerge}_{\text{Attn}} (r=4)$ & 1.9M & 0.2265 & 22.43 & 0.6049 & 0.2213 & 23.59 & 0.6423 \\ %
$\text{OptIMerge}_{\text{Attn}} (r=16)$ & 6.7M & \textbf{0.2193} & \textbf{22.63} & \textbf{0.6169} & 0.2154 & \textbf{24.00} & 0.6549 \\ %
$\text{OptIMerge}_{\text{Attn}} (r=32)$ & 13.3M & 0.2197 & \textbf{22.63} & 0.6162 & \textbf{0.2147} & 23.93  & \textbf{0.6560}\\ %
\midrule
$\text{OptIMerge}$ $(r=4)$ & 3.2M & \textbf{0.2168} & \textbf{22.72} & \textbf{0.6190} & 0.2191 & \textbf{23.91} & \textbf{0.6561} \\ %
$\text{OptIMerge}$ $(r=16)$ & 8.1M & 0.2189 & 22.50 & 0.6118 & 0.2158 & 23.88 & 0.6536 \\ %
$\text{OptIMerge}$ $(r=32)$ & 14.6M & 0.2181 & 22.60 & 0.6145 & \textbf{0.214} & 23.89 & 0.6529 \\ %
\bottomrule
\end{tabular}
}
\caption{\textbf{Ablation over low-rank bottleneck capacity $r$}:  Comparison of $\text{OptIMerge}_{\text{Attn}}$ and $\text{OptIMerge}$ across varying bottleneck capacities. By successfully aligning spatial and semantic representations, the complete OptIMerge framework requires only 3.2M trainable parameters ($r=4$) to achieve peak multi-task fidelity.}
\label{tab:low-rank-capacity-ablation}
\end{table*}

\subsection{Generalizing to three component model merge}
To evaluate the scalability of our framework beyond pairwise unification, \Cref{tab:model_soups_full} presents the quantitative results of a three-way model merge across Super-Resolution (DF2K), Motion Deblurring (OID), and Defocus Deblurring (DPDD). 
As expected, scaling the merging process to three distinct physical restoration tasks severely exacerbates destructive interference. 
The Naive Soup baseline suffers massive performance degradation across all three datasets, dropping to a PSNR of 20.24 on OID and 20.90 on DF2K.   
The introduction of semantic alignment via OptIMerge$_\text{Attn}$ is critical, recovering the majority of the lost performance and establishing strong second-best results. 
Ultimately, our complete formulation, OptIMerge$_\text{Full}$, achieves the highest simultaneous multi-task performance across all evaluated metrics. For instance, on the DPDD dataset, OptIMerge$_\text{Ours}$ reaches a PSNR of 23.00 and an SSIM of 0.6233. 
This demonstrates that our method scales to $N=3$ experts, substantially narrowing the gap to the single-task experts, although a gap remains (e.g., 1.4\,dB on OID), indicating that interference grows with the number of merged experts.

\begin{table*}[t]
\centering
\label{tab:model_results}
\resizebox{\textwidth}{!}{%
\begin{tabular}{cc ccc ccc}
\toprule
\multicolumn{2}{c}{Soup Weights} & \multicolumn{3}{c}{OID} & \multicolumn{3}{c}{DPDD} \\
\cmidrule(lr){1-2} \cmidrule(lr){3-5} \cmidrule(lr){6-8}
DPDD ($w_1$) & OID ($w_2$) & LPIPS ($\downarrow$) & PSNR ($\uparrow$) & SSIM ($\uparrow$) & LPIPS ($\downarrow$) & PSNR ($\uparrow$) & SSIM ($\uparrow$) \\
\midrule
0.10 & 0.90 & 0.2132 & 22.71 & 0.6216 & 0.3627 & 23.30 & 0.6461 \\
0.25 & 0.75 & 0.2305 & 22.08 & 0.6036 & 0.3229 & 23.17 & 0.6434 \\
0.50 & 0.50 & 0.2900 & 20.88 & 0.5601 & 0.2641 & 23.15 & 0.6398 \\
0.75 & 0.25 & 0.3319 & 20.34 & 0.5374 & 0.2227 & 23.43 & 0.6446 \\
0.90 & 0.10 & 0.3498 & 20.00 & 0.5234 & 0.2111 & 23.50 & 0.6440 \\
\bottomrule
\end{tabular}%
}
\caption{Performance metrics of naive model soup for various DPDD and OID weight configurations. Soup weights of $w_1 = 0.5$ and $w_2 = 0.5$ offers good balance in performance on both the datasets.}
\end{table*}

\begin{table*}[t]
\centering
\small
\resizebox{\textwidth}{!}{
\begin{tabular}{l ccc ccc ccc}
\toprule
& \multicolumn{3}{c}{\textbf{DF2K}} & \multicolumn{3}{c}{\textbf{OID}} & \multicolumn{3}{c}{\textbf{DPDD}} \\
\cmidrule(lr){2-4} \cmidrule(lr){5-7} \cmidrule(lr){8-10}
\textbf{Method/Strategy} & \textbf{LPIPS}$\downarrow$ & \textbf{PSNR}$\uparrow$ & \textbf{SSIM}$\uparrow$ & \textbf{LPIPS}$\downarrow$ & \textbf{PSNR}$\uparrow$ & \textbf{SSIM}$\uparrow$ & \textbf{LPIPS}$\downarrow$ & \textbf{PSNR}$\uparrow$ & \textbf{SSIM}$\uparrow$ \\
\midrule
\textit{Single-Task Experts} &  0.3131 & 21.26 & 0.5323 & 0.2113 & 22.92 & 0.6266 & 0.2023 & 23.39 & 0.6392 \\
\textit{Joint-Task Experts} & 0.2989 & 21.53 & 0.5378 & 0.2180 & 22.48 & 0.6046	& 0.2158 & 23.19 & 0.6319 \\
$\text{Naive Soup}$ & 0.3527 & 20.90 & 0.5013 & 0.3021 & 20.24 & 0.5235 & 0.3043 & 22.45 & 0.6118             \\
\midrule
$\text{OptIMerge}_{\text{Conv-KC}}$ & 0.3376 & 20.96 & 0.5098 & 0.2898 & 20.74 & 0.5420 & 0.2749 & 22.64 & 0.6108 \\
$\text{OptIMerge}_{\text{Attn}}$          & \underline{0.3196} & \underline{21.10} & \underline{0.5218} & \underline{0.2563} & \underline{21.44} & \underline{0.5696} & \underline{0.2470} & \underline{22.95} & \underline{0.6231} \\
$\text{OptIMerge}$          & \textbf{0.3192}    & \textbf{21.12}    & \textbf{0.5237}    & \textbf{0.2548}    & \textbf{21.48}    & \textbf{0.5699}    & \textbf{0.2469}    & \textbf{23.00}    & \textbf{0.6233}    \\
\bottomrule
\end{tabular}
}
\caption{\textbf{Performance of three-way model merge:} Quantitative evaluation of model soup merge strategies across DF2K, OID, and DPDD datasets. Best results are in \textbf{bold}, and second best are \underline{underlined}.}
\label{tab:model_soups_full}
\end{table*}

\section{Additional Qualitative Results for OptIMerge}

\begin{figure}[h]
\centering
\begin{subfigure}{\textwidth}
    \zHeaderRow
\end{subfigure}
\hfill
\begin{subfigure}{\textwidth}
    \setzoom{0.50}{0.30}{0.79, 0.21}{yellow}
    \zoomimg{oid_dpdd/oid/4/input.jpeg}
    \zoomimg{oid_dpdd/oid/4/target.jpeg}
    \zoomimg{oid_dpdd/oid/4/expert.jpeg}
    \zoomimg{oid_dpdd/oid/4/naive_soup.jpeg}
    \zoomimg{oid_dpdd/oid/4/optmerge.jpeg}
\end{subfigure}
\hfill
\begin{subfigure}{\textwidth}
    \setzoom{0.30}{0.55}{0.79, 0.21}{yellow}
    \zoomimg{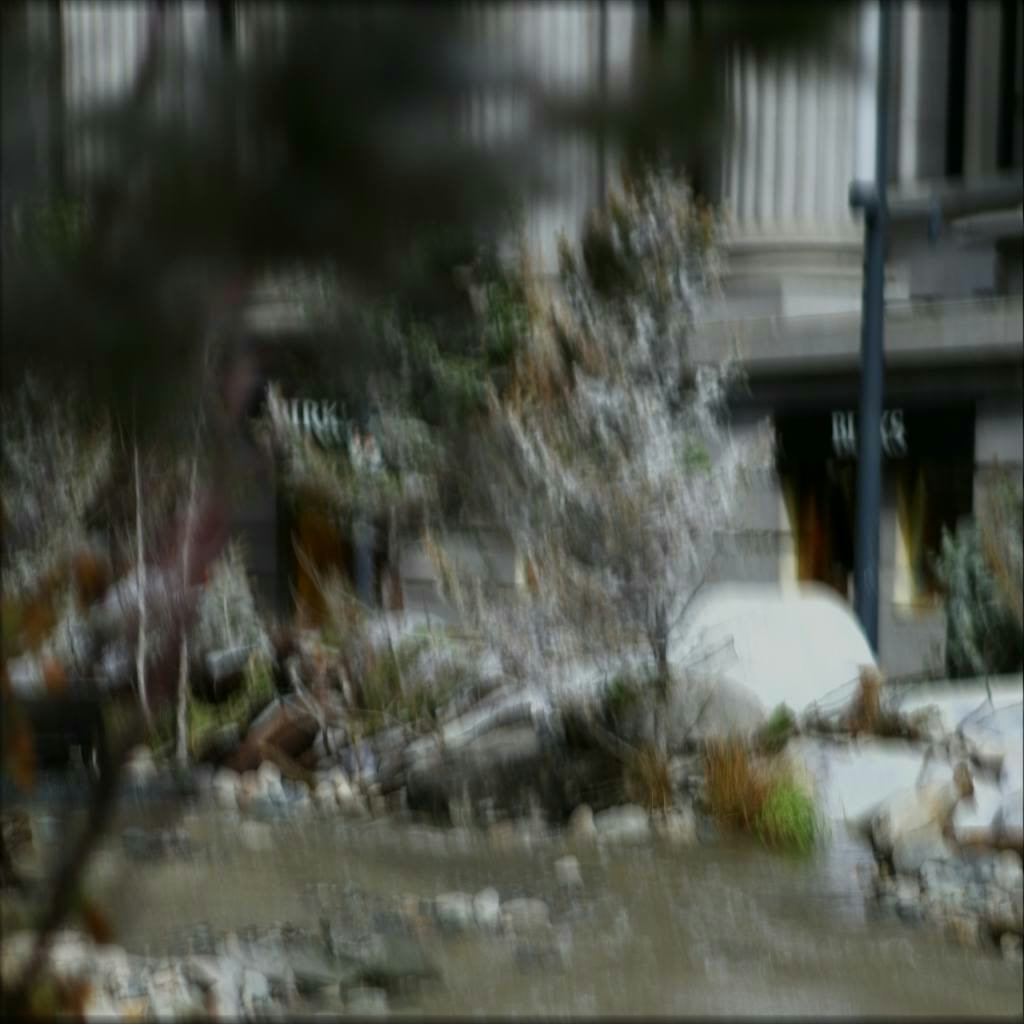}
    \zoomimg{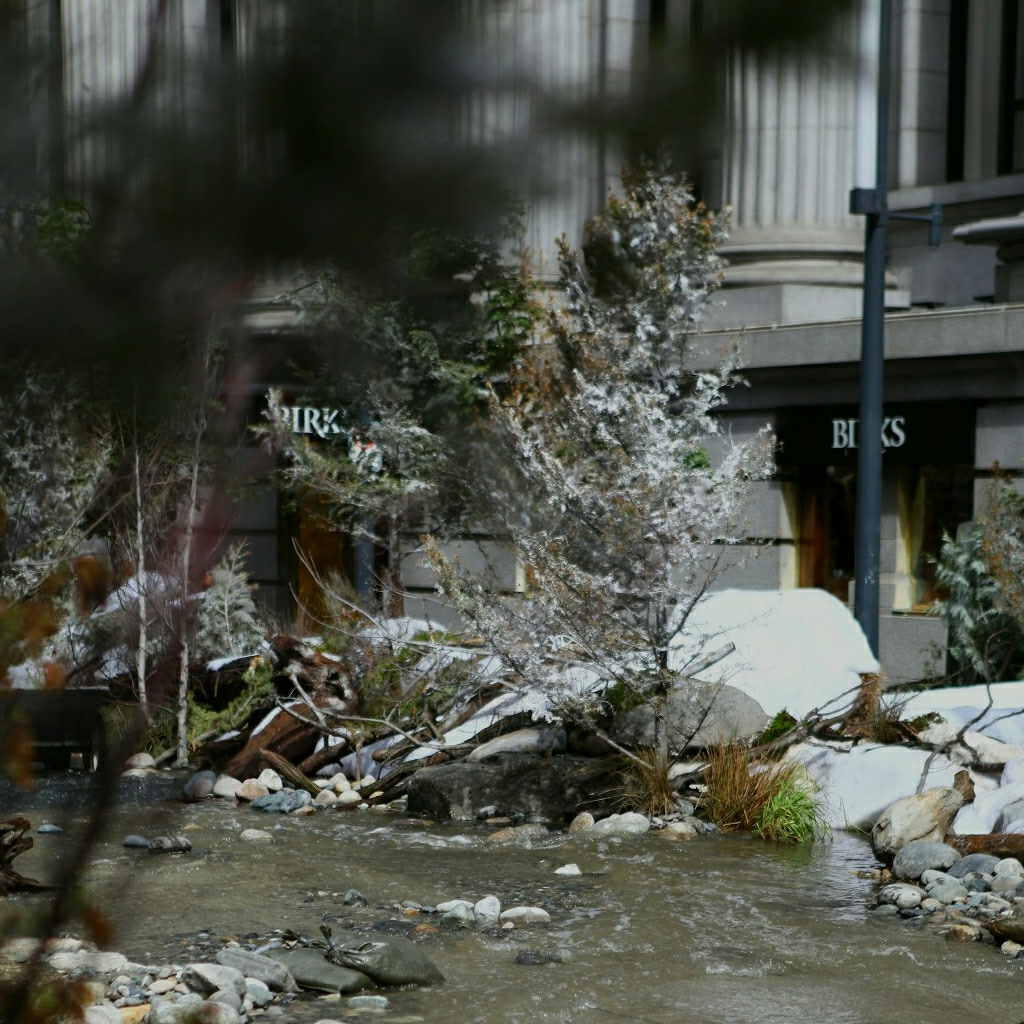}
    \zoomimg{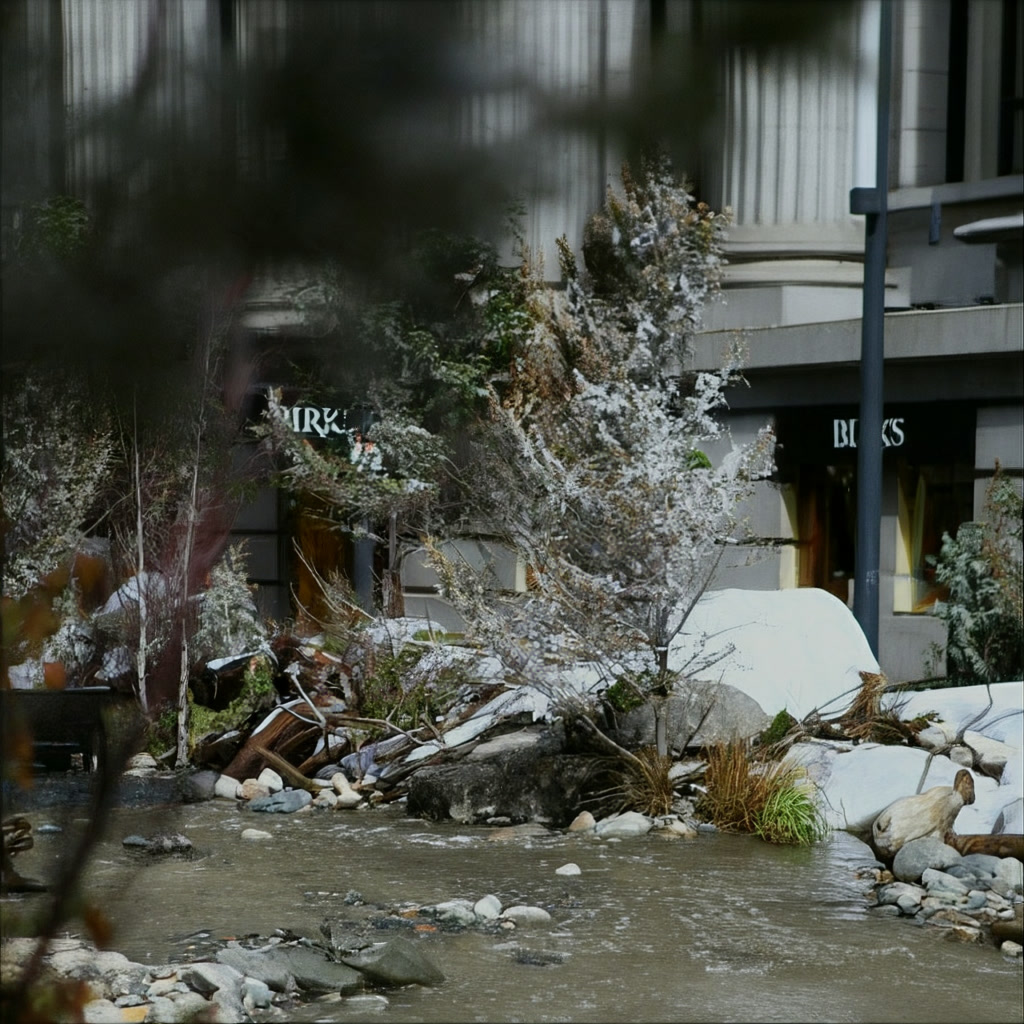}
    \zoomimg{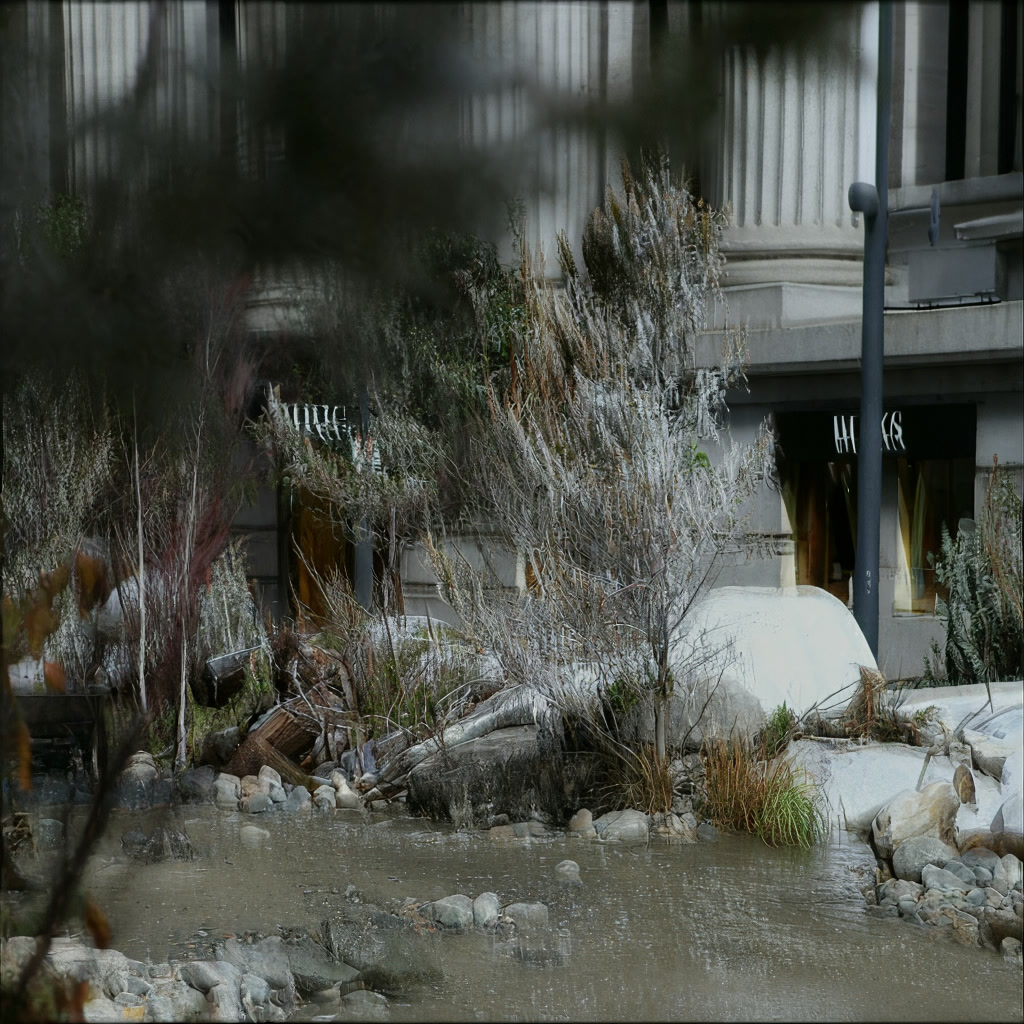}
    \zoomimg{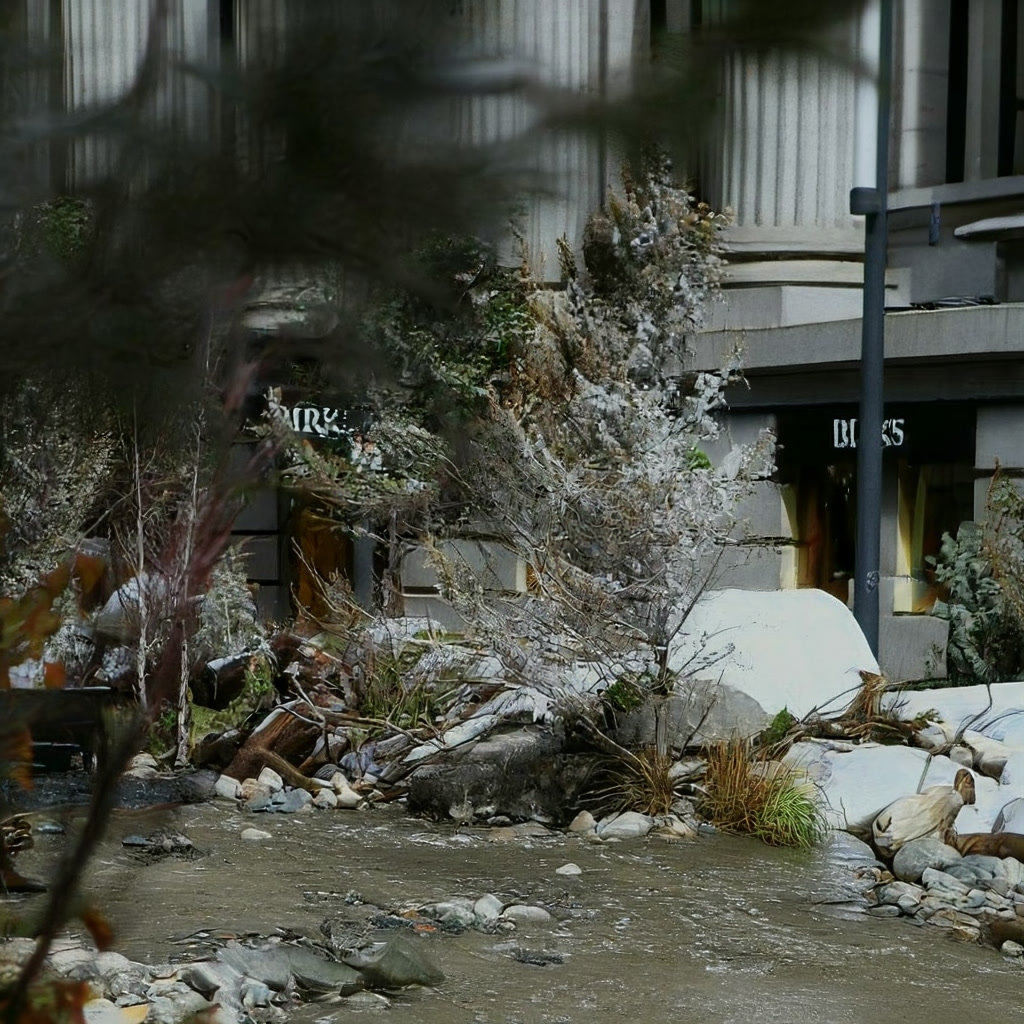}
\end{subfigure}
\hfill
\begin{subfigure}{\textwidth}
    \setzoom{0.50}{0.60}{0.79, 0.21}{yellow}
    \zoomimg{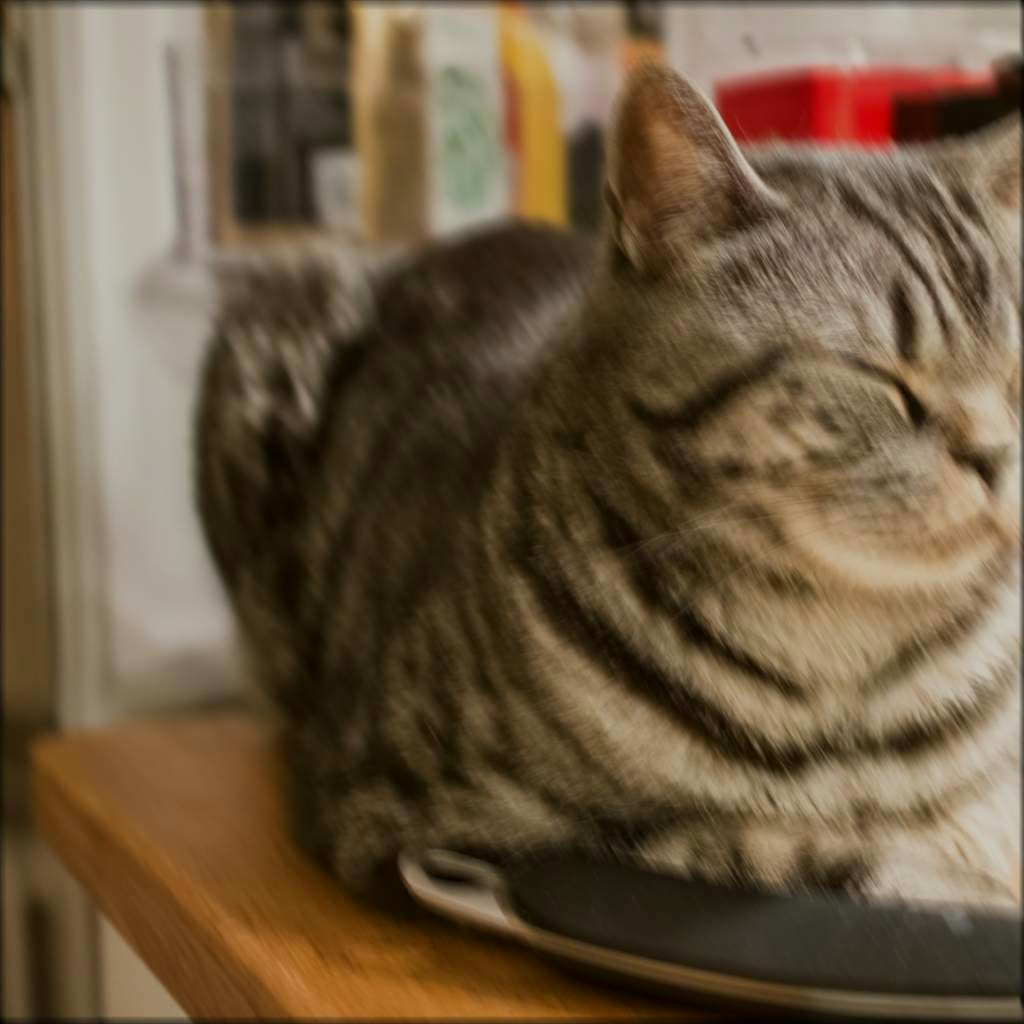}
    \zoomimg{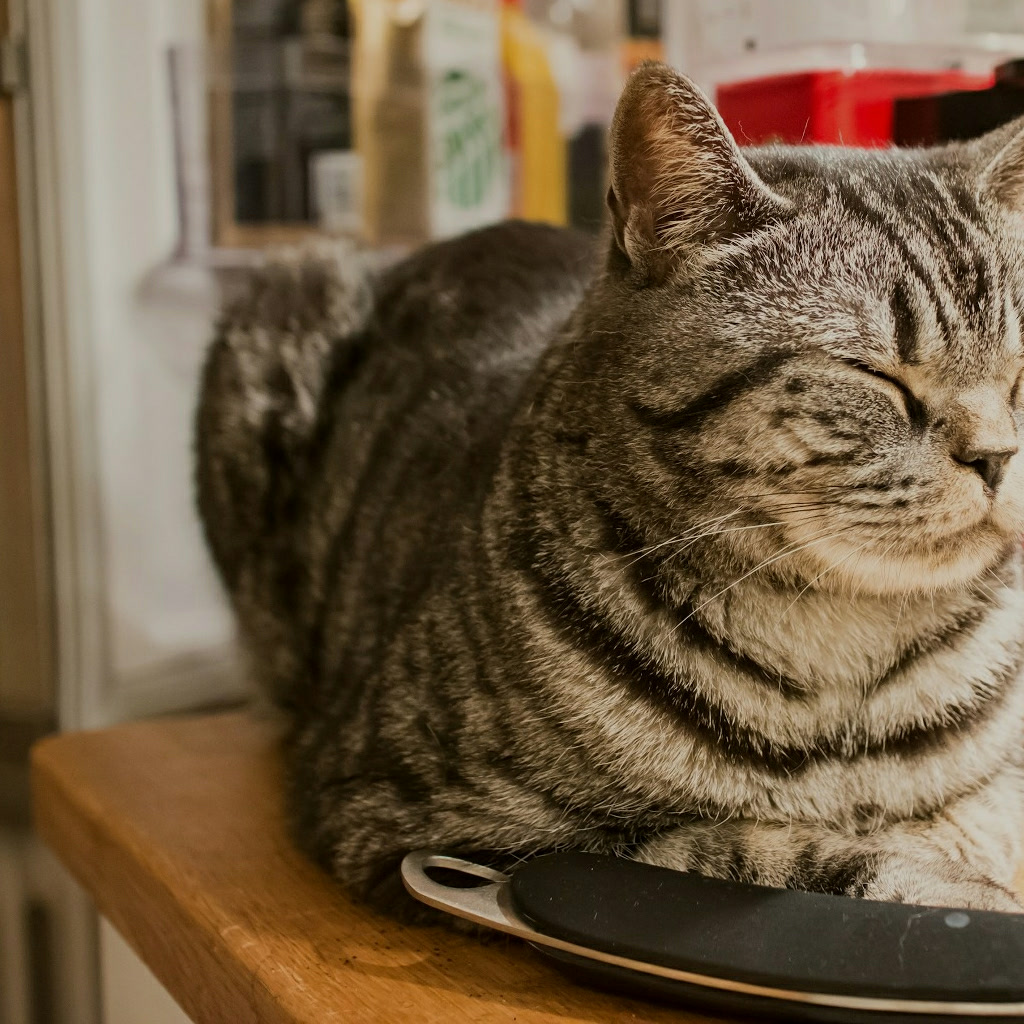}
    \zoomimg{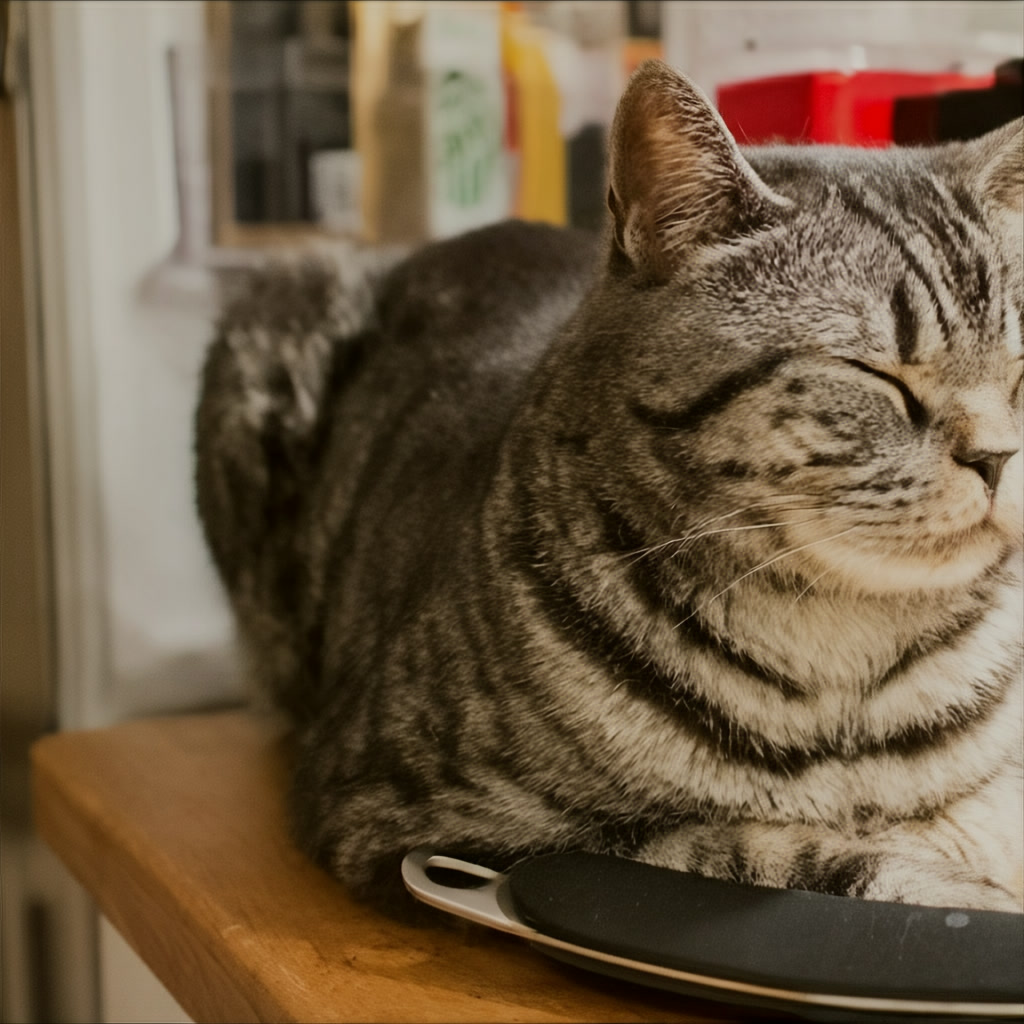}
    \zoomimg{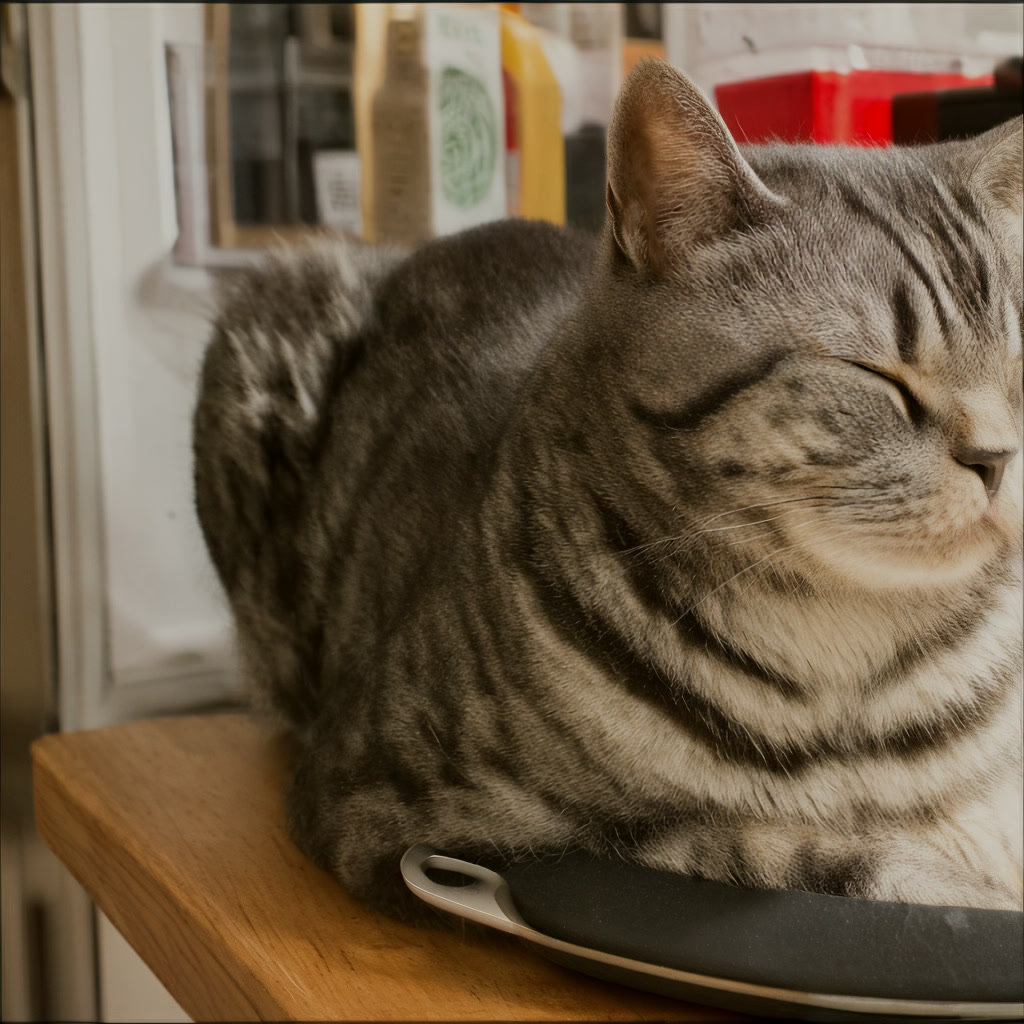}
    \zoomimg{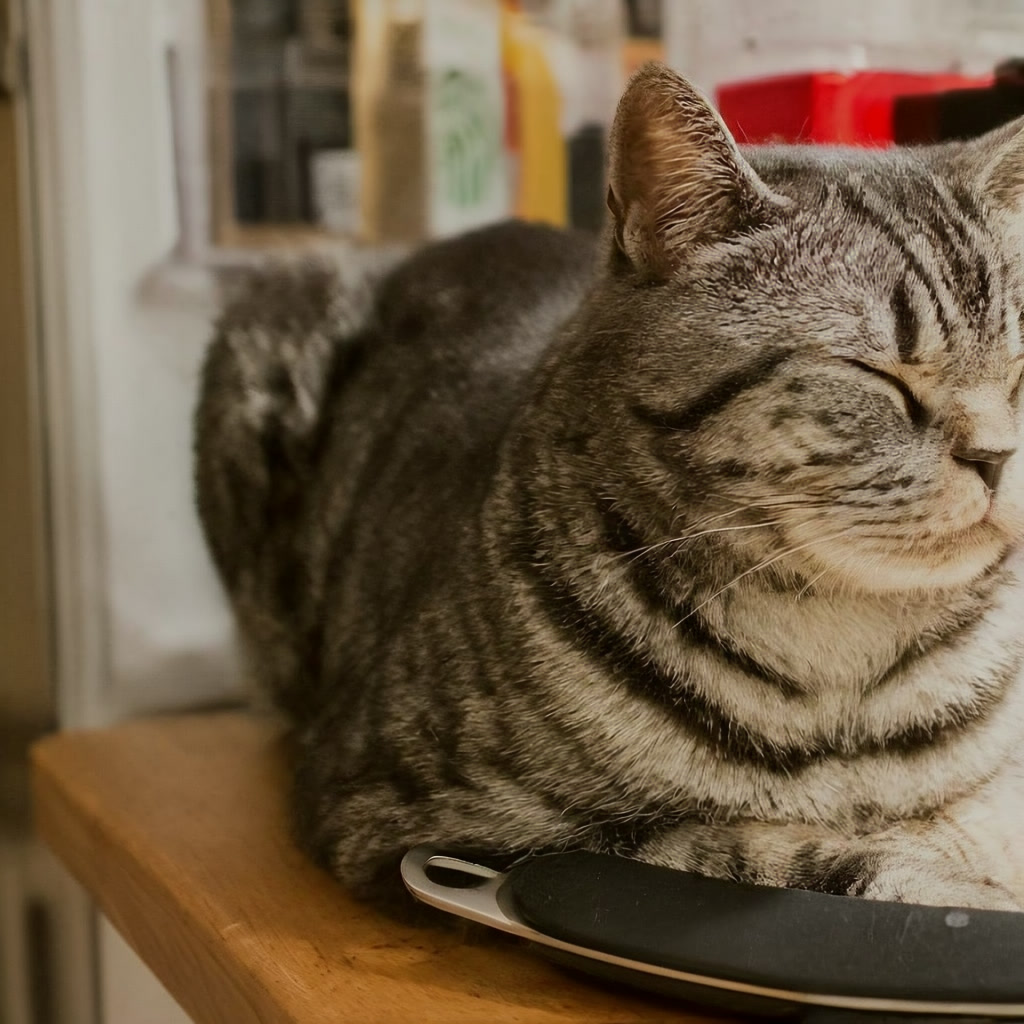}
\end{subfigure}
\hfill
\caption{\textbf{OID-DPDD model soup results with OptIMerge on OID data}. (left-to-right) We provide input, target, and images generated from the expert model, naive model soup, and OptIMerge.}
\label{fig:oid_dpdd_soup_on_oid_data}
\end{figure}

\begin{figure}[h]
\centering
\begin{subfigure}{\textwidth}
    \zHeaderRow
\end{subfigure}
\hfill
\begin{subfigure}{\textwidth}
    \setzoom{0.38}{0.78}{0.79, 0.21}{yellow}
    \zoomimg{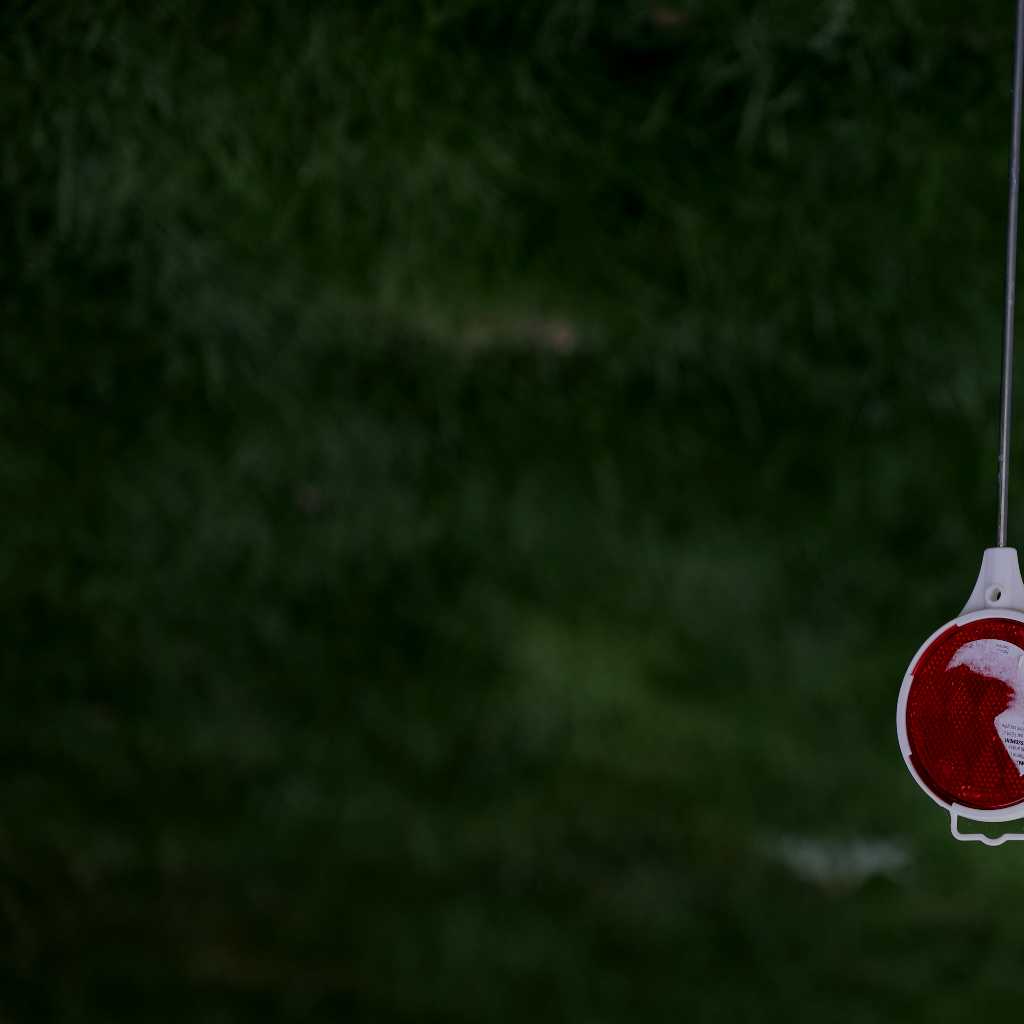}
    \zoomimg{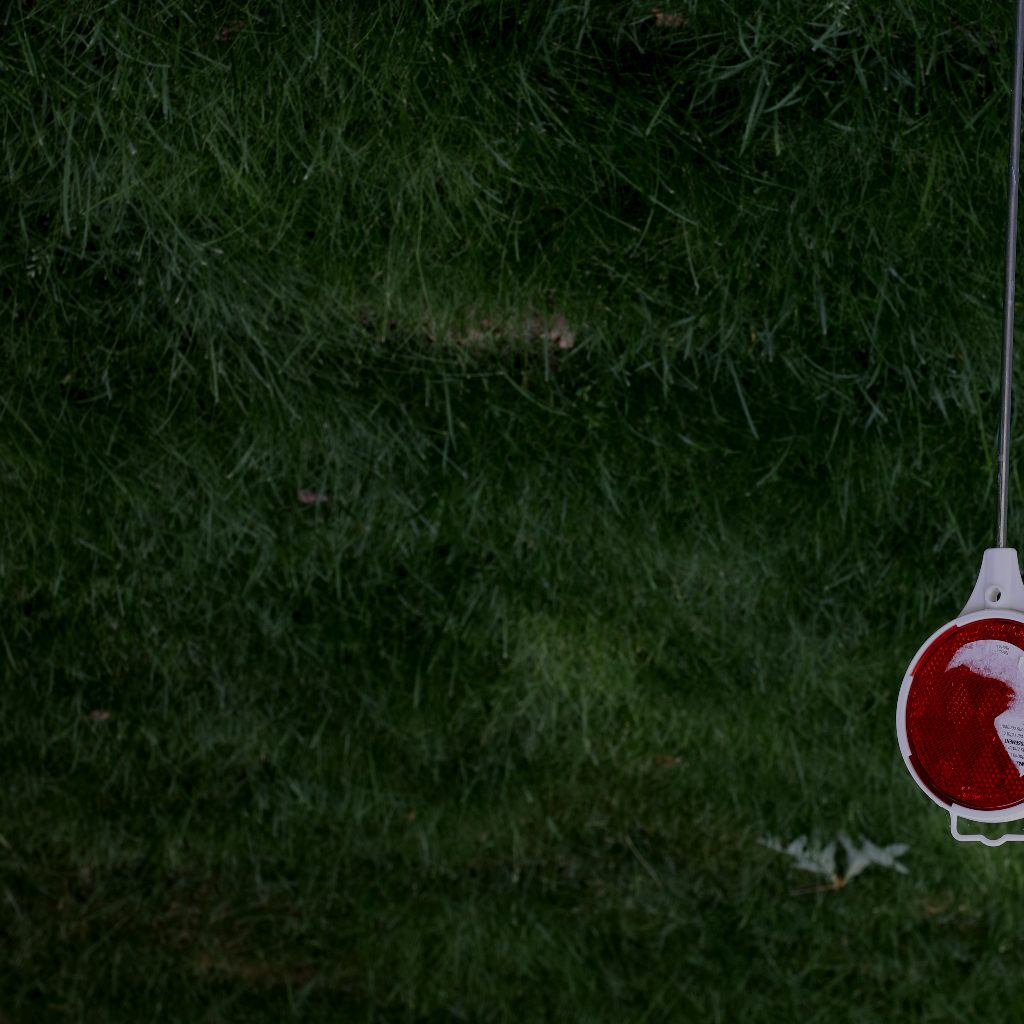}
    \zoomimg{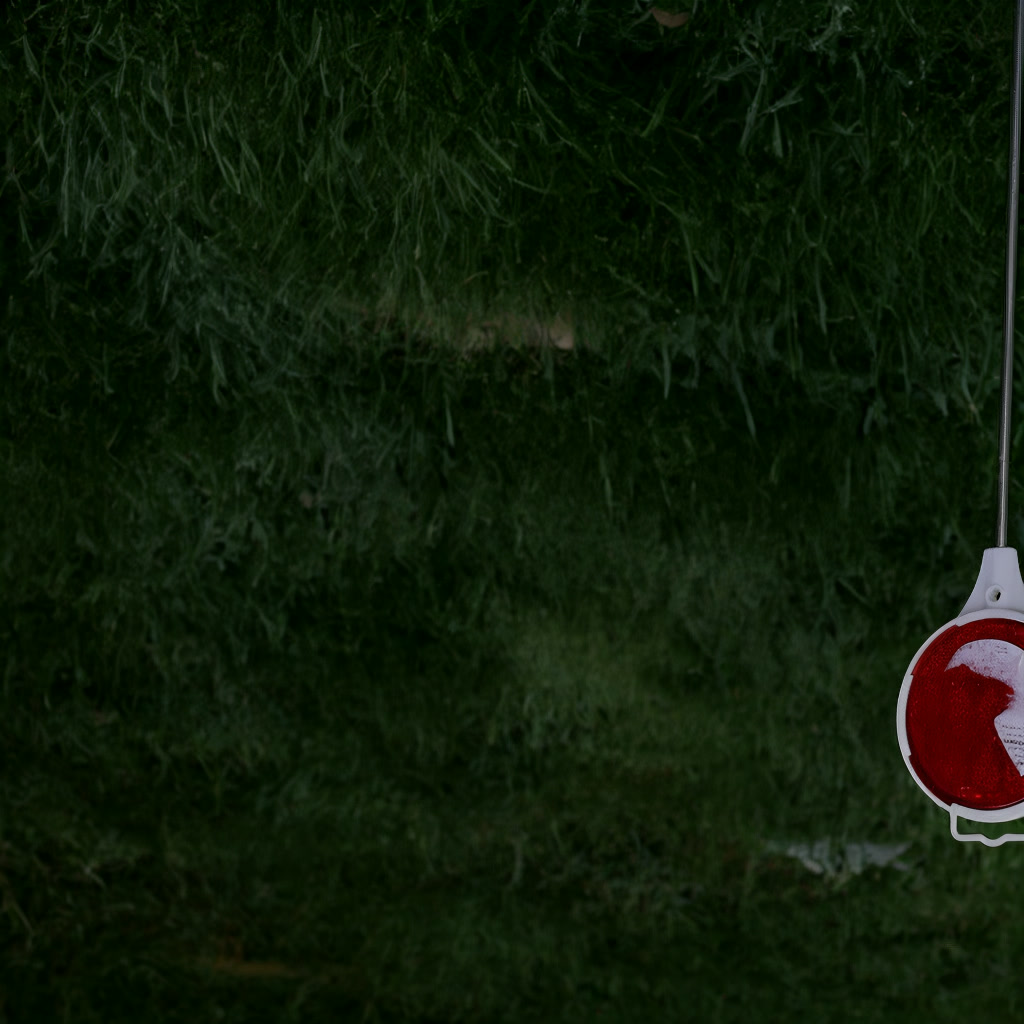}
    \zoomimg{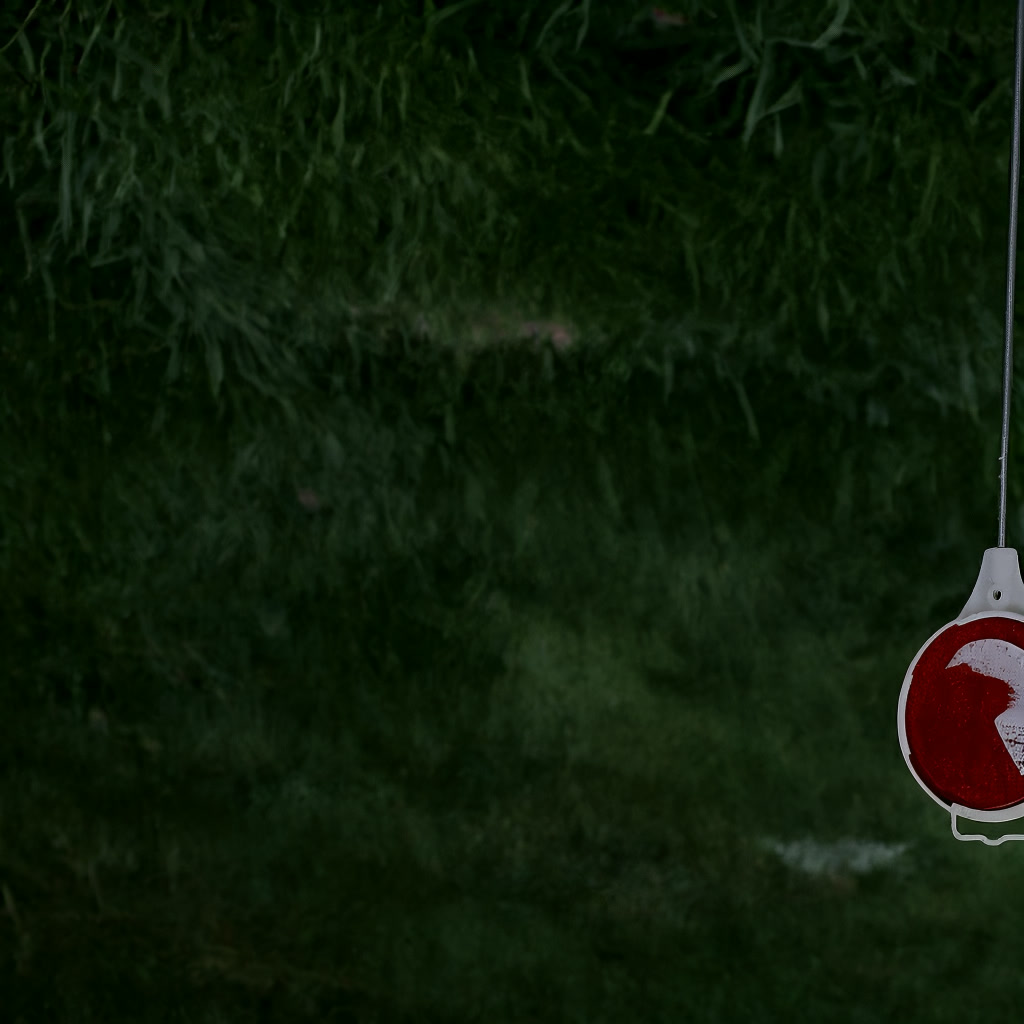}
    \zoomimg{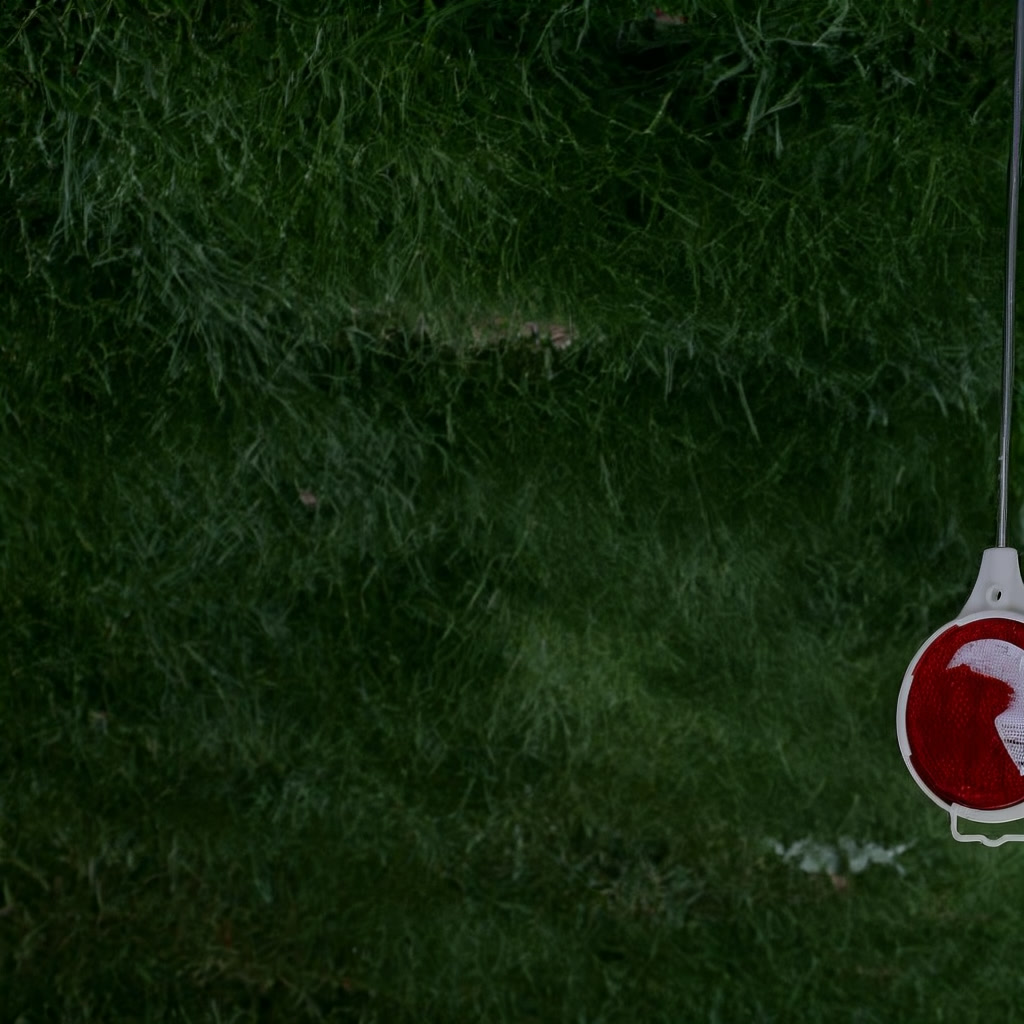}
\end{subfigure}
\hfill
\begin{subfigure}{\textwidth}
    \setzoom{0.35}{0.68}{0.79, 0.21}{yellow}
    \zoomimg{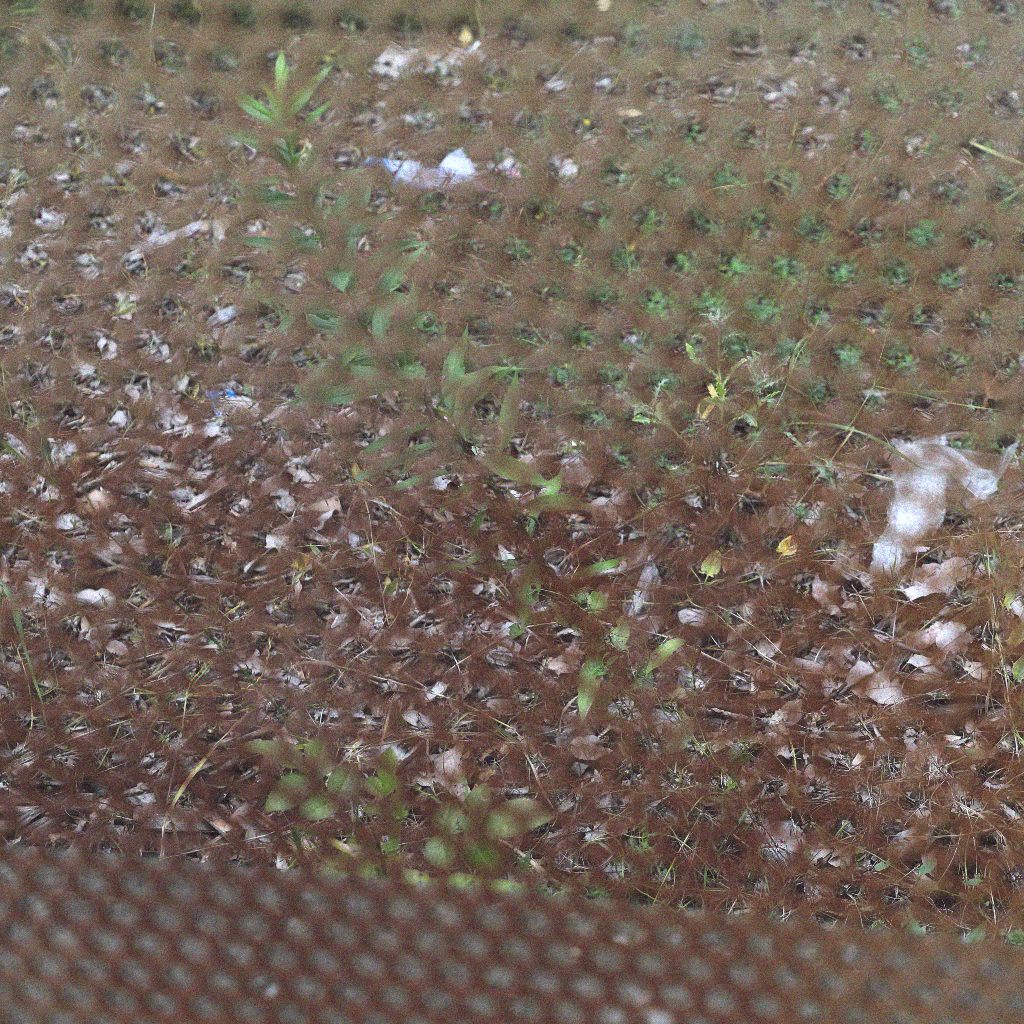}
    \zoomimg{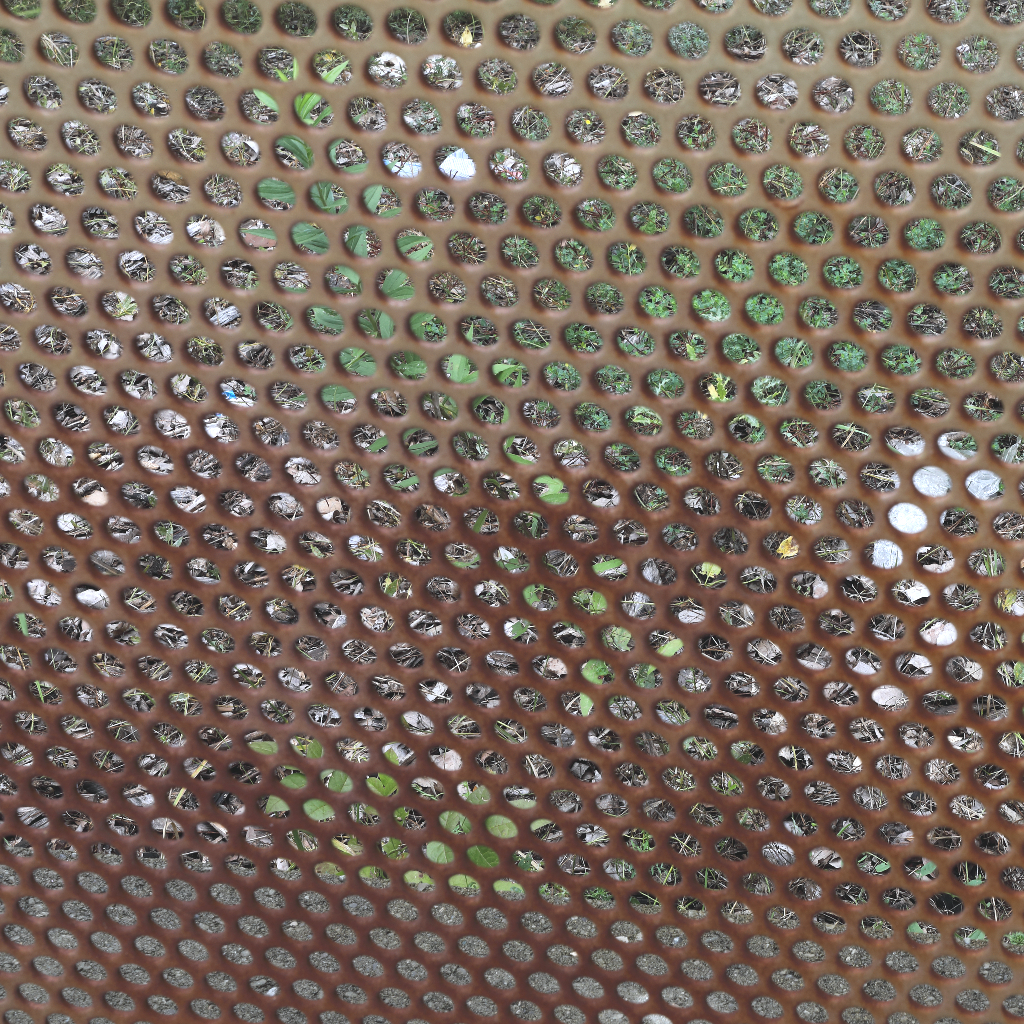}
    \zoomimg{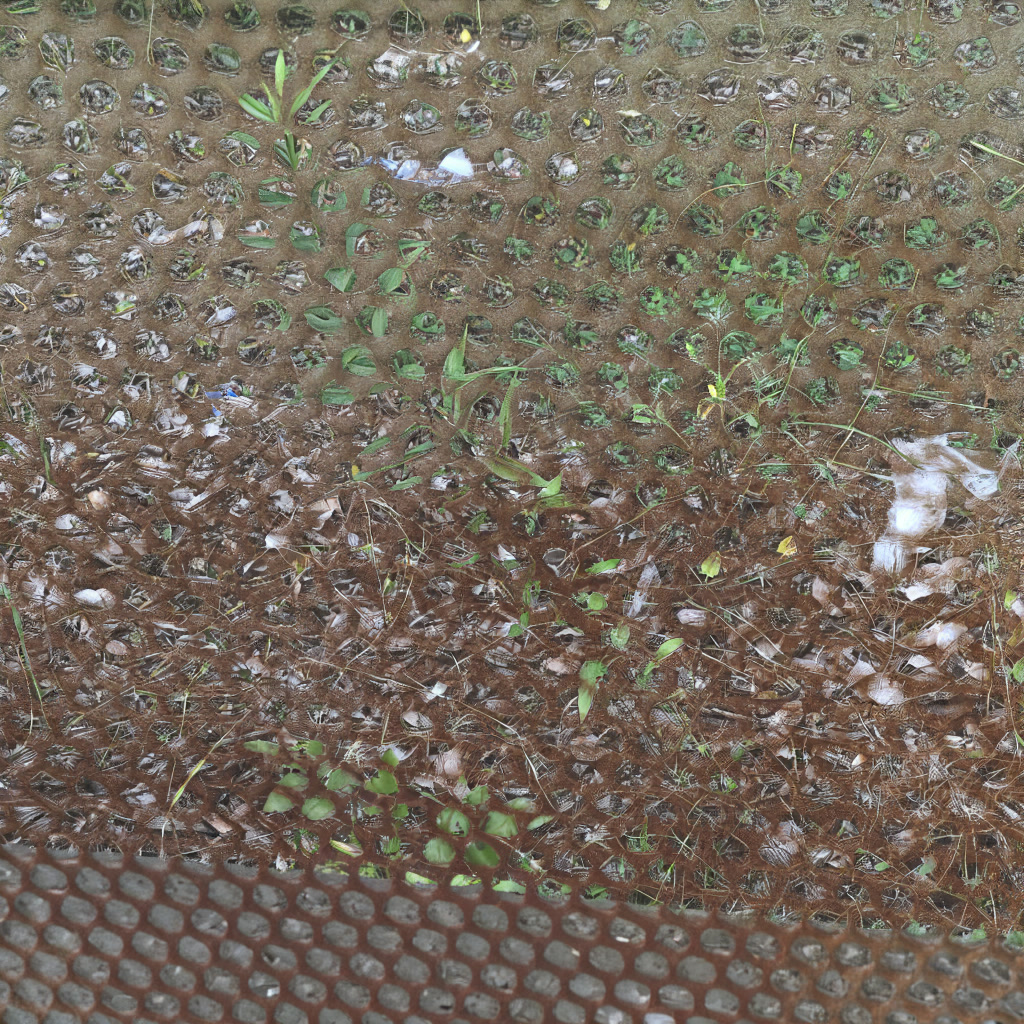}
    \zoomimg{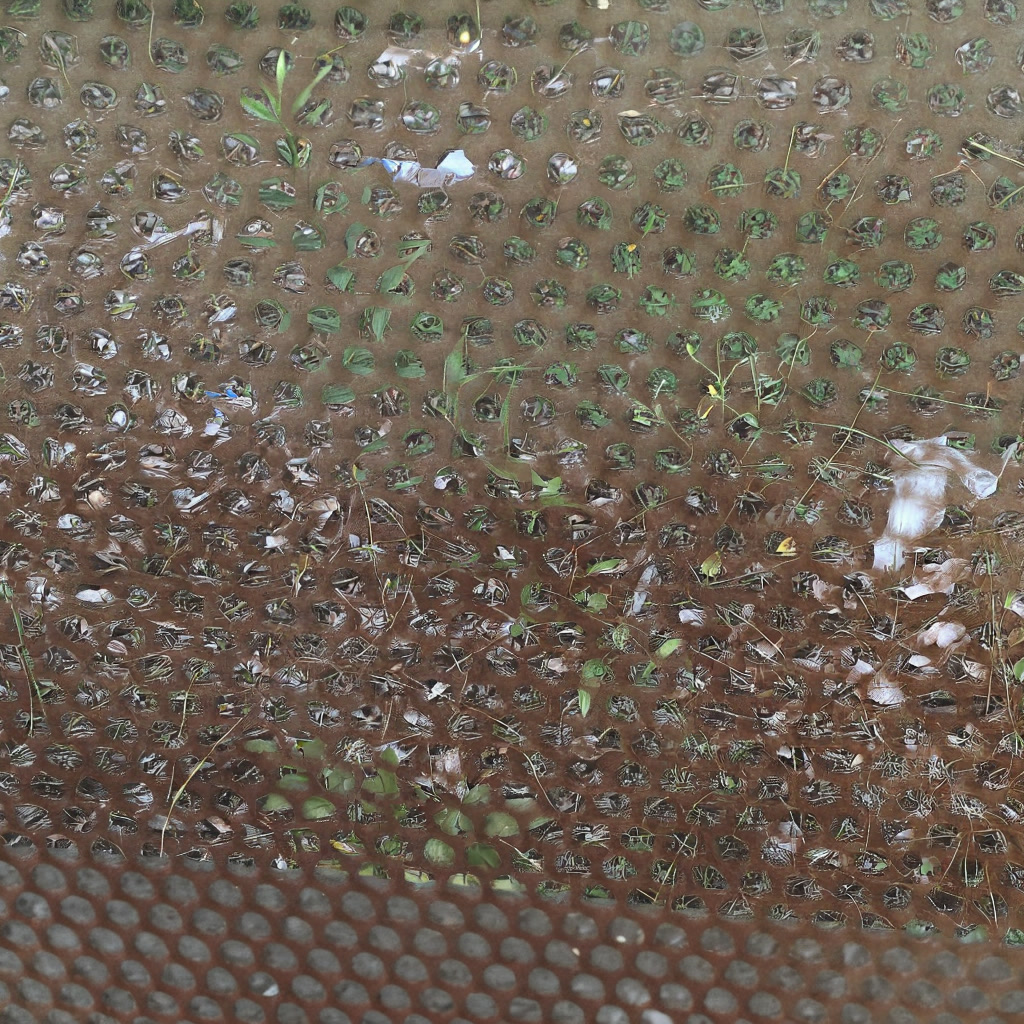}
    \zoomimg{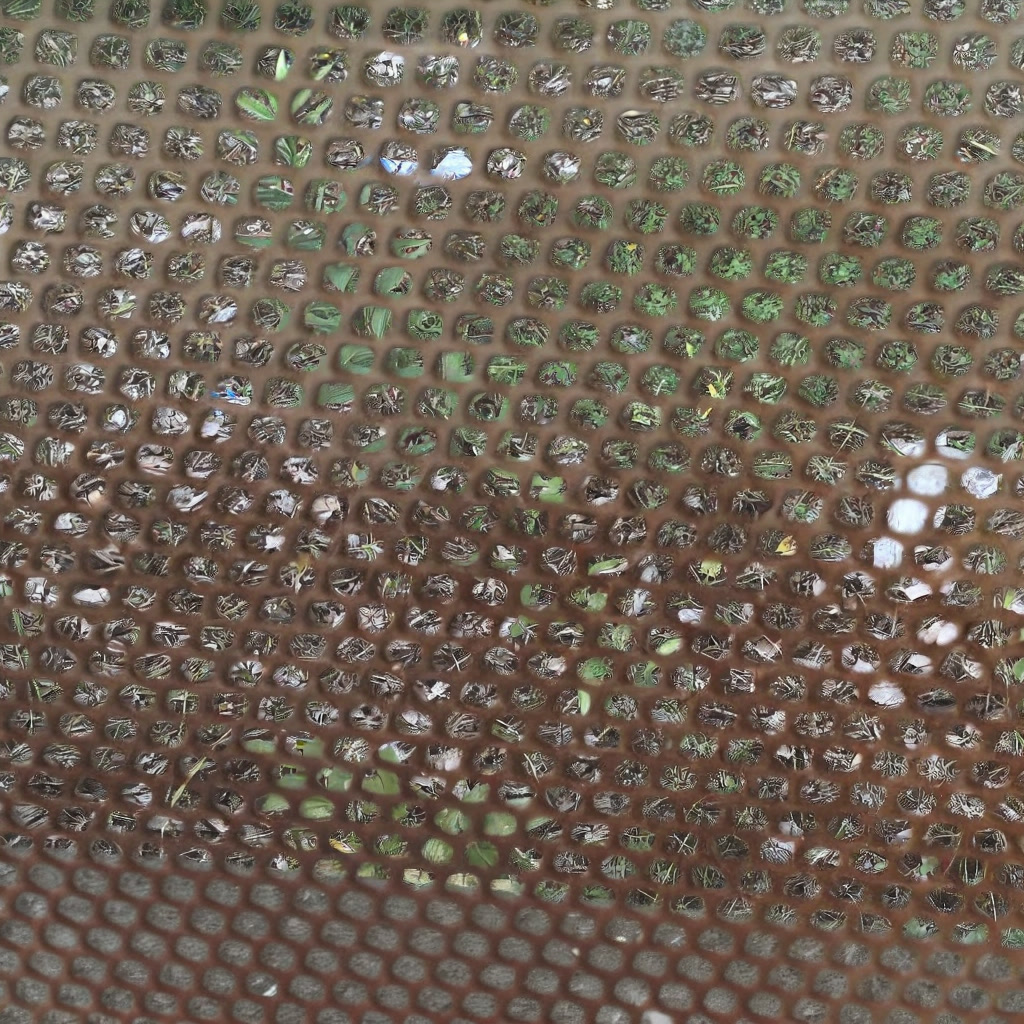}
\end{subfigure}
\hfill
\begin{subfigure}{\textwidth}
    \setzoom{0.50}{0.60}{0.79, 0.21}{yellow}
    \zoomimg[angle=90, origin=c]{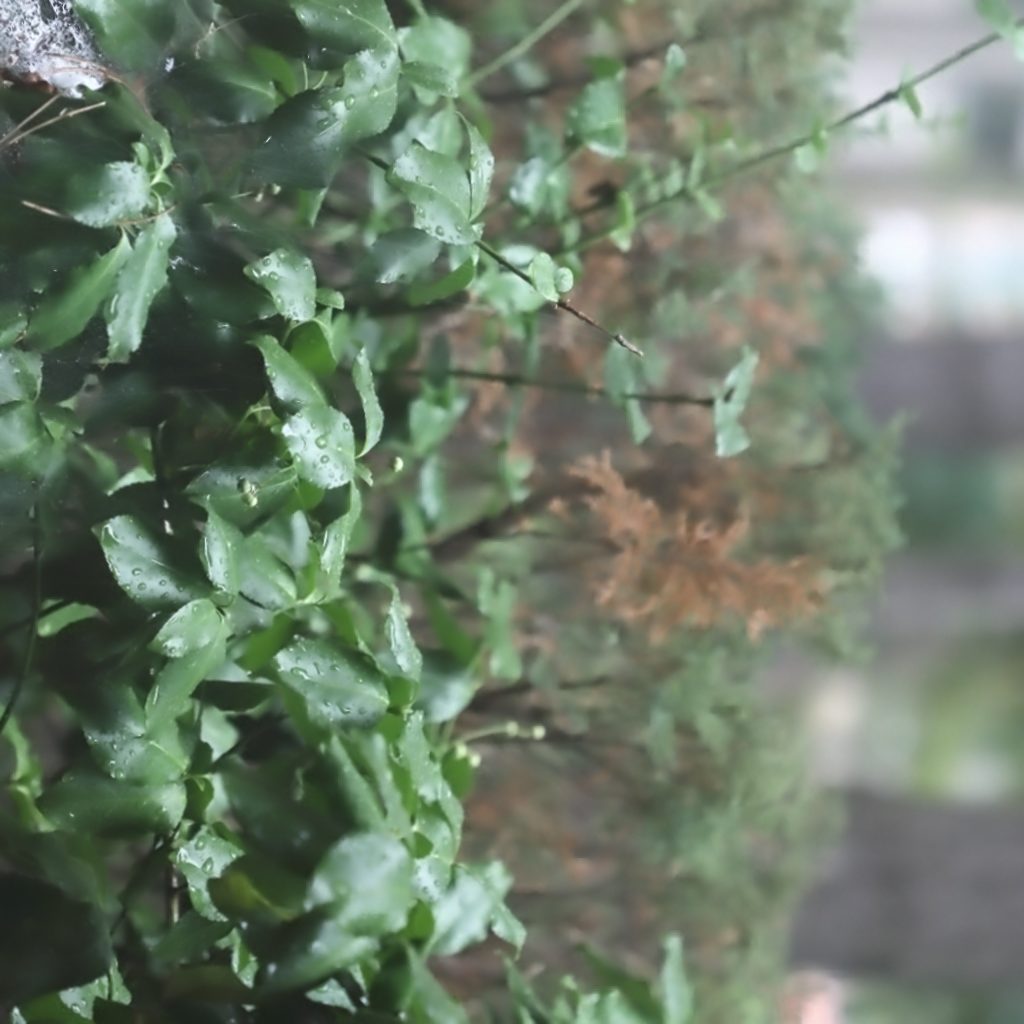}
    \zoomimg[angle=90, origin=c]{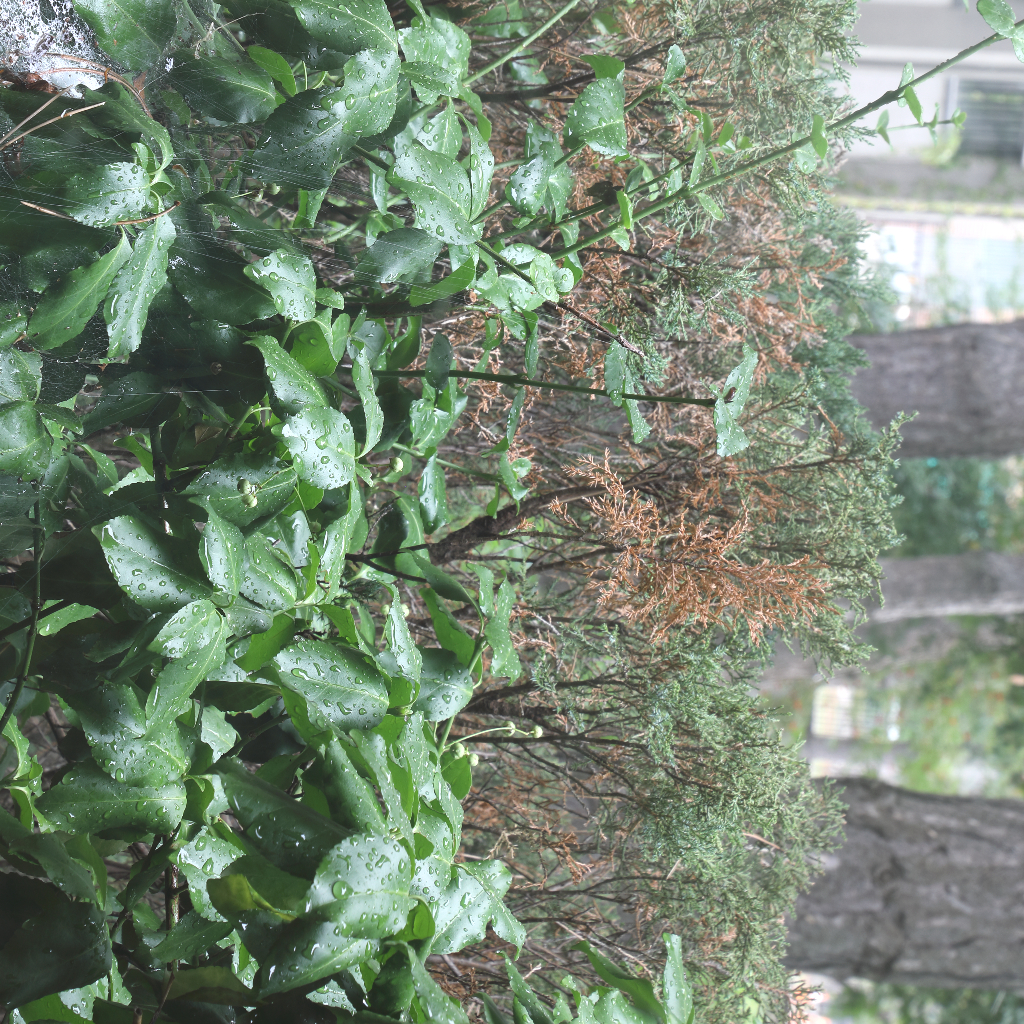}
    \zoomimg[angle=90, origin=c]{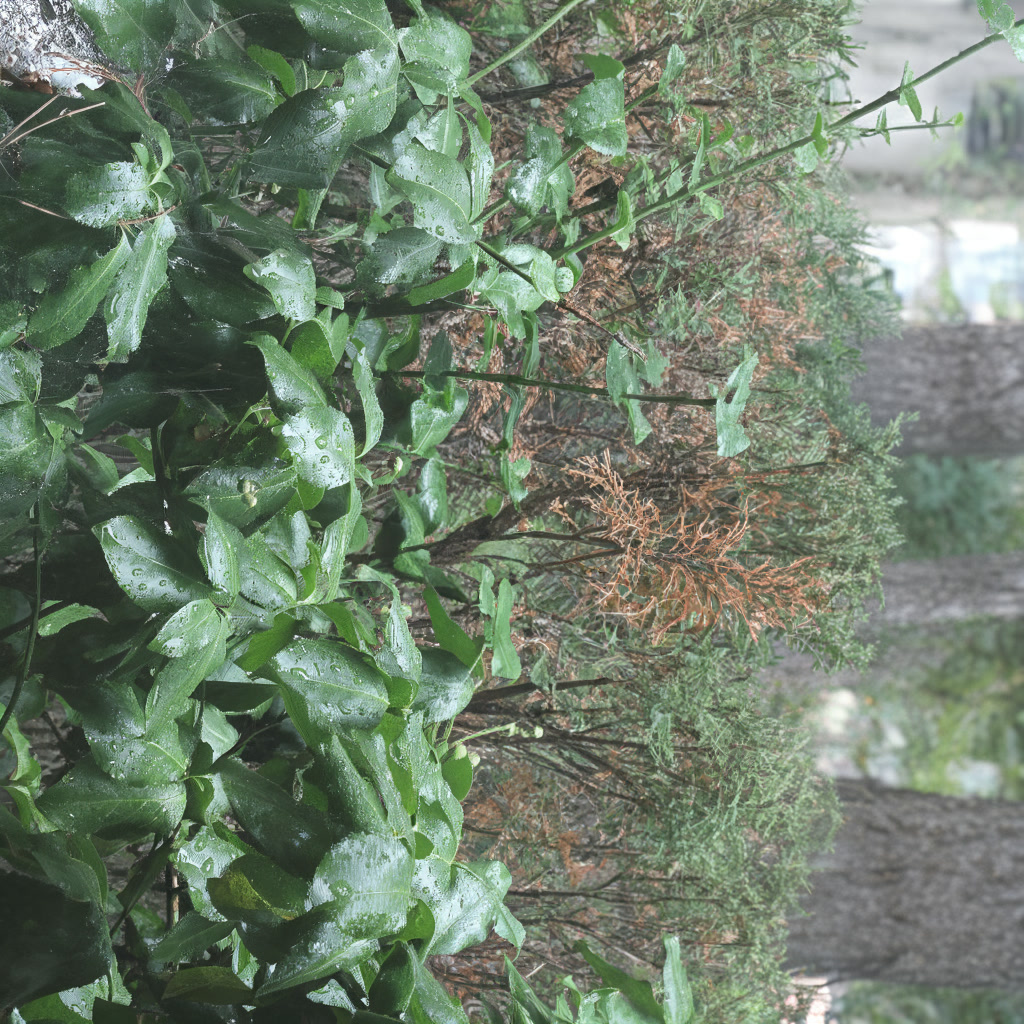}
    \zoomimg[angle=90, origin=c]{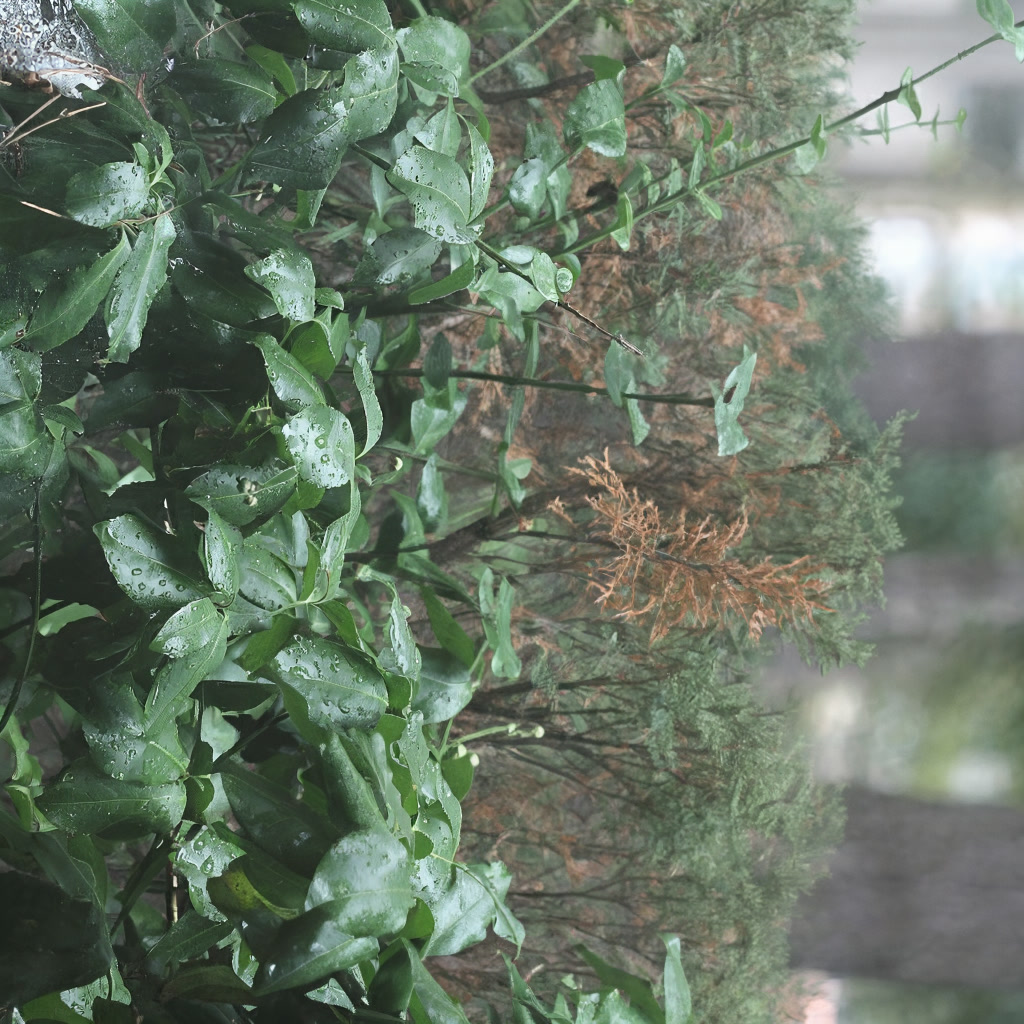}
    \zoomimg[angle=90, origin=c]{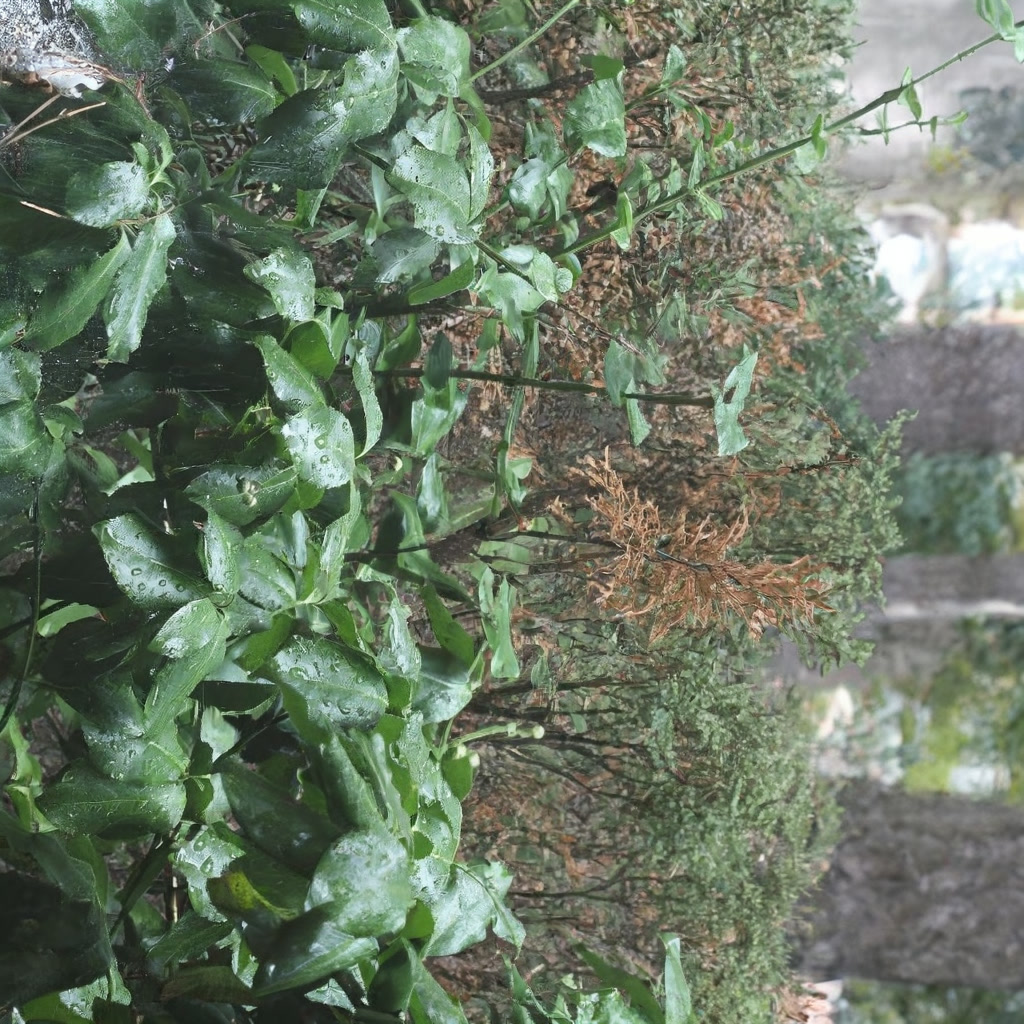}
\end{subfigure}
\hfill
\caption{\textbf{OID-DPDD model soup results with OptIMerge on DPDD data}. (left-to-right) We provide input, target, and images generated from the expert model, naive model soup, and OptIMerge.}
\label{fig:oid_dpdd_soup_on_dpdd_data}
\end{figure}

\begin{figure}[h]
\centering
\begin{subfigure}{\textwidth}
    \zHeaderRow
\end{subfigure}
\hfill
\begin{subfigure}{\textwidth}
    \setzoom{0.50}{0.70}{0.79, 0.21}{yellow}
    \zoomimg{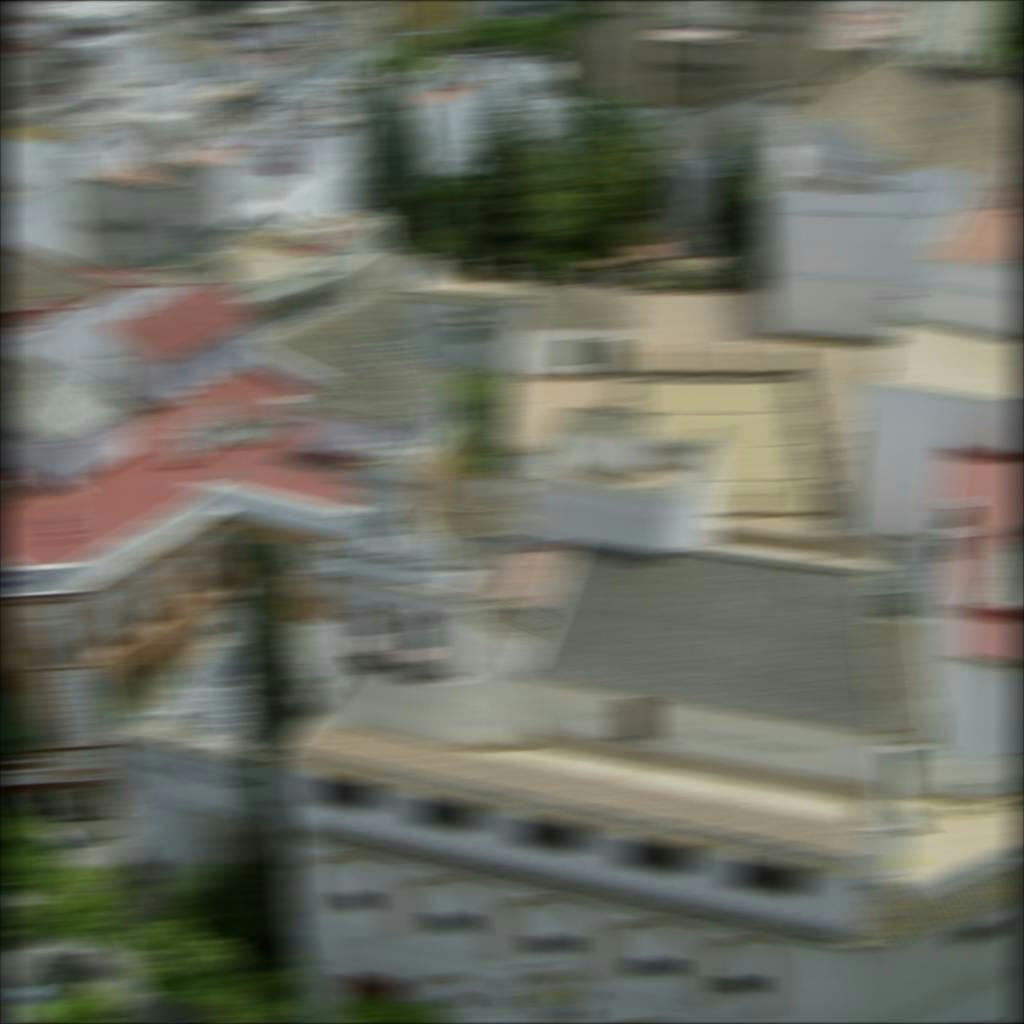}
    \zoomimg{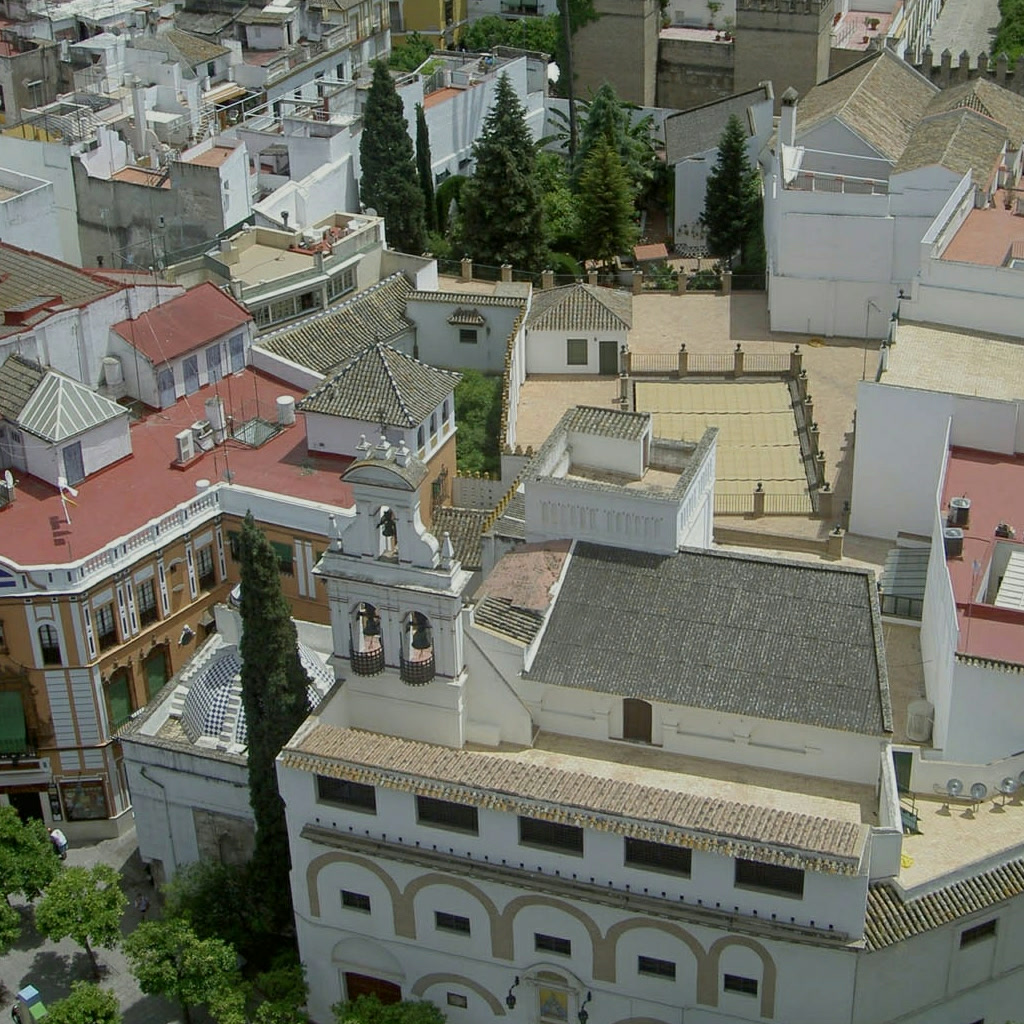}
    \zoomimg{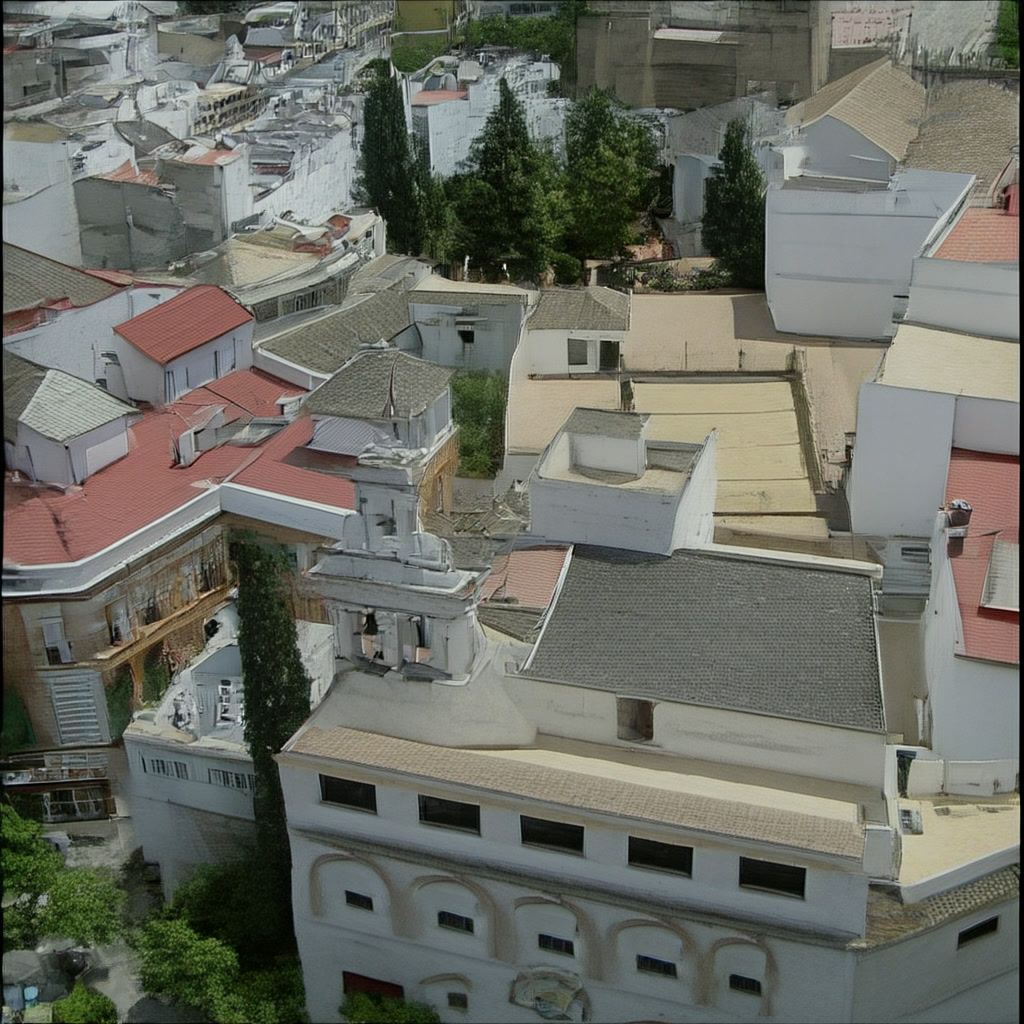}
    \zoomimg{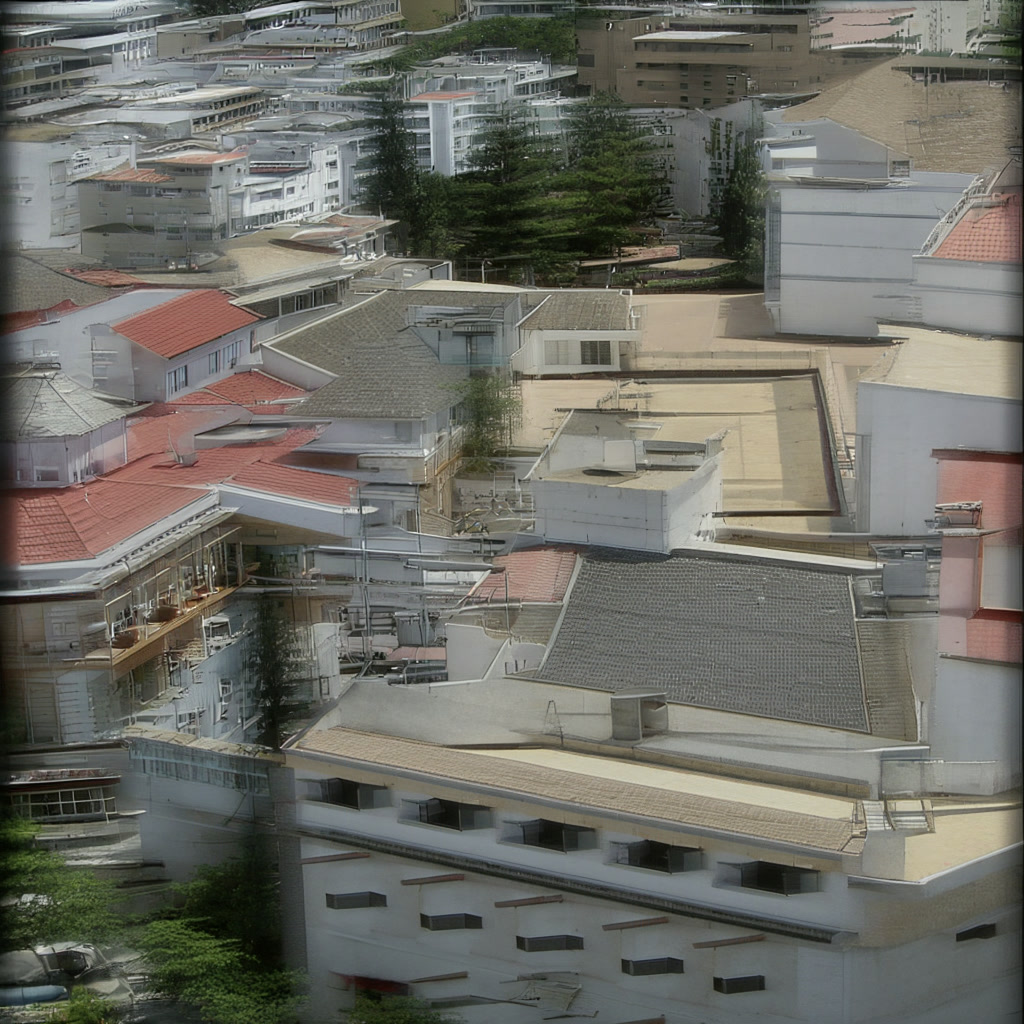}
    \zoomimg{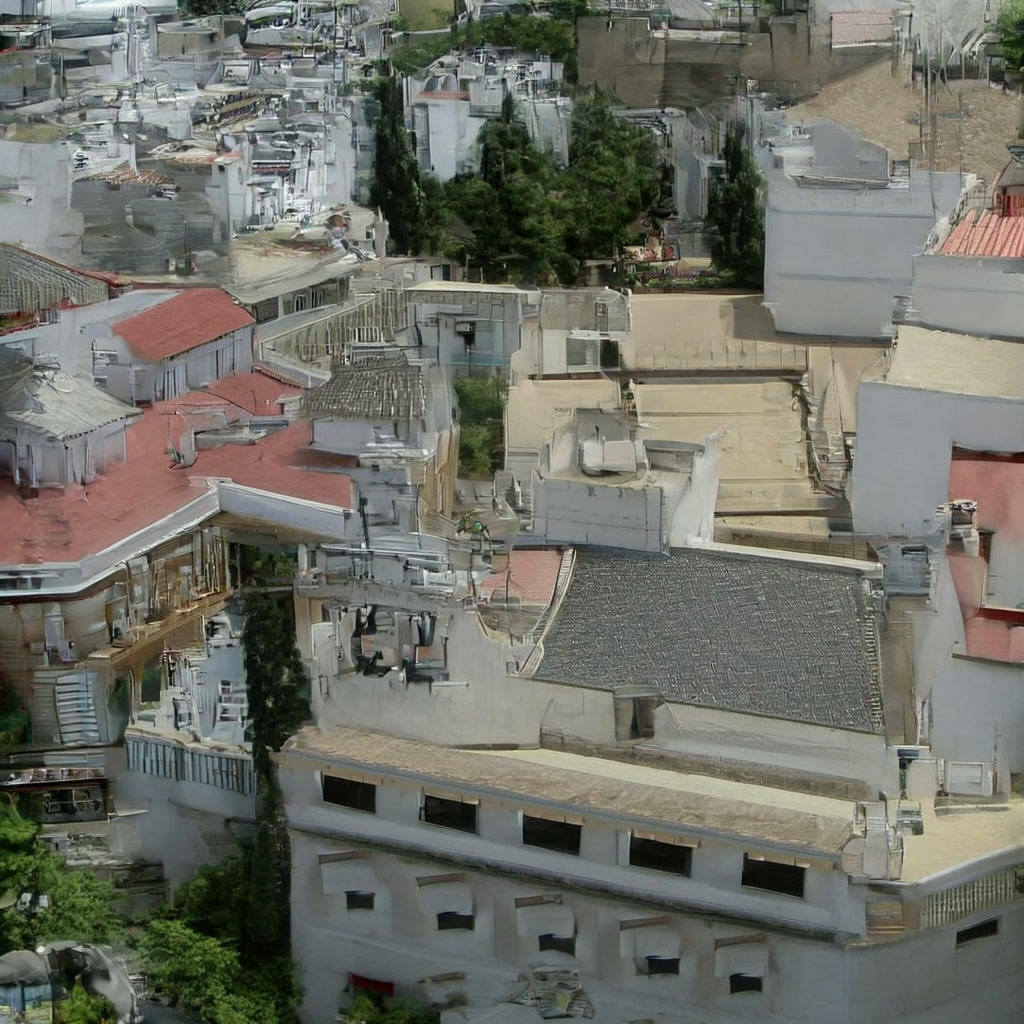}
\end{subfigure}
\hfill
\begin{subfigure}{\textwidth}
    \setzoom{0.60}{0.90}{0.79, 0.21}{yellow}
    \zoomimg{sr_oid_results/oid/2/input.jpeg}
    \zoomimg{sr_oid_results/oid/2/target.jpeg}
    \zoomimg{sr_oid_results/oid/2/expert.jpeg}
    \zoomimg{sr_oid_results/oid/2/naive_soup.jpeg}
    \zoomimg{sr_oid_results/oid/2/optmerge.jpeg}
\end{subfigure}
\hfill
\begin{subfigure}{\textwidth}
    \setzoom{0.10}{0.90}{0.79, 0.21}{yellow}
    \zoomimg{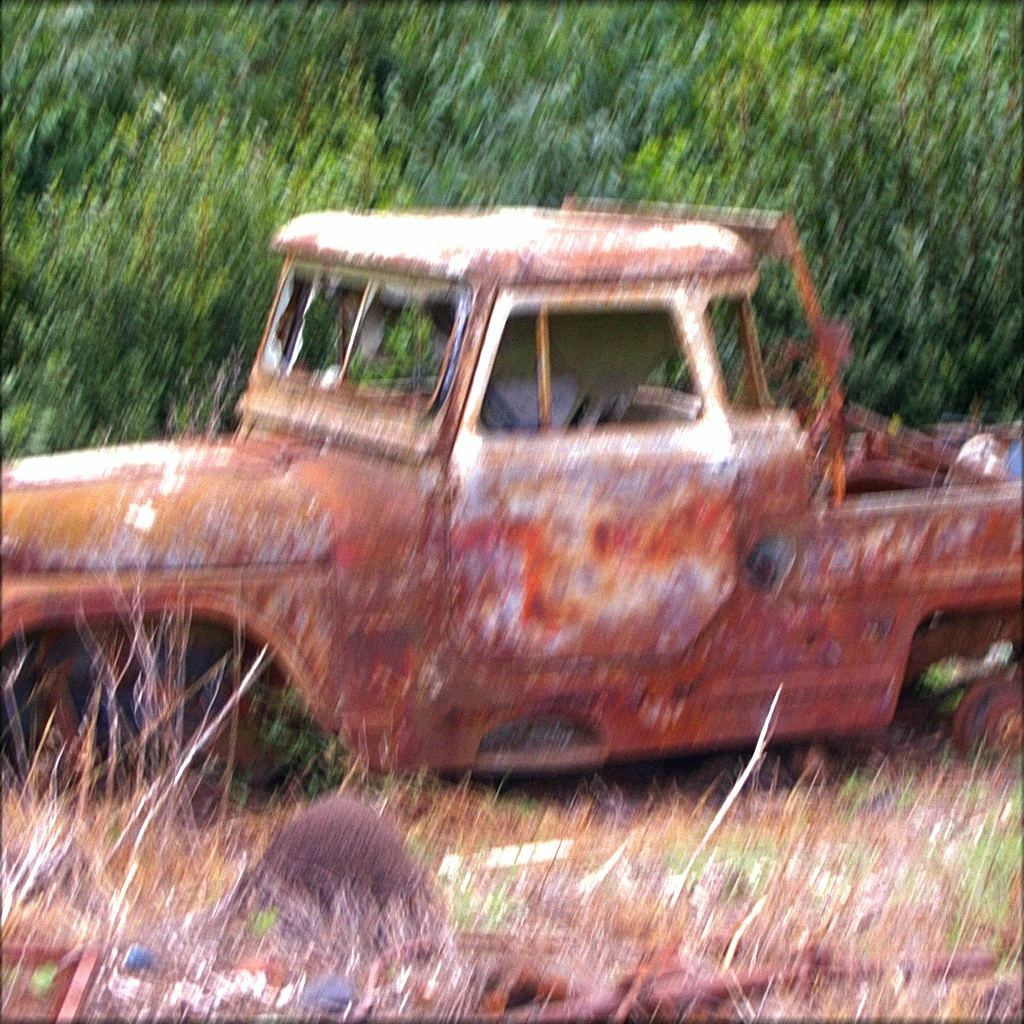}
    \zoomimg{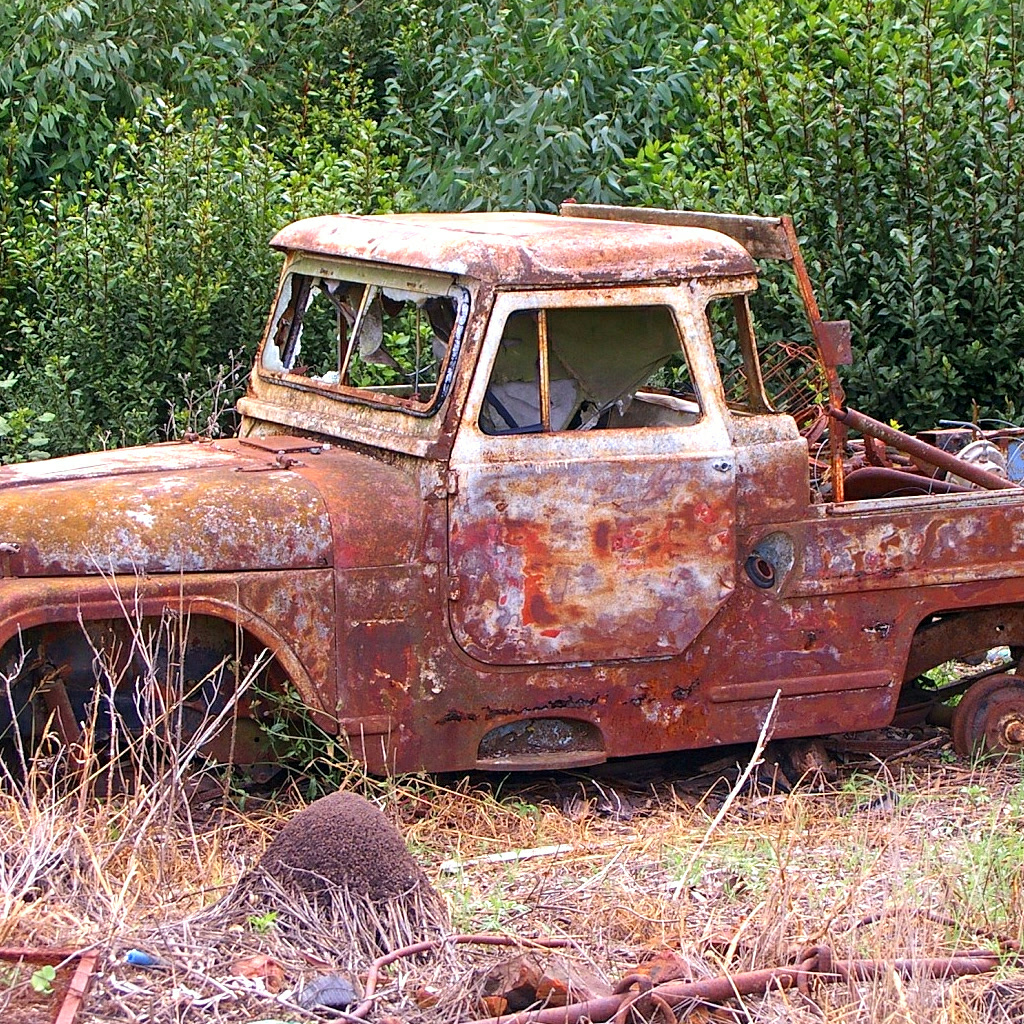}
    \zoomimg{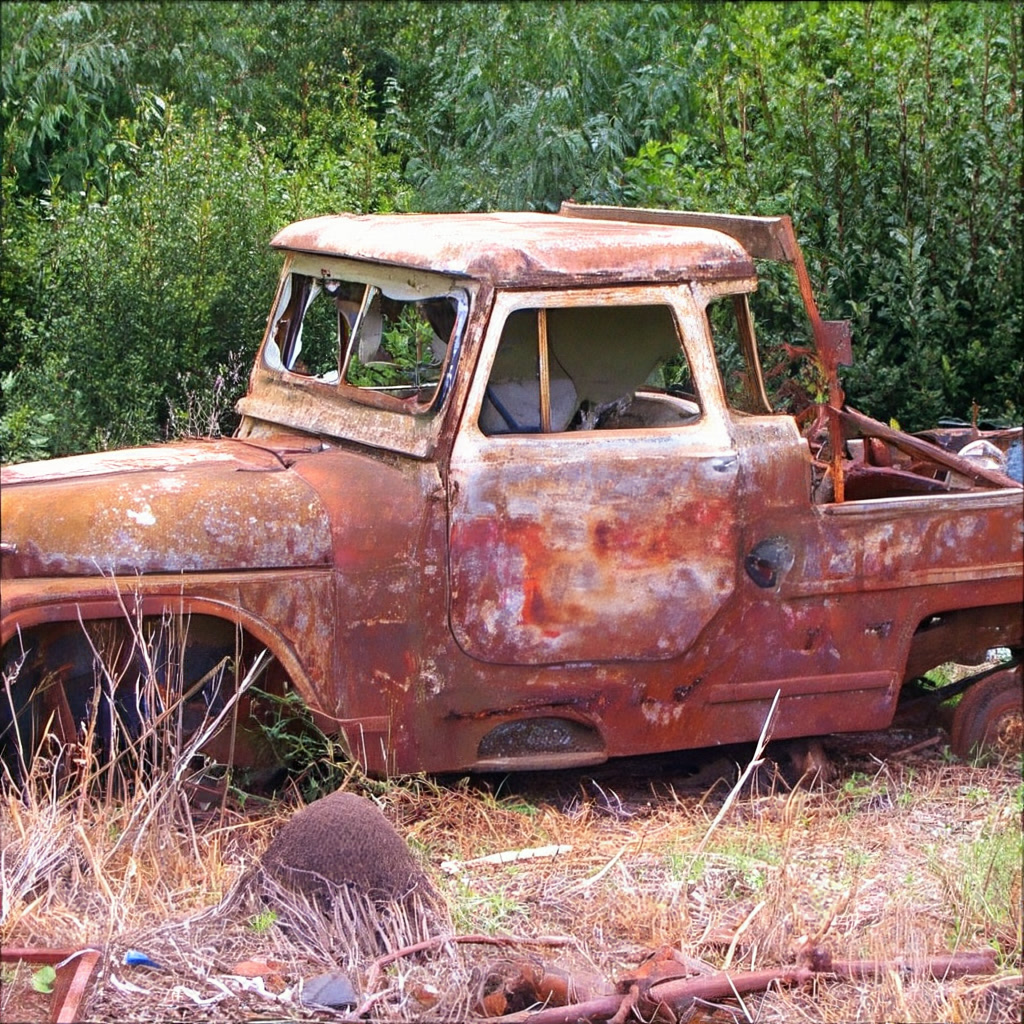}
    \zoomimg{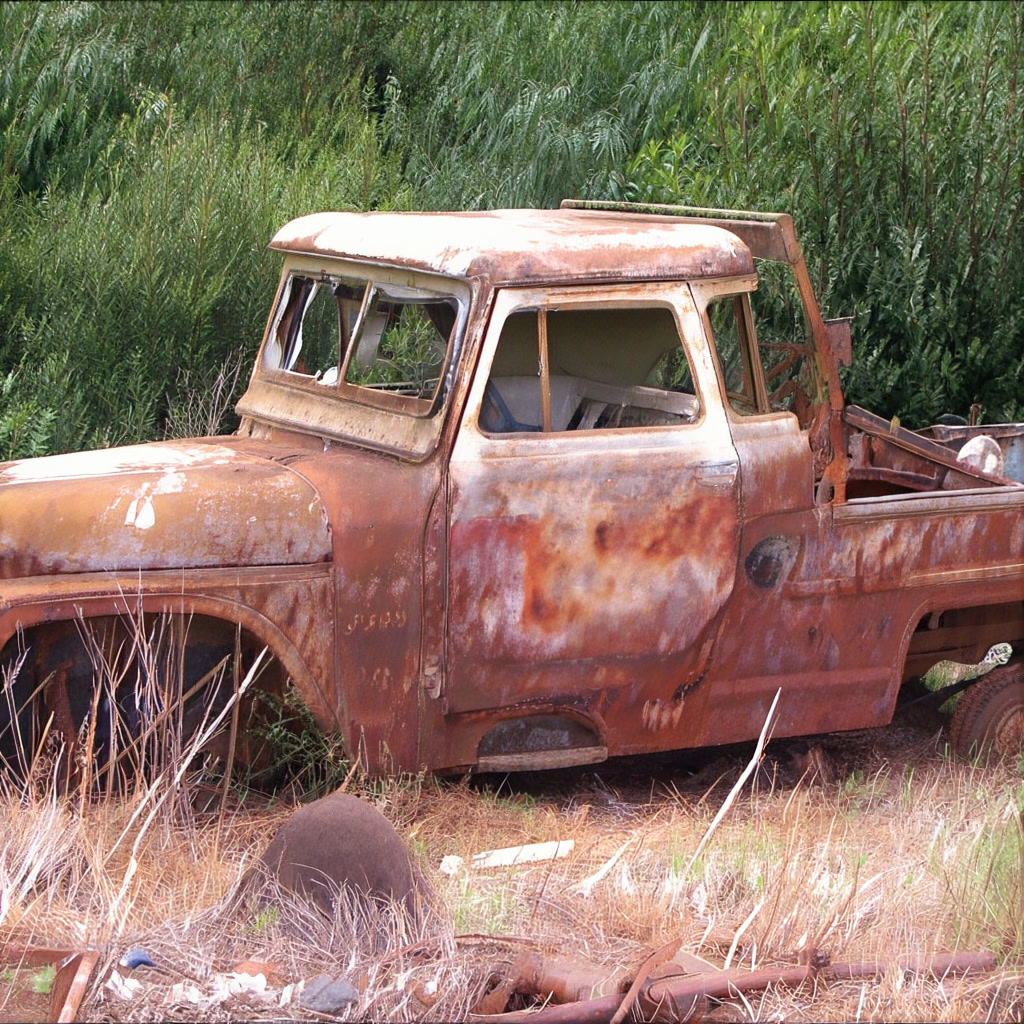}
    \zoomimg{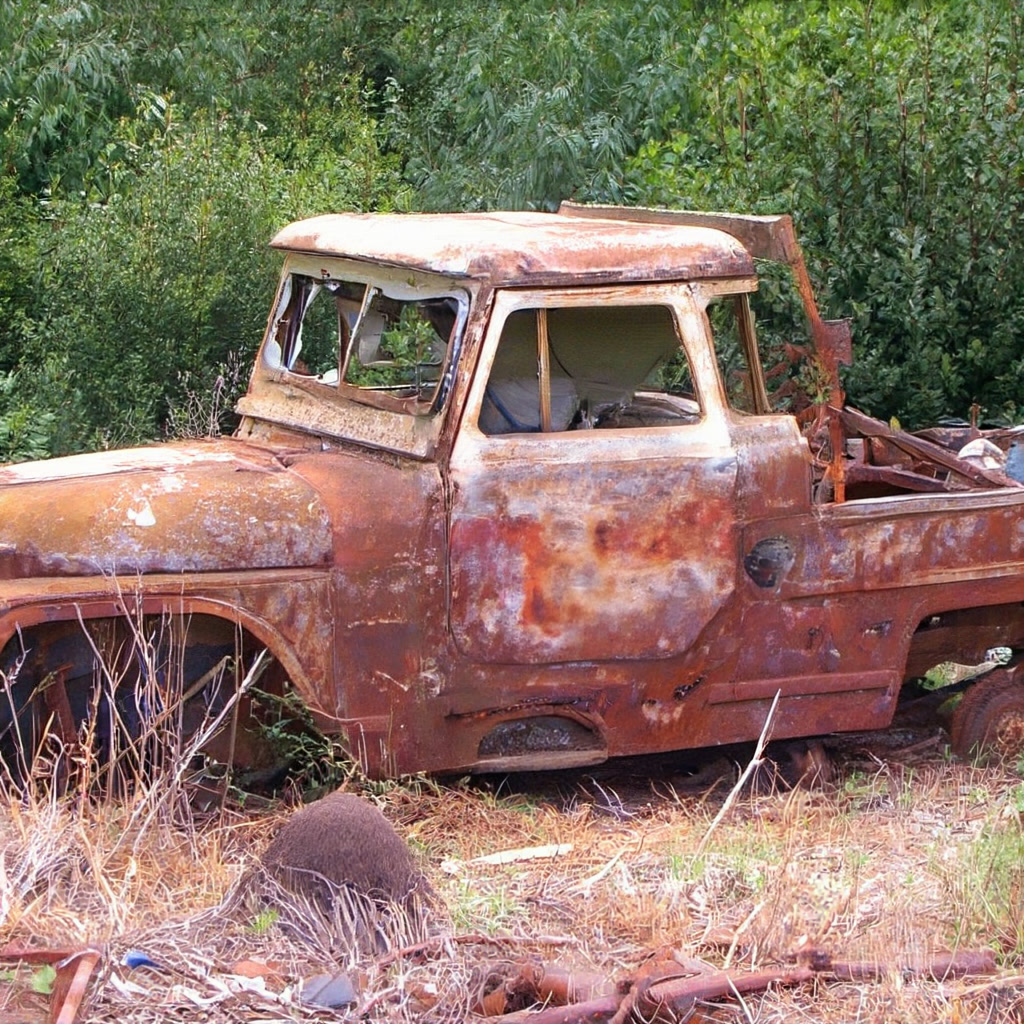}
\end{subfigure}
\caption{\textbf{OID-SR model soup results with OptIMerge on OID data}. (left-to-right) We provide input, target, and images generated from the expert model, naive model soup, and OptIMerge}
\label{fig:sr_oid_soup_on_oid_data}
\end{figure}

\begin{figure}[h]
\centering
\begin{subfigure}{\textwidth}
    \zHeaderRow
\end{subfigure}
\hfill
\begin{subfigure}{\textwidth}
    \setzoom{0.50}{0.90}{0.79, 0.21}{yellow}
    \zoomimg{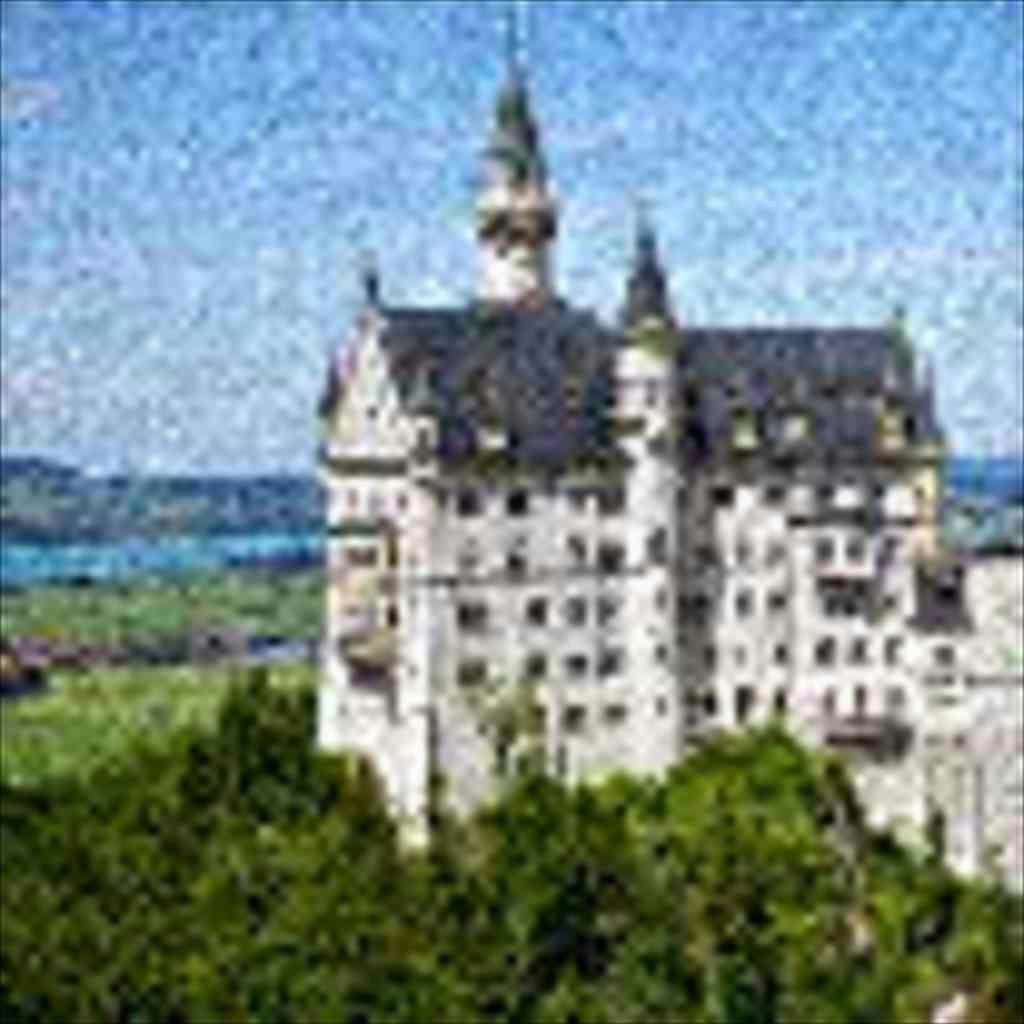}
    \zoomimg{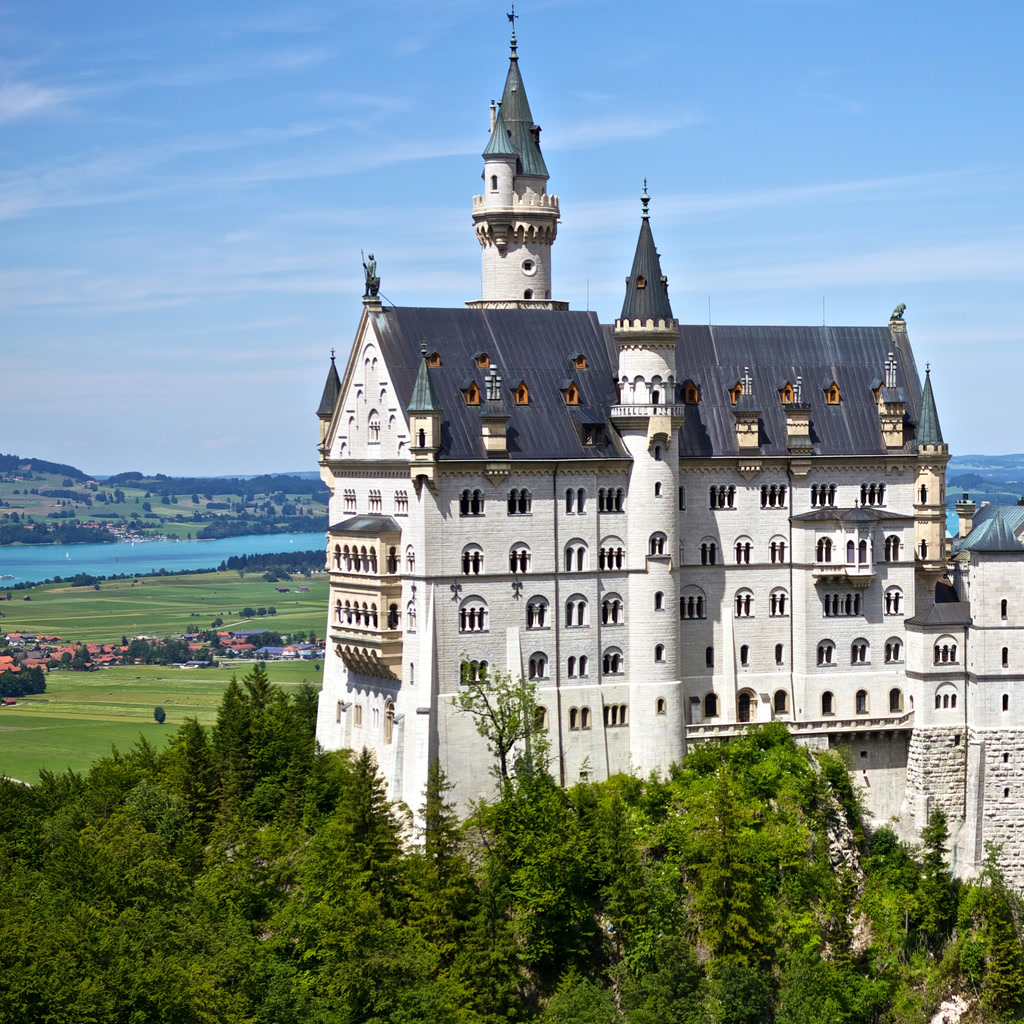}
    \zoomimg{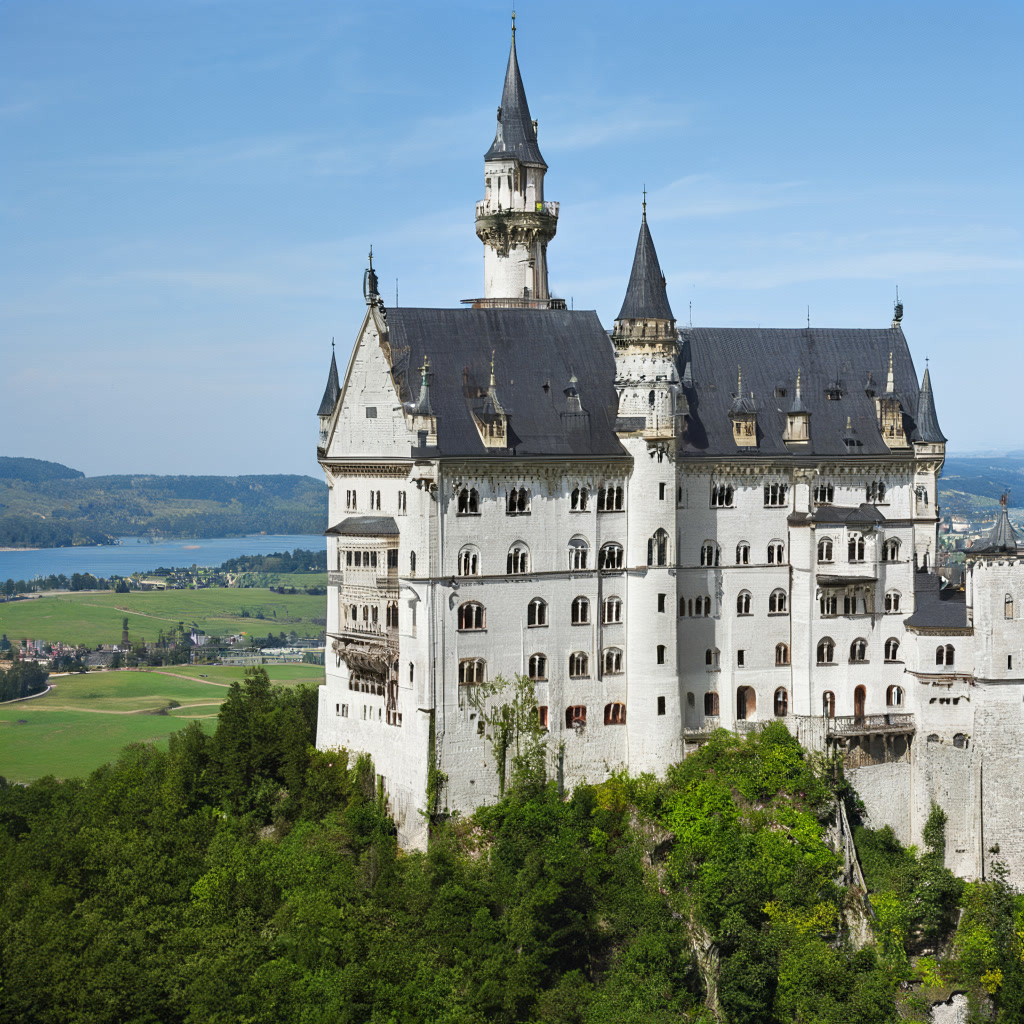}
    \zoomimg{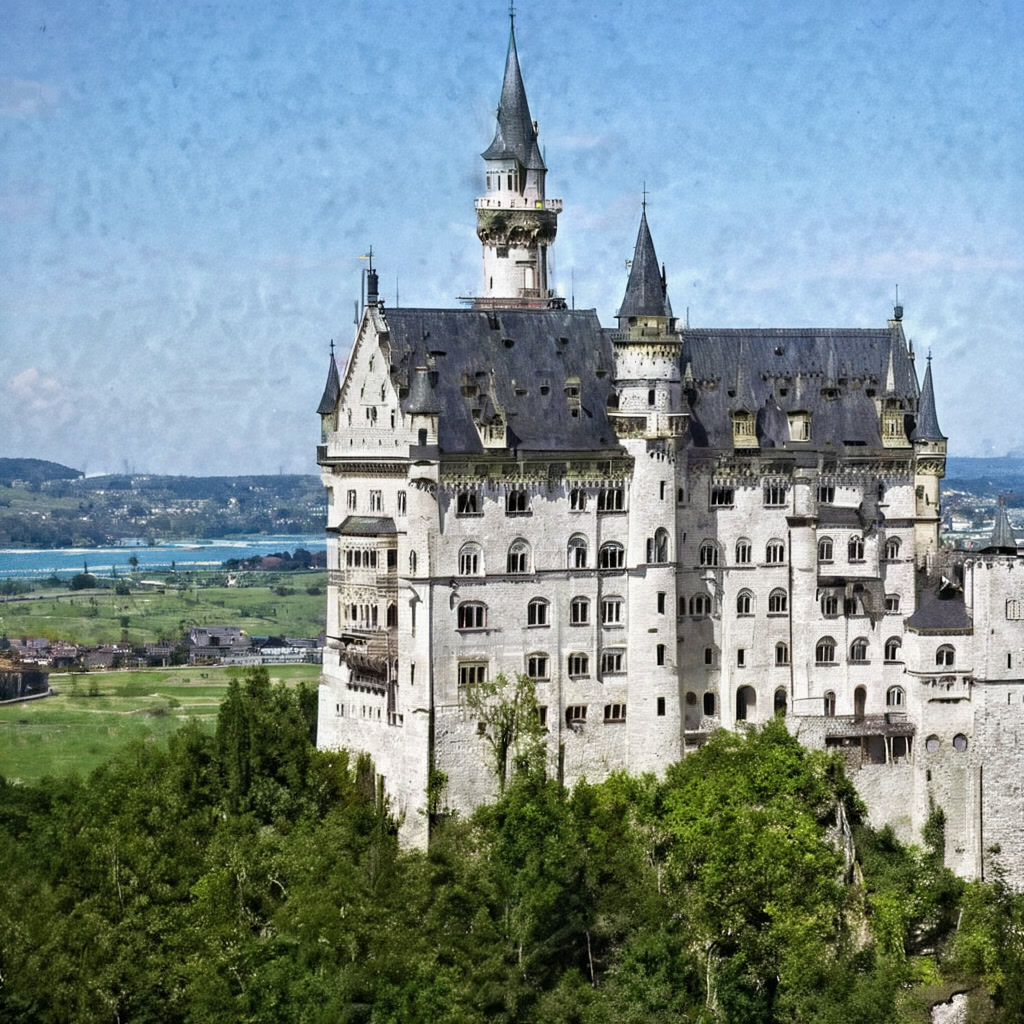}
    \zoomimg{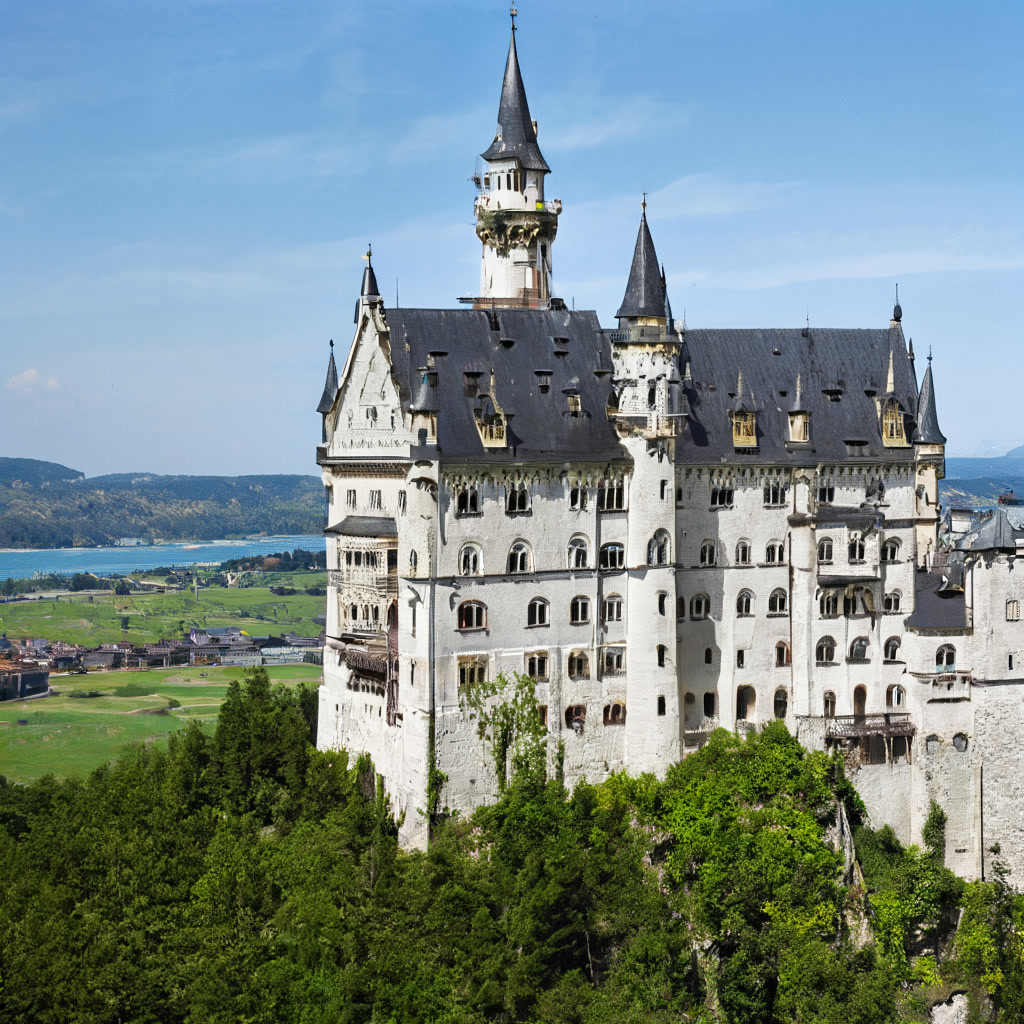}
\end{subfigure}
\hfill
\begin{subfigure}{\textwidth}
    \setzoom{0.50}{0.50}{0.79, 0.21}{yellow}
    \zoomimg{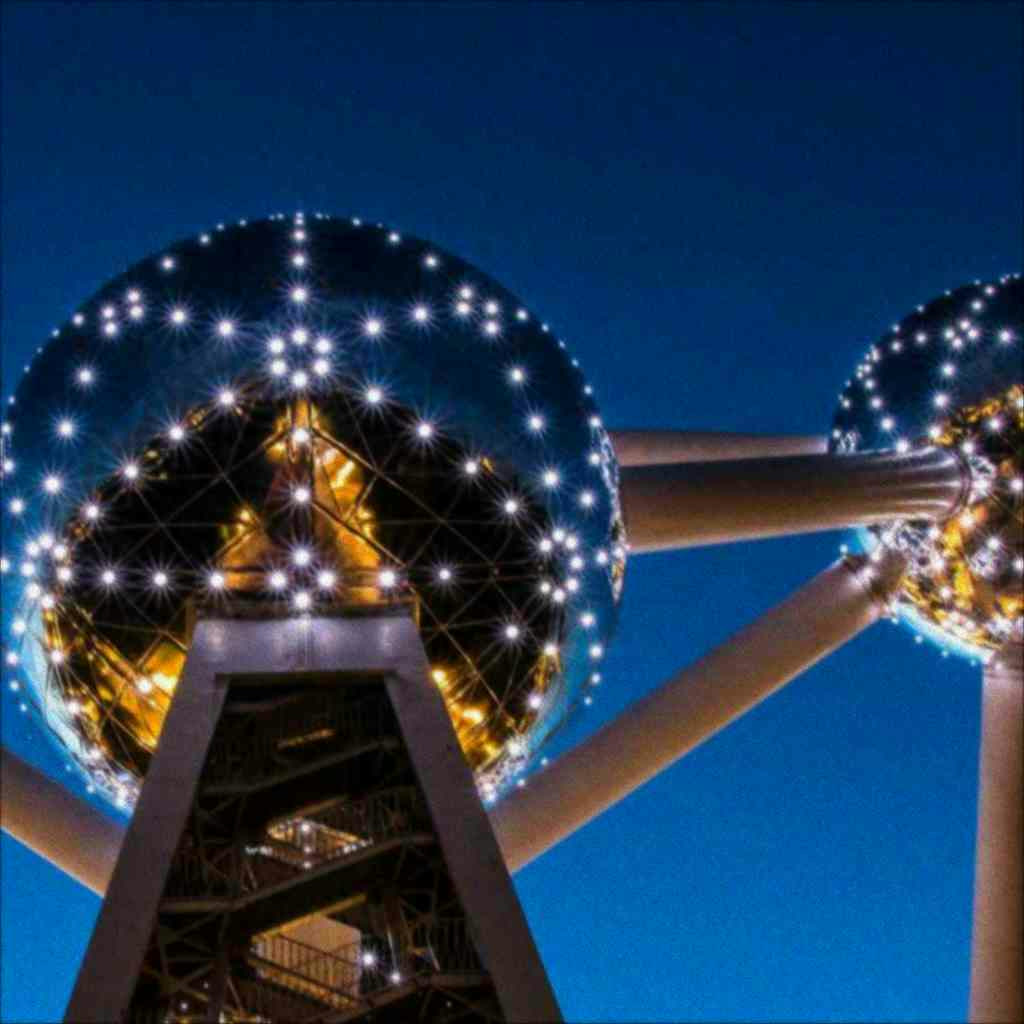}
    \zoomimg{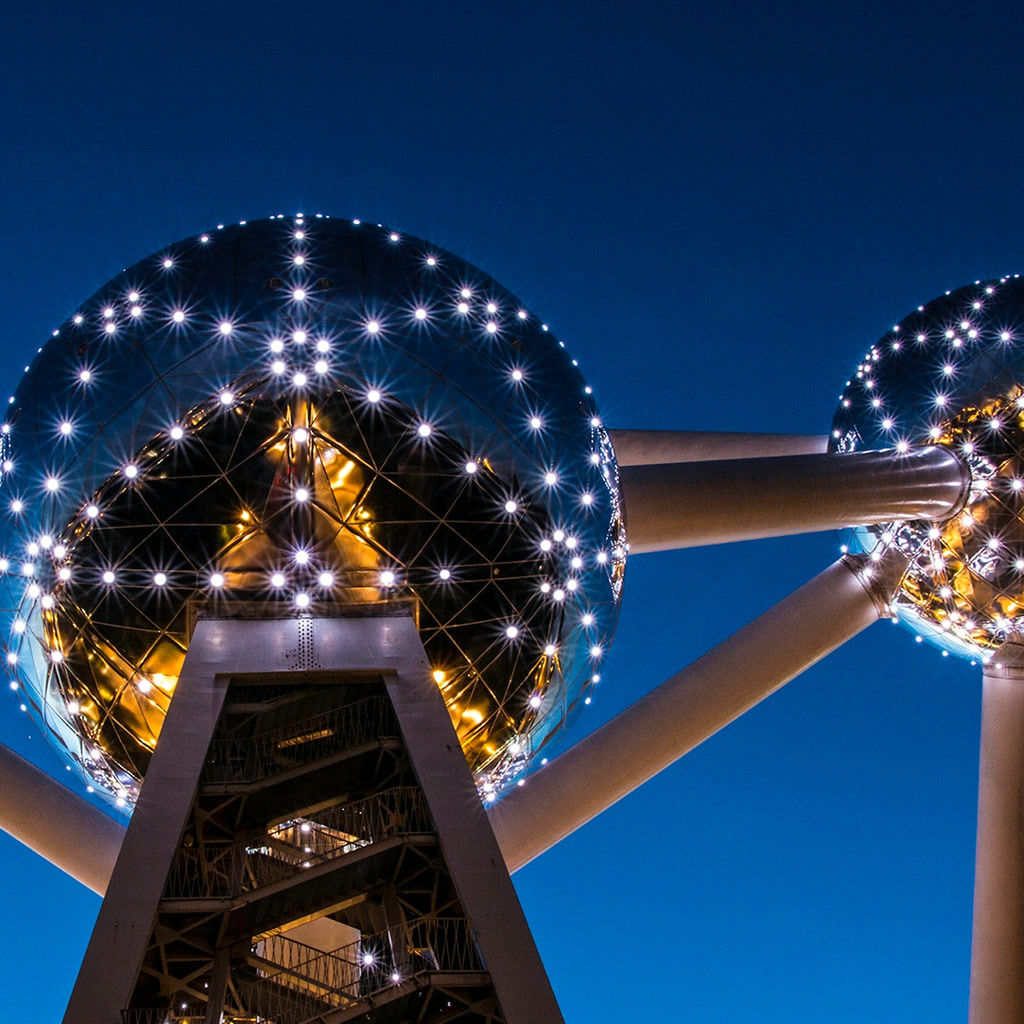}
    \zoomimg{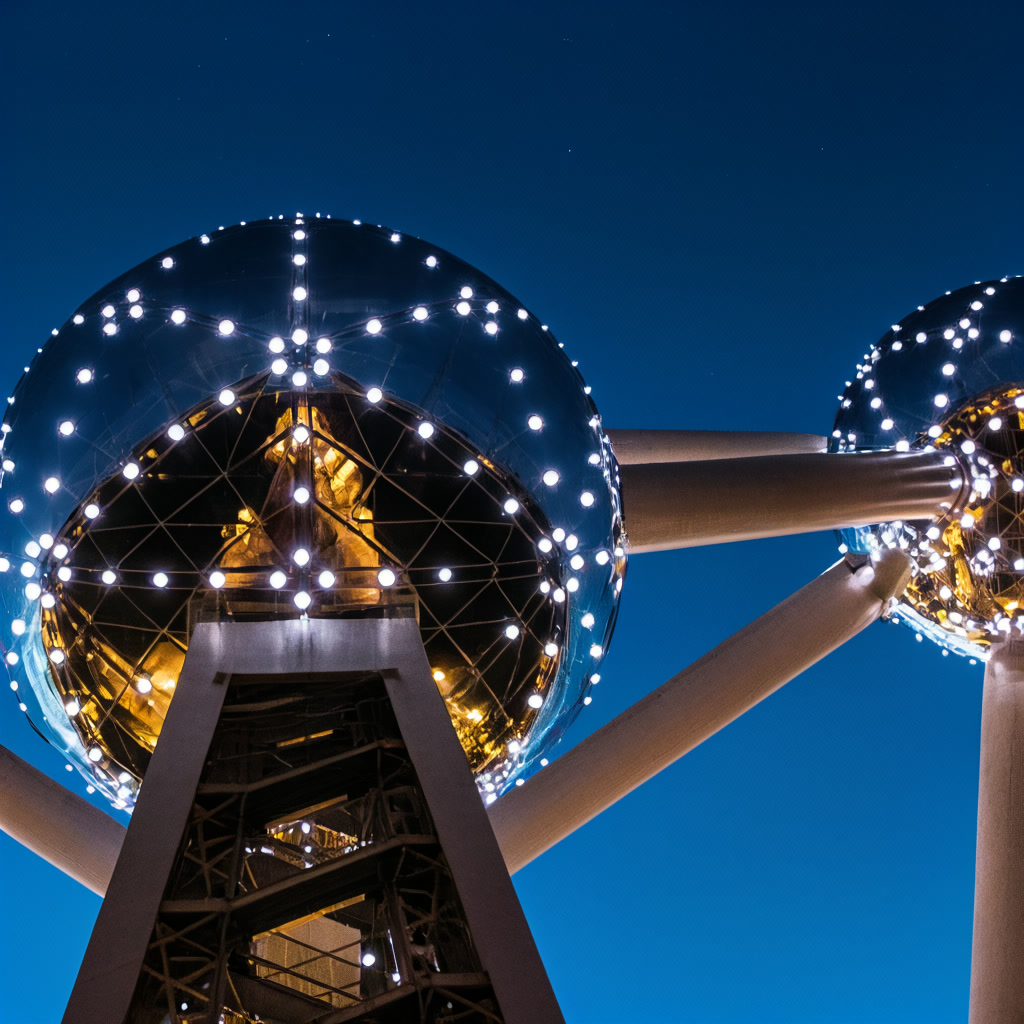}
    \zoomimg{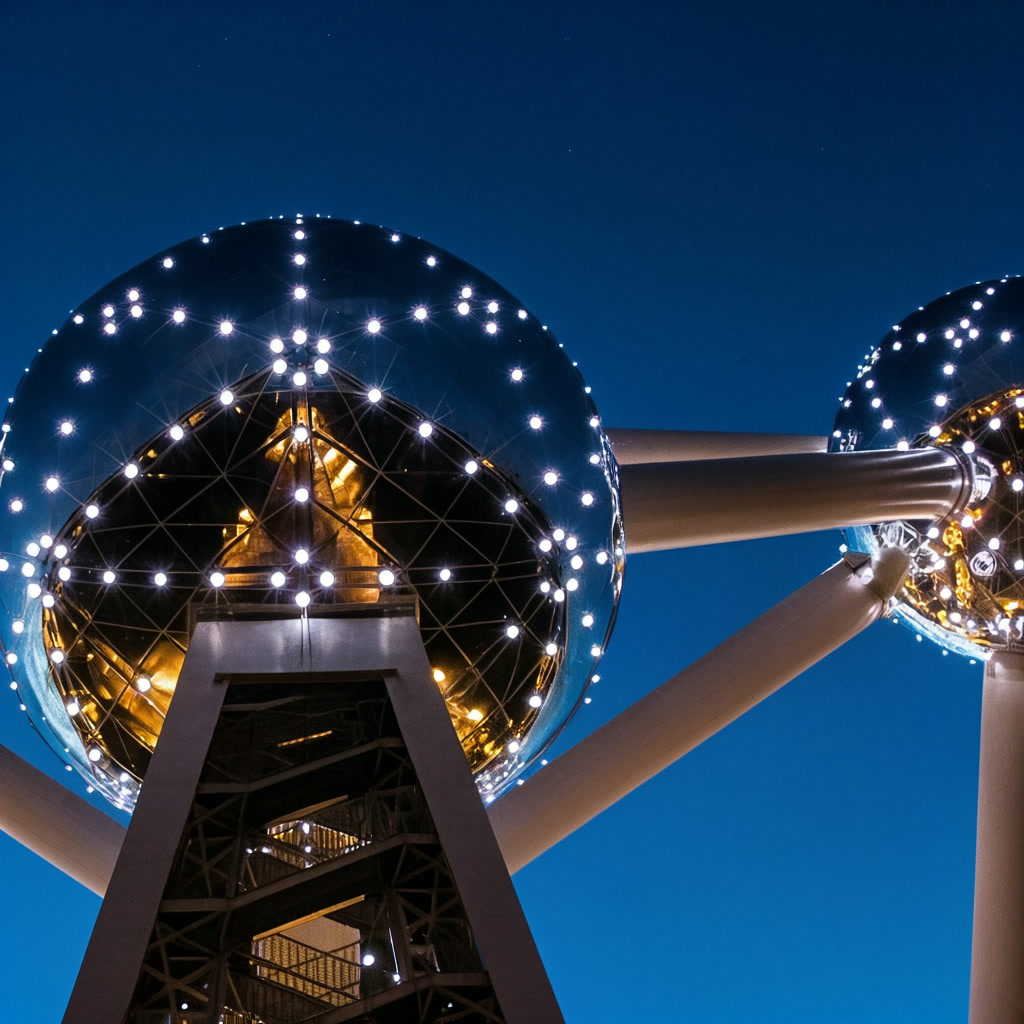}
    \zoomimg{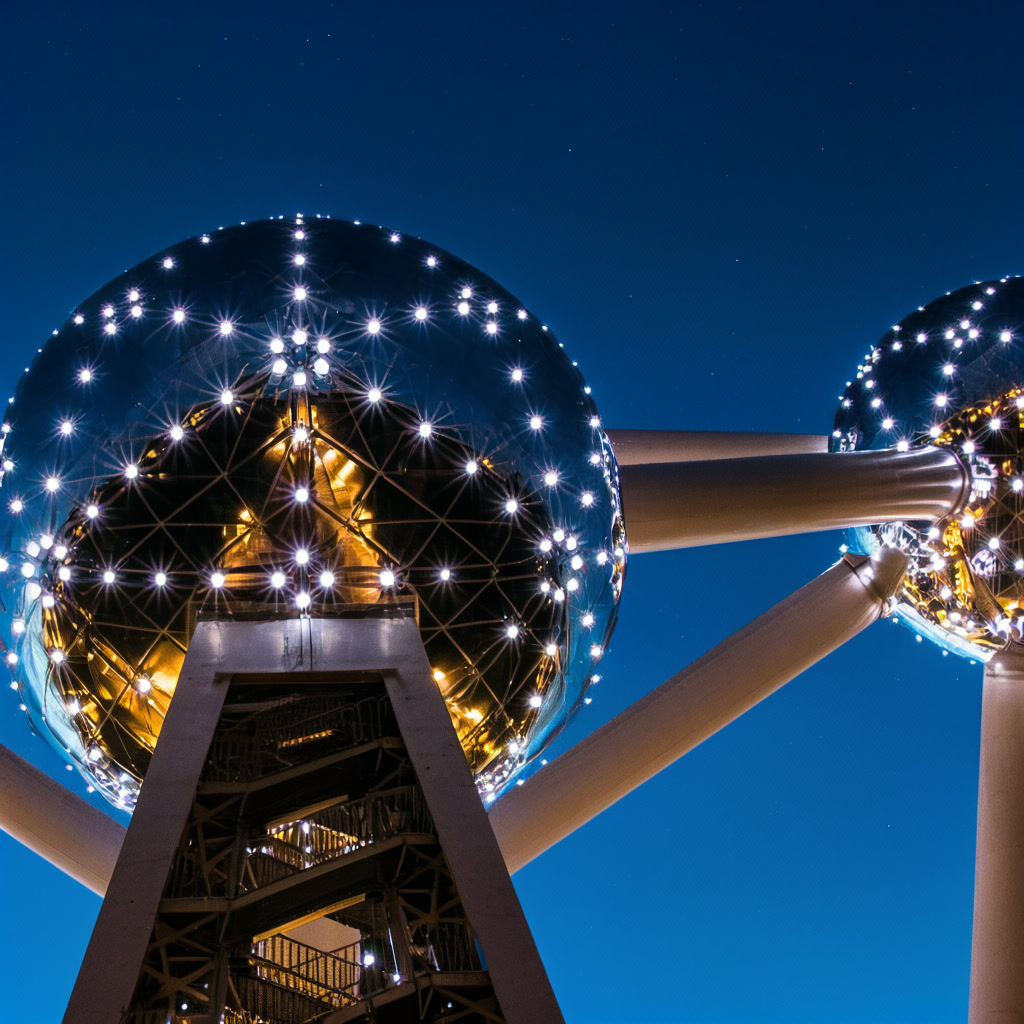}
\end{subfigure}
\hfill
\begin{subfigure}{\textwidth}
    \setzoom{0.40}{0.40}{0.79, 0.21}{yellow}
    \zoomimg{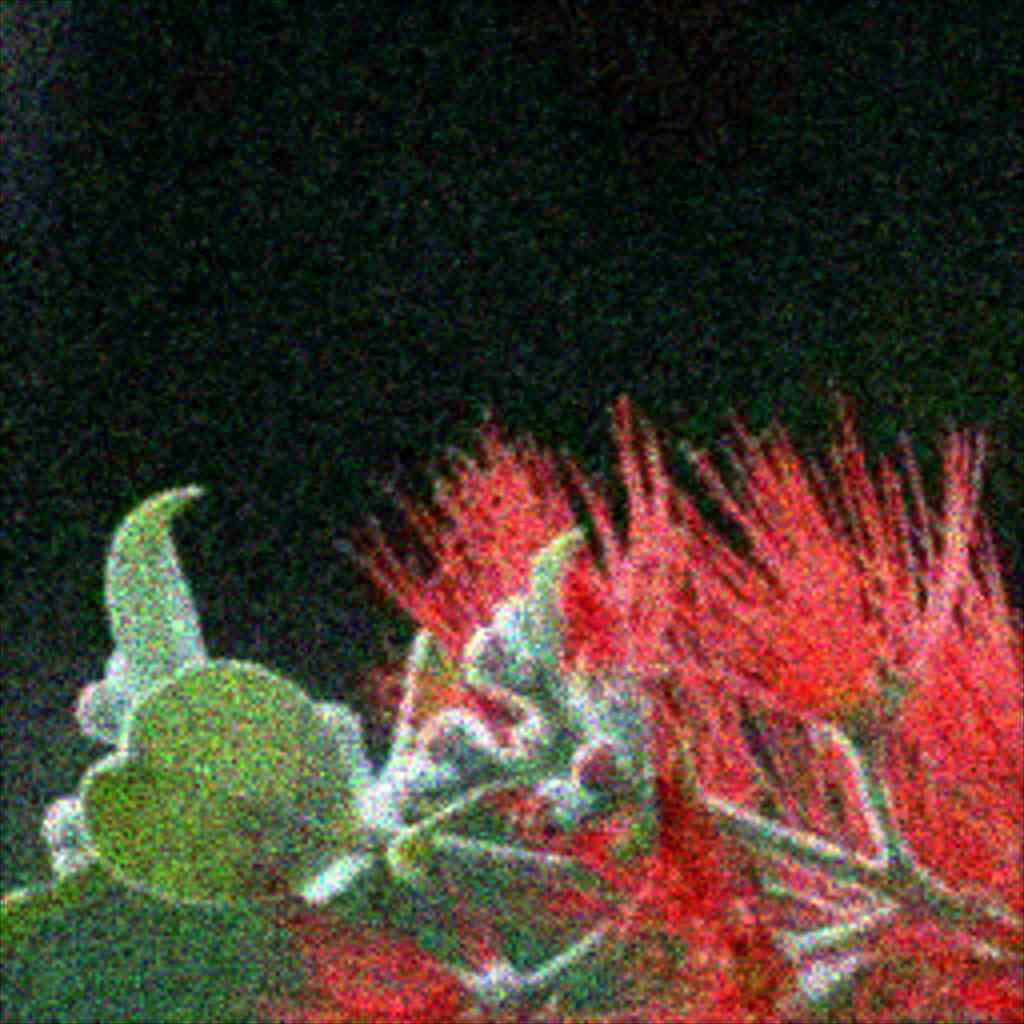}
    \zoomimg{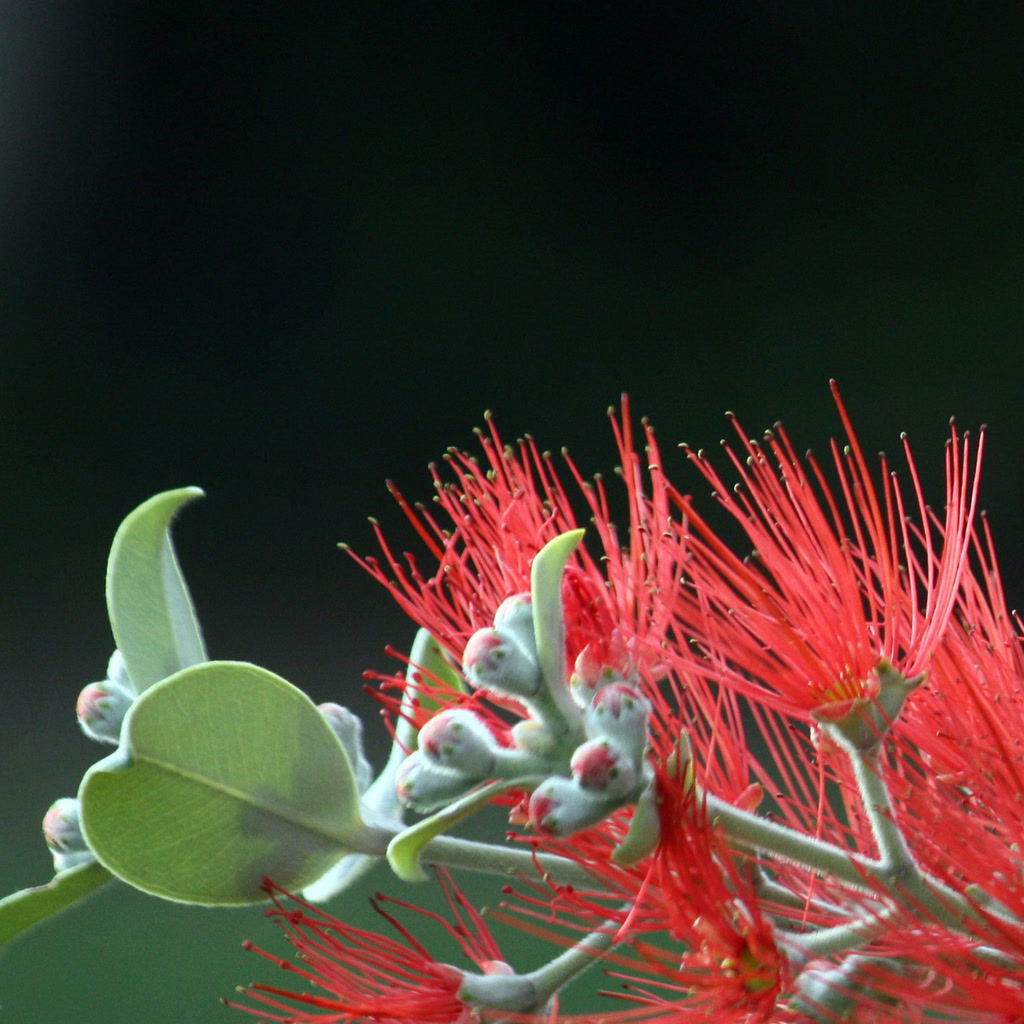}
    \zoomimg{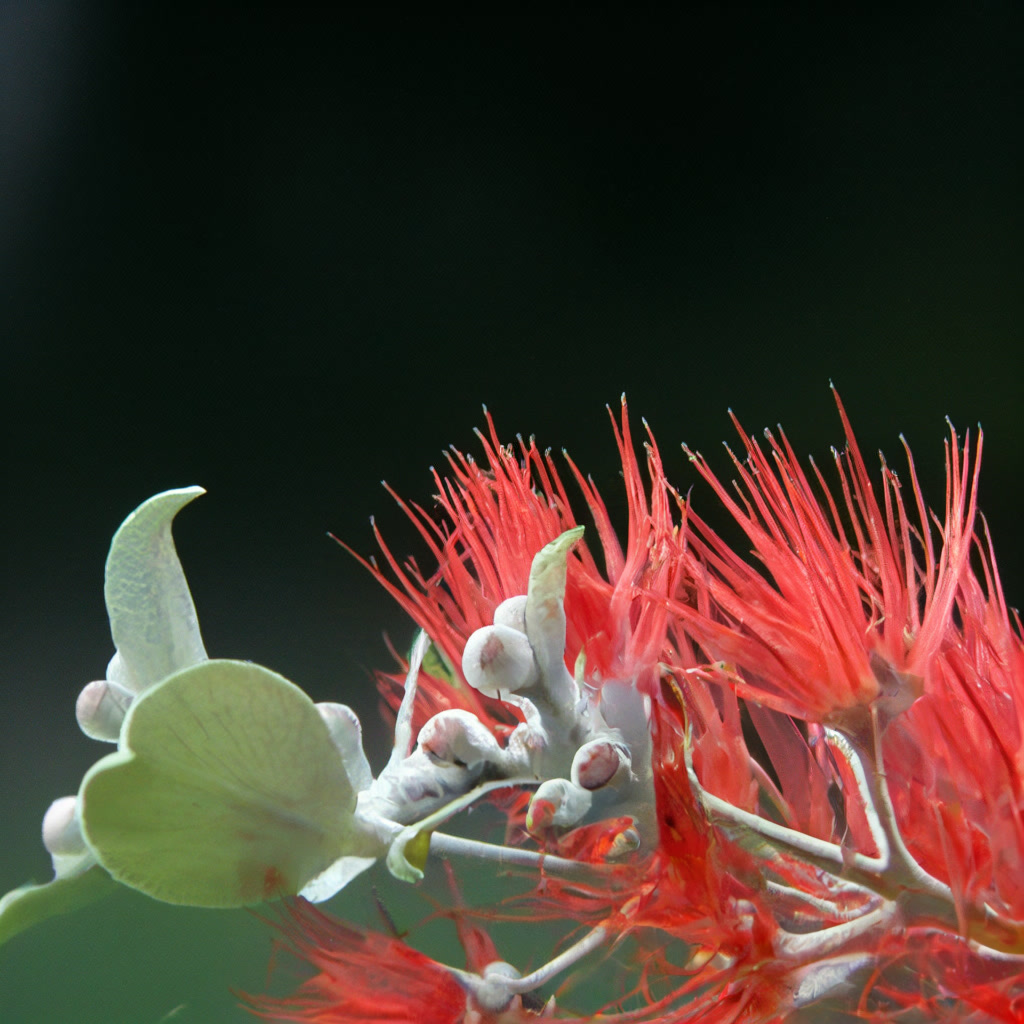}
    \zoomimg{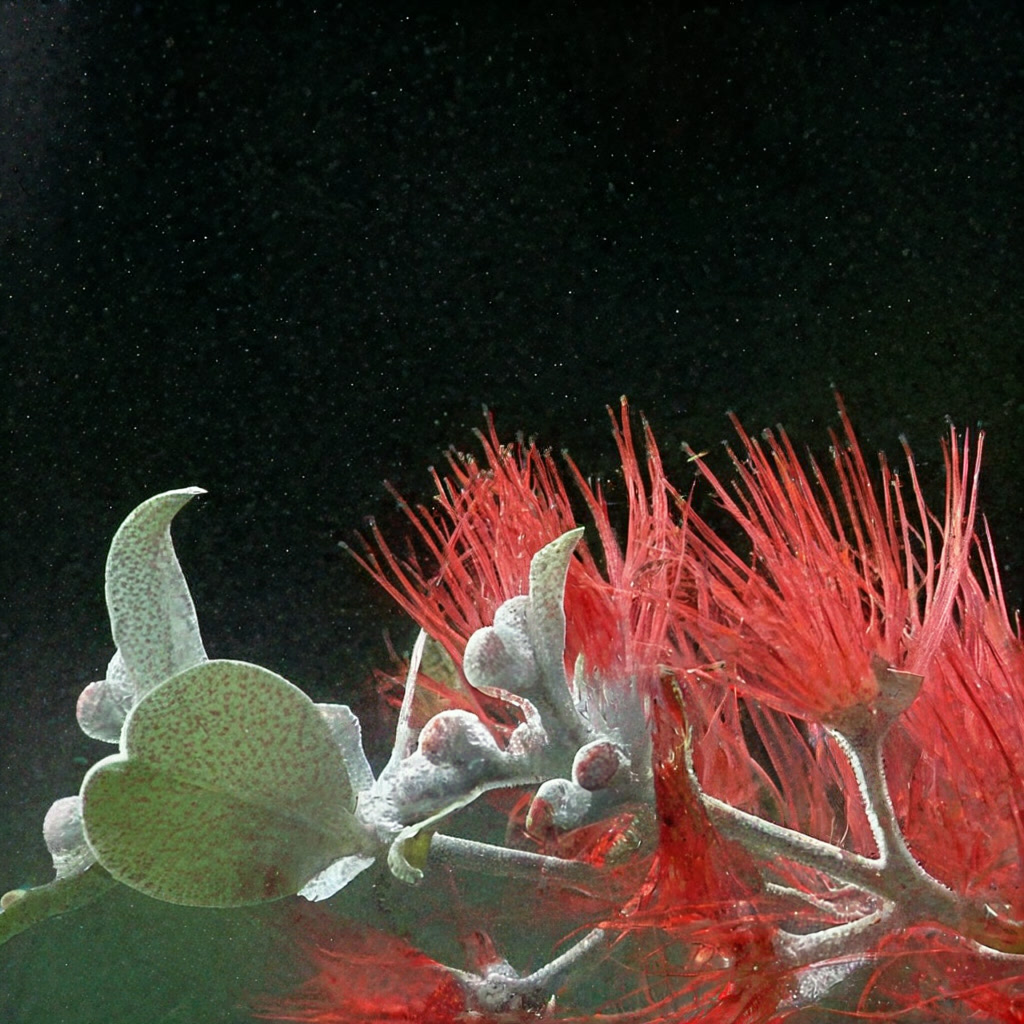}
    \zoomimg{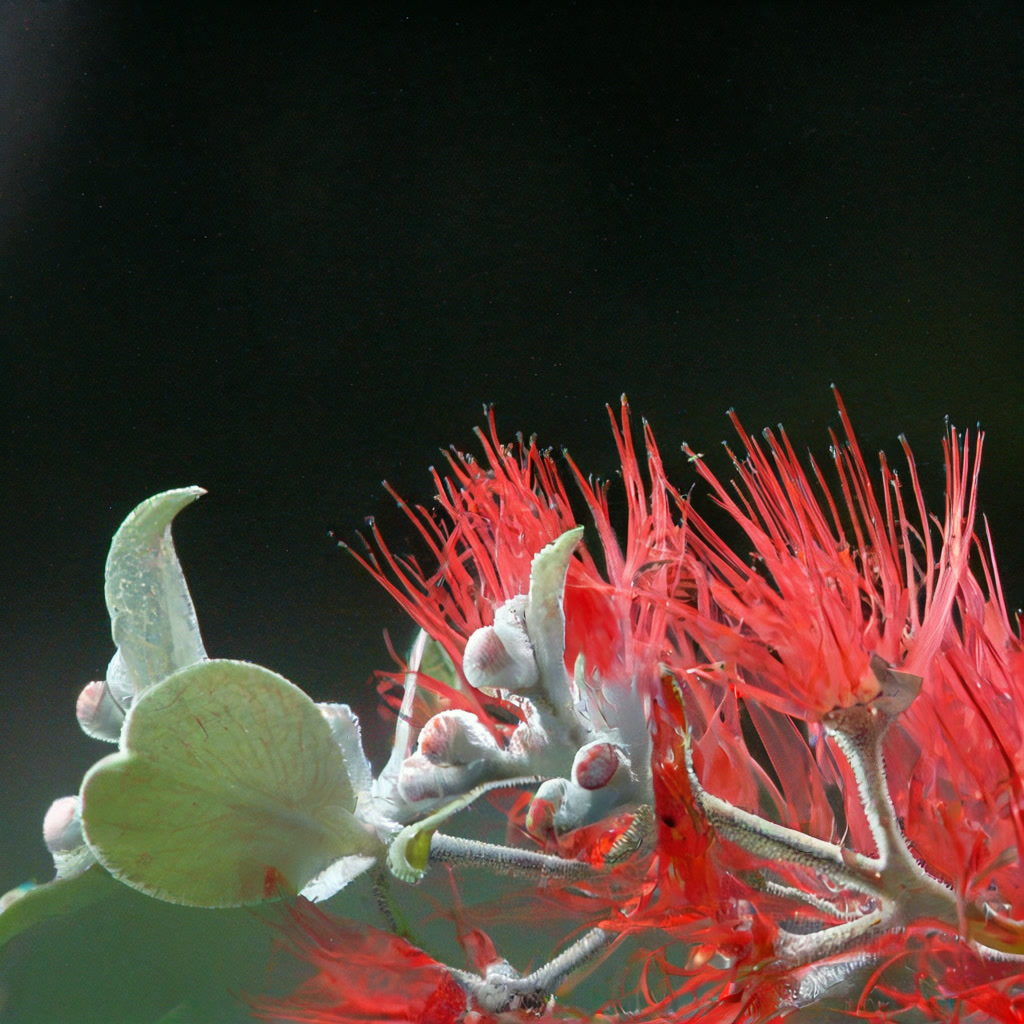}
\end{subfigure}
\caption{\textbf{OID-SR model soup results with OptIMerge on DF2K (SR) data}. (left-to-right) We provide input, target, and images generated from the expert model, naive model soup, and OptIMerge.}
\label{fig:sr_oid_soup_on_sr_data}
\end{figure}

\begin{figure}[h]
\centering
\begin{subfigure}{\textwidth}
    \zHeaderRow
\end{subfigure}
\hfill
\begin{subfigure}{\textwidth}
    \setzoom{0.50}{0.60}{0.79, 0.21}{yellow}
    \zoomimg{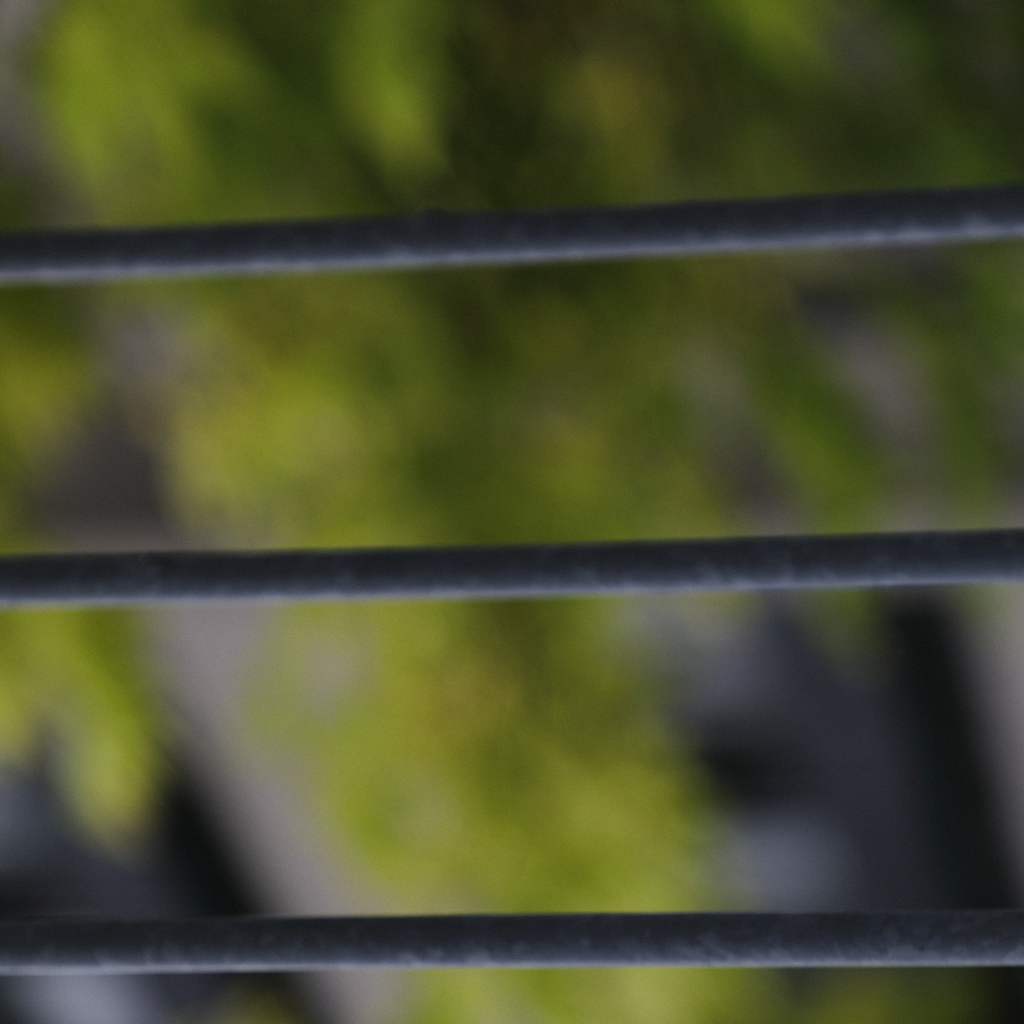}
    \zoomimg{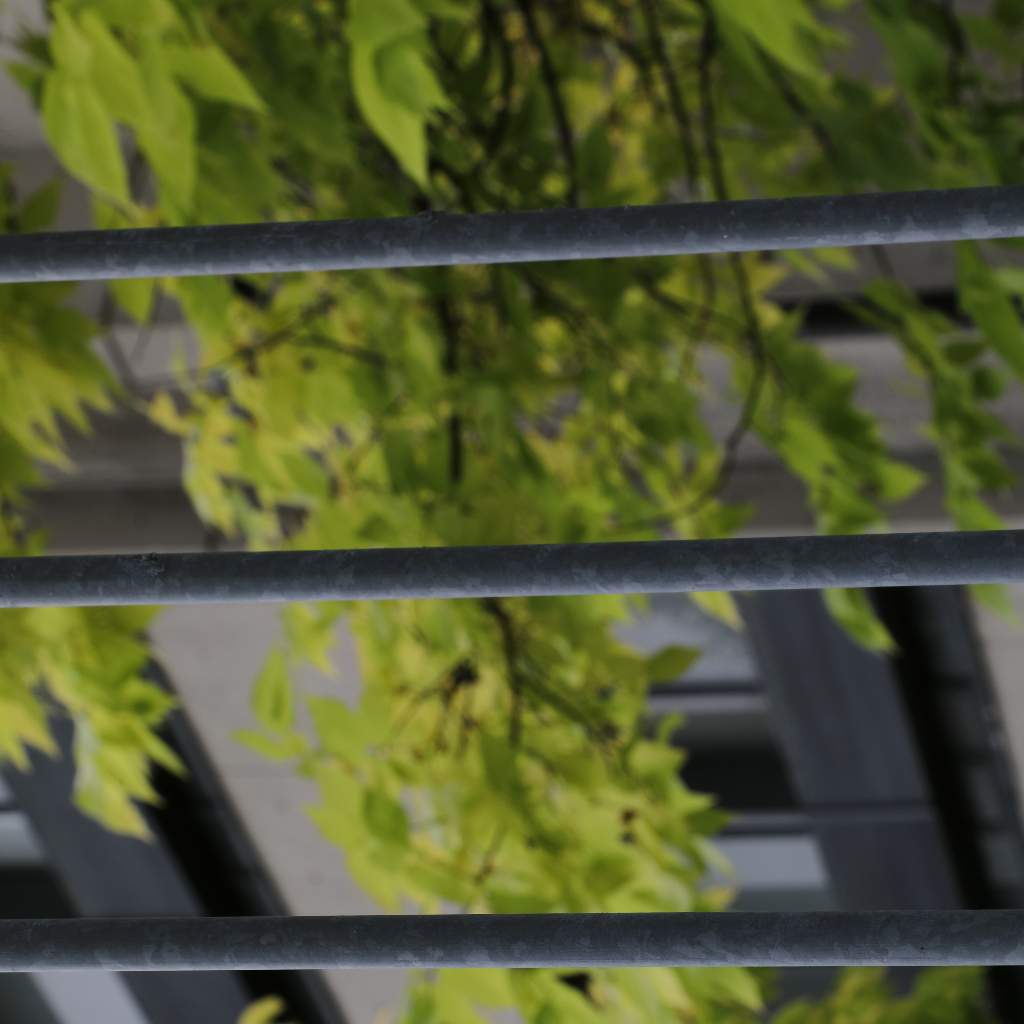}
    \zoomimg{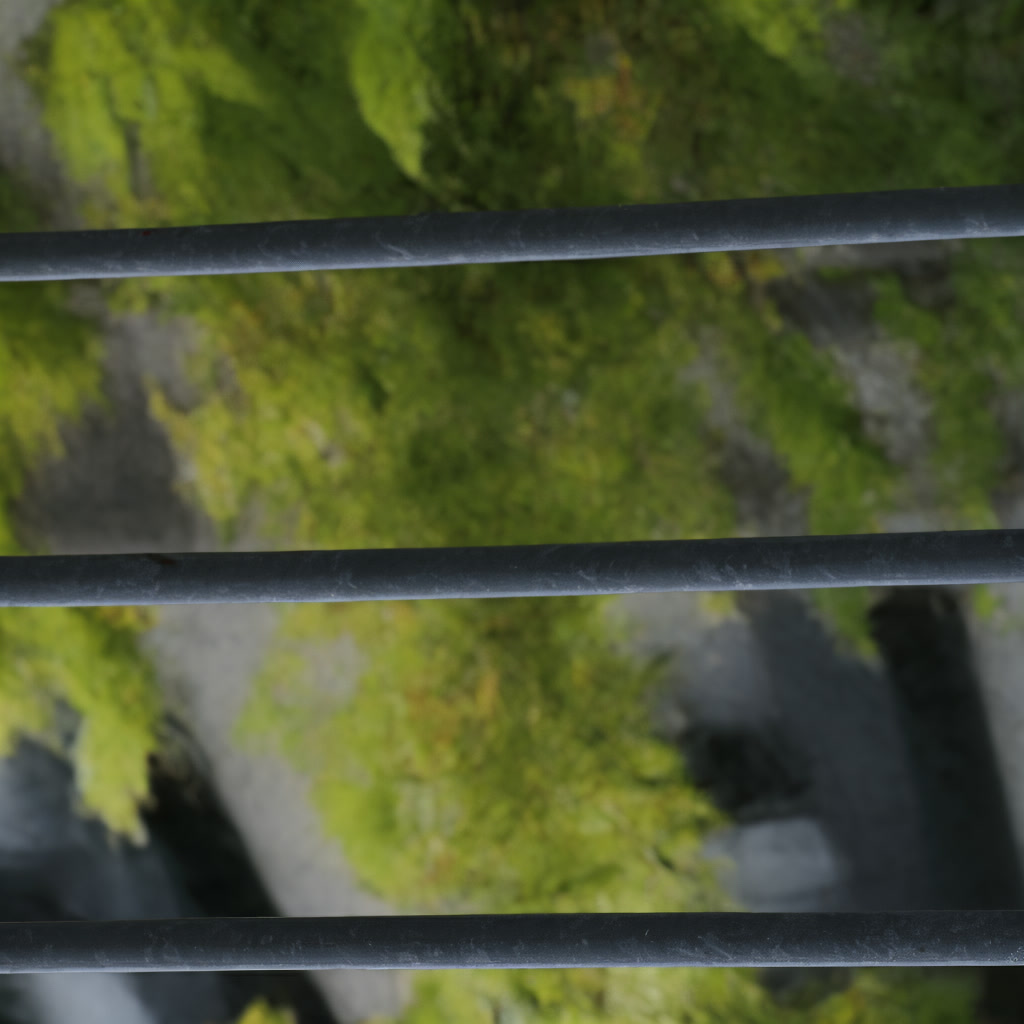}
    \zoomimg{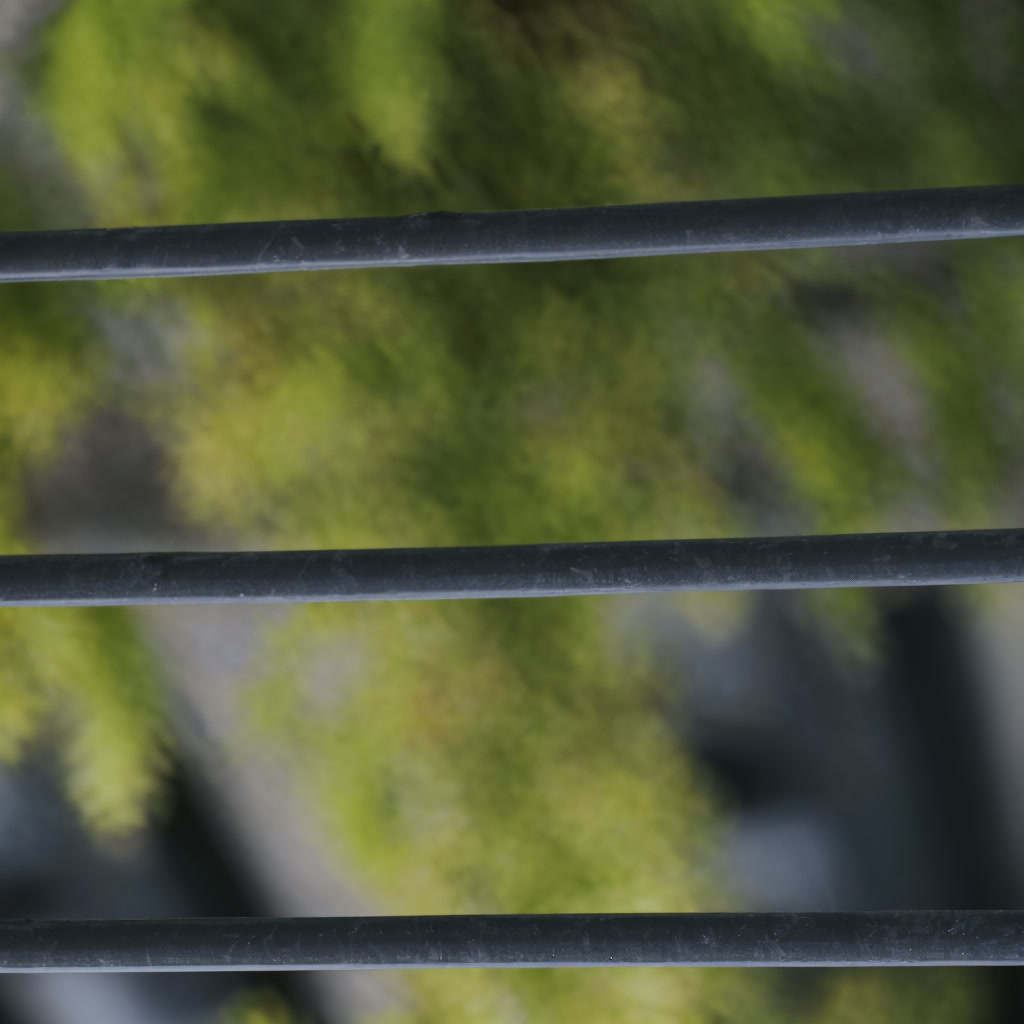}
    \zoomimg{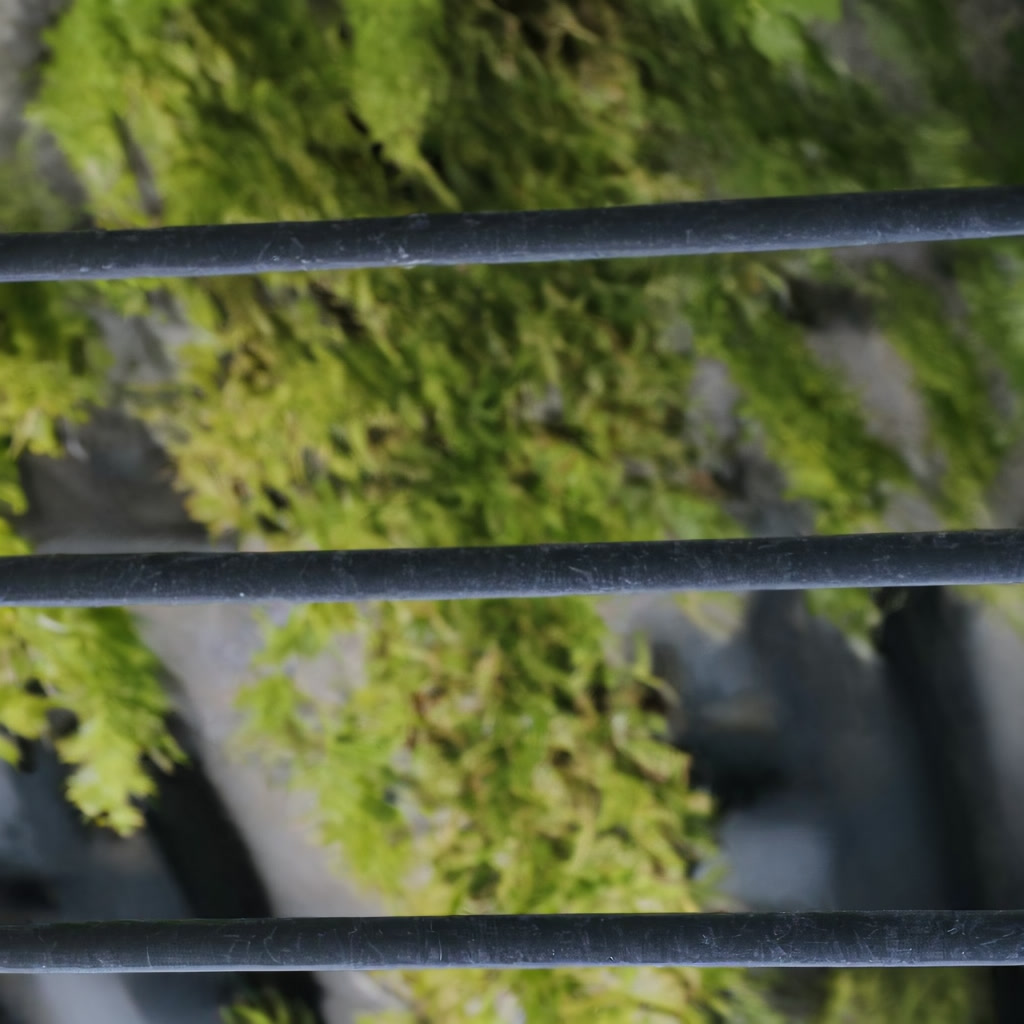}
\end{subfigure}
\hfill
\begin{subfigure}{\textwidth}
    \setzoom{0.38}{0.58}{0.79, 0.21}{yellow}
    \zoomimg{sr_dpdd_results/dpdd/1/input.jpeg}
    \zoomimg{sr_dpdd_results/dpdd/1/target.jpeg}
    \zoomimg{sr_dpdd_results/dpdd/1/expert.jpeg}
    \zoomimg{sr_dpdd_results/dpdd/1/naive_soup.jpeg}
    \zoomimg{sr_dpdd_results/dpdd/1/optmerge.jpeg}
\end{subfigure}
\hfill
\begin{subfigure}{\textwidth}
    \setzoom{0.35}{0.68}{0.79, 0.21}{yellow}
    \zoomimg{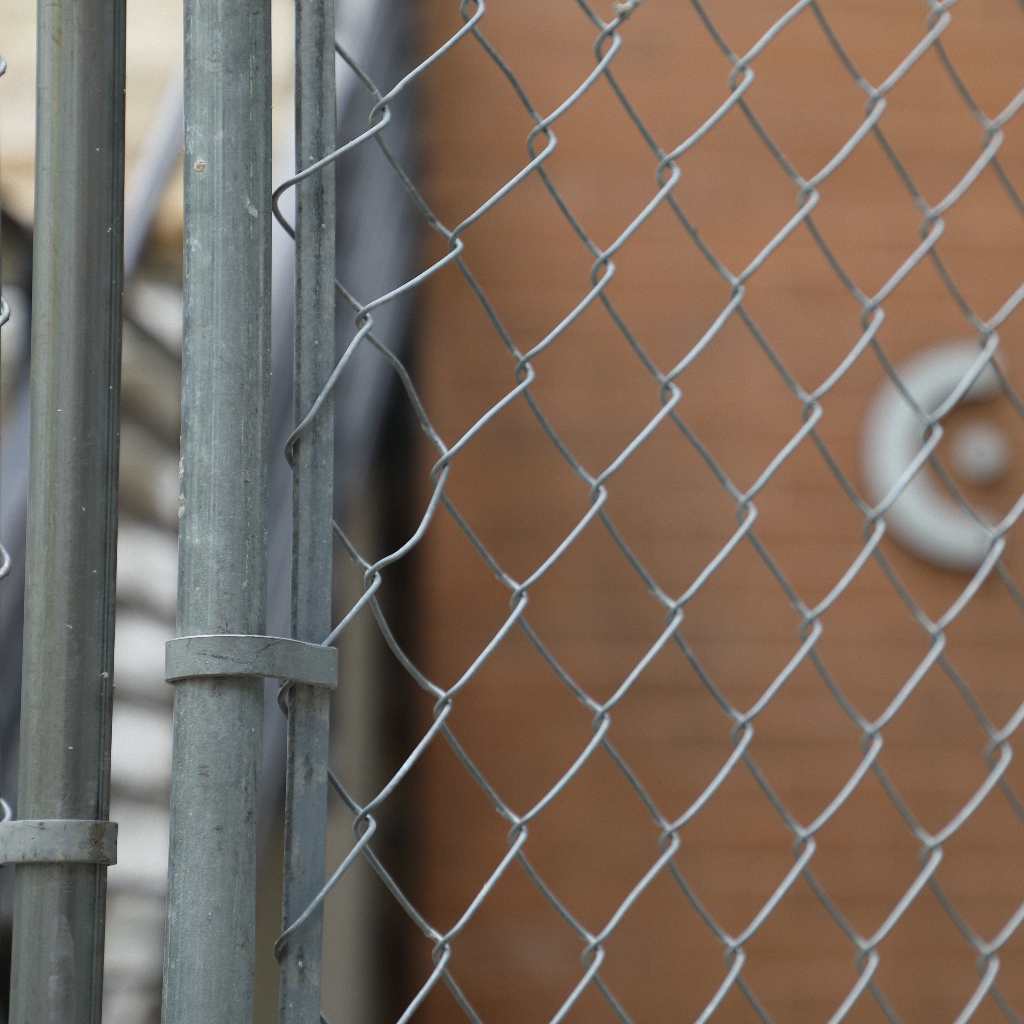}
    \zoomimg{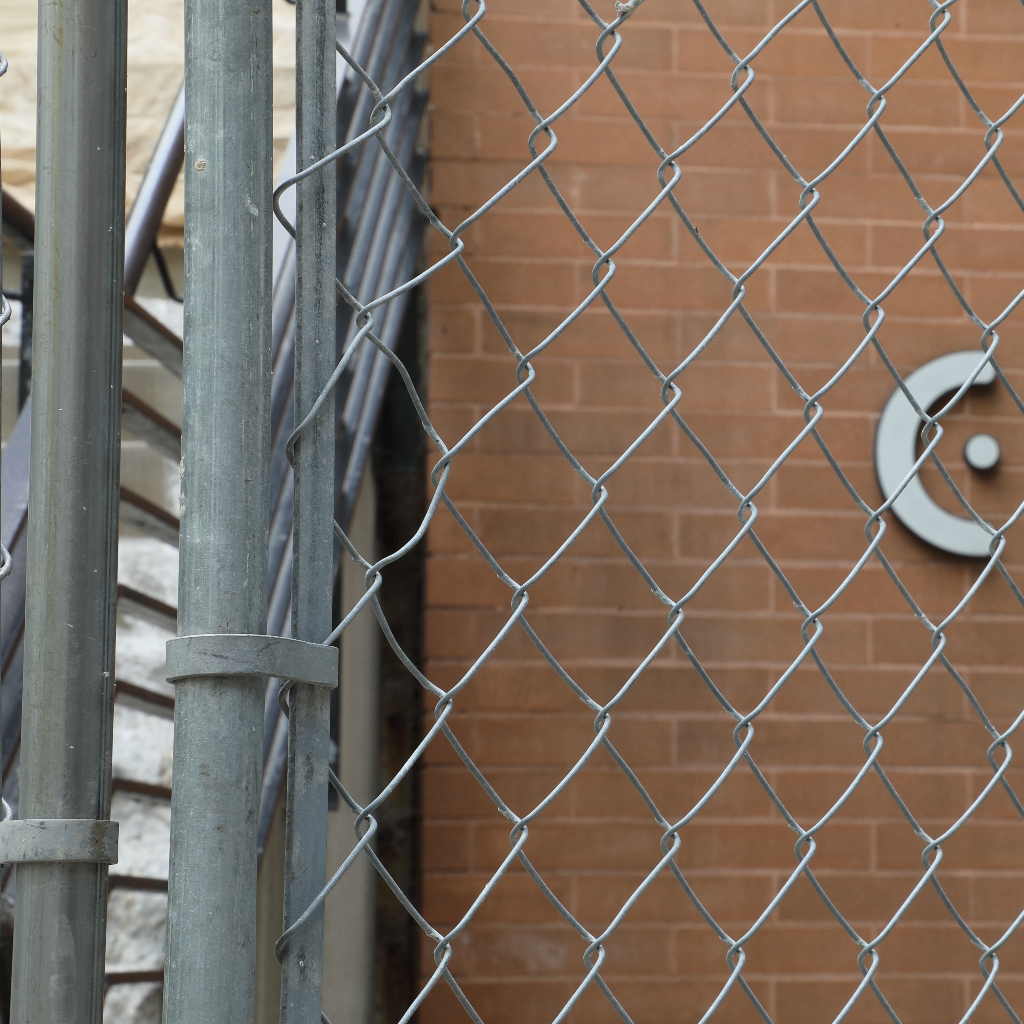}
    \zoomimg{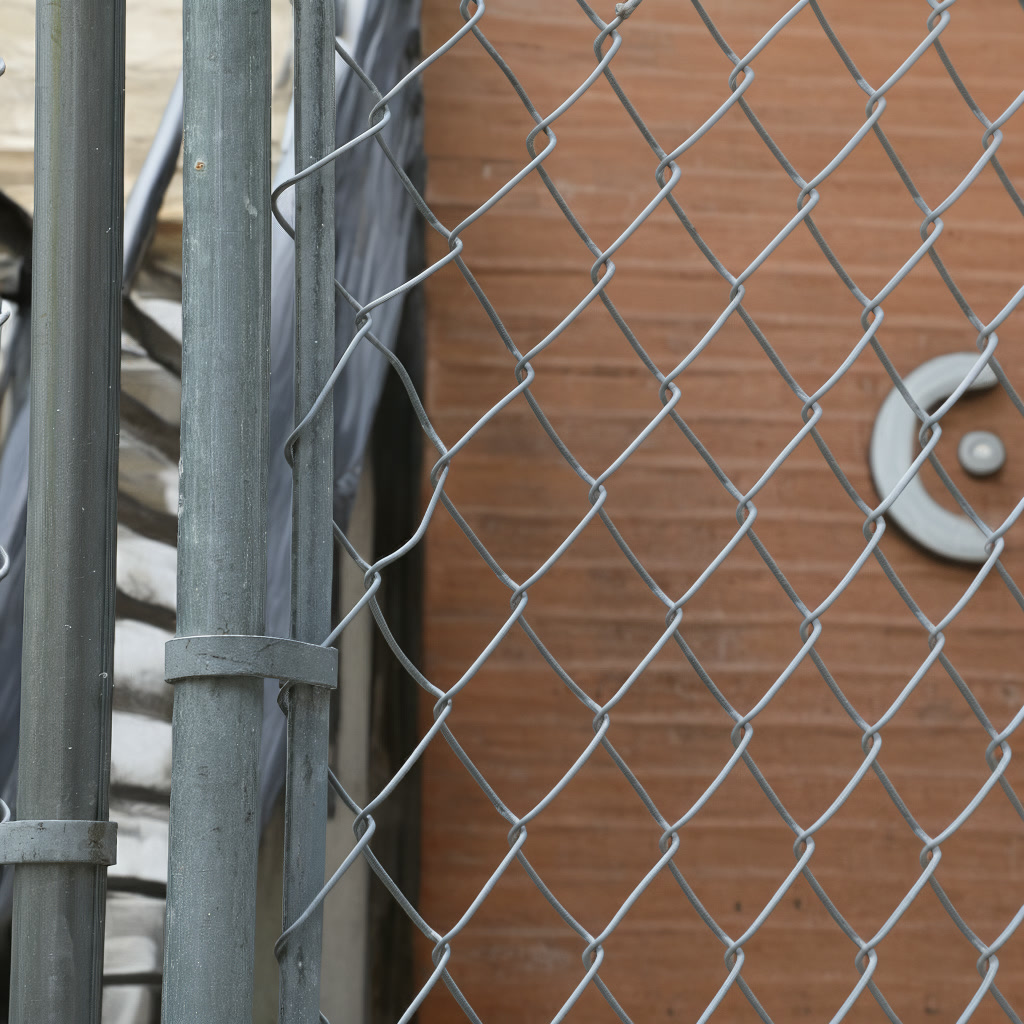}
    \zoomimg{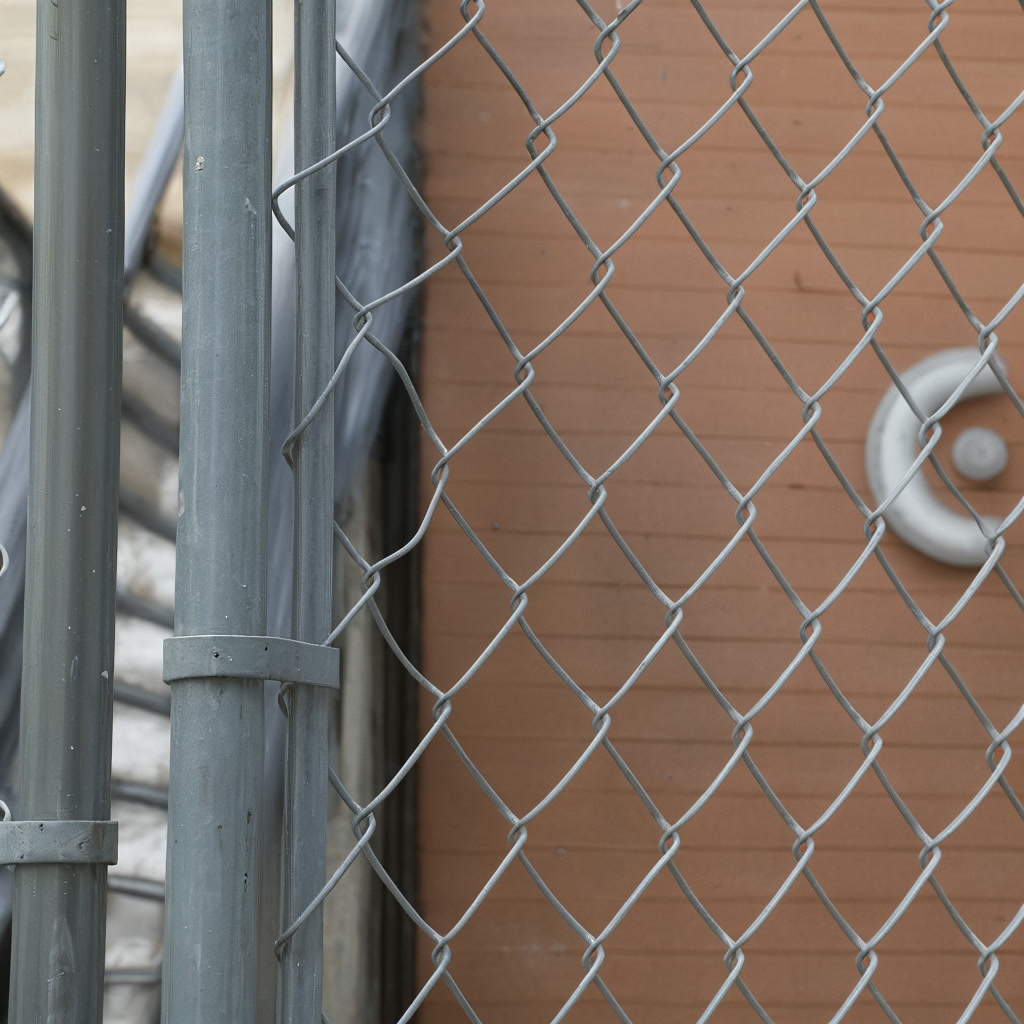}
    \zoomimg{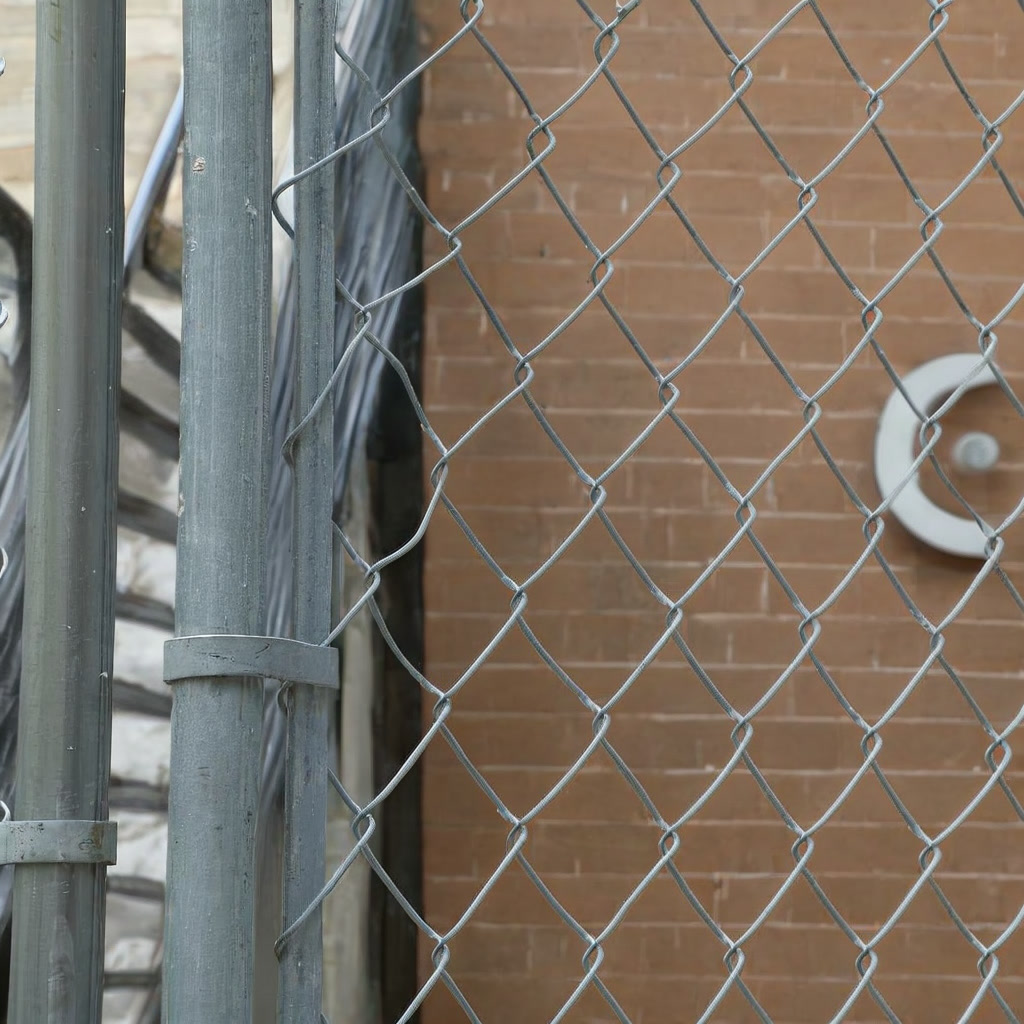}
\end{subfigure}
\caption{\textbf{DPDD-SR model soup results with OptIMerge on DPDD data}. (left-to-right) We provide input, target, and images generated from the expert model, naive model soup, and OptIMerge}
\label{fig:sr_dpdd_soup_on_dpdd_data}
\end{figure}

\begin{figure}[h]
\centering
\begin{subfigure}{\textwidth}
    \zHeaderRow
\end{subfigure}
\hfill
\begin{subfigure}{\textwidth}
    \setzoom{0.50}{0.50}{0.79, 0.21}{yellow}
    \zoomimg{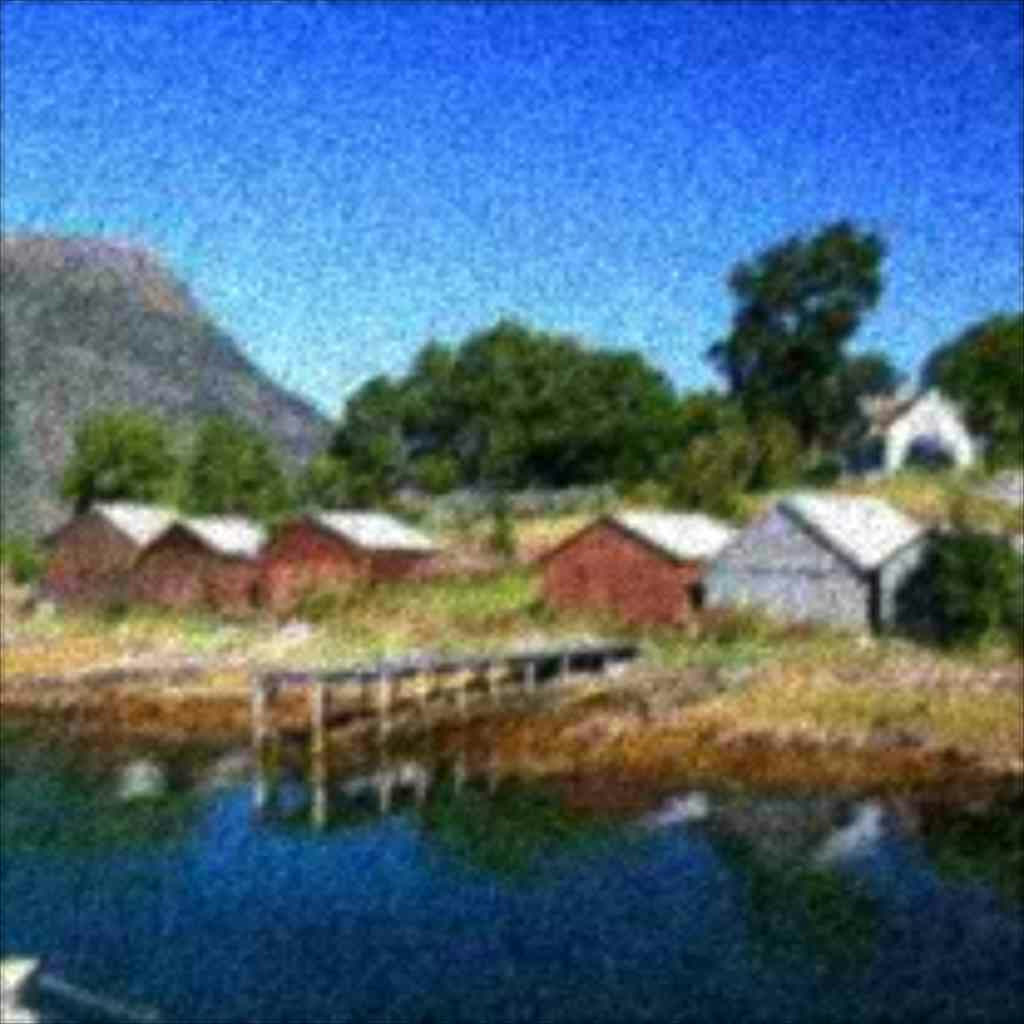}
    \zoomimg{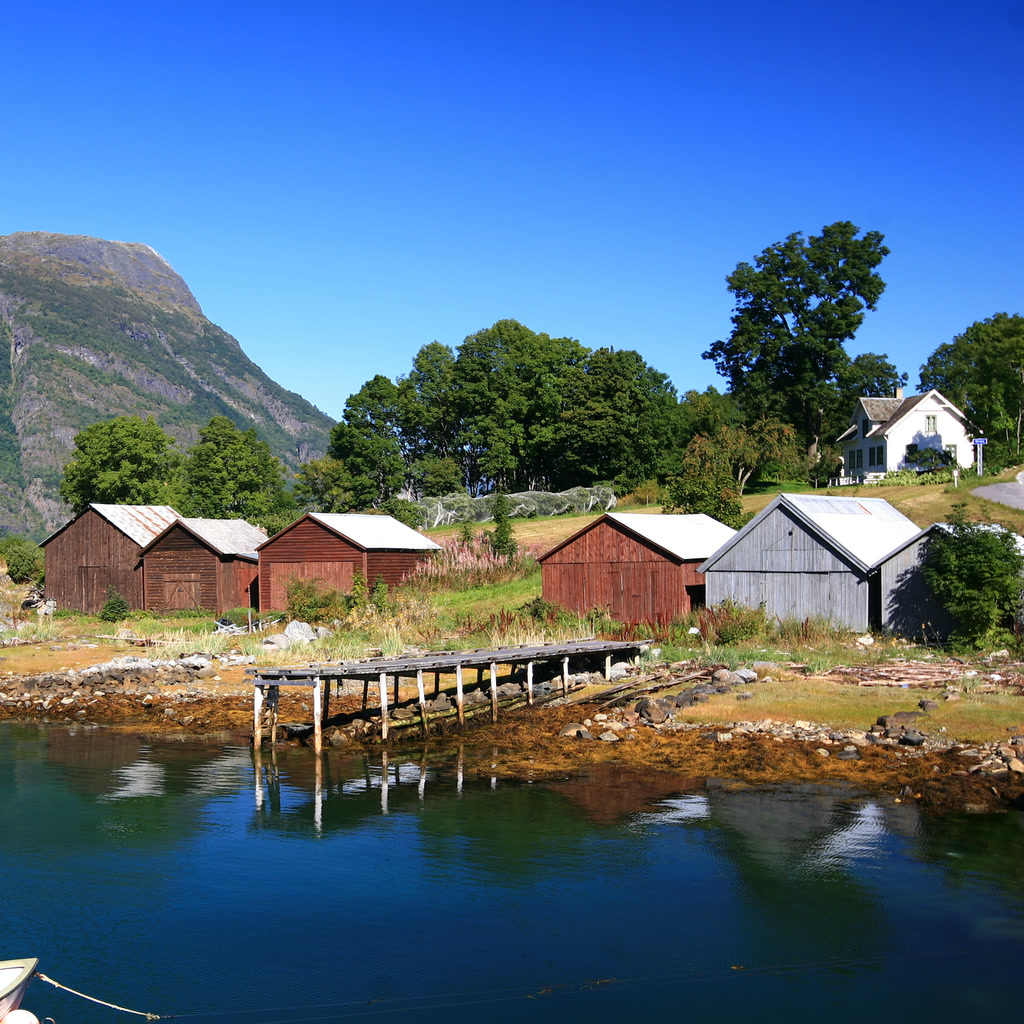}
    \zoomimg{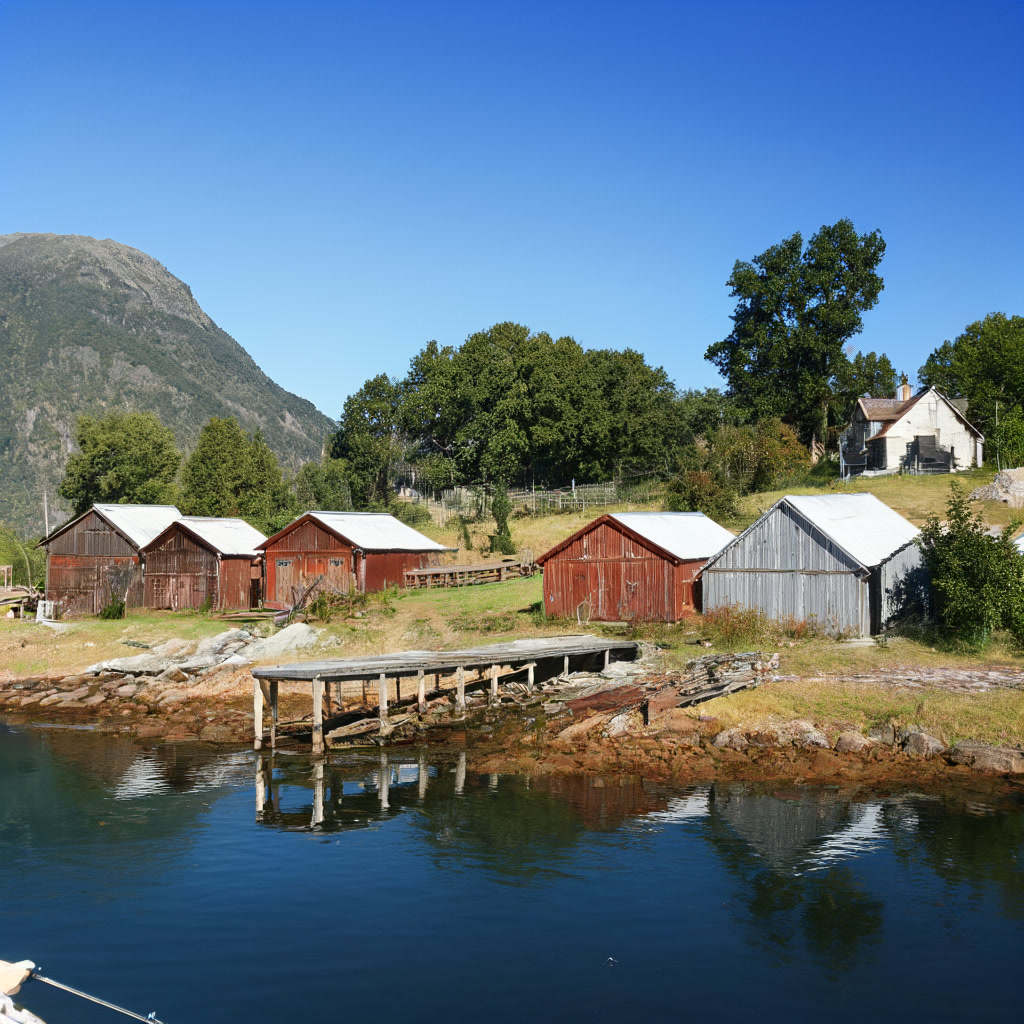}
    \zoomimg{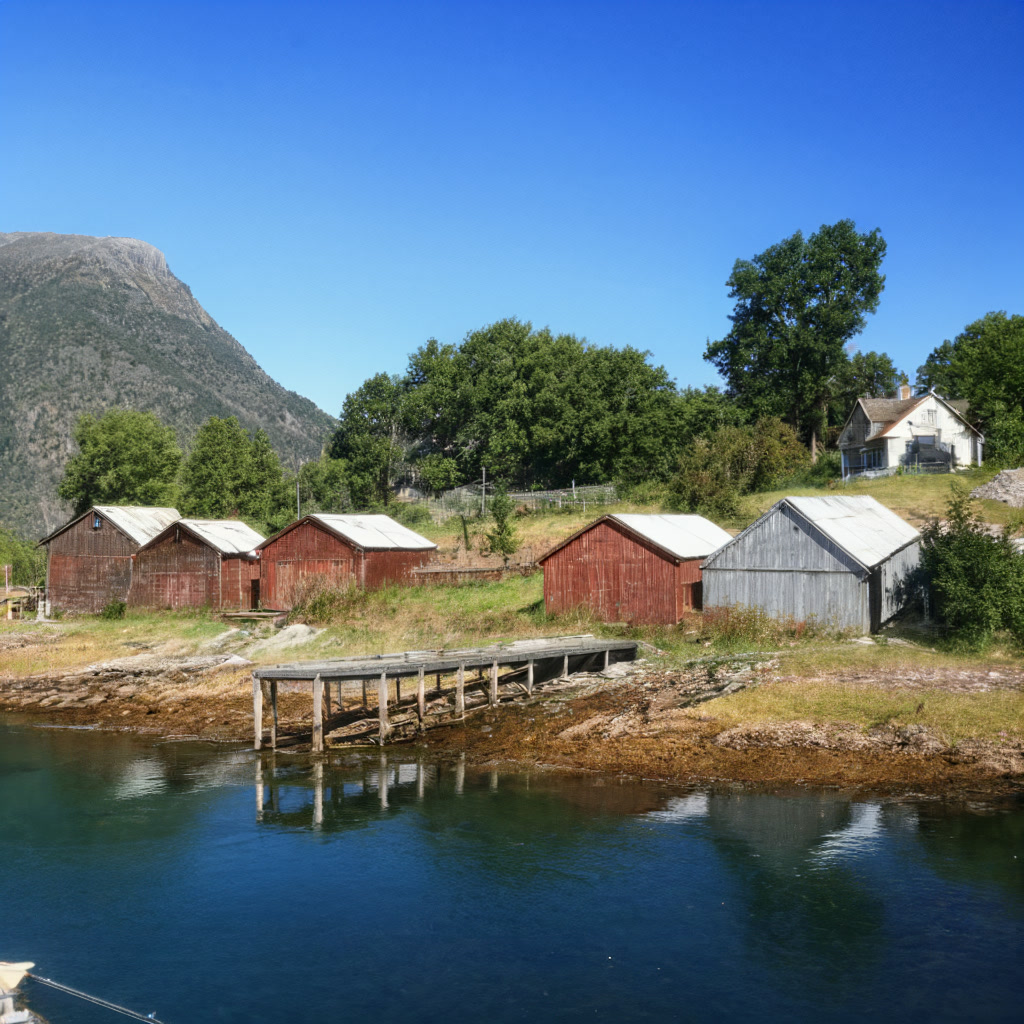}
    \zoomimg{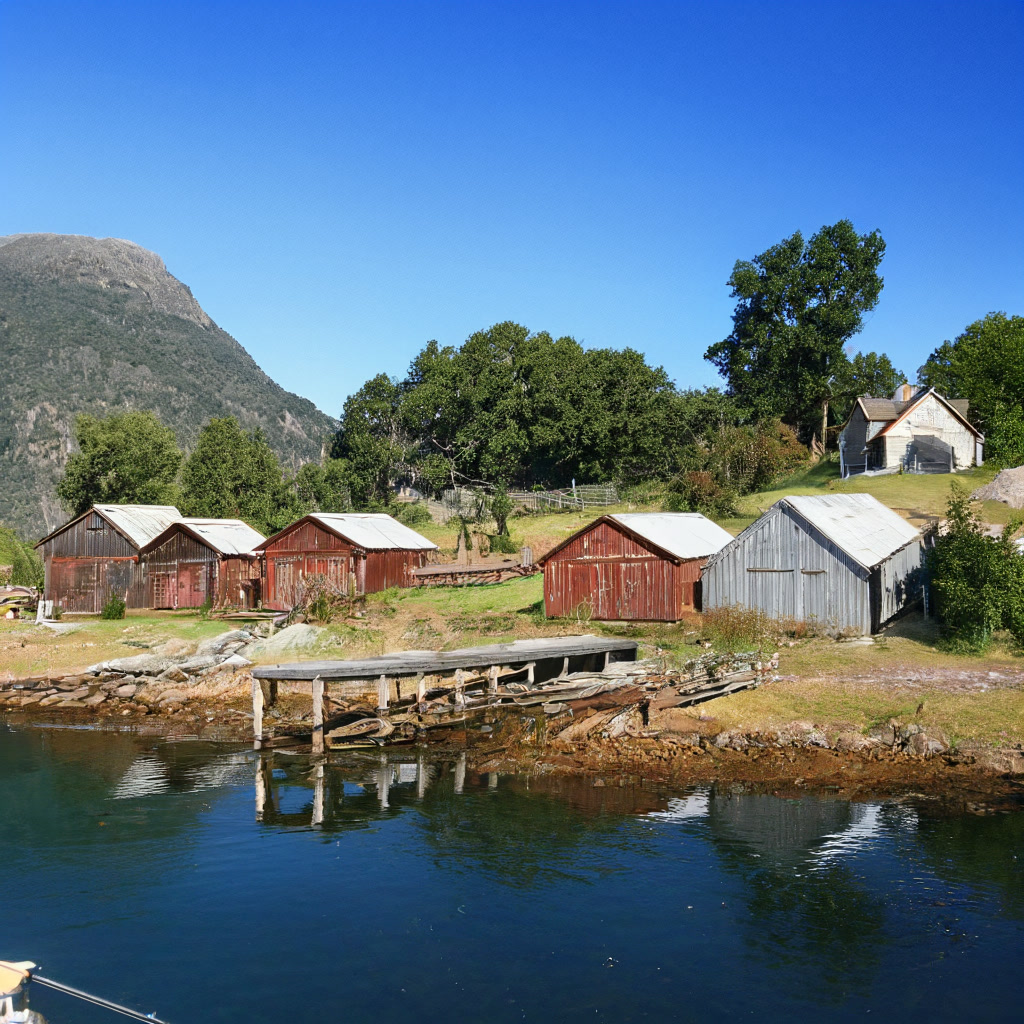}
\end{subfigure}
\hfill
\begin{subfigure}{\textwidth}
    \setzoom{0.40}{0.80}{0.79, 0.21}{yellow}
    \zoomimg{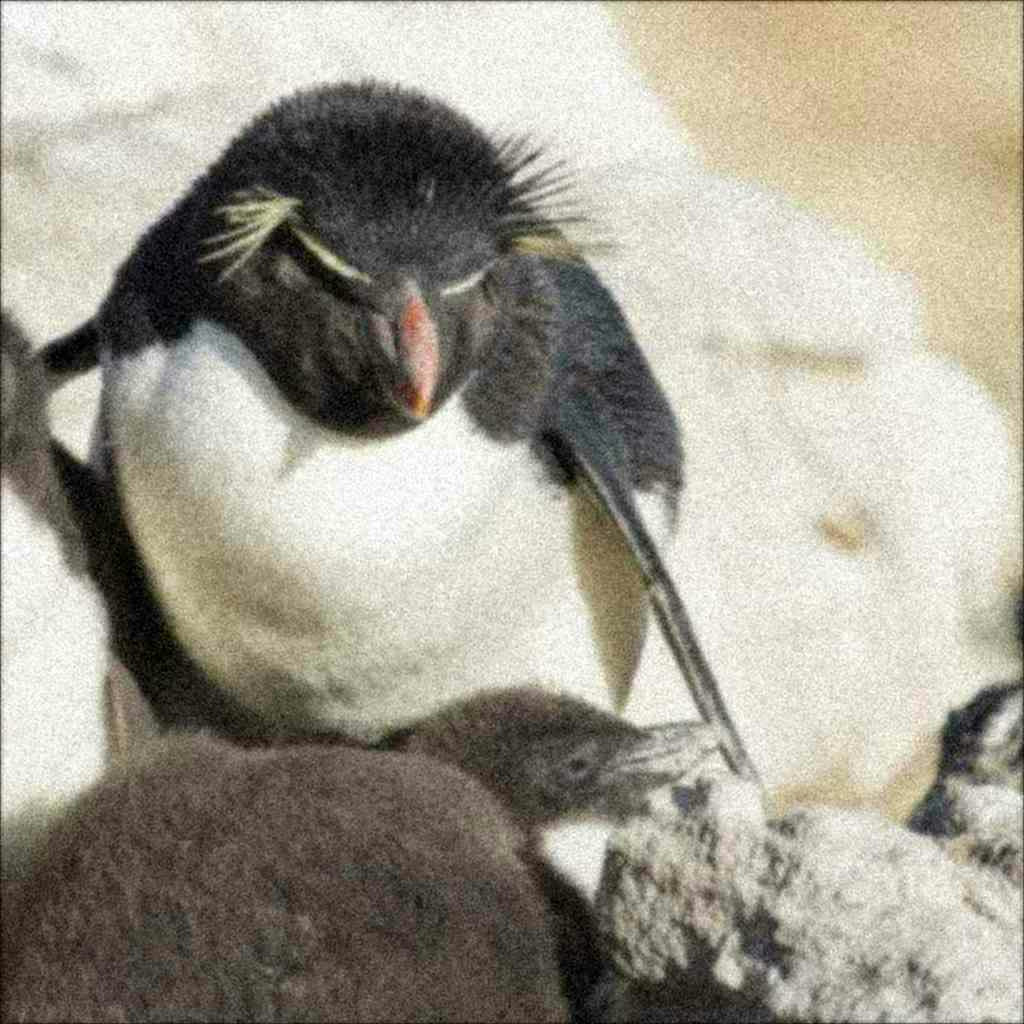}
    \zoomimg{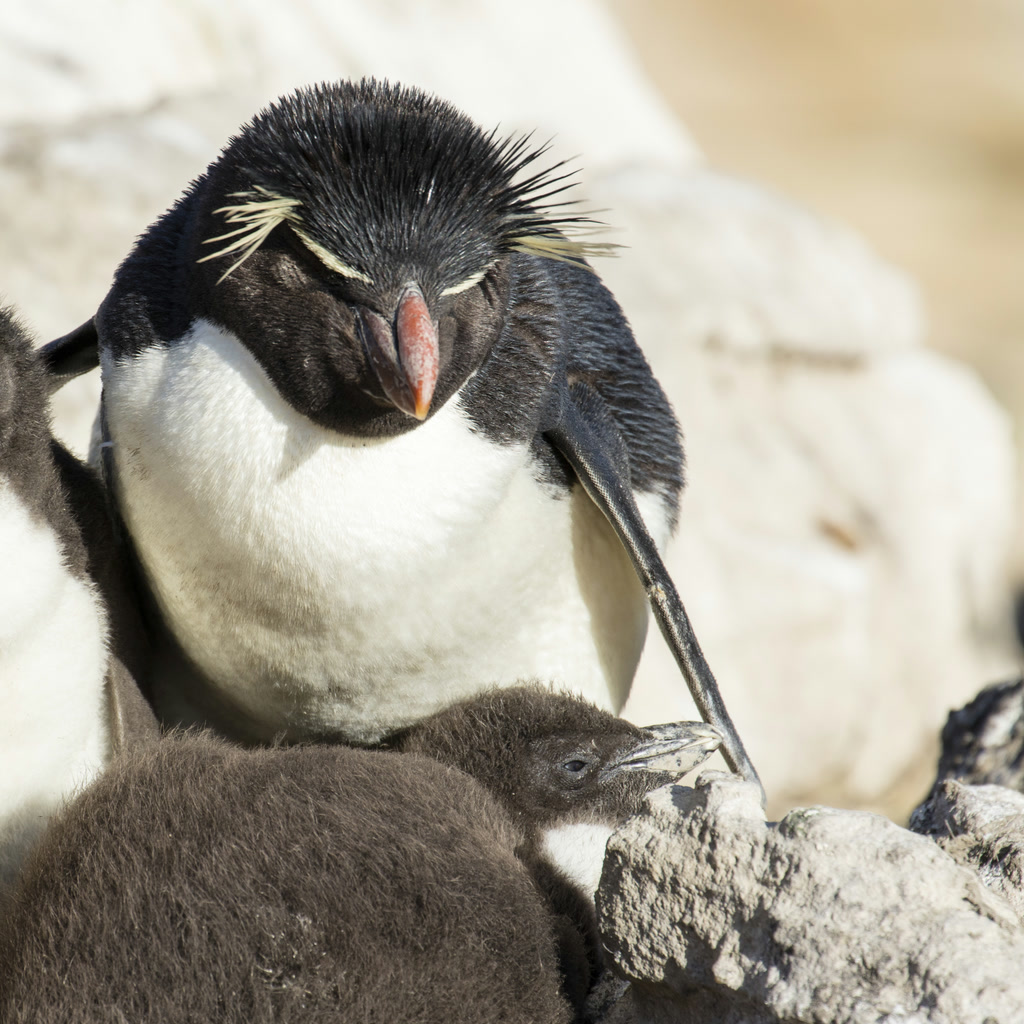}
    \zoomimg{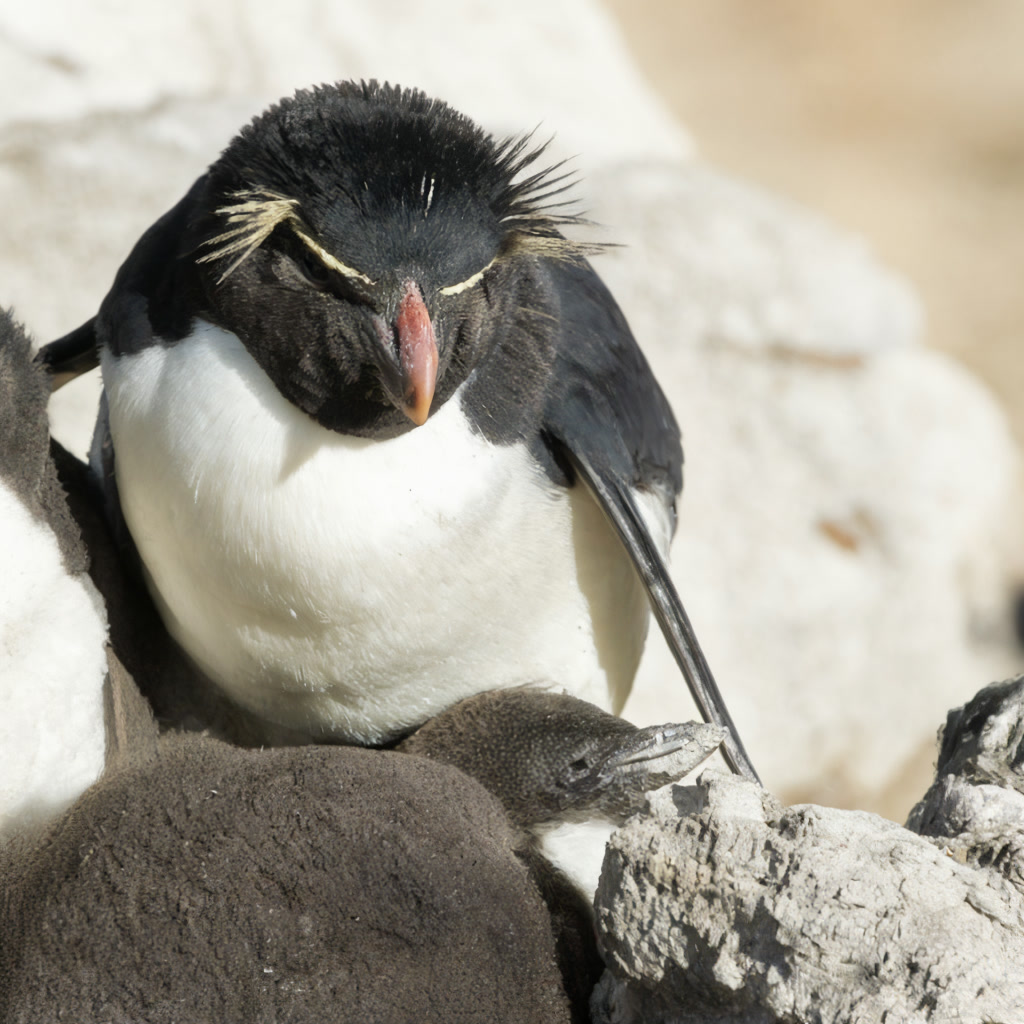}
    \zoomimg{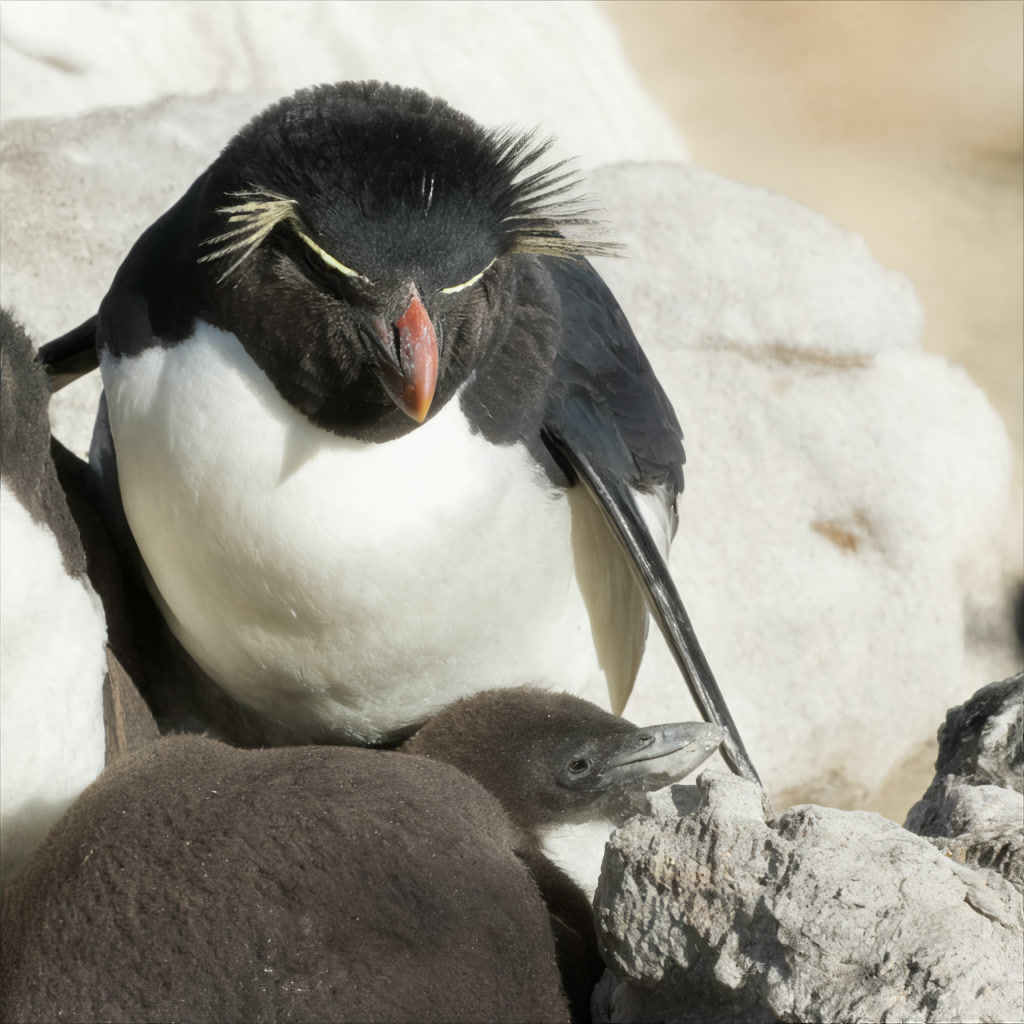}
    \zoomimg{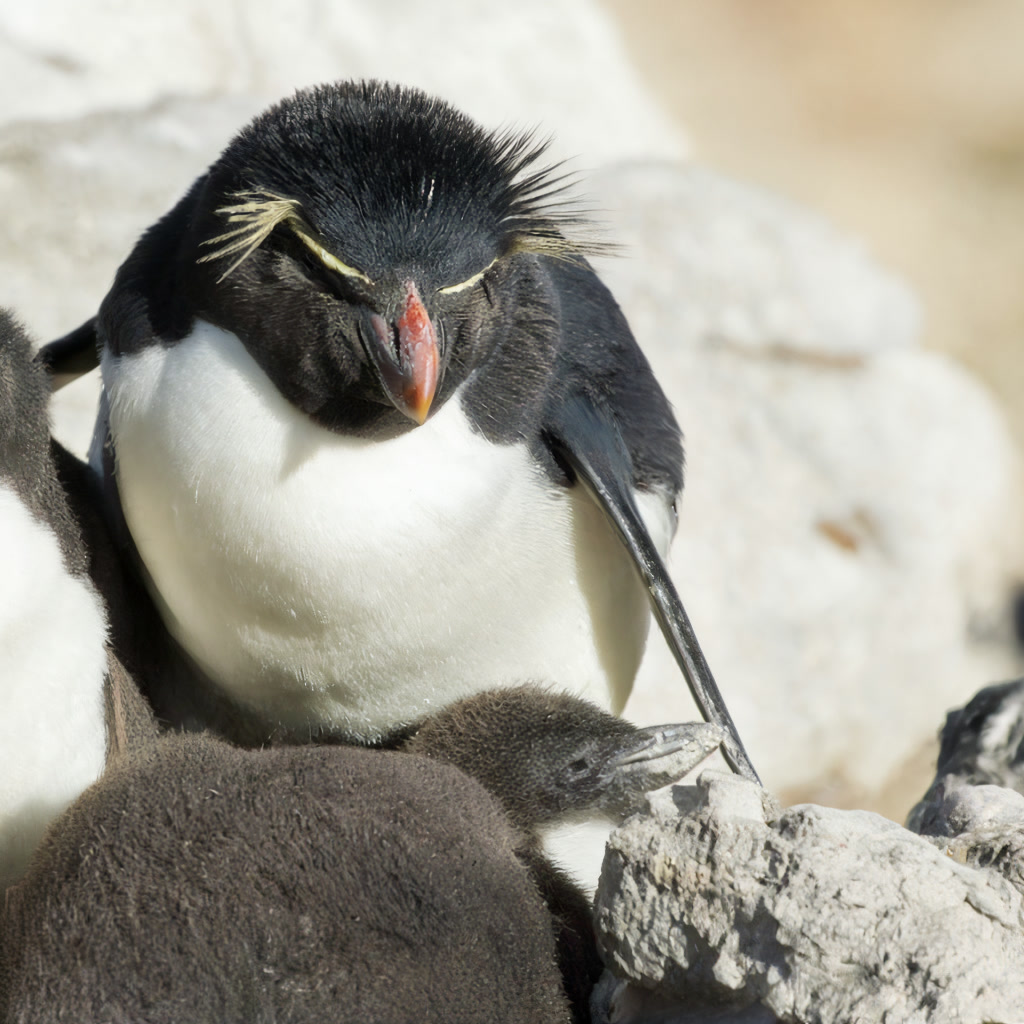}
\end{subfigure}
\hfill
\begin{subfigure}{\textwidth}
    \setzoom{0.30}{0.30}{0.79, 0.21}{yellow}
    \zoomimg{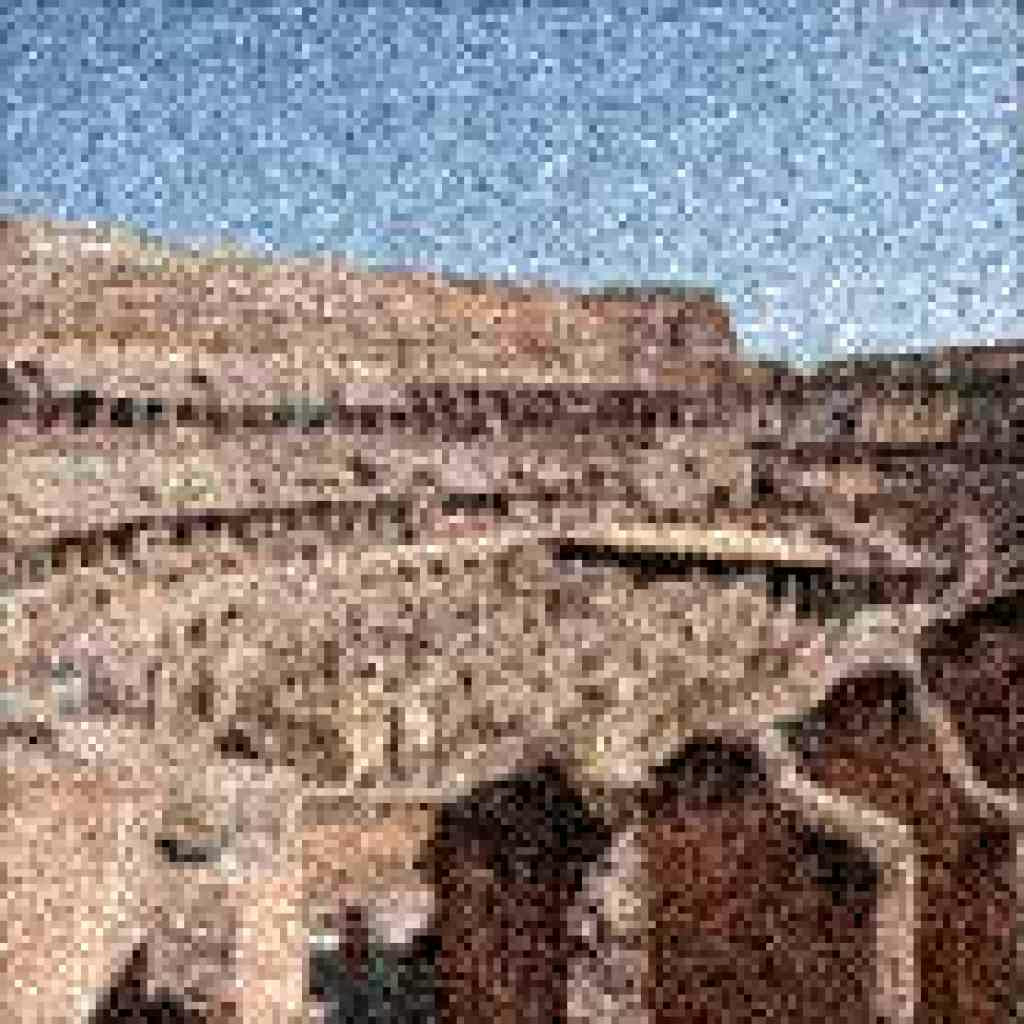}
    \zoomimg{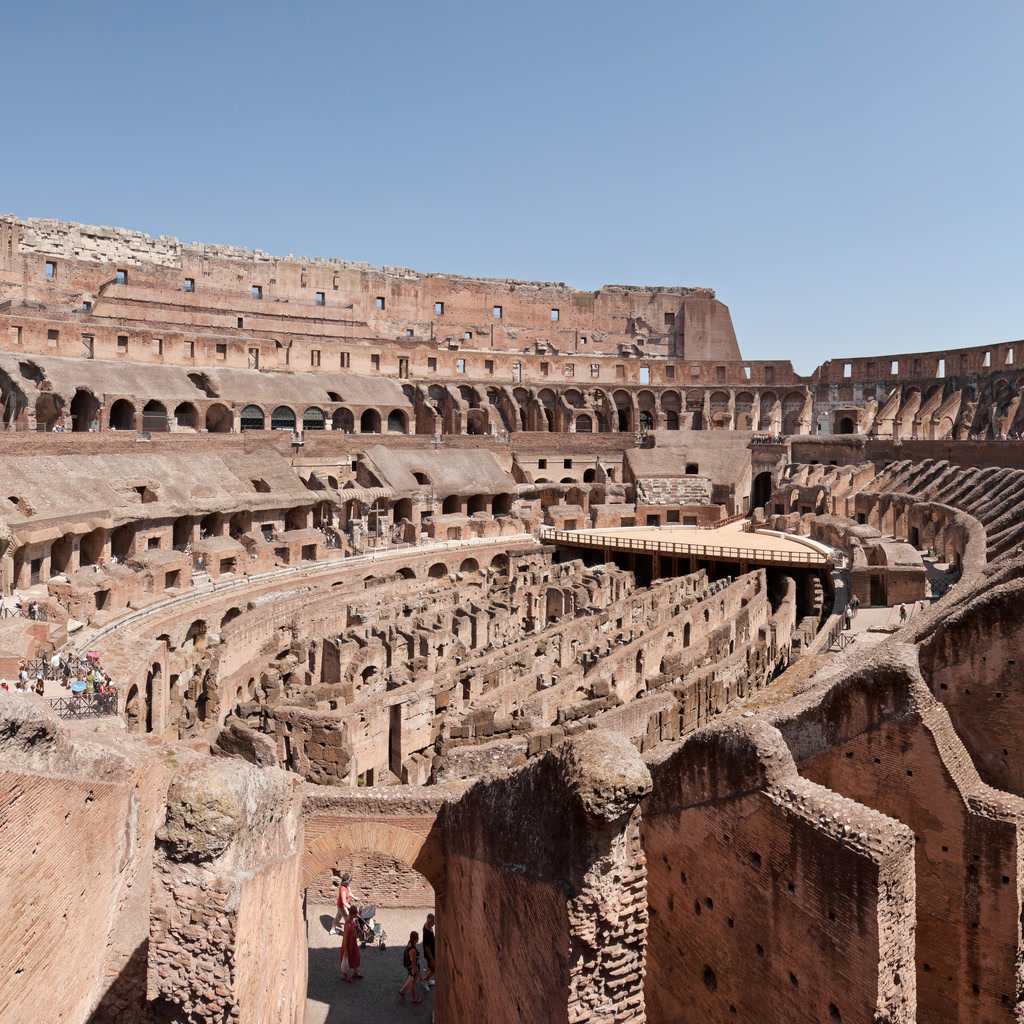}
    \zoomimg{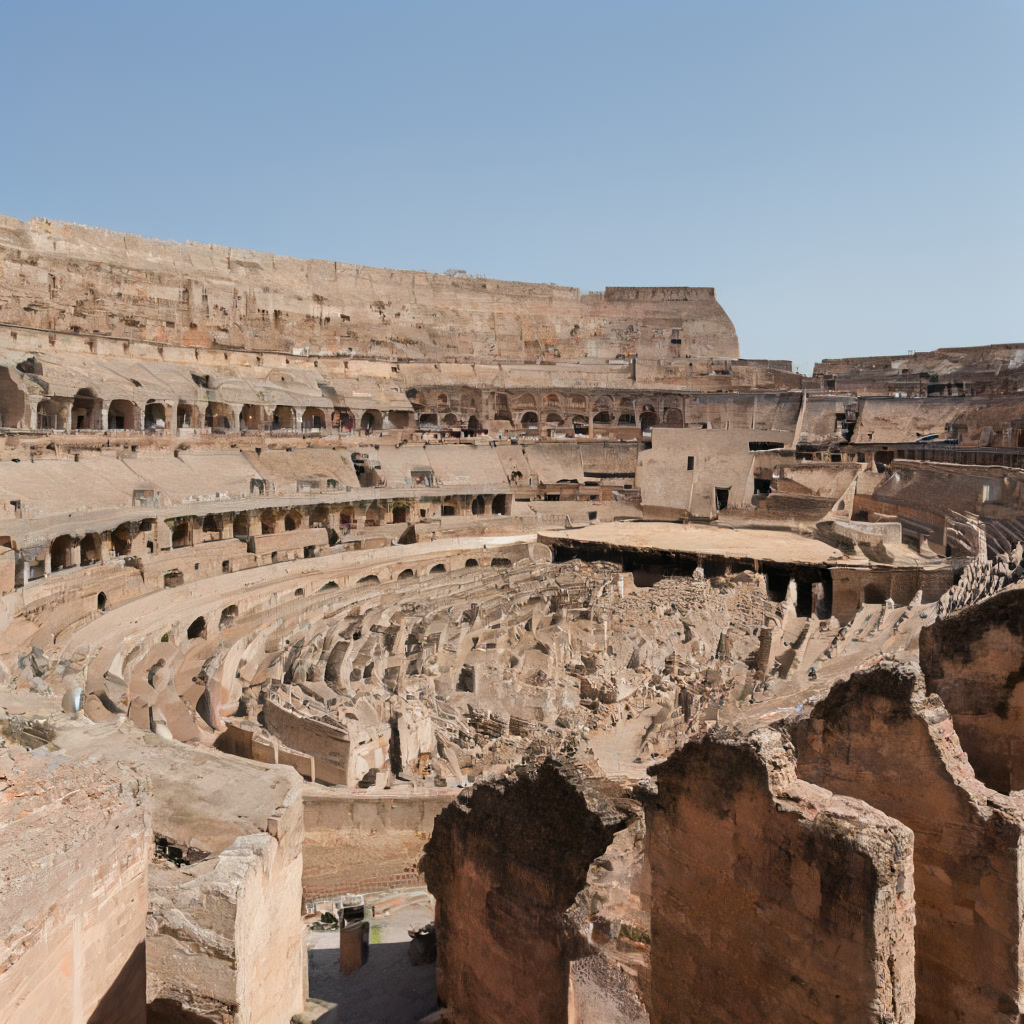}
    \zoomimg{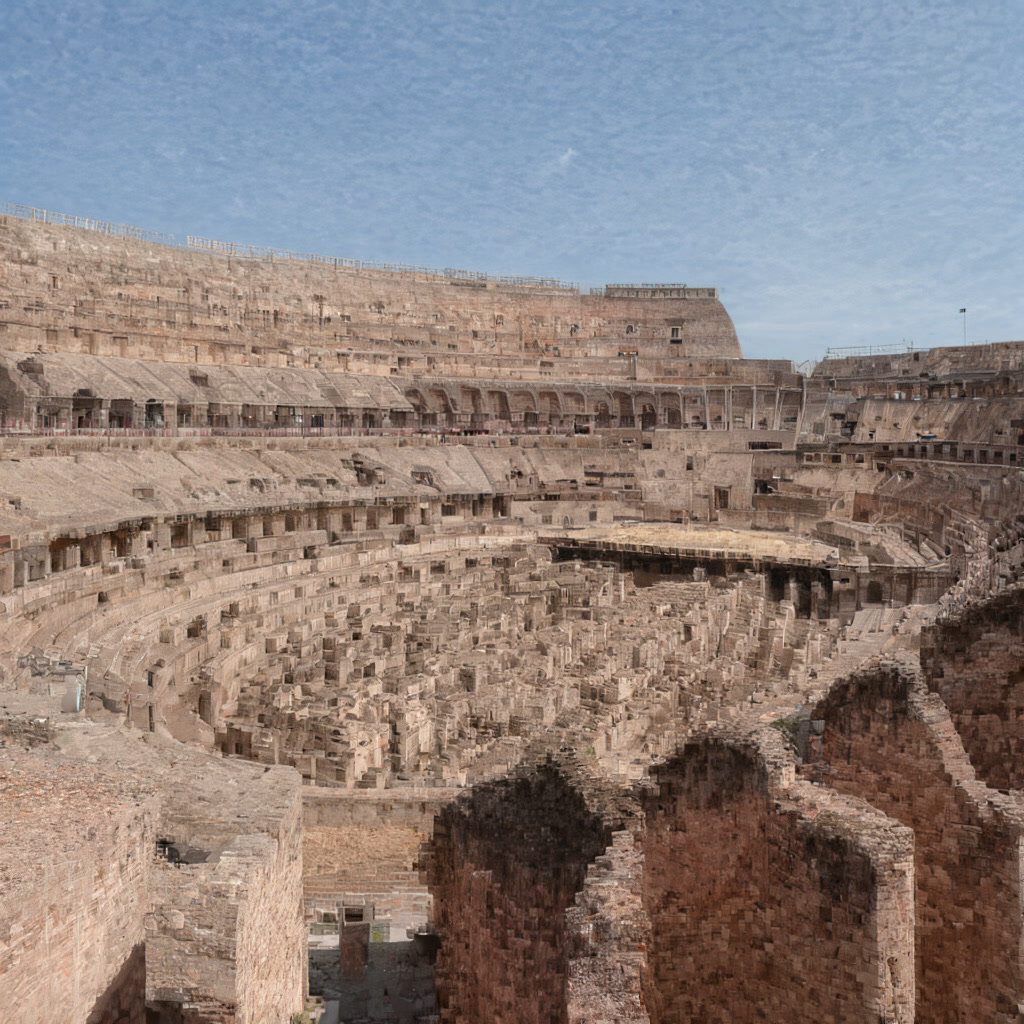}
    \zoomimg{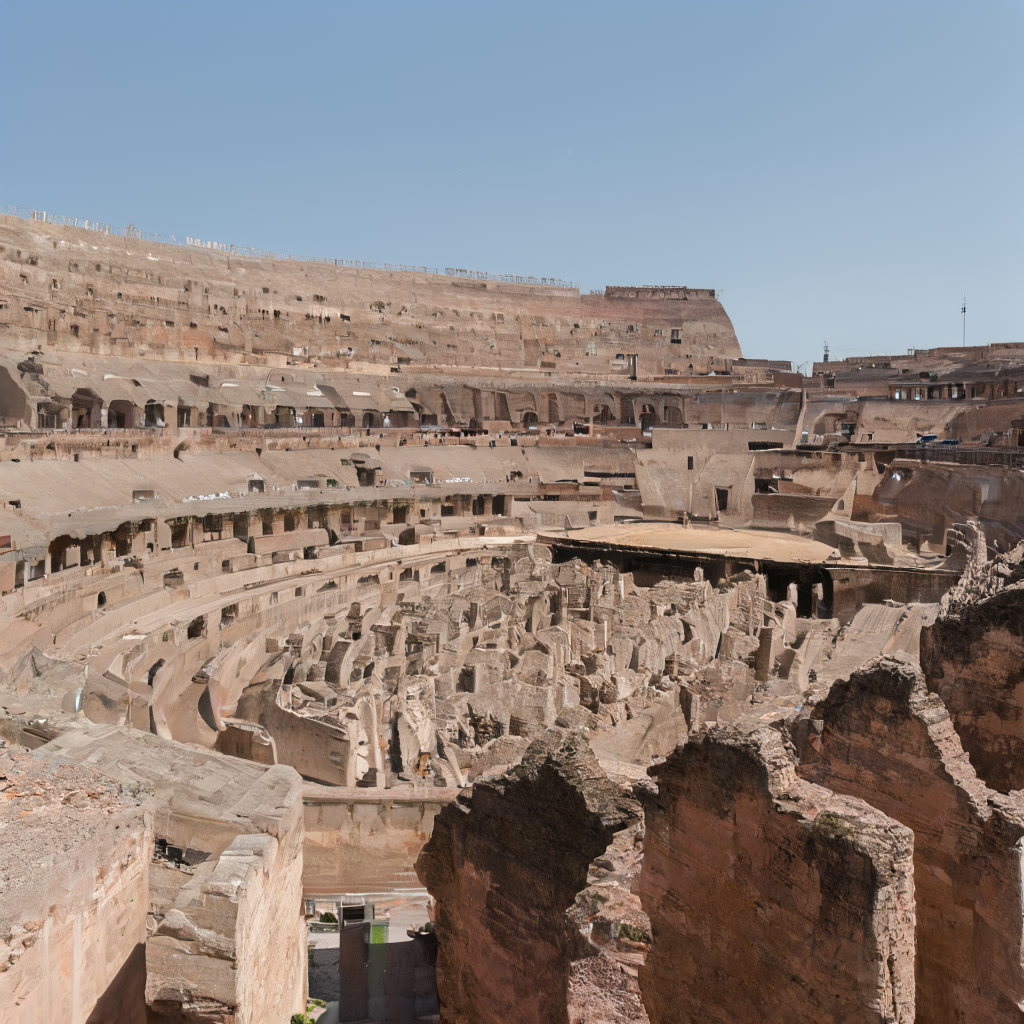}
\end{subfigure}
\caption{\textbf{DPDD-SR model soup results with OptIMerge on DF2K (SR) data}. (left-to-right) We provide input, target, and images generated from the expert model, naive model soup, and OptIMerge}
\label{fig:sr_dpdd_soup_on_sr_data}
\end{figure}

\end{document}